\documentclass[sn-mathphys,Numbered]{sn-jnl}

\usepackage{graphicx}%
\usepackage{multirow}%
\usepackage{amsmath, amssymb, amsfonts}%
\usepackage{amsthm}%
\usepackage{mathrsfs}%
\usepackage[title]{appendix}%
\usepackage{xcolor}%
\usepackage{textcomp}%
\usepackage{manyfoot}%
\usepackage{booktabs}%
\usepackage{algorithm}%
\usepackage{algorithmicx}%
\usepackage{algpseudocode}%
\usepackage{listings}%
\usepackage{caption}%
\usepackage{makecell}%
\usepackage{hyperref}%
\usepackage{url}%
\usepackage{subcaption}%
\usepackage[utf8]{inputenc}%
\usepackage[T1]{fontenc}%
\usepackage[english]{babel}%
\usepackage{geometry}%
\usepackage{fancyhdr}%
\usepackage{microtype}%
\usepackage{parskip}%
\usepackage{tabularx}%
\usepackage{enumitem}%
\usepackage[capitalize,noabbrev]{cleveref}%
\usepackage{float}%
\usepackage{orcidlink}

\theoremstyle{thmstyleone}%
\theoremstyle{thmstyletwo}%

\theoremstyle{thmstylethree}%

\begin{document}

\title[Article Title]{On the Promises and Challenges of Multimodal Foundation Models for Geographical, Environmental, Agricultural, and Urban Planning Applications}



 \author[5]{\fnm{Chenjiao} \sur{Tan}}\email{c.tan@ufl.edu}
 \equalcont{These authors contributed equally to this work.}

 \author[1]{\fnm{Qian} \sur{Cao}}\email{qian.cao1@uga.edu}
 \equalcont{These authors contributed equally to this work.}

 \author[2]{\fnm{Yiwei} \sur{Li}}\email{yiwei.li@uga.edu}
 \equalcont{These authors contributed equally to this work.}

 \author[1]{\fnm{Jielu} \sur{Zhang}}\email{jielu.zhang@uga.edu}
 \equalcont{These authors contributed equally to this work.}

 \author[4]{\fnm{Xiao} \sur{Yang}}\email{xy50573@uga.edu}
 \equalcont{These authors contributed equally to this work.}

 \author[2]{\fnm{Huaqin} \sur{Zhao}}\email{huaqin.zhao@uga.edu}
 \equalcont{These authors contributed equally to this work.}

 \author[2]{\fnm{Zihao} \sur{Wu}}\email{zihao.wu1@uga.edu}
 \equalcont{These authors contributed equally to this work.}

 \author[2]{\fnm{Zhengliang} \sur{Liu}}\email{zl18864@uga.edu}
 \equalcont{These authors contributed equally to this work.}

 \author[1]{\fnm{Hao} \sur{Yang}}\email{haoyang@uga.edu}
 \equalcont{These authors contributed equally to this work.}

 \author[1]{\fnm{Nemin} \sur{Wu}}\email{nemin.wu@uga.edu}
 \equalcont{These authors contributed equally to this work. The order was decided by random seed.}

 \author[6]{\fnm{Tao} 
 \sur{Tang}}\email{Yangtao128@tsinghua.edu.cn}

 \author[3]{\fnm{Xinyue} 
 \sur{Ye} \orcidlink{https://orcid.org/0000-0001-8838-9476} }\email{xinyue.ye@tamu.edu}

 \author*[4]{\fnm{Lilong} 
 \sur{Chai} \orcidlink{https://orcid.org/0000-0002-5378-6727} }\email{lchai@uga.edu}

 \author*[2]{\fnm{Ninghao} 
 \sur{Liu} \orcidlink{https://orcid.org/0000-0002-9170-2424}}\email{ninghao.liu@uga.edu}

 \author*[5]{\fnm{Changying} 
 \sur{Li} \orcidlink{https://orcid.org/0000-0003-2590-4797} }\email{cli2@ufl.edu}

 \author*[1]{\fnm{Lan} 
 \sur{Mu} \orcidlink{https://orcid.org/0000-0003-0199-9509} }\email{mulan@uga.edu}

 \author*[2]{\fnm{Tianming} 
 \sur{Liu} \orcidlink{https://orcid.org/0000-0002-8132-9048} }\email{tliu@uga.edu}

 \author*[1,2]{\fnm{Gengchen} 
 \sur{Mai} \orcidlink{https://orcid.org/0000-0002-7818-7309}}\email{gengchen.mai25@uga.edu}


 \affil[1]{\orgdiv{Department of Geography}, \orgname{University of Georgia}, \orgaddress{\city{Athens}, 
 \state{GA}, \country{US}}}

 \affil[2]{\orgdiv{School of Computing}, \orgname{University of Georgia}, \orgaddress{\city{Athens}, 
 \state{GA}, \country{US}}}

 \affil[3]{\orgdiv{Department of Landscape Architecture and Urban Planning \& Department of Computer Science and Engineering}, \orgname{Texas A\&M University}, \orgaddress{\city{College Station}, 
 \state{TX}, \country{US}}}

 \affil[4]{\orgdiv{School of Electrical and Computer Engineering}, \orgname{University of Georgia}, \orgaddress{\city{Athens}, 
 \state{GA}, \country{US}}}

 \affil[4]{\orgdiv{Department of Poultry Science}, \orgname{University of Georgia}, \orgaddress{\city{Athens}, 
 \state{GA}, \country{US}}}

 \affil[5]{\orgdiv{Department of Agricultural and Biological Engineering}, \orgname{University of Florida}, \orgaddress{\city{Gainesville}, 
 \state{FL}, \country{US}}}

 \affil[6]{\orgdiv{School of Architecture}, \orgname{Tsinghua University}, \orgaddress{\city{Beijing}, 
  \country{China}}}


\abstract{The advent of large language models (LLMs) has heightened interest in their potential for multimodal applications that integrate language and  vision. This paper explores the capabilities of GPT-4V in the realms of geography, environmental science, agriculture, and urban planning by evaluating its performance across a variety of tasks. Data sources comprise satellite imagery, aerial photos, ground-level images, field images, and public datasets. The model is evaluated on a series of tasks including geo-localization, textual data extraction from maps, remote sensing image classification, visual question answering, crop type identification, disease/pest/weed recognition, chicken behavior analysis, agricultural object counting, urban planning knowledge question answering, and plan generation. The results indicate the potential of GPT-4V in geo-localization, land cover classification, visual question answering, and basic image understanding. However, there are limitations in several tasks requiring fine-grained recognition and precise counting. While zero-shot learning shows promise, performance varies across problem domains and image complexities. The work provides novel insights into GPT-4V's capabilities and limitations for real-world geospatial, environmental, agricultural, and urban planning challenges. Further research should focus on augmenting the model's knowledge and reasoning for specialized domains through expanded training. Overall, the analysis demonstrates foundational multimodal intelligence, highlighting the potential of multimodal foundation models (FMs) to advance interdisciplinary applications at the nexus of computer vision and language.}




\maketitle

\newpage 

\section{Introduction}\label{sec:intro}
Large language models (LLMs) and multimodal foundation models (FMs) have shown immense potential for generalized intelligence across diverse tasks and modalities \cite{liu2023summary,zhao2023brain,cao2023comprehensive,liu2023llava}. Leveraging immense data and computational scale, models like GPT-4V \cite{bubeck2023sparks} and DALL-E 2 \cite{ramesh2022hierarchical} acquire broad capabilities from large-scale pretraining, enabling zero-shot transfer to new data and new tasks. However, rigorous analysis is essential to evaluate their applicability to complex scientific domains combining vision, language, and reasoning. This paper provides an empirical assessment of GPT-4V, a state-of-the-art multimodal LLM \cite{liu2023tailoring, liu2023evaluating, dai2023ad, liu2023radonc}, on interdisciplinary challenges spanning geography \cite{mai2022towards,mai2023opportunities,hu2023geo,xie2023geo,rao2023building,roberts2023gpt4geo,zhang2023geogpt,manvi2023geollm,zhang2023text2seg}, environmental science\cite{zhu2023chatgpt,agathokleous2023use}, agriculture \cite{rezayi2022agribert, rezayi2023exploring,lu2023agi}, and urban planning \cite{liu2023transformation,mai2023opportunities}.

As the state-of-the-art multimodal foundation model, GPT-4V leverages a transformer-based \cite{vaswani2017attention} architecture pretrained on image-text data pairs \cite{kim2023medivista}, acquiring joint visual and linguistic representations. This allows combining the strengths of LLMs for natural language understanding with computer vision models adept at image recognition and classification. Prior to this, GPT-4V has been tested in the field of medical images \cite{liu2023holistic} and social media image analysis \cite{lyu2023gpt, liu2023transformation} and proved to have strong professional image understanding capabilities. To investigate the effectiveness of GPT-4V on various geospatial, environmental, agricultural, and urban planning tasks, in this work, we comprehensively analyze GPT-4V's zero-shot performance on various geoscience and agricultural tasks including geo-localization, map data extraction, remote sensing image classification, remote sensing visual question answering, crop type identification, disease/pest/weed recognition, chicken behavior analysis and object counting, urban planning knowledge question answering, and plan generation. The model is tested on diverse real-world data including satellite imagery, aerial photos, ground-level images, field images, and public datasets relevant to the applications.

The advanced capabilities of GPT-4V in geographical, environmental, agricultural, and urban planning applications mark a significant leap in AGI's role in these fields\cite{yang2023dawn}. In these geoscience and agricultural tasks, the model excels in fieldwork assistance, identifying plant and animal species, detecting both point-source and non-point source pollution, and recognizing alarming weather phenomena from photographs and satellite images. It also plays a pivotal role in environmental monitoring, ecosystem change tracking, pollution detection, deforestation identification, glacier retreat recognition, habitat change detection, and wildlife population estimation. This technology is vital for documenting and tracking environmental changes, offering a dynamic approach to analyzing these changes over time, which is essential for environmental assessment and planning.

In geo-localization and remote sensing visual question answering tasks, GPT-4V's proficiency in interpreting complex geographical and environmental data underscores its potential as a powerful tool for geospatial image applications. Its capabilities extend to agriculture, where it revolutionizes the approach to plant health and yield optimization by identifying crop types, diseases, pests, and weeds from visual cues. Additionally, GPT-4V demonstrates its utility in animal welfare and farm productivity through its ability to analyze chicken behavior and perform object counting tasks, thus contributing to the advancement of automated monitoring systems within poultry farms. These diverse applications highlight GPT-4V's transformative impact across various aspects of environmental science and agriculture.

However, our analysis also highlights the limitations of GPT-4V, especially in tasks requiring fine-grained recognition, precise counting, and reasoning. These challenges underscore the need for further refinement in the model's training, particularly in handling complex and nuanced scenarios that are commonplace in scientific domains. The zero-shot and few-shot learning ability of GPT-4V, while impressive, varies in its effectiveness across different problem domains and the complexity of the images it processes.

This research offers a comprehensive view of the capabilities and limitations of GPT-4V in addressing real-world challenges in geospatial, agricultural, environmental, and urban planning contexts. By doing so, it not only contributes to the understanding of the current state of multimodal AI models in scientific applications but also provides a roadmap for future enhancements. The insights gained from this study are instrumental in guiding the development of more robust and specialized AI systems, capable of handling the intricate demands of interdisciplinary scientific research. The cross-disciplinary benchmarking methodology and findings will inform continued progress in developing robust and generalizable AI systems. As foundation models grow in scale and training data diversifies, their transfer learning potential for varied real-world tasks will rapidly expand. This work provides key insights into current capabilities, guiding the advancement of multimodal intelligence for broad scientific applications. The findings in the tested domains and tasks include:  

\begin{itemize}
\item \textbf{Geography:} 
    \begin{itemize}
        \item \textbf{Image Geo-localization \cite{weyand2016planet,seo2018cplanet,muller2018geolocation,izbicki2020exploiting,cepeda2023geoclip}:} GPT-4V shows strong performance for geo-localization in urban areas, accurately identifying landmarks and urban features, but struggles with natural landscapes and lacks precision in areas without distinct artificial objects.
        
        \item \textbf{Text Localization and Extraction from Historical Maps \cite{kim2023mapkurator,liu2023transformation}:} GPT-4V is effective in extracting textual content from historical map images, identifying over half of the texts in map tiles, but faces challenges in generating precise text locations and output consistency.
        
        \item \textbf{Remote Sensing Image Classification \cite{ayush2021geography,manas2021seasonal,mai2023sphere2vec,mai2023csp}:} GPT-4V demonstrates proficiency in distinguishing icebergs from ships in satellite imagery, particularly successful in identifying icebergs but showing limitations in accurately classifying ships, especially in complex scenarios.
        
        \item \textbf{Remote Sensing Visual Question Answering \cite{lobry2020rsvqa,lobry2021rsvqa,chappuis2022prompt}:} The model excels in recognizing land cover types and understanding geographic scales in remote sensing images but displays limitations in counting tasks and fine-grained image segmentation.
    \end{itemize}

\item \textbf{Environmental Science:} 
    \begin{itemize}
        \item \textbf {Environment Monitoring:} In our experiments of air quality evaluation to GPT-4V, it is demonstrated that GPT-4V showed a certain degree of speculative capability in estimating Air Quality Index(AQI) categories when provided with reference images. However, it lacks the ability to predict precise AQI values. For future work, we are going to test more about environmental monitoring, such as assessing the health of ecosystems\cite{lackey2001values}, tracking pollution levels\cite{sicard2023trends}, monitoring deforestation\cite{wang2023siamhrnet}, glacier retreat\cite{cauvy2019global}, habitat changes\cite{chen2023habitat}, and observing wildlife populations\cite{taylor2023associations}. 

        \item \textbf{Future Work for Fieldwork Assistance:} GPT-4V showed enhanced capability to potentially identify plant and animal species\cite{ge2023mllm} and alarming weather phenomena\cite{youvan2023interwoven} in prior research. This notable capability holds great promise for expanding its utility in diverse fieldwork scenarios, such as habitat assessment, early detection of invasive species, and biodiversity surveys.

         \item \textbf{Future Work for Documenting and Tracking Environmental Changes:} Environmental assessment and planning necessitate the documentation of pre- and post-scenarios\cite{ray2015confronting}. In future endeavors, leveraging GPT-4V can enhance the process by seamlessly aiding in the real-time identification and comparison of visual data. This approach offers a more dynamic means to report and analyze environmental changes over time, providing a versatile and efficient tool for comprehensive assessments in environmental research and planning.
         
    \end{itemize}

\item \textbf{Agriculture:} 
    \begin{itemize}
        \item \textbf{Remote Sensing-based Fine-Grained Crop Type Identification \cite{cai2018high,lu2022fine}:}
         GPT-4V's performance in identifying cropland types showed significant variability across different regions. It is more efficient in areas with distinct patterns, clear textures, and regular shapes. However, GPT-4V faced challenges in complex surroundings like forests or irregular paths, leading to difficulties in cropland type identification. 

        \item \textbf{Detection of Nutrition Deficiency in Crops \cite{chiu20201st,Chiu_2020_CVPR_Workshops,mccauley2009plant,wulandhari2019plant,feng2020advances,barbedo2019detection}:}
        GPT-4V showed effectiveness in identifying nutrient deficiencies, particularly nitrogen and chlorophyll, through signs and pixel analysis. However, it requires detailed instructions and domain knowledge for effective use. 

        \item \textbf{Plant Disease, Weeds Recognition, and Phenotyping\cite{hasan2021survey,li2021plant,song2021high}:} 
        GPT-4V successfully recognized cotton diseases and pests using symptom identification, as well as correctly classified weeds based on leaf features, which showed its effectiveness and, to a certain degree, its zero-shot learning abilities in disease, pest, and weed recognition. Additionally, it showed potential in counting tasks under simple scenarios but was inconsistent in challenging scenarios with occlusions.

        \item \textbf{Poultry Science \cite{yang2023computer,subedi2023tracking}:} 
        We conduct a series of experiments on various poultry science tasks such as eggshell issue identification, 
        chicken number quantification, and chicken behavior monitoring.

        The analyses 
demonstrate GPT-4V's adeptness in identifying various issues related to egg quality and chicken behavior. It successfully diagnosed potential problems with eggshells, such as staining and structural weaknesses, indicative of environmental and dietary factors. GPT-4V's keen observation of a hen's behavior provided insights into typical actions like egg-laying and brooding. Moreover, its capability to count and categorize chickens, coupled with the assessment of their social interactions and environmental conditions, highlights a valuable application of AI in improving productivity and welfare practices on poultry farms. GPT-4V's precision in these analyses underscores its utility as a tool for enhancing the management of poultry health and welfare..

    \end{itemize}

\item \textbf{Urban Planning:} 
    \begin{itemize}
        \item \textbf{Urban Planning Knowledge Question Answering \cite{yin2019nlp,feng2021intelligent,martinez2022development}:} GPT-4V demonstrates an excellent command of urban planning theories and practices, evidenced by its interpretation of various visual materials.

        \item \textbf{Master Plan \cite{dunham1958city,peter2019urban,altschuler2018goals}:} The model exhibits satisfactory performance in interpreting land use plans, provided that detailed prompts with legend information are given. However, it fails to extract and generate data about other urban systems, such as the transportation system, from these plans. 

         \item \textbf{Urban Street Design \cite{bevan2007sustainable,hassen2016examining,dover2013street}:} The model demonstrates proficiency in understanding and analyzing street-scale urban design illustrations. It is also capable of proposing specific and reasonable improvement solutions. However, GPT-4V struggles with balancing constraints and proposed solutions, which can sometimes lead to impractical or unrealistic urban designs.
    \end{itemize}
\end{itemize}


Based on a systematic investigation of GPT-4V's capabilities on various geoscience tasks, our core discoveries are listed below:
    \begin{itemize}
        \item GPT-4V shows strong performances in tasks such as image geo-localization, land cover classification, diagnostic analysis, and basic image understanding.
        \item The limitations of GPT-4V can be seen in fine-grained recognition and counting tasks, particularly in complex backgrounds and with occlusions.
        \item We observe variable performances across different problem domains and image complexities.
        \item These experiments indicate the promising capabilities of GPT-4V but also imply a need for more specialized domain training.
    \end{itemize}

\section{Background}\label{sec:background}
\subsection{LLMs and Multimodal Models}
Transformer-based \cite{vaswani2017attention} models have significantly advanced \cite{10.1007/978-3-031-45673-2_46,liu2023radiologyllama2,tang2023policygpt,liu2023radoncgpt,liuradiology,LIU2023100045,dou2023artificial,holmes2023evaluating,gong2023evaluating,liu2023transformation,holmes2023benchmarking,shi2023mededit,zhong2023chatabl,zhao2023generic,zhou2023fine,liao2023mask} the fields of natural language processing (NLP), computer vision (CV), and other application domains, laying the groundwork for the development of multimodal models such as LLaVA \cite{liu2023llava} and GPT-4V. The transformer architecture 
employs self-attention mechanisms that enable 
parallel processing of the sequential data and weighting mechanism across different parts of the input data. This feature has been crucial in handling the complexities of human language and visual data \cite{liu2023summary}.

In NLP, Bidirectional Encoder Representations from Transformers (BERT) \cite{devlin2018bert} emerged as a major breakthrough that leverages bidirectional training to understand the context of a word based on all its surroundings in a sentence. Following BERT, models like RoBERTa \cite{liu2019roberta} and domain-specific variants such as AgriBERT \cite{rezayi2022agribert}, ClinicalRadioBERT \cite{rezayi2022clinicalradiobert} and BioBERT \cite{lee2020biobert} improved upon this foundation, offering more nuanced language understanding. GPT-3 \cite{brown2020language}, a generative pretrained model, further expanded these capabilities with 175 billion parameters, marking a substantial leap with 175 billion parameters, enabling it to generate pages of human-like text and achieving significant adoption for text generation tasks.

InstructGPT \cite{ouyang2022instructgpt}, a version of GPT-3 fine-tuned through reinforcement learning from human feedback (RLHF), was launched in January 2022 to address issues such as offensive language and misinformation, aiming to provide more helpful responses. A further refined version of InstructGPT was ChatGPT \cite{openaiIntroducingChatGPT}, which became particularly notable for its conversation-focused capabilities, reaching 100 million users within two months of its release. The introduction of GPT-4 \cite{openai2023gpt4} in March 2023 marked a further enhancement of these capabilities, especially for complex tasks. All these kinds of large language models have demonstrated success \cite{10.1007/978-3-031-21014-3_28,PMID:36097765,liao2023maskguided,rezayi2023exploring} in numerous domains \cite{dai2023auggpt,liu2023deidgpt,ma2023impressiongpt,liao2023differentiate,dai2023adautogpt,liu2023summary,guan2023cohortgpt,cai2022coarse,liu2023pharmacygpt,shi2023mededit,gong2023evaluating} and have transformed the landscape of AI\cite{liu2023transformation,ZHAO2023100005,Holmes_2023,wu2023exploring,rezayi2022agribert,liu2023radiologygpt,Liu_2023,wang2023review,li2023artificial,cai2023coarsetofine,holmes2023benchmarking}. 

Computer vision has experienced parallel advancements through the development of models such as the Vision Transformer (ViT) \cite{dosovitskiy2020image} and Masked Autoencoders (MAE) \cite{he2022mae}. ViT, applying transformer principles to image classification tasks, divides an image into patches and processes them as a sequence, enabling the model to capture both local and global features within the image. This approach has found wide applications across various domains \cite{wang2023review,dai2023samaug,zhang2023segment,xiao2023instructionvit,liu2022survey}, including fields traditionally dominated by convolutional neural networks (CNNs) \cite{zhao2022embedding,dai2022graph,liu2022discovering,zhang2023beam,liu2020survey,bi2023community,ding2023deep,ding2022accurate}. More recent developments such as the Segment Anything Model (SAM) \cite{kirillov2023segment} showcases impressive generalization capabilities and zero-shot learning, making it suitable for processing diverse aerial and orbital images. 
Osco et al \cite{osco2023segment} and Zhang et al \cite{zhang2023text2seg} 
highlighted SAM's potential for remote sensing applications but also noted its limitations in complex scenarios with lower spatial resolutions. A new technique combining text-prompt with one-shot training was found to improve SAM’s accuracy, demonstrating its adaptability to remote sensing tasks and reducing the need for manual annotations. Despite challenges, SAM's adaptability makes it a promising tool for future research in remote sensing image processing.

Building upon the foundation established by LLMs and ViT, there is a growing demand for the development of multimodal models \cite{liu2023artificial,chen2023ma,holmes2023evaluating,cai2023multimodal,holmes2023evaluating2,liu2023llava,mai2023opportunities}. These models are designed to utilize multimodal inputs, enhancing robustness in the pursuit of Artificial General Intelligence (AGI). Specifically, multimodal models integrate tasks from Computer Vision and Natural Language Processing (NLP) to improve perception and interaction across diverse environments and scenarios \cite{xiao2023instruction}. They are particularly adept at aligning with human multimodal instructions, which may encompass not just text but also audio, images, video, and even brain signals such as EEG or fMRI, facilitating brain-computer interaction \cite{zhao2023brain}.

Specifically, these newly released multimodal models harness pretrained ViT as image encoders and LLMs such as LLaMA 2 \cite{touvron2023llama} and Vicuna \cite{vicuna2023} as interfaces for various vision-language inputs.  By leveraging large-scale, multimodal instruction-following datasets along with the advanced reasoning capabilities inherent in LLMs, these models demonstrate enhanced alignment with human preferences and exhibit emergent abilities when encountering never-before-seen data \cite{chen2023minigpt,wu2023next,han2023imagebind,liu2023visual}.

The evolution of these models culminated in the development of GPT-4V \cite{openaiGPT-4VisionSystem}, a large multimodal model that integrates visual understanding with language processing. GPT-4V is trained on a vast corpus of multimodal data and has shown robust visual comprehension abilities. This model significantly expands the capabilities of previous models by processing both text and images, a key feature distinguishing it from ChatGPT, which primarily focuses on text.

\subsection{Applications of FMs on Geography and Environmental Science}
Recent advancements in Geographic Information System (GIS) and Geospatial Artificial Intelligence (GeoAI) \cite{janowicz2020geoai,gao2023handbook} technology have been significantly influenced by the integration of Foundation Models (FMs) like ChatGPT and GPT-4V, leading to transformative developments across various geospatial domains \cite{mai2022towards,mai2023opportunities,xie2023geo,rao2023building,hu2023geoknowgpt,zhang2023geogpt,manvi2023geollm,roberts2023gpt4geo,lacoste2023geobench,balsebre2023cityfm}.

In the context of disaster response, Geo-Knowledge-Guided GPT models \cite{hu2023geo} have marked a significant advancement in location description recognition from disaster-related tweets. Traditional place name recognition tools \cite{alex2019geoparsing,karimzadeh2019geotxt} often struggle with extracting complex location descriptions such as full street addresses from text data such as social media messages. To overcome this, a method combining geo-knowledge with Generative Pre-trained Transformer models, such as ChatGPT and GPT-4, has been introduced. This approach, requiring minimal training examples but rich in geo-knowledge, excels in extracting location descriptions from disaster-related messages, thereby significantly enhancing the precision in locating disaster victims and improving the overall efficiency of disaster response strategies.

Another key development is the creation of GeoLLM \cite{manvi2023geollm}, a method that extracts geospatial knowledge from LLMs using auxiliary map data from sources like OpenStreetMap. This approach, which diverges from traditional methods reliant on expensive or limited satellite imagery, demonstrates the untapped potential of LLMs in geospatial prediction tasks, showcasing superior performance over existing methods.

However, the development of multimodal foundation models for geospatial AI also faces significant challenges, particularly due to the multimodality of geospatial tasks \cite{mai2023opportunities}. While FMs show prowess in text-based tasks, they often underperform in geospatial tasks involving multiple data modalities, especially tasks that involve vector data \cite{mai2023spatialrl}. The proposal of a multimodal foundation model is a promising pathway to address these challenges. Such a model would enable effective reasoning across various types of geospatial data, potentially bridging the gap in current limitations and enhancing the overall capability of geospatial AI.

Alongside these advancements, the privacy and security considerations \cite{rao2023building} in the development of GeoAI foundation models, encompassing language, vision, and multimodal models, have become increasingly critical. A comprehensive framework to mitigate privacy and security risks during GeoAI foundation model development highlights the necessity of responsible development practices in applications like geographic question answering, remote sensing image understanding, and location-based services.

In addition, a recently proposed concept of Autonomous GIS \cite{li2023autonomous} represents a major stride in AI integration within the GIS industry. The proposed LLM-Geo system, utilizing the GPT-4 API in a Python environment, exemplifies this advancement, aiming for minimal human intervention and capable of self-generation, organization, verification, execution, and growth. This approach not only demonstrates the potential of AI in revolutionizing GIS technology but also sets a precedent for future self-reliant and intelligent GIS systems.

The integration of large language models and large vision models in remote sensing represents a groundbreaking shift in the field. Projects like Text2Seg \cite{zhang2023text2seg} are at the forefront, pioneering the use of text prompts to guide visual foundation models for semantic segmentation tasks. This approach, which leverages advanced models such as Grounding DINO \cite{liu2023grounding} and SAM \cite{kirillov2023segment}, illustrates the immense potential of combining textual and visual data. The synergy of these modalities enables a deeper understanding of remote sensing imagery, facilitating more accurate and detailed image analysis. This method provides an intuitive and effective way of processing complex semantic segmentation tasks in remote sensing.

In parallel, the Tree-GPT framework \cite{du2023tree} is revolutionizing forestry remote sensing data analysis. This innovative approach integrates LLMs into the domain, enhancing data analysis capabilities through a combination of image understanding modules, domain-specific knowledge bases, and specialized toolchains. Tree-GPT empowers LLMs to comprehend images and perform accurate data analysis, signifying a significant advancement in the application of AI in forestry remote sensing. The convergence of LLMs with vision models in these areas not only increases the efficiency and accuracy of remote sensing data interpretation but also opens up new possibilities for environmental monitoring and management \cite{yang2023dawn}.

\subsection{Applications of FMs on Agriculture}
In the field of agriculture, the application of LLMs is ushering in a new era of innovation and efficiency \cite{yang2023dawn,lu2023agi}. A key development in this area is AgriBERT \cite{rezayi2022agribert}, a transformer-based language model fine-tuned specifically for agricultural NLP tasks. This model excels in semantic matching between food descriptions and nutrition data, demonstrating the versatility and power of GPT-based models in the agricultural domain \cite{rezayi2023exploring}. The use of ChatGPT as an external knowledge source further expands the scope of agricultural NLP, enabling more precise and contextually relevant analyses.

In addition, GPT-4 was integrated with an object detection model to facilitate disease diagnosis in the agricultural domain \cite{qing2023gpt}. A mapping method between detection results and text was proposed to simplify image information to match the text processing capability of LLMs. The results showed that the developed approach provided reasonable diagnosis reports and solutions with good generalization capability \cite{yang2023dawn}.

Furthermore, LLMs are transforming agricultural extension services by simplifying scientific knowledge and providing personalized, location-specific, and data-driven recommendations \cite{lu2023agi}. This approach overcomes the traditional limitations in agricultural extension, such as limited institutional capacity. By utilizing pre-trained transformers \cite{bubeck2023sparks}, these models offer a new way to disseminate agricultural knowledge, making it more accessible and relevant to farmers \cite{rezayi2022agribert}. This technological leap in agriculture is not only enhancing the efficiency of data processing and analysis but also contributing significantly to the advancement of sustainable farming practices and food security.

\subsection{Applications of FMs on Urban Planning}
The emergence of large language models and multimodal foundation models has equipped urban planners with efficient tools that enhance the entire urban planning process. This includes pre-planning, planning generation, and planning evaluation.

In the stage of pre-planning, urban planners utilize models to conduct domain knowledge retrieval \cite{wang2023optimizing} and urban region profiling \cite{yan2023urbanclip} as the preparation of planning. Based on LLMs such as ChatGPT and ChatGLM, researchers generate question-answering datasets using domain literature corpora and fine-tuning the model to create an LLM for urban planning, supporting planners to enhance knowledge for specific regions or subfields. UrbanCLIP \cite{yan2023urbanclip} is another model for learning the site before planning. It is capable of predicting urban indicators in metropolitan areas from remote sensing images by integrating textual modality into urban imagery analysis.

In the stage of planning generation, text-to-text and text-to-image models are employed to generate texts and images depicting planning. A 15-minute City concept tool \cite{deshpandegenerative} is developed after fine-tuning the pre-trained language model with structured contents of the 15-minute City, automatically generating community plan texts that reach qualified levels evaluated by human experts. DALL-E \cite{ramesh2021dalle} is another promising tool for urban design, and it is especially proficient in illustrating the urban design to the public for walkable streets via changing street elements on street view images to the public \cite{walker2022dalle}. 

In the stage of planning evaluation, researchers use ChatGPT to evaluate multiple plans and compare the model's evaluation results with trained human coders' results, finding that the model gets an average of 68\% accuracy when compared with the results from human evaluators \cite{fu2023can}. The research reveals ChatGPT's limitation in comprehending urban planning terminology but also shows its potential to assist human experts by capturing details in complex documents.

\subsection{Benefits and Risks of Multimodal FMs}
LLMs like ChatGPT pose both direct and indirect environmental impacts \cite{rillig2023risks,shi2023thinking}. While they can contribute to carbon emissions due to the energy used for training and running \cite{touvron2023llama}, they may also offset this by reducing the need for more energy-intensive activities like video conferencing. Moreover, foundation model development also has raised lots of ethical \cite{gehman2020realtoxicityprompts,zhao2018gender,wang2023decodingtrust} and privacy concerns \cite{rao2023building}. There's a risk of misinformation and biases in training data, which could affect environmental literacy. 
On the research front, LLMs and FMs could aid scientists but also perpetuate biases and be misused in academic misconduct.
To tackle this problem, various efforts have been contributed to quantify the toxicity \cite{deshpande2023toxicity}, gender bias \cite{zhao2018gender}, geographic and geopolitical Biases \cite{faisal2022geographicbias}, and trustworthiness \cite{wang2023decodingtrust} in foundation models.

\section{Multimodal FMs Applications on Geography} \label{sec:geo}
In the following, we will systematically investigate the effectiveness of the state-of-the-art multimodal foundation model, GPT-4V, on a series of geographic tasks.

\subsection{Image Geo-localization} \label{sec:img-local}
\subsubsection{Data Source}
Image geo-localization \cite{weyand2016planet,seo2018cplanet,muller2018geolocation,izbicki2020exploiting,cepeda2023geoclip}, also known as image retrieval, is a task that predicts the geolocation of images solely based on their visual information. When dealing with street view images, remote sensing images, or landscape photos, people can discern the geolocation from architectural style, road signs, terrain, flora, or fauna. In this section, we utilize the Yahoo-Flickr Creative Commons 100 Million (YFCC100M) dataset to evaluate the performance of GPT-4V on the task of image geo-localization. YFCC is a public collection containing nearly 100 million photos with metadata from Flickr \cite{thomee2016yfcc100m}. We manually choose geo-informed outdoor photos from various countries to examine GPT-4V’s capability of image geo-localization. 

\subsubsection{Analysis and Results}
\begin{figure}[H]
\centering
\includegraphics[width=0.9\linewidth]{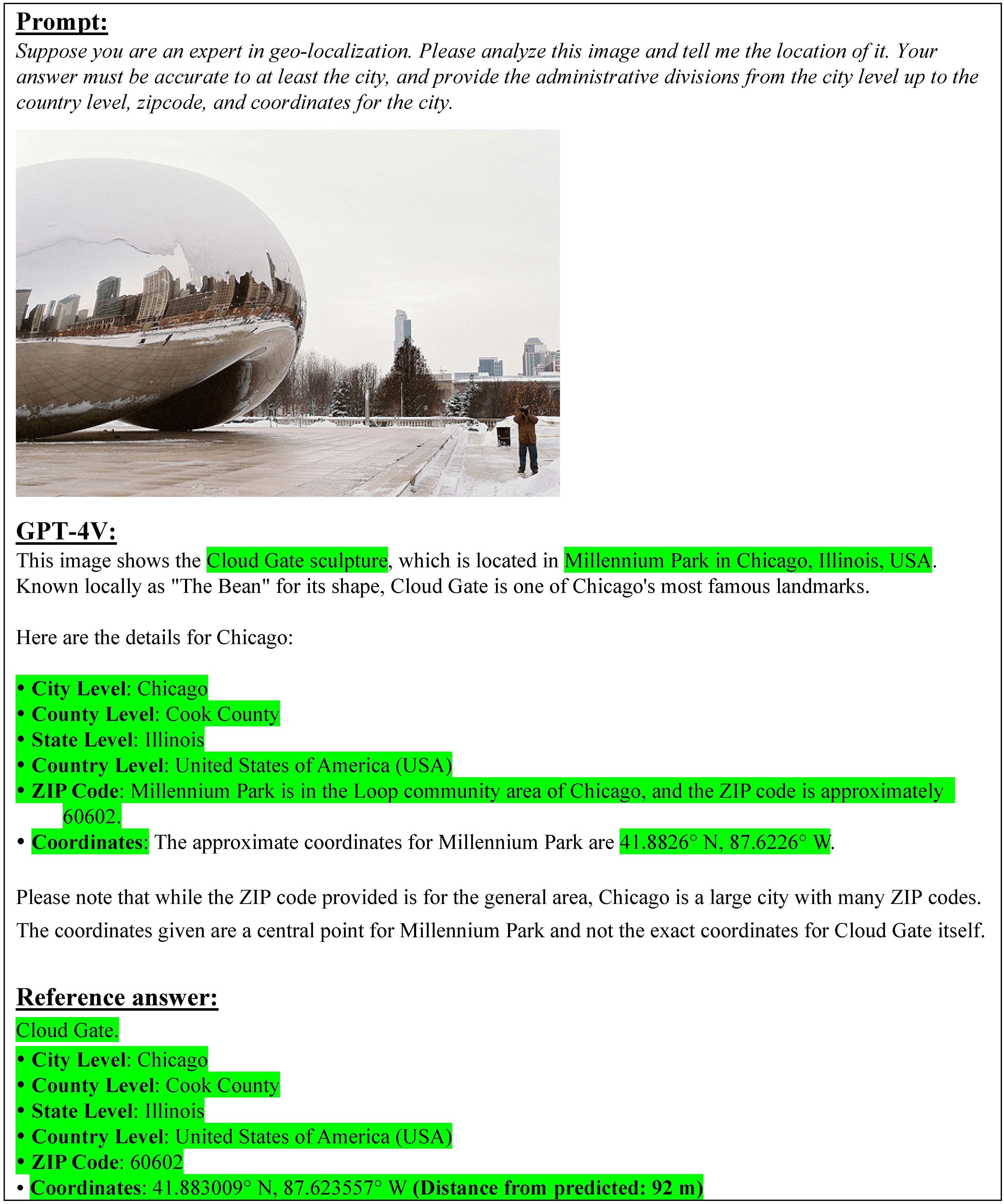}
\captionsetup{justification=raggedright,singlelinecheck=false}
\caption{Geo-localization of the Cloud Gate sculpture, USA. For all prompts we listed in this paper, we use different colors to indicate the correctness of GPT-4V's outputs. Here, texts highlighted in \textbf{green} color indicate correct answers from GPT-4V and texts with \textbf{red} color indicate wrong answers. And \textbf{yellow} color indicate uncertain results.}
\label{fig:geoloc1}
\end{figure}

\begin{figure}[H]
\centering
\includegraphics[width=0.9\linewidth]{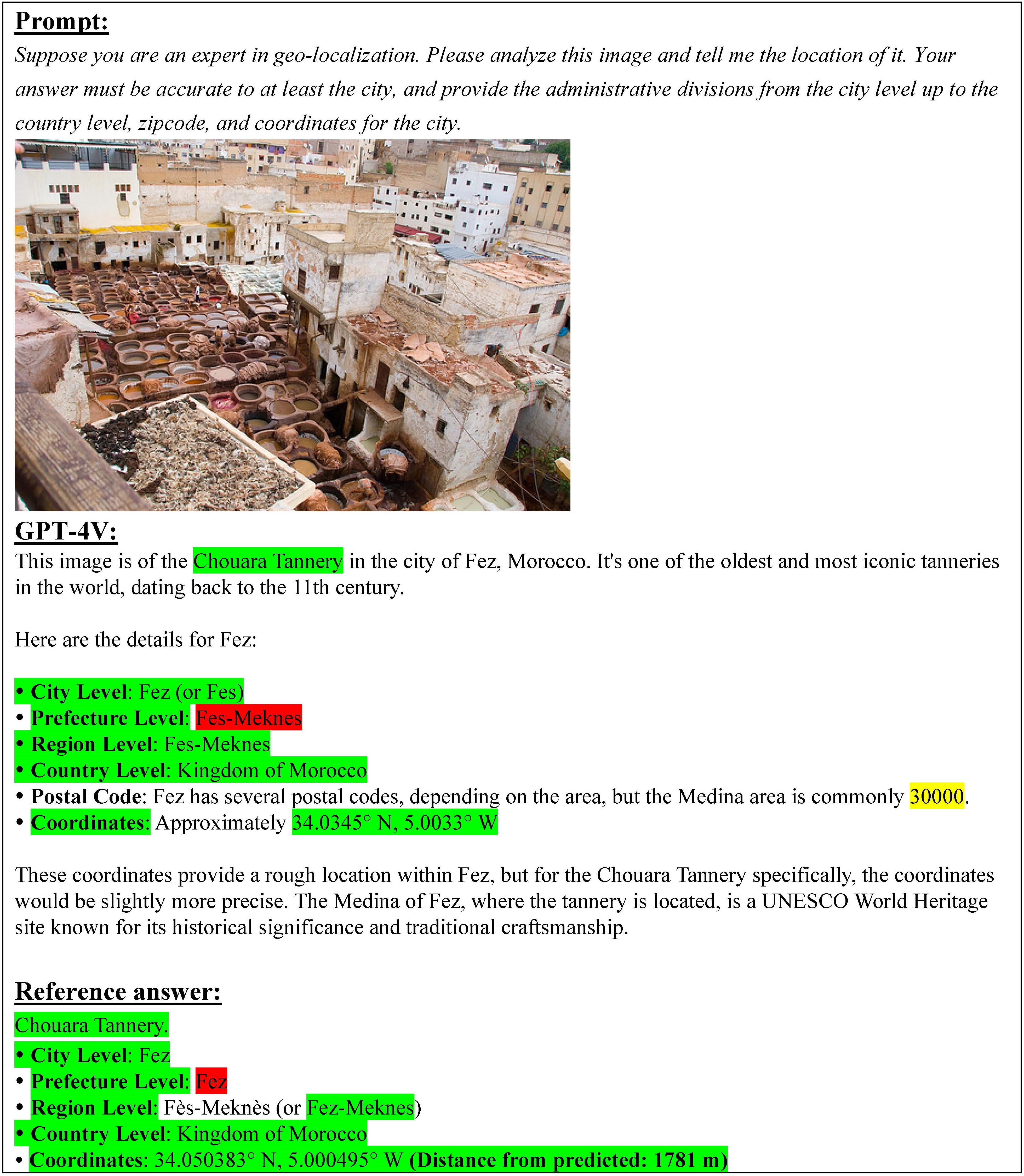}
\captionsetup{justification=raggedright,singlelinecheck=false}
\caption{Geo-localization of the Chouara Tannery, Morroco.}
\label{fig:geoloc2}
\end{figure}

\begin{figure}[H]
\centering
\includegraphics[width=0.9\linewidth]{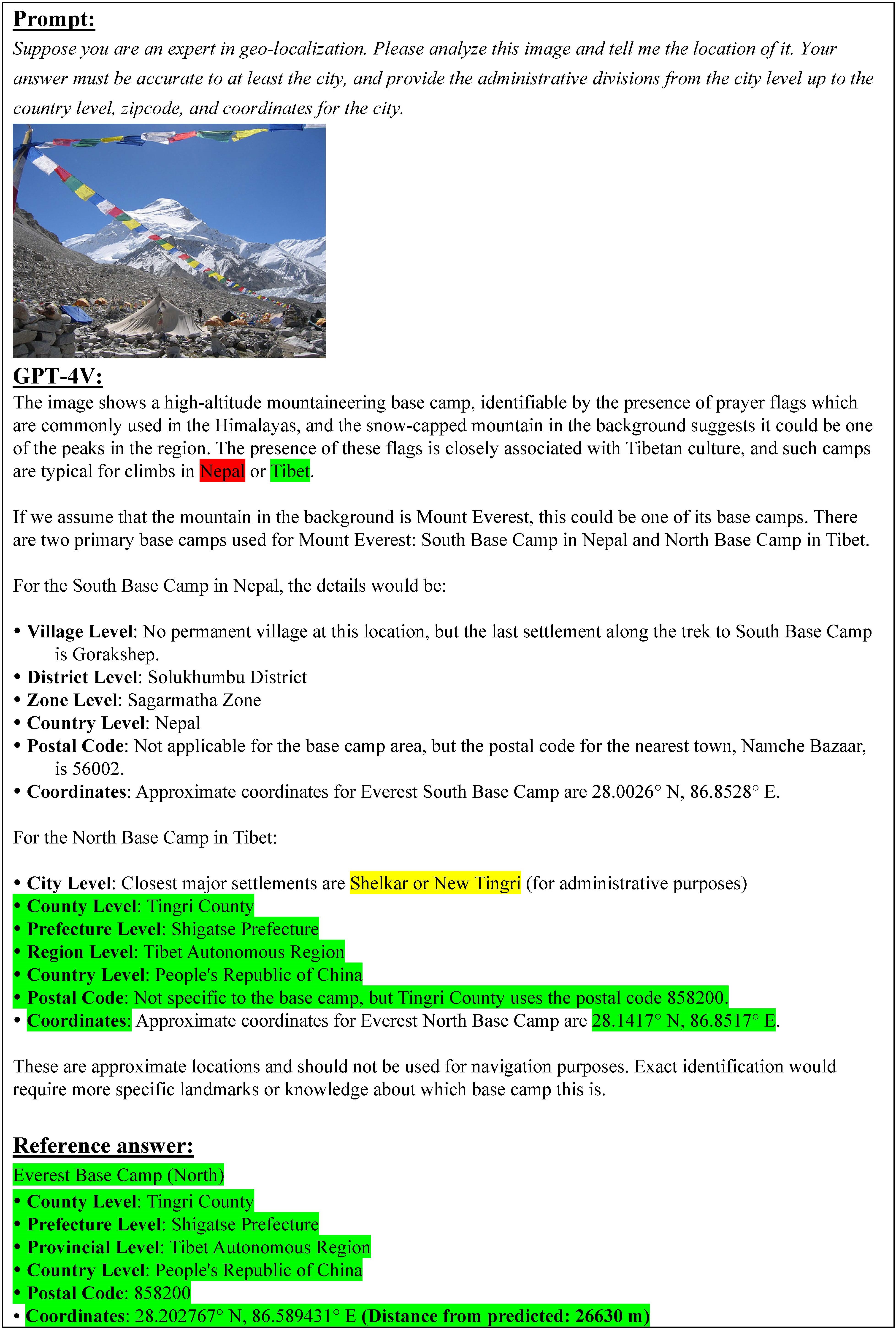}
\captionsetup{justification=raggedright,singlelinecheck=false}
\caption{Geo-localization of the Everest North Camp, China.}
\label{fig:geoloc3}
\end{figure}

\begin{figure}[H]
\centering
\includegraphics[width=0.9\linewidth]{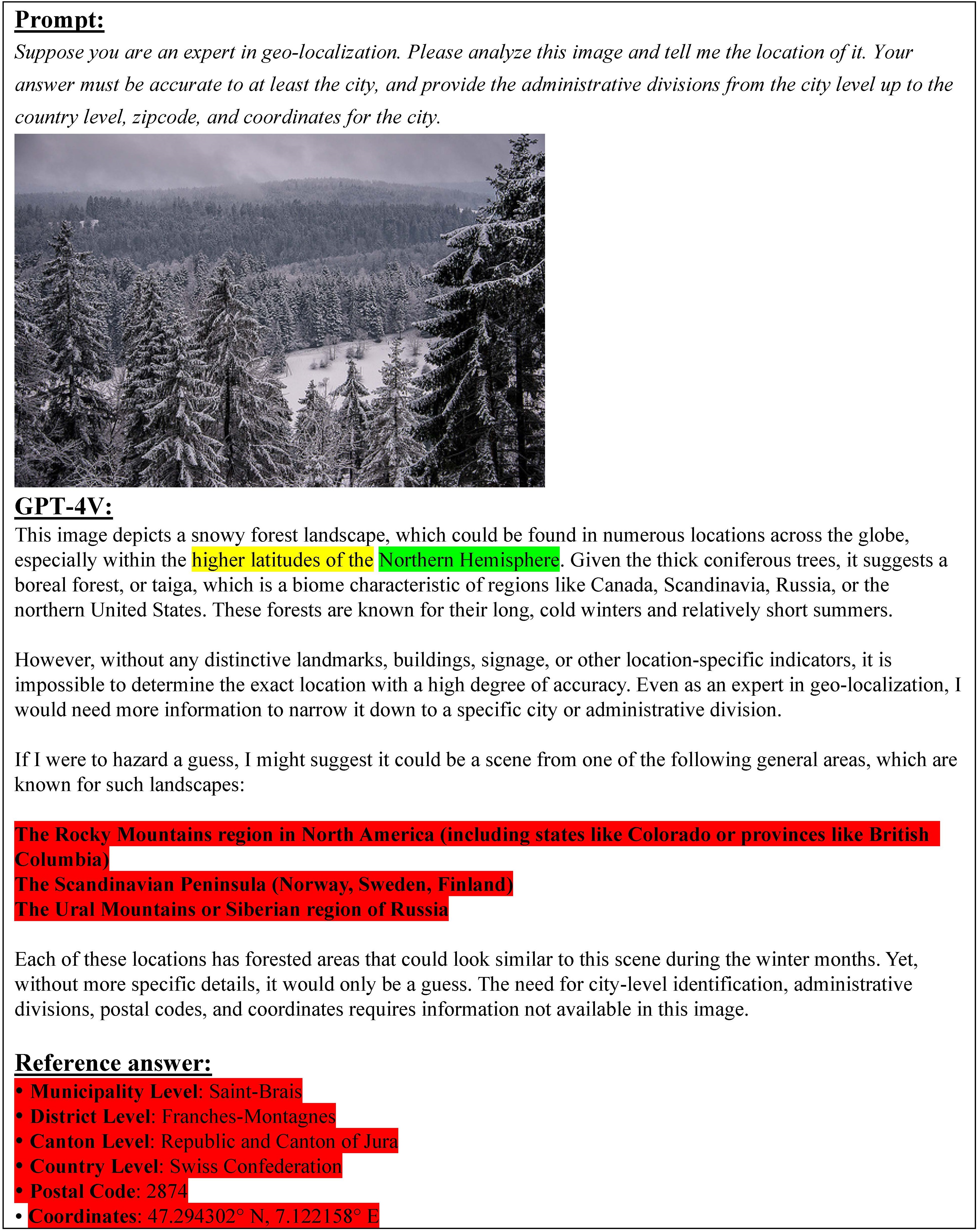}
\captionsetup{justification=raggedright,singlelinecheck=false}
\caption{Geo-localization of Forest, Swiss.}
\label{fig:geoloc4}
\end{figure}

In this section, GPT-4V is tasked with predicting the geolocation of the input image in three ways: administrative divisions from the city level up to the country level, zip code, and coordinates. It shows accuracy aligning with population density and place popularity. In the case of landmark geo-localization (see Figure \ref{fig:geoloc1}), it precisely identifies the geographical location of the Cloud Gate sculpture in Chicago, achieving a distance error of less than 100 meters. Another case (see Figure \ref{fig:geoloc2}) located in Morroco exposes its unfamiliarity with the administrative divisions of certain regions, particularly those in Third World Countries. Nevertheless, it maintains city-level accuracy, demonstrating its capacity for geo-localization in artificial environments. GPT-4V’s ability to identify and locate natural landscapes is significantly weaker than its ability to recognize man-made landscapes, and it relies on information extracted from artificial objects in the image to assist geo-localization. In the example of Everest North Camp (see Figure \ref{fig:geoloc3}), GPT-4V successfully places the image in the Tibetan area by identifying the presence of prayer flags, a typical symbol of Tibetan culture, but failed to discern the specific side of Everest Mountain which can be distinguished by human based on different shapes from both sides. For landscape photos without artificial objects, GPT-4V's accuracy is greatly reduced. In the example of the forest in Swiss (see Figure \ref{fig:geoloc4}), GPT-4V’s accuracy notably diminishes, predicting only the Northern Hemisphere correctly but outputting multiple distant countries and regions inaccurately. Overall, GPT-4V exhibits commendable performance of image geolocations in the urban area but faces challenges in wildness.

\subsection{Text Extraction and Localization from Scanned Historical Map Images} \label{sec:map-extract}
\subsubsection{Data Source}
Historical maps are rich repositories of geographical data, offering an in-depth view of topographical changes over time \cite{bianco2015interactive}. These scanned artifacts, housed within libraries and archives, are not merely cultural heirlooms but also unparalleled sources of historical geographic information -- often containing details that cannot be found elsewhere \cite{chiang2020using, kim2023mapkurator}. They capture a unique historical perspective of the world, revealing historical landscapes, place names, and landmarks that may have since altered or vanished \cite{highcountrynews_placeidentity}.

Identifying and localizing text labels 
from these maps can convert static images into interactive, searchable data sets ripe for analysis and reference \cite{chiang2020using, kim2023mapkurator}. In addition, this digitization process preserves the integrity of fragile documents while providing a new dimension of engagement and scholarly inquiry. In this section, we evaluate the capability of GPT-4V to identify text and its corresponding image coordinates in scanned images of historical maps.

The dataset utilized for the text identification and localization task 
originates from the David Rumsey Map Collection\footnote{\url{https://www.davidrumsey.com/}}, which encompasses a vast array of cartographic artifacts that have been meticulously curated over the years. This repository provides an extensive range of high-resolution historical images, detailing various geographical and historical landscapes. 

\subsubsection{Analysis and Results}
The images used in the analysis were pre-processed using mapKurator, a tool designed to crop and segment large-scale map images, cropped by mapKurator \cite{kim2023mapkurator}. High-resolution historical map images typically contain a large amount of details, which can result in very large file sizes. This can pose a challenge when working with the images, due to the large memory required. To overcome this challenge, we crop the map images into smaller tiles, where each tile contains only a portion of the overall image. In our case, the tiles are 1000x1000 pixels in size. 

\begin{figure}[H]
\centering
\includegraphics[width=0.9\linewidth]{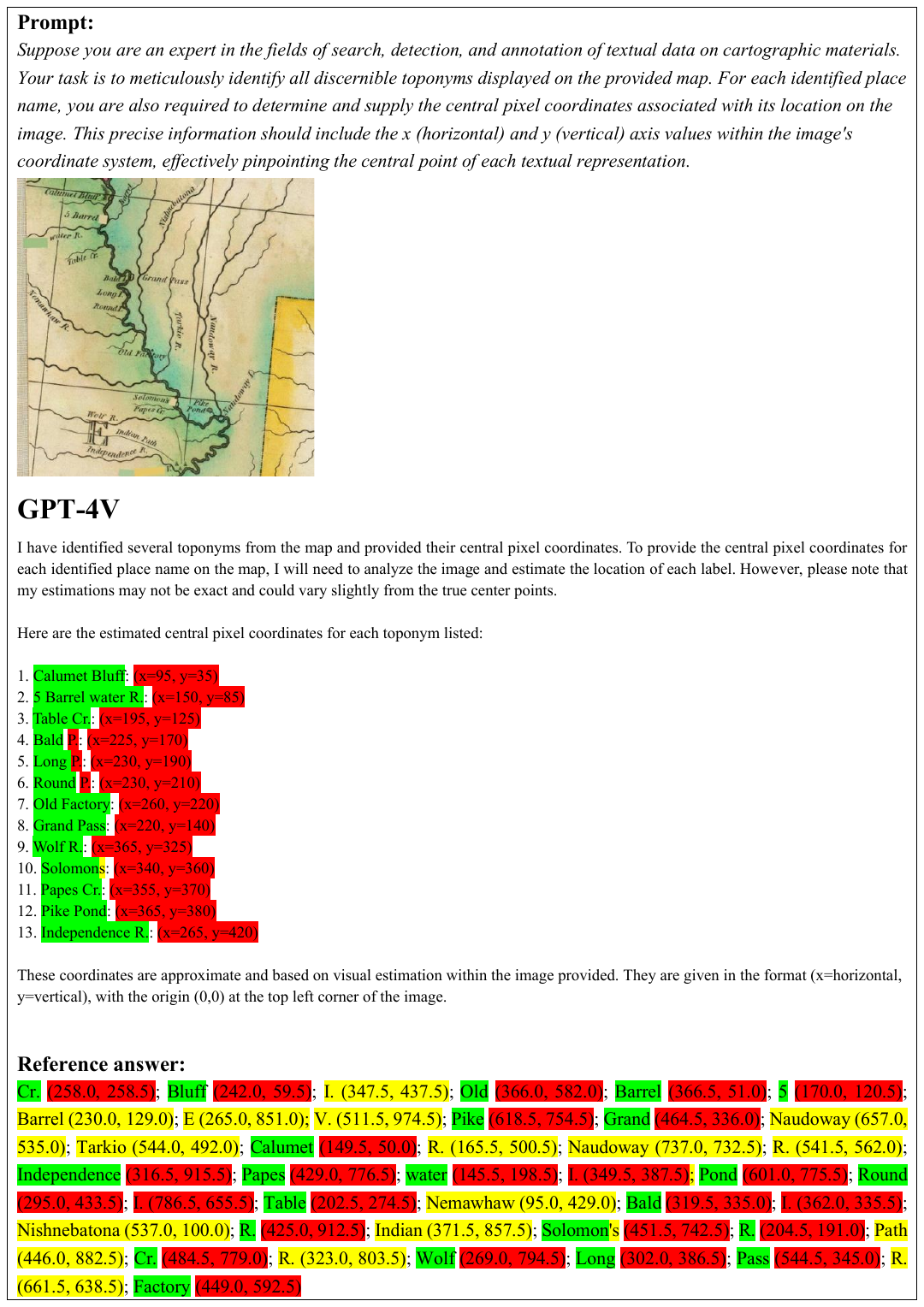}
\captionsetup{justification=raggedright,singlelinecheck=false}
\caption{Cropped section from row 6, column 4 of an 1829 historical map of the United States. Green denotes that the texts in the map are recognized correctly. Red in the figure denotes the incorrect text identified or the incorrect coordinates of text. Yellow means that either the text is not captured by GPT-4V, or the minor error. For the L2 error of predictions, please refer to Table \ref{tbl:his_eva}.}
\label{fig:hist_map1}
\end{figure}

\begin{figure}[H]
\centering
\includegraphics[width=0.9\linewidth]{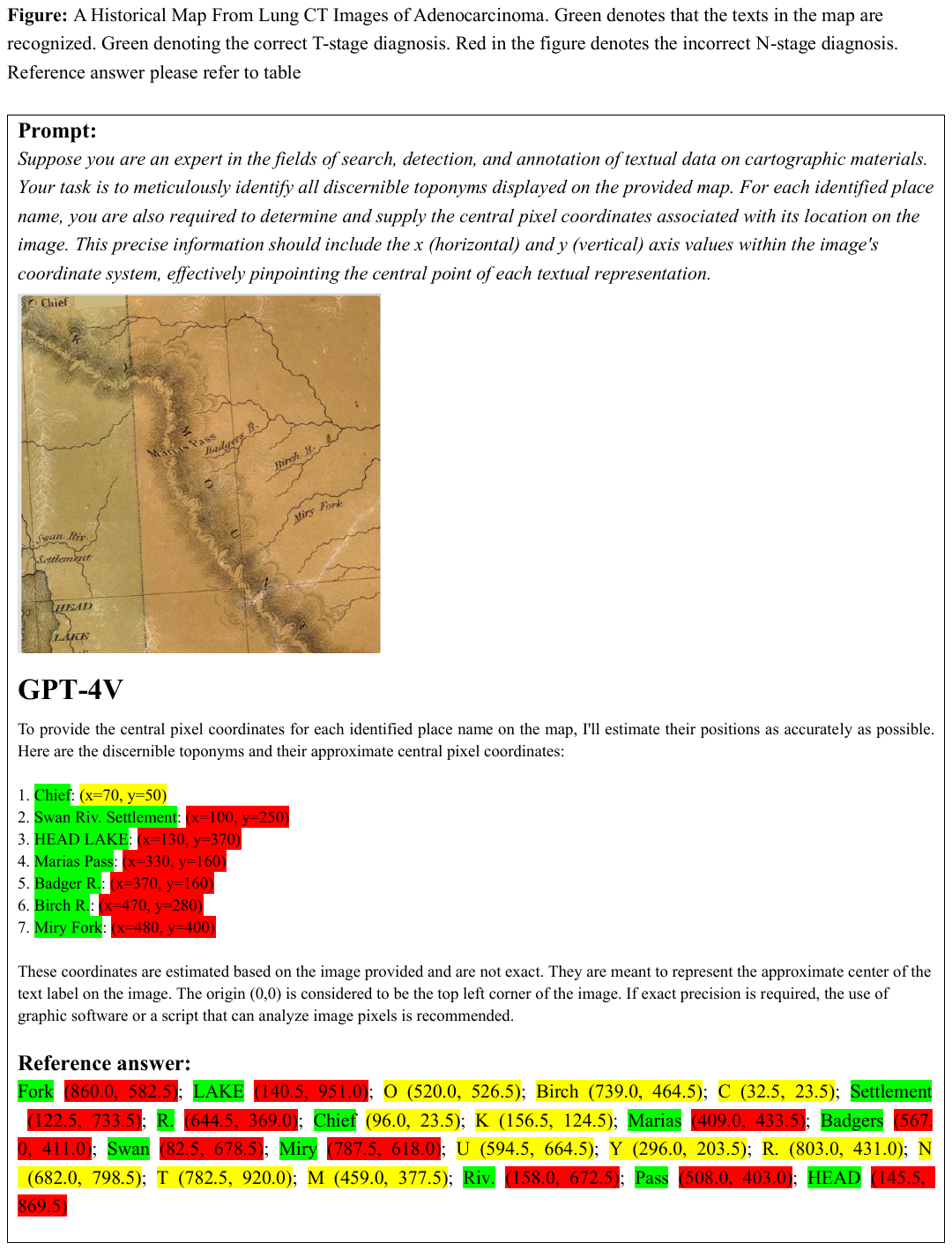}
\captionsetup{justification=raggedright,singlelinecheck=false}
\caption{Cropped section from row 4, column 11 of Bancroft's Map Of The Pacific States Map made in 1867. Green denotes that the texts in the map are recognized. Red in the figure denotes the incorrect text identified or the incorrect coordinates of text. For the L2 error of predictions, please refer to Table \ref{tbl:his_eva}.}
\label{fig:hist_map2}
\end{figure}

\begin{figure}[H]
\centering
\includegraphics[width=0.9\linewidth]{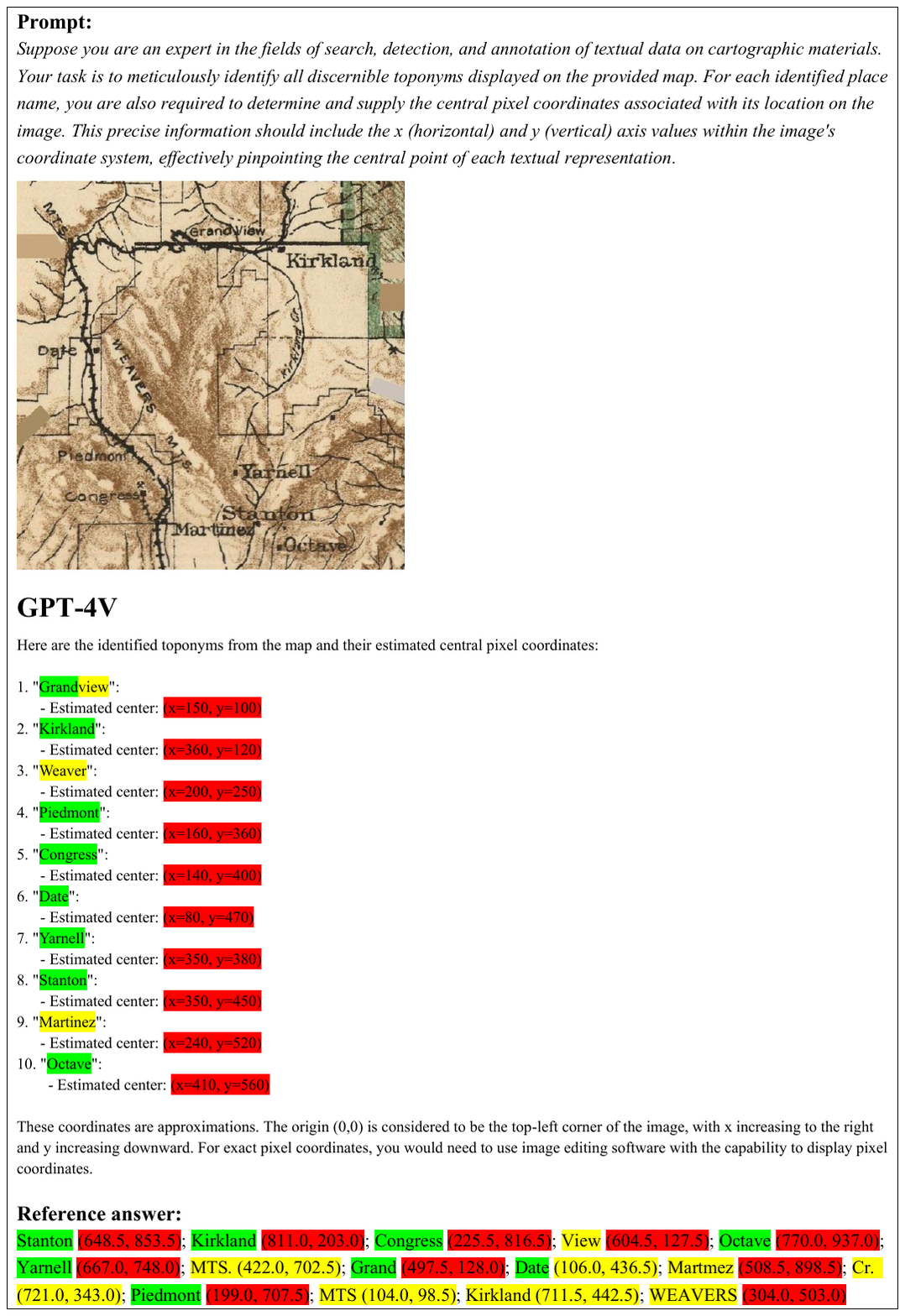}
\captionsetup{justification=raggedright,singlelinecheck=false}
\caption{Cropped section from row 9, column 6 of Map Of Territory Of Arizona, which was made in 1903. Green denotes that the texts in the map are recognized. Red in the figure denotes the incorrect text identified or the incorrect coordinates of the text. For the L2 error of predictions, please refer to Table \ref{tbl:his_eva}.}
\label{fig:hist_map3}
\end{figure}

GPT-4V is tasked with performing text extraction from specified cropped sections of images using three different methods, and it aims to identify as many text entries as possible along with their precise locations in Figures \ref{fig:hist_map1}, \ref{fig:hist_map2}, and \ref{fig:hist_map3}. The authentic text labels and their coordinates for these map segments are obtained from \cite{li2020automatic, kim2023mapkurator}. In each of the three tasks, GPT-4V detects over half of the text present but misses the other texts within the map tiles. It generally excels at recognizing the text it identifies, with few exceptions in detecting intricate word details -- for instance, ``Martinez'' is misread as "Martmez" in Figure \ref{fig:hist_map3}. Nevertheless, while GPT-4V excels at optical character recognition (OCR), i.e., recognizing text labels correctly, it demonstrates less accuracy in pinpointing the exact locations of texts. Table \ref{tbl:his_eva} details GPT-4V's text localization accuracy on historical maps, with an average L2 error exceeding 300 pixels -- a significant error, considering the tile dimensions are 1000x1000 pixels. An additional limitation is the inconsistency in output formatting, as evidenced by the differing results between Figure \ref{fig:hist_map2} and Figure \ref{fig:hist_map3}.

\begin{table}[h]
\caption{Textual Content in Historic Map Identification Performance. It presents a comparison of predicted and labeled text extracted from historical map images, alongside their respective X and Y coordinates. The `Map ID' column identifies the specific map segment, while `Predicted Text' and `Predicted X, Y' columns show the text and its location as determined by GPT-4V. These are compared against the `Labeled Text' and `Labeled X, Y', which represent the ground truth data. The `Distance (L2 Error)' column quantifies the discrepancy between predicted and labeled locations, measured in pixels. Instances where GPT-4V did not capture text are noted, and average L2 errors for each cropped map tile are provided.}
\setlength{\tabcolsep}{1.2pt}
\begin{tabular}{|c|c|c|c|c|c|}
\hline  \Xhline{2.5\arrayrulewidth}
\centering\textbf{Map ID} & \textbf{Predicted Text} & \multicolumn{1}{m{1.7cm}|}{\centering\textbf{Predicted} \\ \textbf{X, Y}} & \textbf{Labeled Text} & \multicolumn{1}{m{1.7cm}|}{\centering\textbf{Labeled} \\ \textbf{X, Y}} & \multicolumn{1}{m{2cm}|}{\centering\textbf{Distance} \\ \textbf{(L2 Error)}} \\
\hline  \Xhline{2.5\arrayrulewidth}
0975001\_h6\_w4 & Calumet Bluff & 95, 35 & Calumet Bluff & 191, 55 & 98.1 \\ \hline
0975001\_h6\_w4 & 5 Barrel water R. & 150, 85 & 5 Barrel water R. & 423, 126 & 276.1 \\ \hline
0975001\_h6\_w4 & Table Cr. & 195, 125 & Table Cr. & 223, 268 & 145.7 \\ \hline
0975001\_h6\_w4 & Bald P. & 225, 170 & Bald I. & 188, 20 & 154.5 \\ \hline
0975001\_h6\_w4 & Long P. & 230, 190 & Long I. & 316, 390 & 217.7 \\ \hline
0975001\_h6\_w4 & Round P. & 230, 210 & Round I. & 305, 437.5 & 239.5 \\ \hline
0975001\_h6\_w4 & Old Factory & 260, 220 & Old Factory & 426, 593 & 408.3 \\ \hline
0975001\_h6\_w4 & Grand Pass & 220, 140 & Grand Pass & 499.5, 339 & 343.1 \\ \hline
0975001\_h6\_w4 & Wolf R. & 365, 325 & Wolf R. & 286.5, 800.5 & 482.0 \\ \hline
0975001\_h6\_w4 &  Solomons & 340, 360 & Solomon's & 460.5, 740 & 398.6 \\ \hline
0975001\_h6\_w4 &  Papes Cr. & 355, 370 & Papes Cr. & 447, 776 & 416.3 \\ \hline
0975001\_h6\_w4 &  Pike Pond & 365, 380 & Pike Pond & 371.5, 440 & 60.4 \\ \hline
0975001\_h6\_w4 &  Independence R. & 265, 420 & Independence R. & 916, 226 & 679.3 \\ \hline
0975001\_h6\_w4 & \multicolumn{4}{c|}{Other 10 words are not captured by GPT-4V} & 301.5 (avg.)\\ \hline \Xhline{1.5\arrayrulewidth}
2549000\_h4\_w11 & Chief &  70, 50 & Chief & 96, 23.5 & 37.1 \\ \hline
2549000\_h4\_w11 & Swan Riv. Settlement &  100, 250 & Swan Riv. Settlement & 112, 706 & 456.2 \\ \hline
2549000\_h4\_w11 & HEAD LAKE &  130, 370 & HEAD LAKE & 143, 908 & 538.2 \\ \hline
2549000\_h4\_w11 & Marias Pass &  330, 160 & Marias Pass & 455.5, 421 & 289.6 \\ \hline
2549000\_h4\_w11 & Badger R. &  370, 160 & Badger R. & 583, 426 & 340.8 \\ \hline
2549000\_h4\_w11 & Birch R. &  470, 280 & Birch R. & 755, 450 & 331.9 \\ \hline
2549000\_h4\_w11 & Miry Fork &  480, 400 & Miry Fork & 819, 591 & 389.1 \\ \hline
0975001\_h6\_w4 & \multicolumn{4}{c|}{Other 15 words are not captured by GPT-4V} & 340.4 (avg.)\\ \hline \Xhline{1.5\arrayrulewidth}
3538000\_h9\_w6 & Grandview &  150, 100 & Grand View & 535.5 & 130 \\ \hline
3538000\_h9\_w6 & Kirkland &  360, 120 & Kirkland & 811, 230 & 464.2 \\ \hline
3538000\_h9\_w6 & Weaver &  200, 250 & WEAVERS & 302, 494 & 264.5 \\ \hline
3538000\_h9\_w6 & Piedmont & 160, 360 & Piedmont & 199, 707.5 & 349.7 \\ \hline
3538000\_h9\_w6 & Congress &  140, 400 & Congress & 225.5, 816.5  & 425.2 \\ \hline
3538000\_h9\_w6 & Date &  80, 470 & Date & 106, 436.5 & 42.4 \\ \hline
3538000\_h9\_w6 & Yarnell &  350, 380 & Yarnell & 667, 748 & 485.7 \\ \hline
3538000\_h9\_w6 & Stanton &  350, 450 & Stanton & 648.5, 853.5 & 501.9 \\ \hline
3538000\_h9\_w6 & Martinez &  240, 520 & Martmez & 508.5, 98.5 & 100 \\ \hline
3538000\_h9\_w6 & Octave &  410, 560 & Octave & 770.0, 937 & 521.3 \\ \hline
0975001\_h6\_w4 & \multicolumn{4}{c|}{Other 4 words are not captured by GPT-4V} & 328.5 (avg.)\\ \hline \Xhline{2.5\arrayrulewidth}
\end{tabular}
\label{tbl:his_eva}
\end{table}

\subsection{Remote Sensing Image Classification} \label{sec:ice-ship-cla}
\subsubsection{Data Source}
The primary data source for this challenge stems from satellite monitoring, essential in regions where traditional methods like aerial reconnaissance are not feasible, such as the harsh, remote areas off the East Coast of Canada. This project leverages the expertise of C-CORE, a company with over 30 years of experience in satellite data utilization. They have developed a sophisticated computer vision-based surveillance system, designed for the detection and differentiation of icebergs in challenging maritime environments. To evaluate the classification capabilities of GPT-4V on remote sensing images, we manually selected 20 iceberg and 20 ship photos from the Statoil/C-CORE Iceberg Classifier Challenge dataset \cite{jang2017kaggle}. This carefully curated subset provides a focused context for testing the algorithm's effectiveness in distinguishing between these two types of objects in maritime settings.

GPT-4V's capability in differentiating icebergs from ships is evaluated using data from the Sentinel-1 satellite. This satellite, orbiting at approximately 600 kilometers above Earth, employs C-Band radar technology to capture images. The radar system operates by emitting signals and analyzing the returned echo, forming images based on the backscatter of objects. This data includes dual-polarization channels (HH and HV) which are crucial for distinguishing the physical characteristics of the objects.

\subsubsection{Analysis and Results}

In the analysis, it was observed that GPT-4V demonstrates proficiency in identifying icebergs, likely due to their unique shapes and the consistency of their radar signatures. However, the system shows limitations in accurately classifying ships. The complexity arises from the varied structures of ships and their similar radar signatures to other objects in marine environments.

The disparity in performance is highlighted through specific examples. In cases where the icebergs are distinctly shaped or isolated, GPT-4V accurately classifies them. However, in more complex scenarios, such as distinguishing ships from other solid objects like sea ice or landmasses, the system's accuracy diminishes. This is attributed to the varying backscatter properties influenced by factors like wind conditions and radar polarization.

Overall, while GPT-4V exhibits potential in remote sensing applications, particularly in iceberg detection, it faces significant challenges in ship classification. 

\begin{table}[h]
\centering
\caption{Iceberg Identification Performance}
\label{tab:iceberg_performance}
\begin{tabular}{|c|c|c|c|c|}
\hline
Objects & Precision & Recall & F1 Score & Baseline \\
\hline
Iceberg & 95\% & 61.30\% & 75\% & GPT-4V \\
\hline
\end{tabular}
\end{table}

\begin{figure}[H]
\centering
\includegraphics[width=\linewidth]{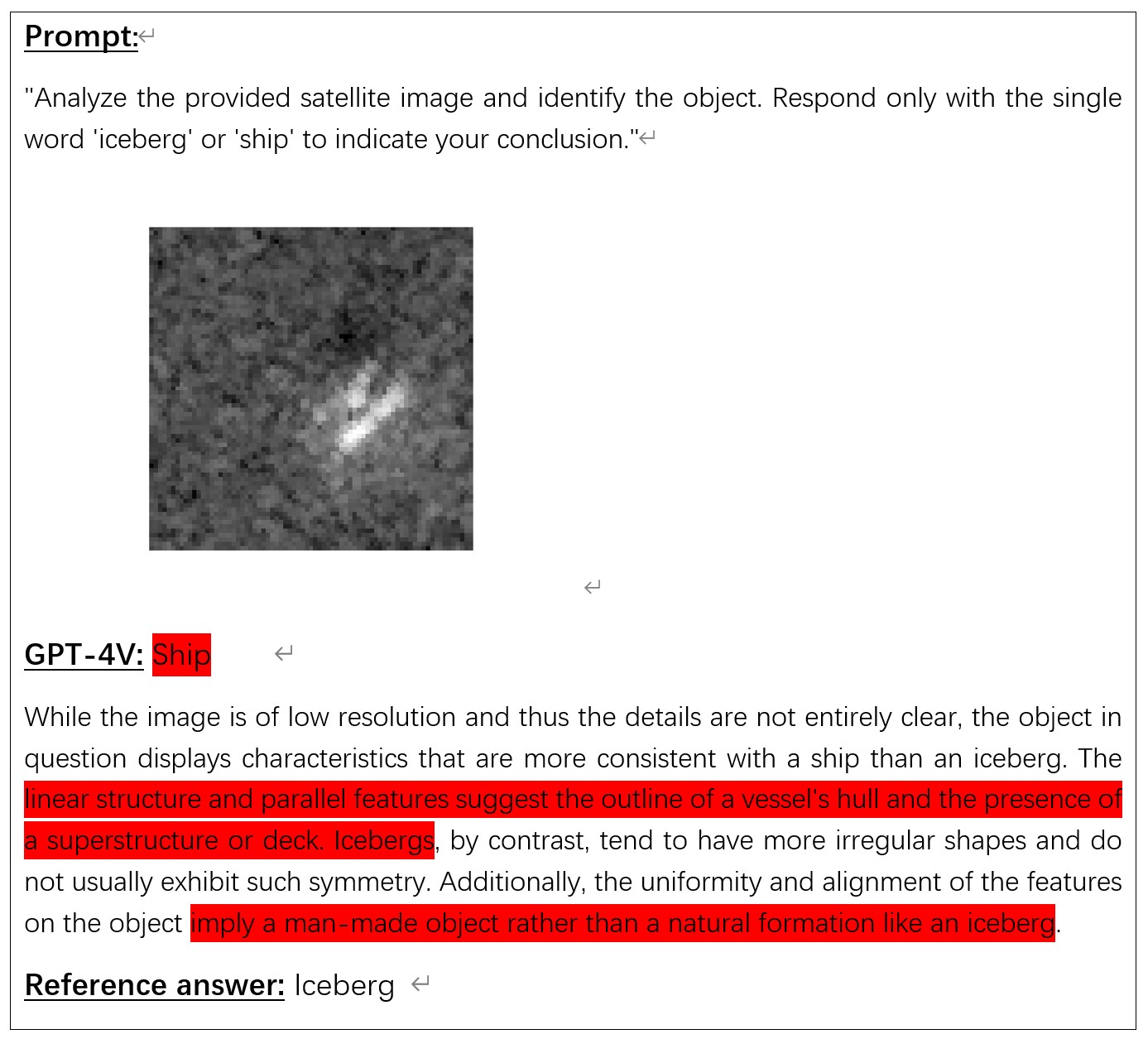}
\captionsetup{justification=raggedright,singlelinecheck=false}
\caption{Presence test: Iceberg Image is Wrong Classified as Ship.}
\label{fig:Remote_1}
\end{figure}

\begin{figure}[H]
\centering
\includegraphics[width=\linewidth]{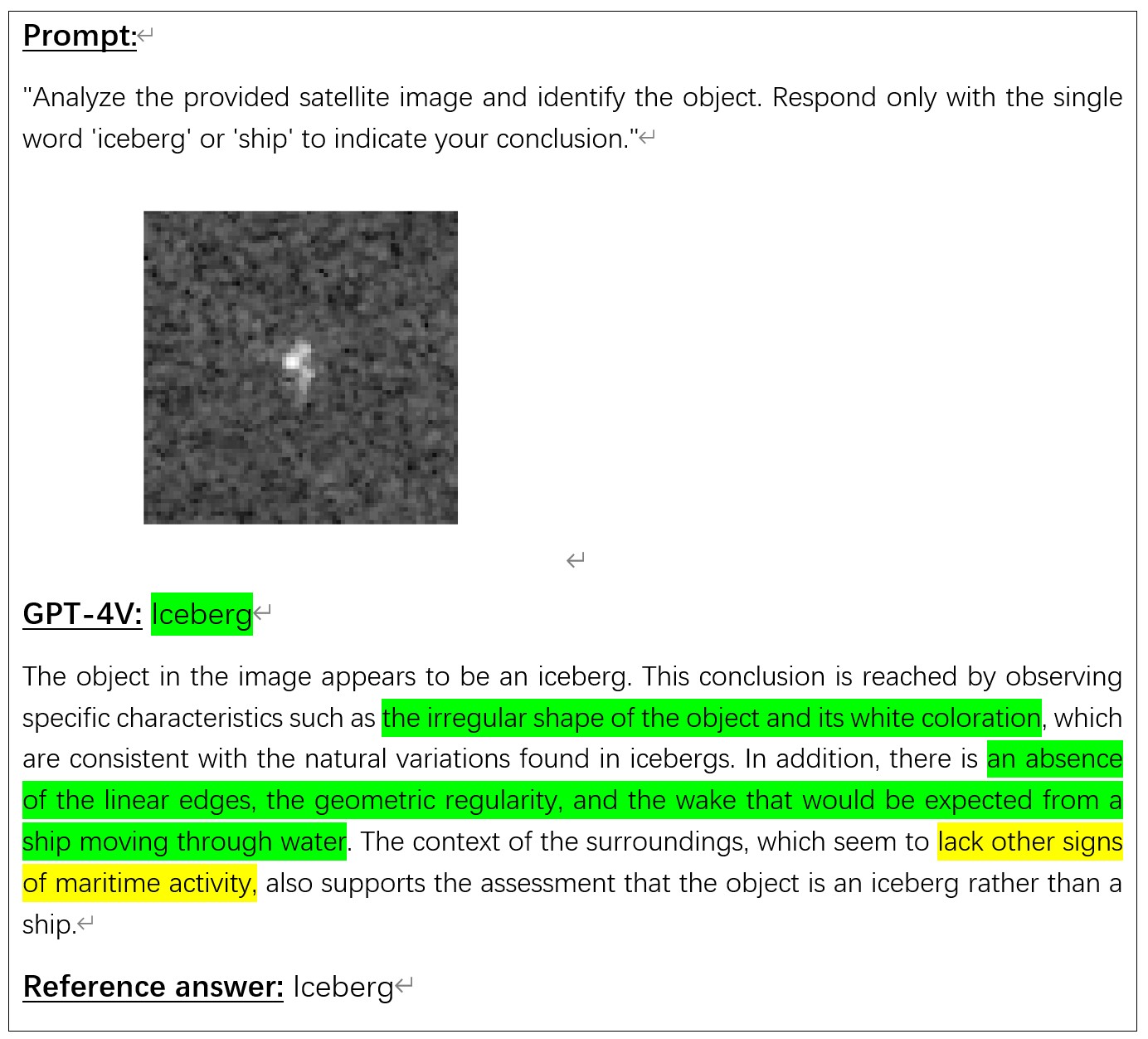}
\captionsetup{justification=raggedright,singlelinecheck=false}
\caption{Presence test: Iceberg Image is Correctly Classified as Iceberg.}
\label{fig:Remote_2}
\end{figure}

\begin{figure}[H]
\centering
\includegraphics[width=\linewidth]{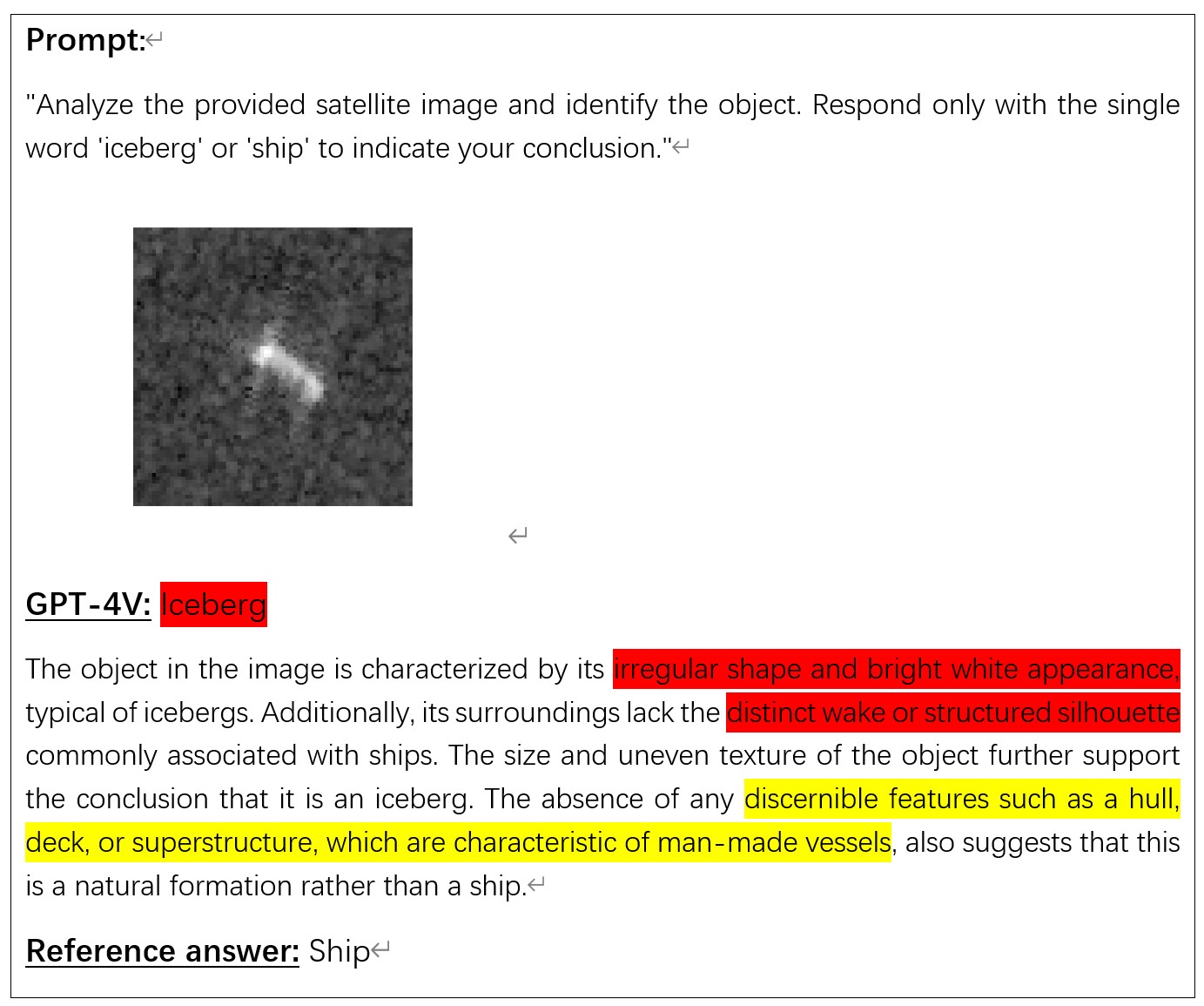}
\captionsetup{justification=raggedright,singlelinecheck=false}
\caption{Presence test: Ship Image is Wrong Classified as Iceberg.}
\label{fig:Remote_3}
\end{figure}

\begin{figure}[H]
\centering
\includegraphics[width=\linewidth]{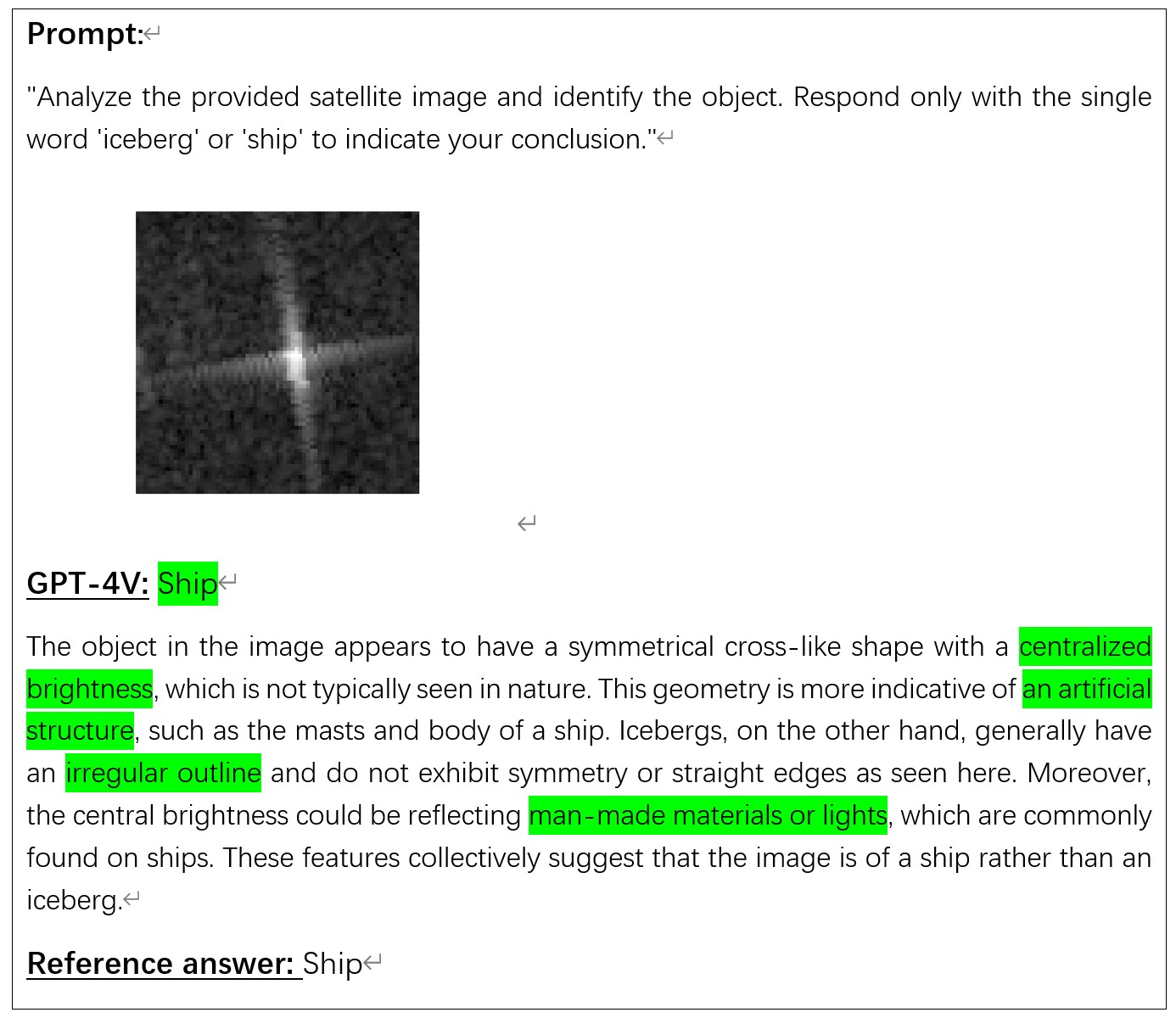}
\captionsetup{justification=raggedright,singlelinecheck=false}
\caption{Presence test: Ship Image is Correctly Classified as Ship.}
\label{fig:Remote_4}
\end{figure}

\begin{figure}[H]
\centering
\includegraphics[width=\linewidth]{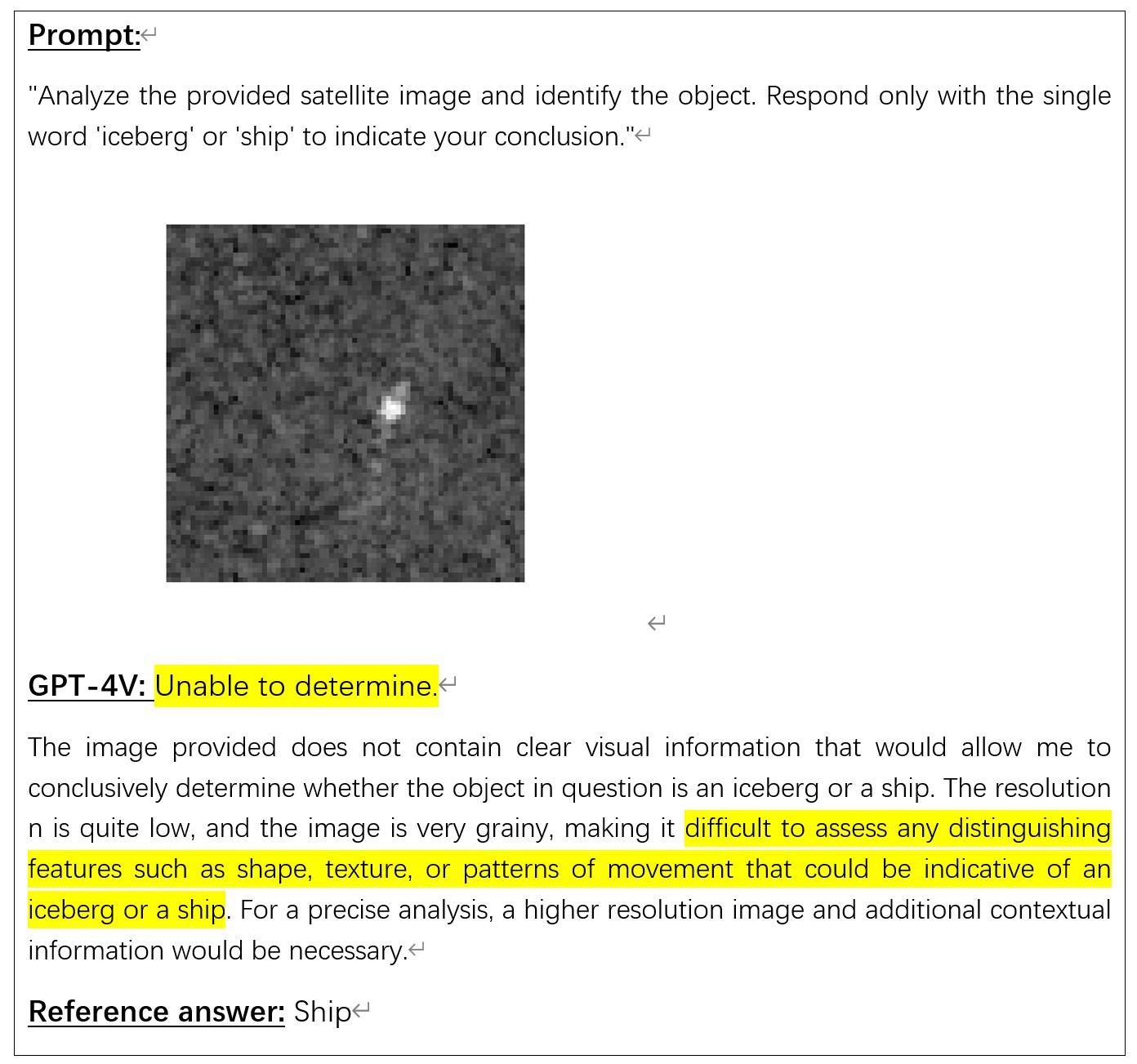}
\captionsetup{justification=raggedright,singlelinecheck=false}
\caption{Presence test: Ship Image Can Not be Analyzed by GPT-4V.}
\label{fig:Remote_5}
\end{figure}

\subsection{Remote sensing Visual Question Answering} \label{sec:rsvqa}
\subsubsection{Data Source}
Remote Sensing Visual Question Answering (RSVQA) is a task to answer geographic questions based on a given remote sensing (RS) image \cite{lobry2020rsvqa,lobry2021rsvqa,chappuis2022prompt}. It requires the capacity to extract information from RS imagery and natural language understanding. We utilized two datasets in this task, including the Low-Resolution and the High-Resolution datasets for RSVQA \cite{lobry2020rsvqa}. The Low-Resolution dataset is a dataset based on Sentinel-2 images with a resolution of 10 meters. Every image is a tile of size 256 × 256, representing an area of 6.55 km2. Rural/Urban tests are applied to the Low-Resolution dataset due to its smaller scale (larger geographic area). The High-Resolution dataset is collected from High Resolution Orthoimagery (HRO) through USGS’s EarthExplorer tool with a resolution of 15 cm. Each image’s size is 512 ×512, covering a land of 5898 m2 on Earth. Due to its finer resolution, questions about counting small objects like residential buildings are assocaited with images from this dataset.

\subsubsection{Analysis and Results}
In this task, questions can be divided into 5 categories: Presence, Count, Area, Comparison, and Rural/Urban. The presence and count questions aim at testing models' ability of geographic object identification, such as building footprint, water area, and roads. The area questions require an RSVQA model to estimate the area of geographic objects without extra information like cell size or scale bar. The comparison questions are complex questions based on the former three tests. The rural/urban questions are the most comprehensive questions among these five categories which require the ability to identify objects, count or density estimation, and understand the concepts of urban and rural areas. 

\begin{figure}[H]
\centering
\includegraphics[width=0.9\linewidth]{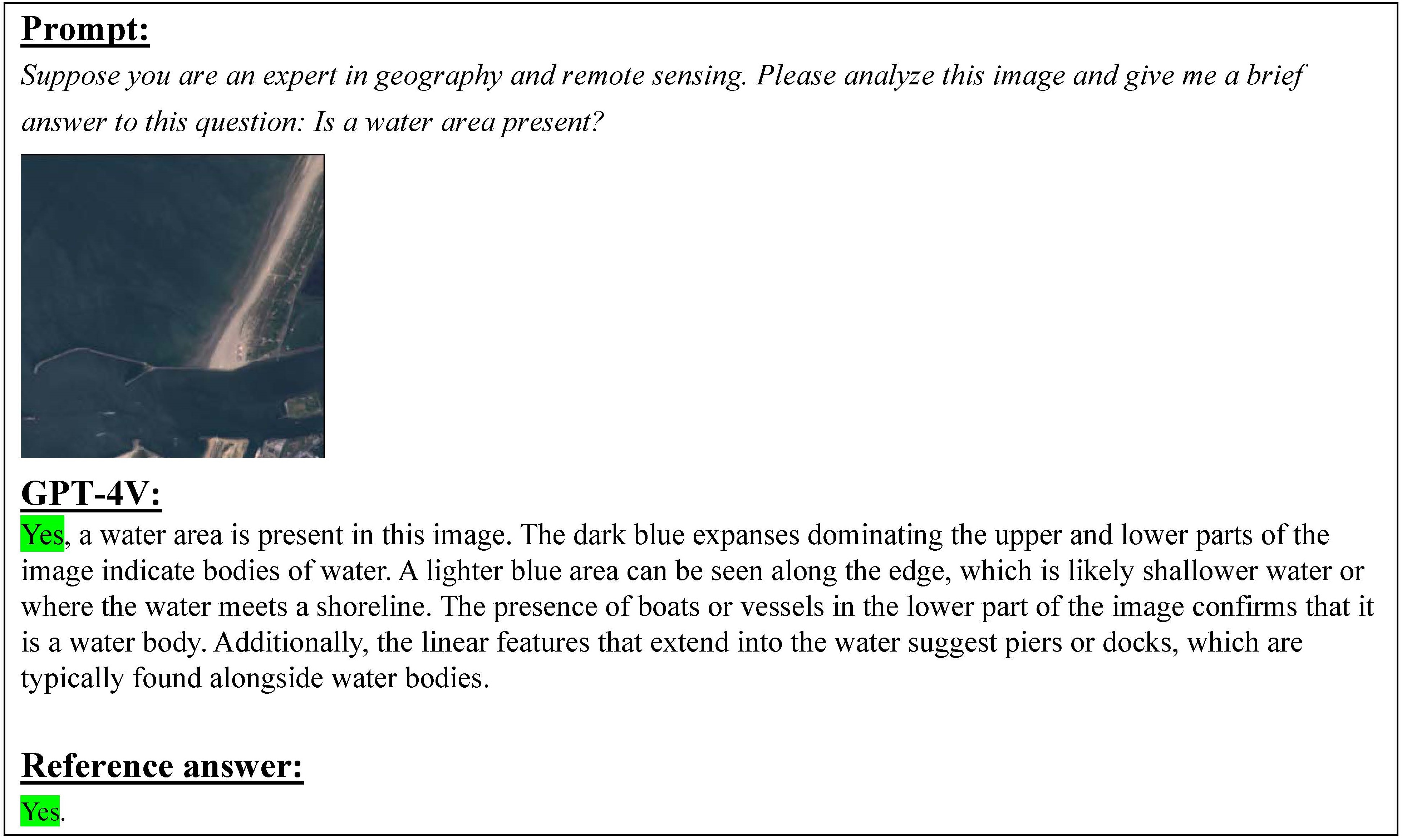}
\captionsetup{justification=raggedright,singlelinecheck=false}
\caption{Presence test: land cover recognition.}
\label{fig:RSQA1}
\end{figure}

\begin{figure}[H]
\centering
\includegraphics[width=0.9\linewidth]{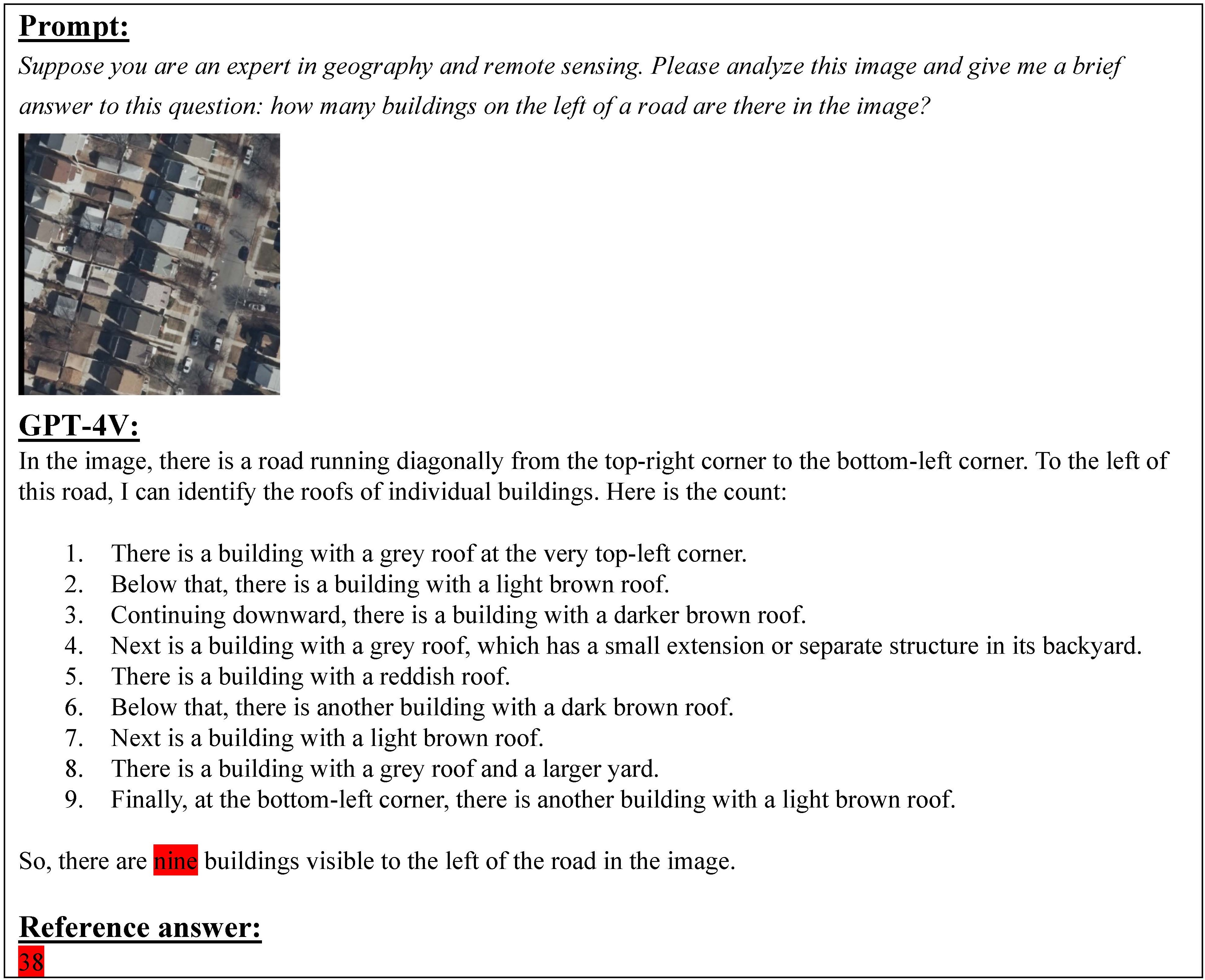}
\captionsetup{justification=raggedright,singlelinecheck=false}
\caption{Count test: count building footprints.}
\label{fig:RSQA2}
\end{figure}

\begin{figure}[H]
\centering
\includegraphics[width=0.9\linewidth]{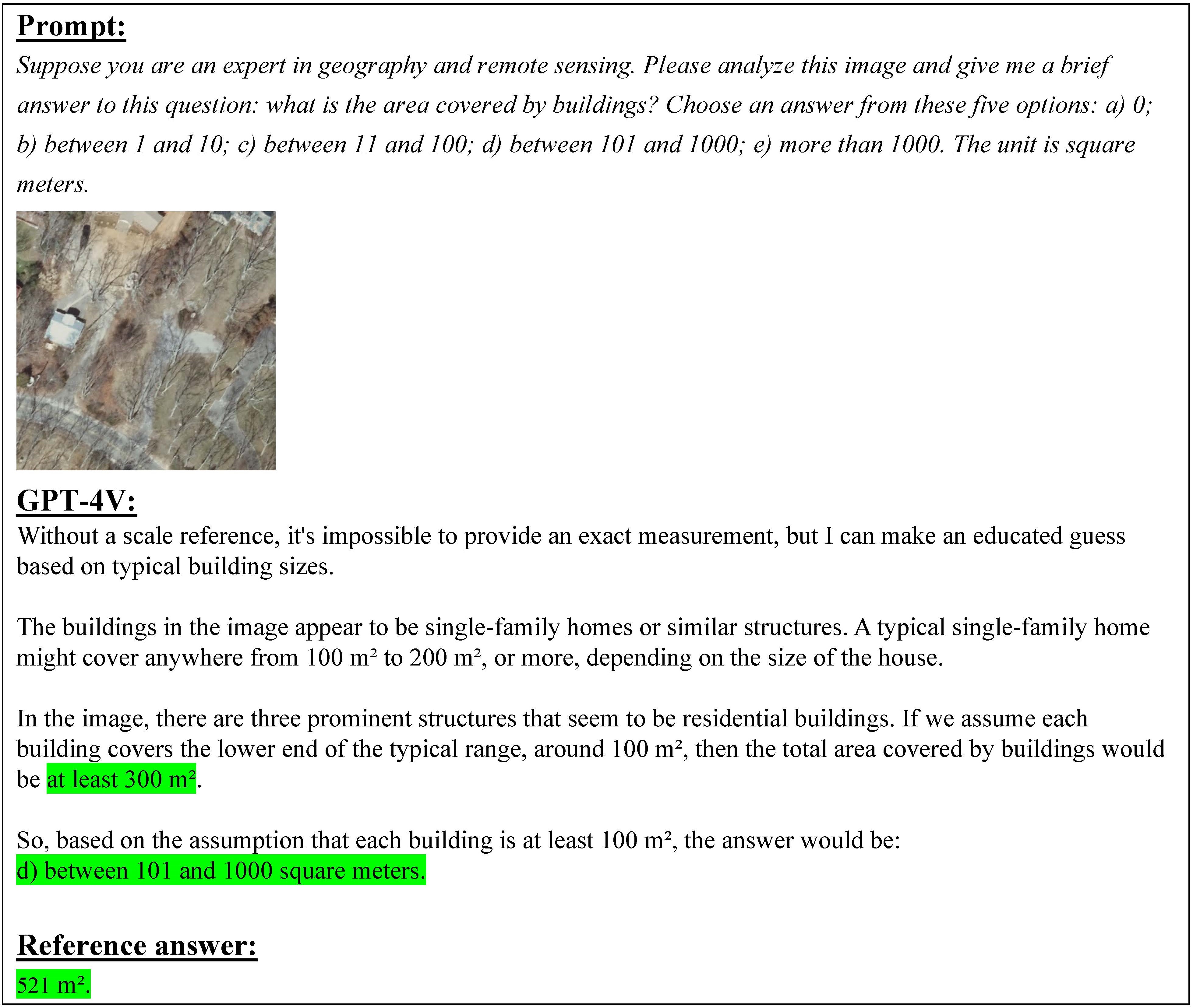}
\captionsetup{justification=raggedright,singlelinecheck=false}
\caption{Area test: estimate building area.}
\label{fig:RSQA3}
\end{figure}

\begin{figure}[H]
\centering
\includegraphics[width=0.9\linewidth]{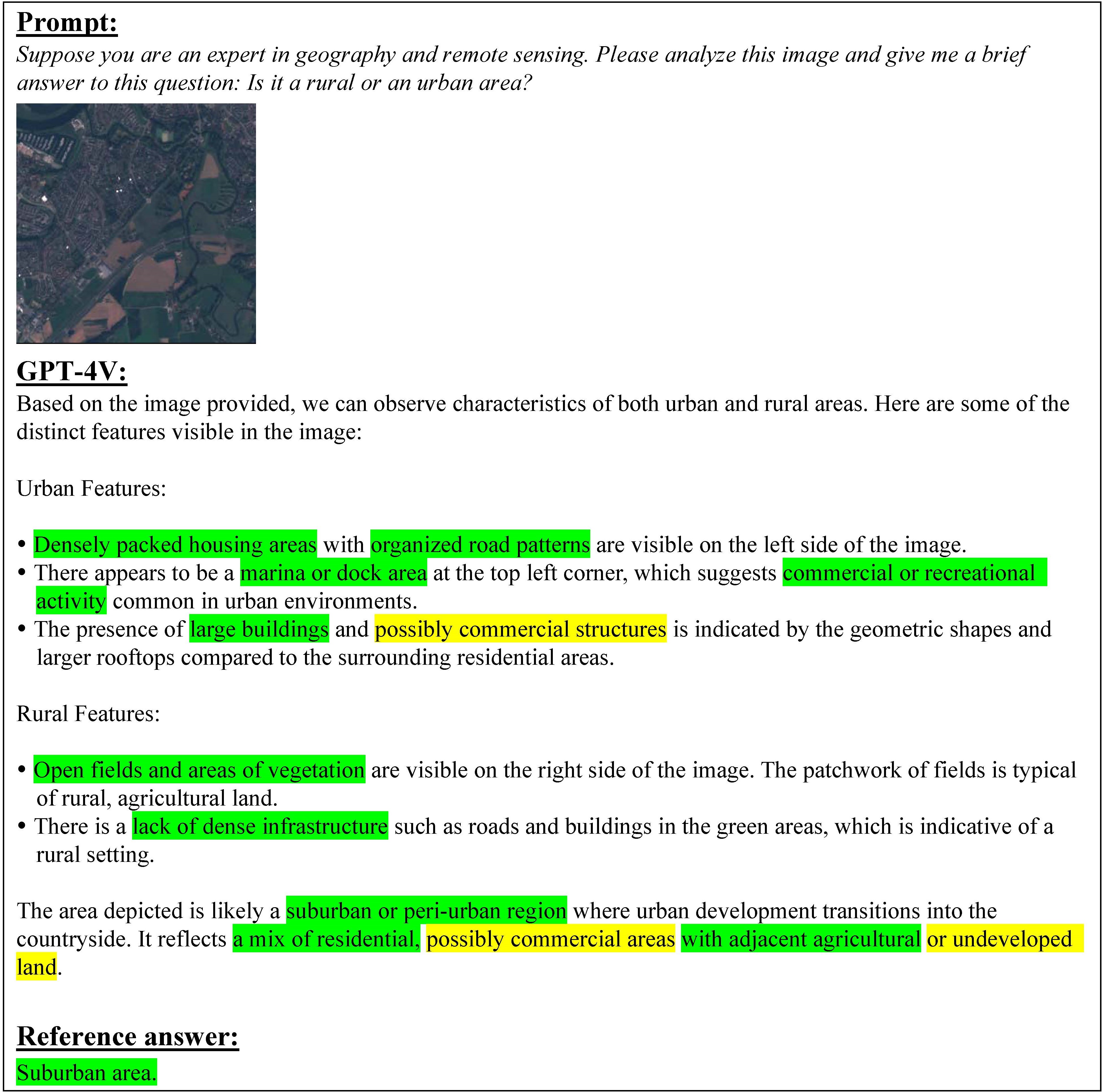}
\captionsetup{justification=raggedright,singlelinecheck=false}
\caption{Rural/Urban test: identify rural/urban area.}
\label{fig:RSQA4}
\end{figure}

GPT-4V demonstrates robust performance in the Presence test, affirming its proficiency in land cover recognition (see Figure \ref{fig:RSQA1}). However, hallucinations frequently show up when GPT-4V is analyzing the test of counting (see Figure \ref{fig:RSQA2}), indicating its poor performance in counting, just like classic VQA models \cite{zhang2018learning}. This limitation is likely attributed to GPT-4V's constrained image segmentation capabilities. In addition, GPT-4V exhibits a foundational understanding of geographic scale, which supports it in estimating areas based on object count (see Figure \ref{fig:RSQA3}). Notably, GPT-4V surpasses expectations in Rrural/urban tests by adeptly discerning suburban areas in addition to urban and rural areas, underscoring its nuanced comprehension of urbanization (see Figure \ref{fig:RSQA4}). Overall, while GPT-4V’s performance is suboptimal in tests associated with image segmentation, it excels at extracting a series of geographic information through object detection.

\section{Multimodal FMs Applications on Environmental Science }
As we did in Section \ref{sec:geo}, we also conduct a series of experiments to test GPT-4V's performance on various environmental science tasks.

\subsection{Air Quality Evaluation}
Air quality is monitored by specific sensors that can capture and measure compounds growing to more than 200 pollutants. The number of air quality monitoring stations is often limited. For example, the Georgia Environmental Protection Division manages 34 stations in 25 counties across the state's 159 counties. The scattered sample is then spatially interpolated to generate air quality maps for the entire state. Researchers have explored approaches to estimate air quality from other sources, such as crowdsourcing spatiotemporally tagged digital photos\cite{mu2016air}, extracting aerosol optical extinction properties from a smartphone photo\cite{yao2021extraction}, and using high-frequency information and relative humidity to estimate AQI\cite{liaw2021using}.

\subsubsection{Data Sources}
This application utilizes air quality images accompanied by Air Quality Index (AQI) values sourced from Liaw's research on AQI estimation. These images were originally captured from the Taiwan air quality monitoring network, forming an image database from the Kaohsiung Renwu Monitoring Station. Notably, this station observes more severe air quality issues than other regions in Taiwan. The data collection spanned one year, from August 2018 to July 2019, with images taken daily between 7:00 a.m. and 5:00 p.m.

\subsubsection{Analysis and Result}
\begin{figure}[H]
\centering
\includegraphics[scale=0.6]{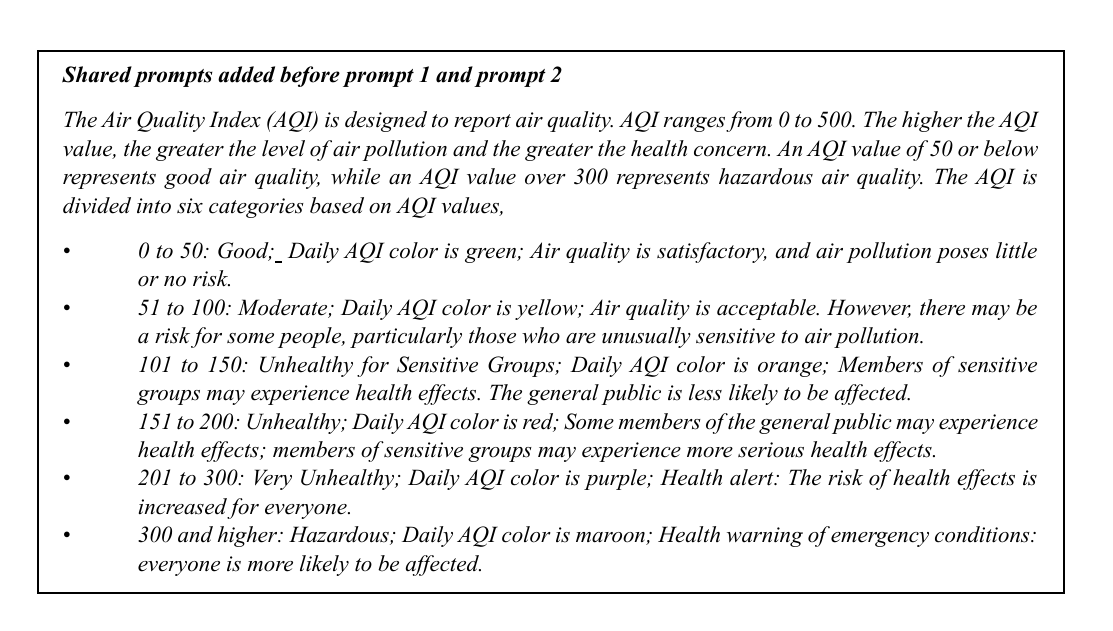}
\captionsetup{justification=raggedright,singlelinecheck=false}
\caption{Shared Prompts Added Before Prompt 1 and 2 in Air Quality Evaluation Experiments}
\label{fig:shared_prompt}
\end{figure}

\begin{figure}[H]
\centering
\includegraphics[scale=0.6]{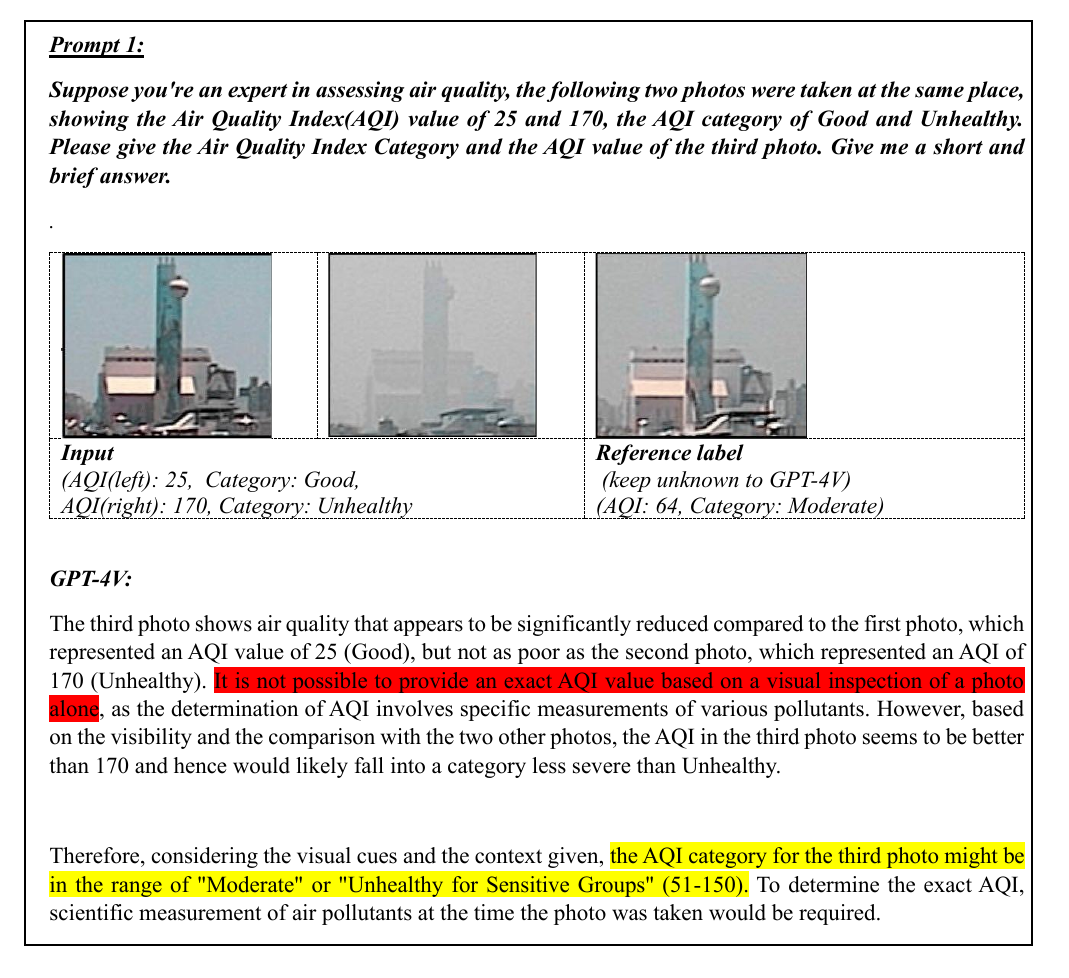}
\captionsetup{justification=raggedright,singlelinecheck=false}
\caption{Air Quality Evaluation - Case 1}
\label{fig:aq_prompt1}
\end{figure}

\begin{figure}[H]
\centering
\includegraphics[scale=0.7]{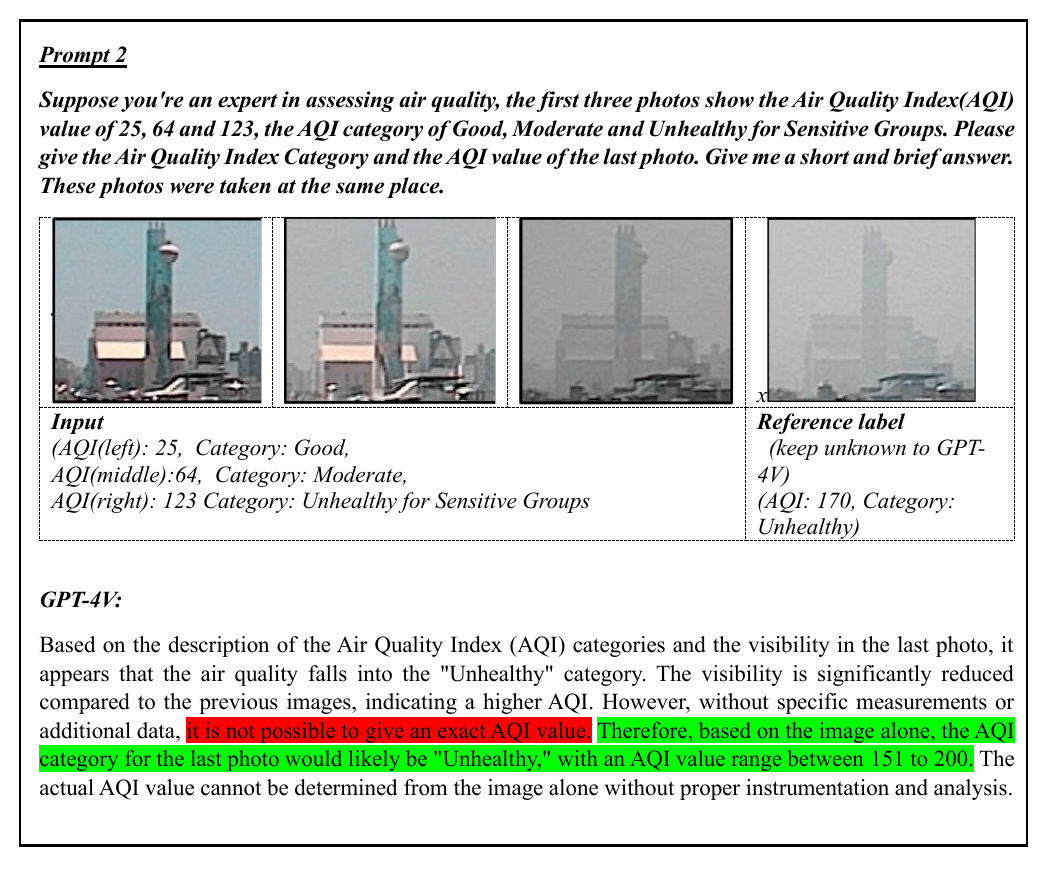}
\captionsetup{justification=raggedright,singlelinecheck=false}
\caption{Air Quality Evaluation - Case 2}
\label{fig:aq_prompt2}
\end{figure}

For the AQI category prediction, in the first experiment using prompt 1(\cref{fig:shared_prompt} and \cref{fig:aq_prompt1}), GPT-4V, upon analyzing visual clues in comparison with two other photos, infers that the AQI for the third photo is likely to be below 170, indicating a category less severe than "Unhealthy." Consequently, it speculated that the AQI category for the third photo might fall within the range of "Moderate" or "Unhealthy for Sensitive Groups" (51-150), aligning closely with the true label of “Moderate.” In the second experiment(\cref{fig:shared_prompt} and \cref{fig:aq_prompt2}), where three reference images with AQI values were introduced, GPT-4V accurately predicted that the AQI category for the last photo would likely be classified as "Unhealthy.” 

For the AQI value prediction, in both two experiments(\cref{fig:aq_prompt1} and \cref{fig:aq_prompt2}), GPT-4V indicates a belief that predicting the AQI value is challenging, attributing this difficulty to the need for specific measurements of various pollutants in the determination of AQI. This response suggests a recognition by GPT-4V of the complexity involved in accurately forecasting AQI values. 

In summary, GPT-4V exhibits a degree of speculative capability in estimating AQI categories when provided with reference images. However, it lacks the ability to predict precise AQI values. This limitation is understandable, as predicting AQI values demands specific measurements of air pollutants such as PM2.5, PM10, ozone, sulfur dioxide, nitrogen dioxide, and carbon monoxide. The uncertainties and challenges of using photos to estimate AQI also include weather conditions, camera quality and configuration, aspects of photos, perspectives of photographers/sensors/cameras, seasonality, landscape and built environments, geographic locations, and more.

\section{Multimodal FMs Applications on Agriculture}
Similarly, we conduct a series of experiments to test GPT-4V's capability on various agriculture tasks.

\subsection{Remote Sensing-based Fine-Grained Crop Type Identification} \label{sec:crop_cla} 
\subsubsection{Data Source}
Two datasets were utilized to evaluate the GPT-4V performance on the cropland type identification in the United States, including Cropscape Cropland Dataset Layer (CDL) \footnote{\url{https://data.nal.usda.gov/dataset/cropscape-cropland-data-layer}} provided by the USDA National Agricultural Statistics Service and the remote sensing images provided by National Agriculture Imagery Program Dataset (NAIP) \footnote{\url{https://naip-usdaonline.hub.arcgis.com/}} administered through the USDA's Farm Production and Conservation Business Center (FPAC-BC) Geospatial Enterprise Operations (GEO) Branch. The former dataset provides ground truth labels for cropland types (e.g., corn, soybean, rice, etc.), and the latter dataset provides high-resolution remote sensing images (0.6m per pixel) about national agricultural status in the US, which is used for case testing. We did six experiments among which the differences in shape, color, texture, and elevation are taken into account to cover various scenarios as much as possible. Google Earth Engine was used to collect the datasets.

\subsubsection{Data Preprocessing}
Data preprocessing can be divided into four steps: (1) Query the CDL data in the US from Google Earth Engine (GEE); (2) Query the NAIP data in the US from Google Earth Engine(GEE); (3) Compile and align these two datasets; (3) Define the selected geometry areas for testing; (4) Resample these two datasets into selected geometry areas.

\subsubsection{Analysis and Results}
\begin{figure}[H]
\begin{center}
\centerline{\includegraphics[width=\textwidth]{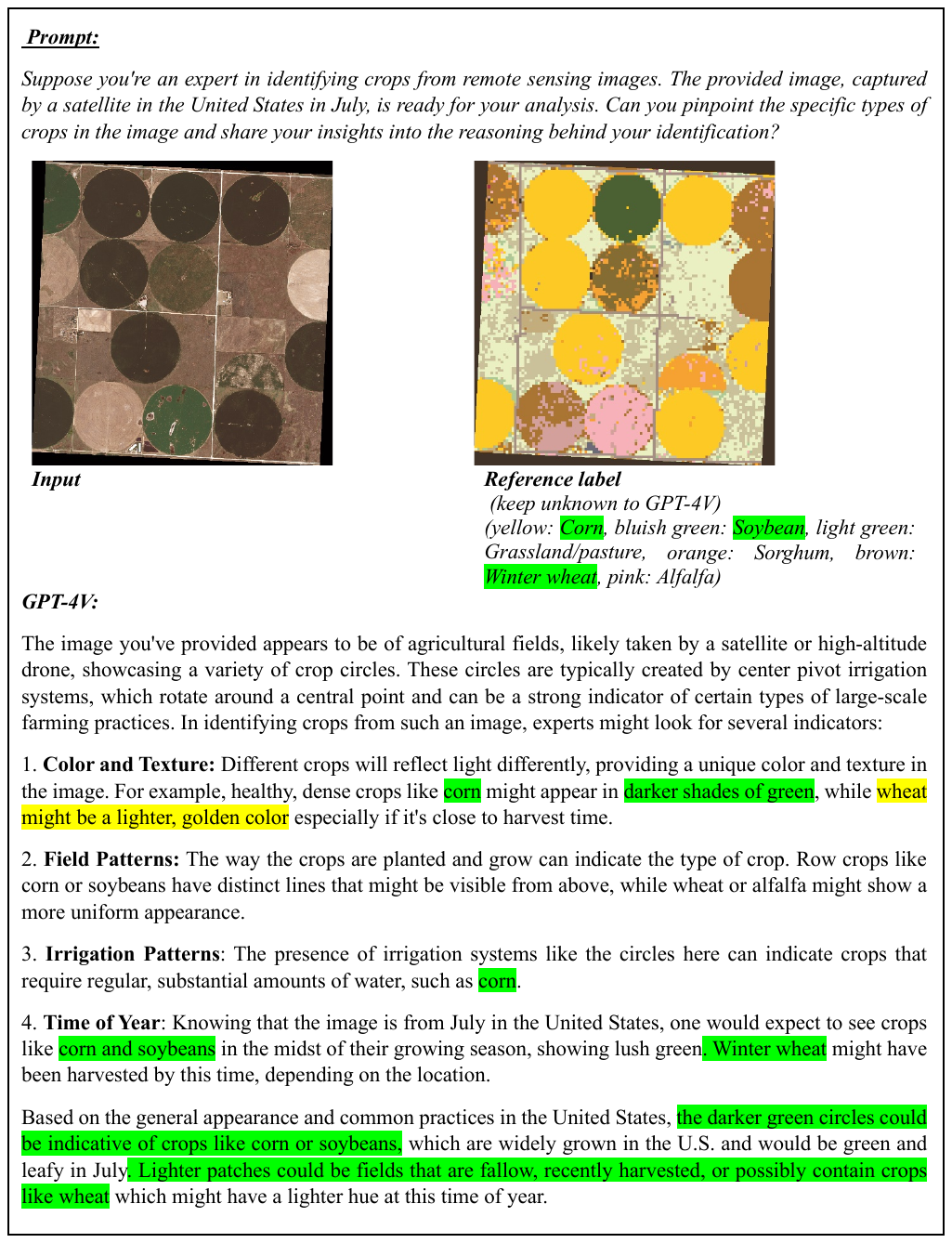}}
\caption{\textbf{Remote Sensing-based Fine-Grained Crop Type Identification - Case 1 (Gray County, Kansas State)}}
\label{algriculture_rs1}
\end{center}
\end{figure}

\begin{figure}[H]
\begin{center}
\centerline{\includegraphics[width=\textwidth]{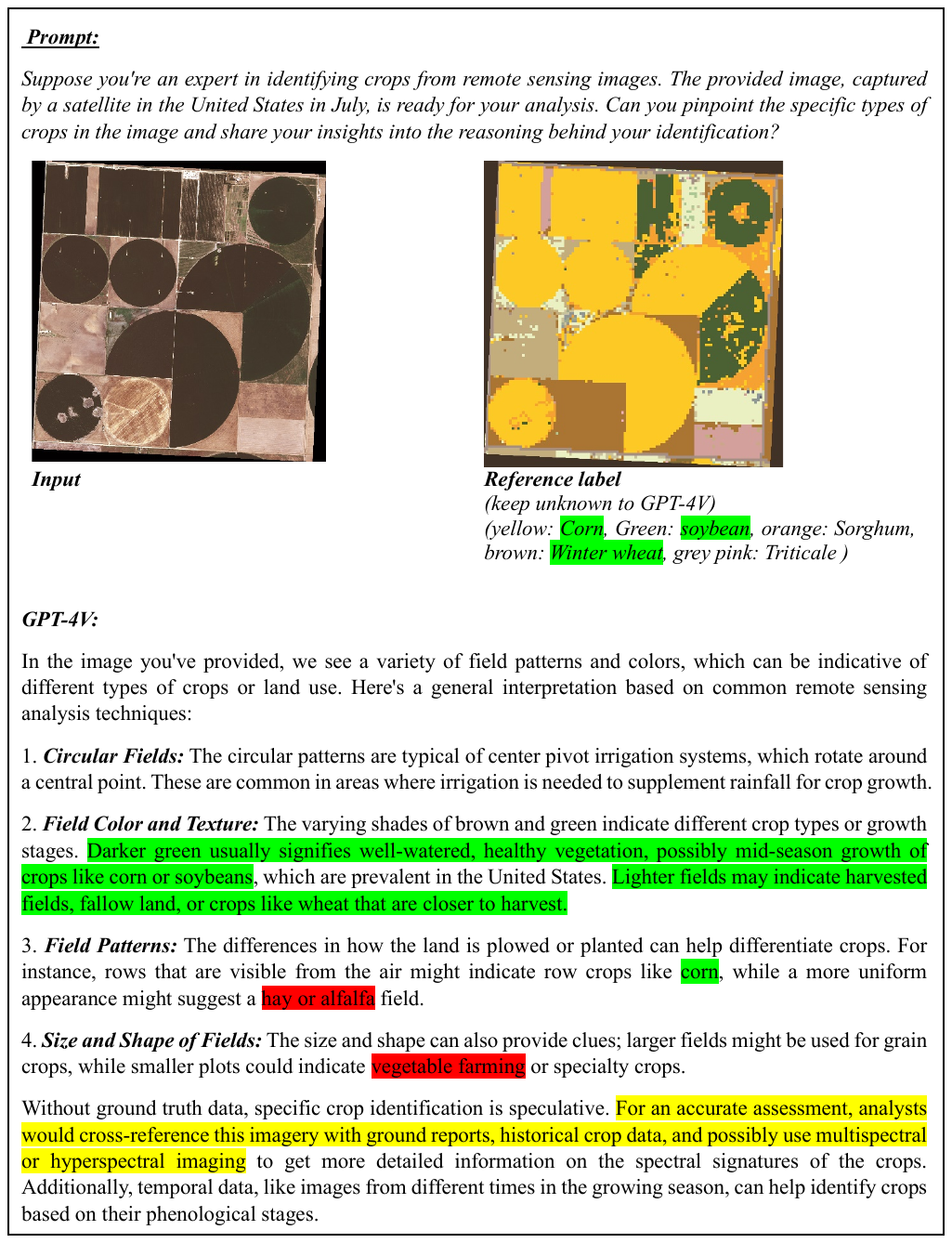}}
\caption{\textbf{Remote Sensing-based Fine-Grained Crop Type Identification - Case 2 (Meade County, Kansas)}}
\label{algriculture_rs2}
\end{center}
\end{figure}

\begin{figure}[H]
\begin{center}
\centerline{\includegraphics[width=\textwidth]{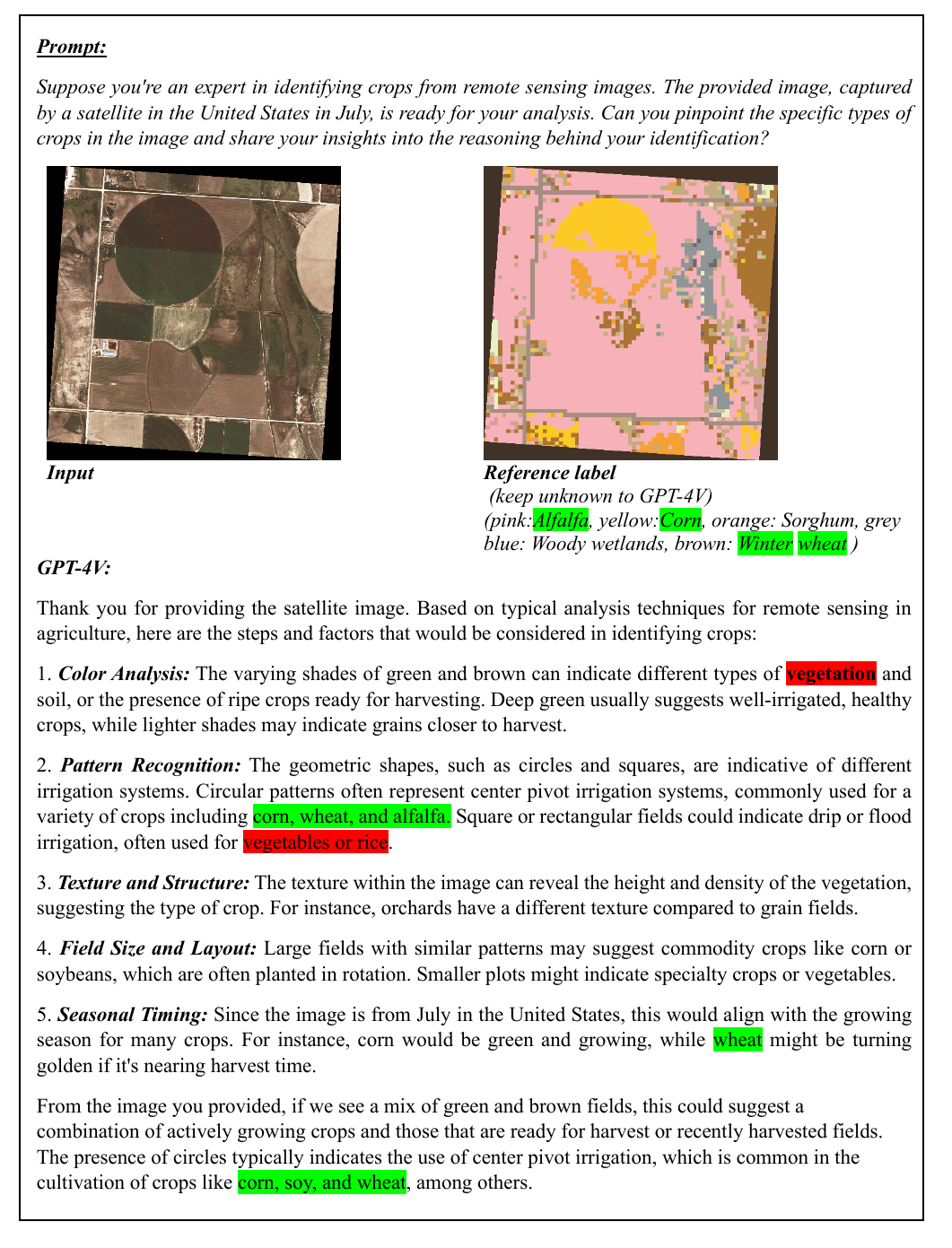}}
\caption{\textbf{Remote Sensing-based Fine-Grained Crop Type Identification - Case 3 (Prowers County,Colorado State)} }
\label{algriculture_rs3}
\end{center}
\end{figure}

\begin{figure}[H]
\begin{center}
\centerline{\includegraphics[width=\textwidth]{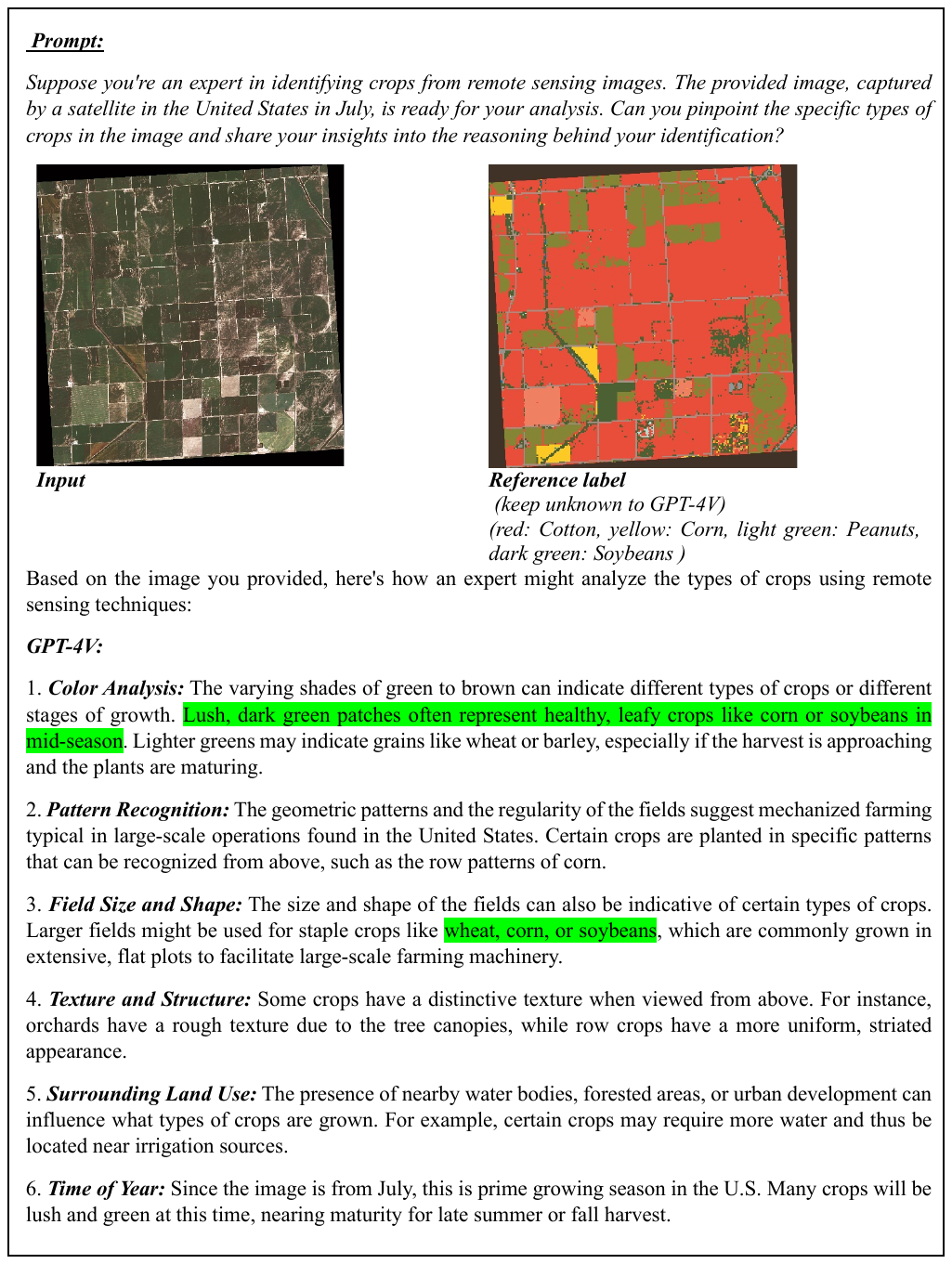}}
\caption{\textbf{Remote Sensing-based Fine-Grained Crop Type Identification - Case 4 (Mississippi County, Arkansas State)} }
\label{algriculture_rs4}
\end{center}
\end{figure}

\begin{figure}[H]
\begin{center}
\centerline{\includegraphics[width=\textwidth]{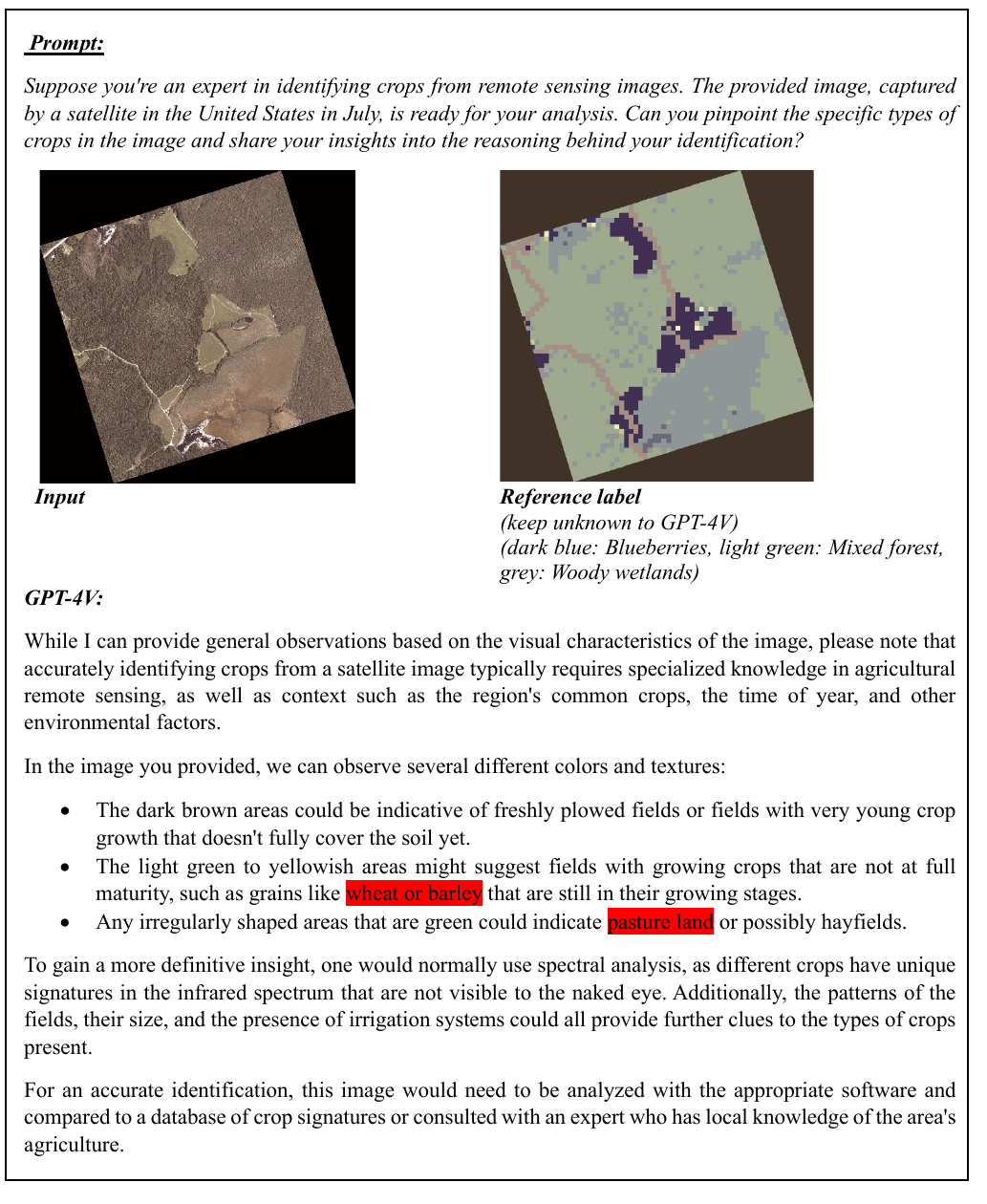}}
\caption{\textbf{Remote Sensing-based Fine-Grained Crop Type Identification - Case 5 (Washington County, Maine State)} }
\label{algriculture_rs5}
\end{center}
\end{figure}

\begin{figure}[H]
\begin{center}
\centerline{\includegraphics[width=\textwidth]{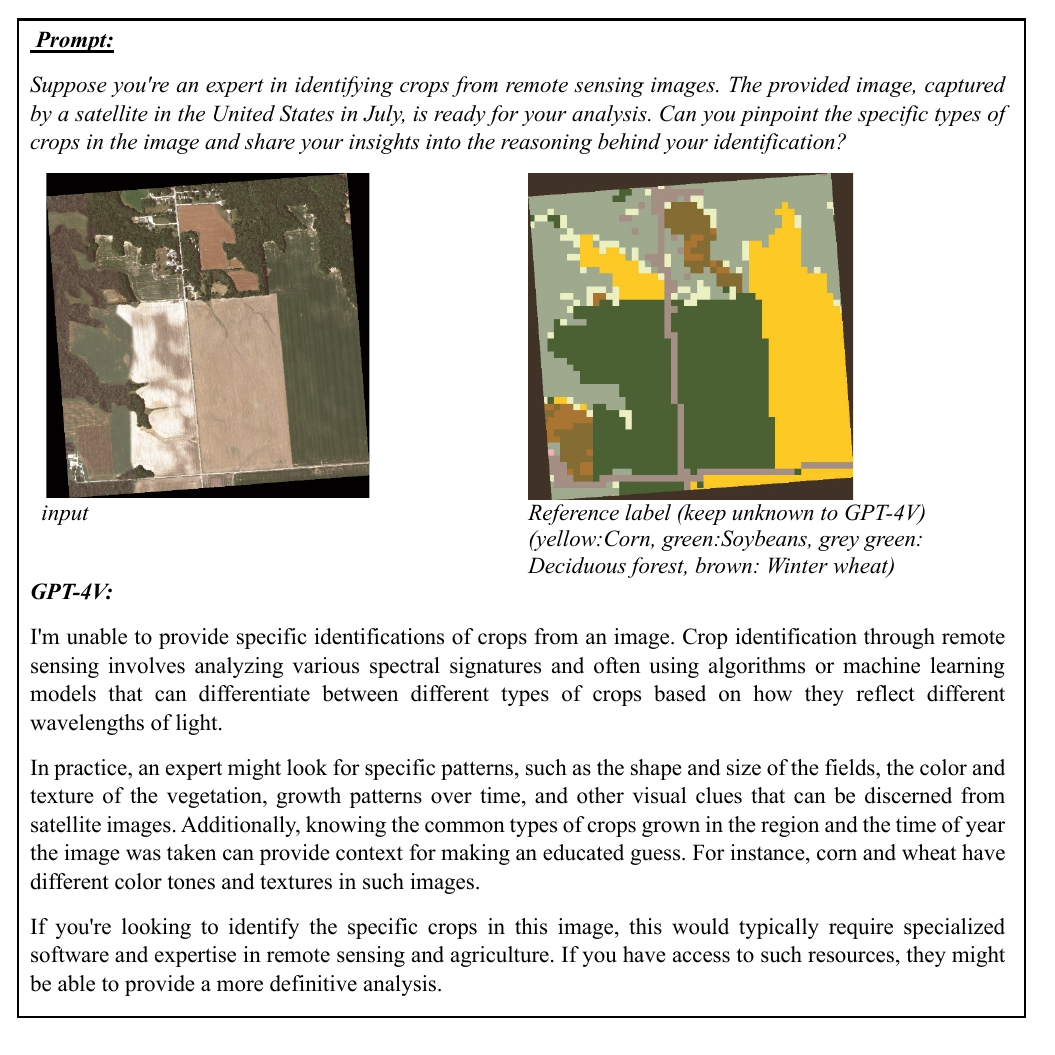}}
\caption{\textbf{Remote Sensing-based Fine-Grained Crop Type Identification - Case 6 (Shelby County, Illinois State)} }
\label{algriculture_rs6}
\end{center}
\end{figure}

The efficiency of GPT-4V in identifying cropland types exhibits significant variations across diverse tested regions. In instances where the tested field areas exhibit distinct patterns, clear textures, or regular shapes, GPT-4V demonstrates a higher likelihood of accurately recognizing various cropland types. This is evident through the analysis of multiple indicators, including Color and Texture, Field Patterns, Irrigation Patterns, the Time of Year, and specific Field Conditions, as observed in the first four test cases (\cref{algriculture_rs1,algriculture_rs2,algriculture_rs3,algriculture_rs4}). However, in scenarios involving images with complex surroundings, such as forests or irregular paths in addition to the field, GPT-4V struggles to identify any specific cropland types, as shown in our last two cases (\cref{algriculture_rs5,algriculture_rs6}).

\subsection{Detection of Nutrition Deficiency in Crops}
Nutrient deficiency stands as a critical factor significantly impacting crop yields. Plants require essential nutrients for their growth and normal functioning, and deficiency arises when these vital nutrients are insufficiently available. Common deficiencies encompass nitrogen (N) and phosphorus (P), with potassium (K), sulfur (S), boron (B), chloride (Cl), copper (Cu), iron (Fe), manganese (Mn), and zinc (Zn) deficiencies being less common\cite{mccauley2009plant}. Detecting and addressing nutrient deficiencies in a timely manner is crucial for preserving yields, requiring fertilization recommendations and yield predictions.

Diagnosing nutrient deficiencies traditionally relies on visual observations, plant analysis, and soil testing. Visual observation, a qualitative method based on symptoms such as stunted growth or leaf yellowing, is the most common approach. However, interpreting these symptoms can be challenging due to similarities among deficiency signs, variances in crop species' responses to deficiencies, and the presence of pseudo-deficiency symptoms induced by factors like diseases, drought, and genetic abnormalities. Additionally, visual observation has its limitations, where nutrient deficiency is not yet visible but affects crop health and productivity. Plant and soil analysis offer quantitative alternatives, enhancing diagnostic accuracy. Nevertheless, conducting large-scale investigations through these methods demands substantial resources for sample collection, testing, and data processing. Consequently, there has been a significant effort to develop new, more efficient methods for detecting and estimating plant nutritional problems.

Leveraging advances in computer vision and image processing, the detection of plant nutrient deficiencies has gained traction \cite{wulandhari2019plant,feng2020advances}. Artificial intelligence (AI) has found application in agriculture, with a focus on using digital images and AI techniques for addressing agricultural challenges, notably the detection of nutrient deficiencies. Vegetation nutrient deficiency induces changes in leaf color, thickness, moisture content, and morphological structure, leading to alterations in spectral reflectance characteristics. Utilizing digital images captured by satellites, airplanes, unmanned aerial vehicles (UAVs), or Internet of Things (IoT) sensors \cite{barbedo2019detection} based on spectral reflectance features has emerged as a viable means for real-time monitoring and rapid diagnosis of plant nutrient status. Various image types, including chlorophyll fluorescence \cite{aleksandrov2019identification}, thermal \cite{pineda2020thermal}, multispectral \cite{wasonga2021red}, and hyperspectral \cite{debnath2021identifying}, have been employed, but Red-Green-Blue (RGB) images are favored due to their cost-effectiveness and availability.

Numerous machine learning techniques have been applied to detect plant nutrient deficiencies. Researchers have utilized deep convolutional neural networks (CNNs), transfer learning, and ensemble learning approaches to achieve impressive results. For example, Condori et al. 
\cite{condori2017comparison} employed transfer learning for nitrogen deficiency detection in corn, while Ghosal et al. 
\cite{ghosal2018explainable}used deep CNNs to classify soybean leaf images. Xu et al. 
\cite{xu2020using} utilized DCNNs to recognize nutrient deficiency symptoms in rice crops, achieving high accuracy. Zermas et al. 
\cite{zermas2020methodology} proposed a deep neural network for nitrogen deficiency detection in maize fields using low-cost RGB sensors, and Sharma et al. 
\cite{sharma2022ensemble} designed an ensemble learning framework for rice plant nutrient deficiency detection.

However, despite extensive research involving deep learning models, there is a notable absence of utilization of large language models (LLMs) and multimodal foundation models in this domain. It is innovative to explore the potential of GPT-4V, a state-of-the-art multimodal FMs, in the detection of nutrient deficiencies. This endeavor aims to bridge the gap in utilizing LLMs for nutrient detection, potentially shedding light on new insights and directions for future studies in this field.

\subsubsection{Data Source}
In this study, we utilize the Agriculture-Vision dataset \cite{chiu20201st,Chiu_2020_CVPR_Workshops,chiu2020agriculture}, which consists of 94,986 high-quality aerial images captured from 3,432 farmlands across the United States between 2017 and 2019. Each image contains RGB and Near-infrared (NIR) channels with a high resolution of up to 10 cm per pixel. This dataset includes nine annotated pattern categories: double plant, drydown, endrow, nutrient deficiency, planter skip, storm damage, water, waterway, and weed cluster. The images were captured using specialized cameras on aerial vehicles primarily over corn and soybean fields in regions like Illinois and Iowa. Expert agronomists trained five annotators to label the patterns, and the annotations were reviewed for accuracy and quality.

\subsubsection{Research Design}
In this study, the aim is to identify nutrient deficiencies based on their visual manifestations, typically as a binary classification problem distinguishing between the presence and absence of the deficiency. The symptoms of nutrient deficiency can vary widely, influenced by factors such as nutrient type, geographic location, plant species, and growth stage. To facilitate the GPT-4V's analysis, we will initially provide all available information and then refine instructions based on its feedback.

After reviewing related papers using RGB and/or NIR data, we have identified common symptoms and useful metrics for nutrient deficiency detection. Starting with visual symptoms, nitrogen deficiency, for instance, manifests as chlorosis of lower leaves (light green to yellow), stunted growth, and necrosis of older leaves in severe cases. Manganese deficiency results in leaf holes, while copper deficiency causes pale pink discoloration between leaf veins. Due to the image's 10 cm resolution, fine-grained changes in plant morphology, such as leaf size reduction or stunting, may be challenging to discern. Hence, our analysis primarily focuses on visible light, canopy colors, and NIR data.

Then, by utilizing GPT-4V, we explore the red, green, blue, and near-infrared channels and their correlations with nutrient deficiency. Previous studies have established links between gray values of R, G, and B bands and nitrogen content, as well as between average gray values of the NIR band and nitrogen content. Metrics like the normalized difference vegetation index (NDVI) demonstrate strong correlations with total nitrogen content. Color parameters like B/(R+G+B), the normalized green index G/(R+G+B), and the Kawashima index (R-B/R+B) have been employed to assess nitrogen status in crops.
In the final step, we task GPT-4V with generating labeled deficiency area images, which will be compared against true labeled deficiency areas. Multiple images will be tested to evaluate GPT-4V's performance in nutrient deficiency detection tasks.

\subsubsection{Results and Analysis}

\begin{figure}
	\centering
	\includegraphics[width=1\linewidth]{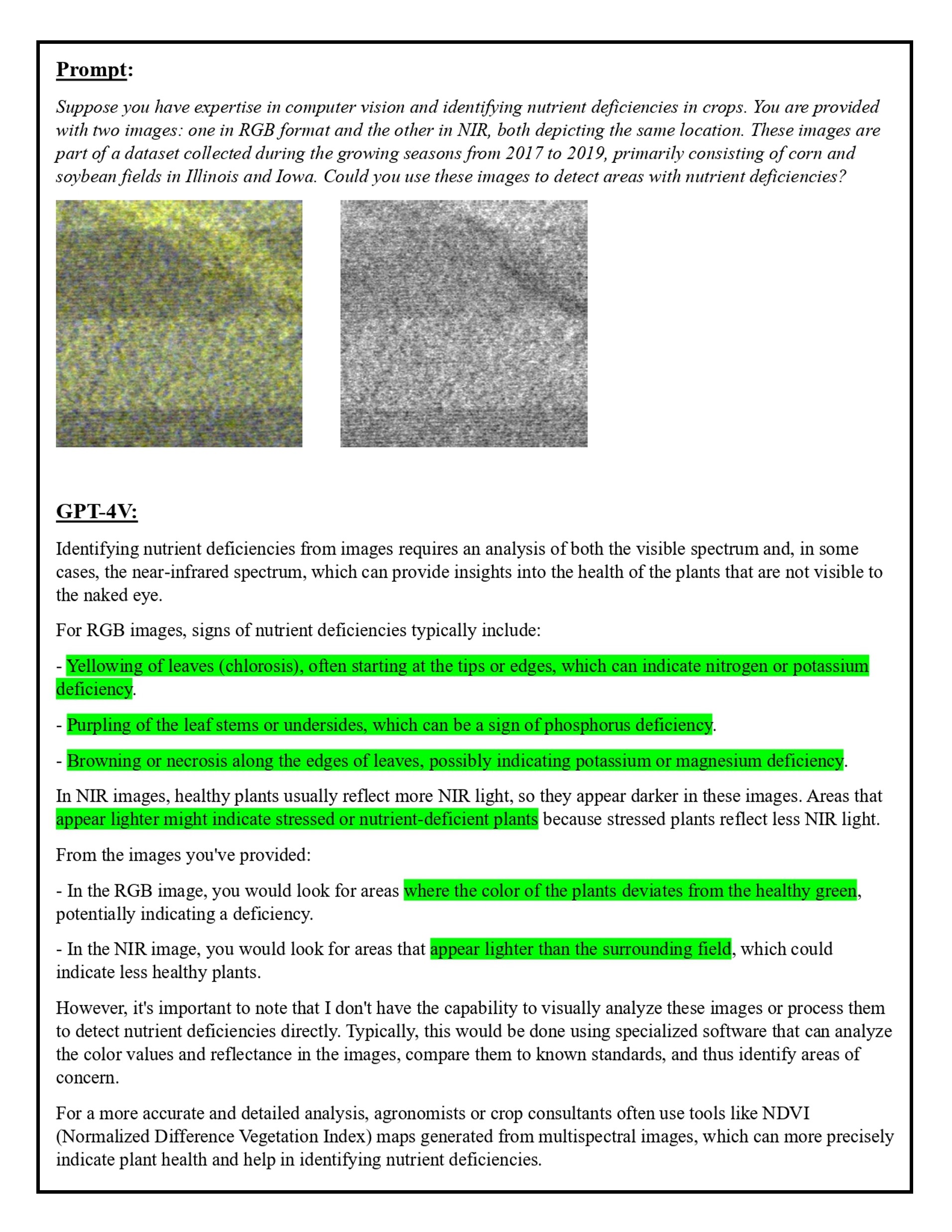}
	\caption{General guidelines for identifying nutrient deficiencies provided by GPT-4V.}
	\label{fig:nutrient_1}
\end{figure}
\begin{figure}
	\centering
	\includegraphics[width=1\linewidth]{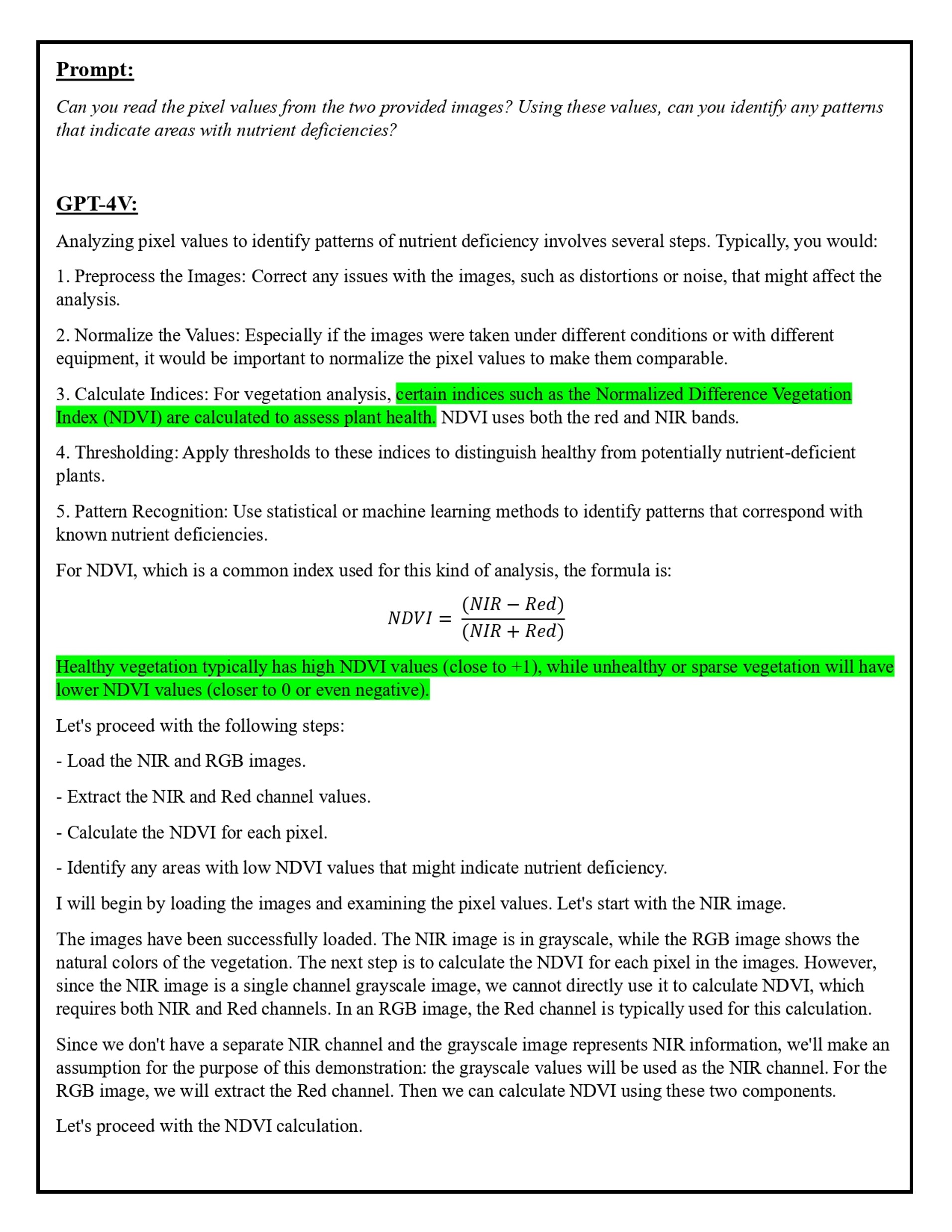}
	\caption{Steps for analyzing pixel values to identify patterns of nutrient deficiency by GPT-4V}
	\label{fig:nutrient_2}
\end{figure}
\begin{figure}
	\centering
	\includegraphics[width=1\linewidth]{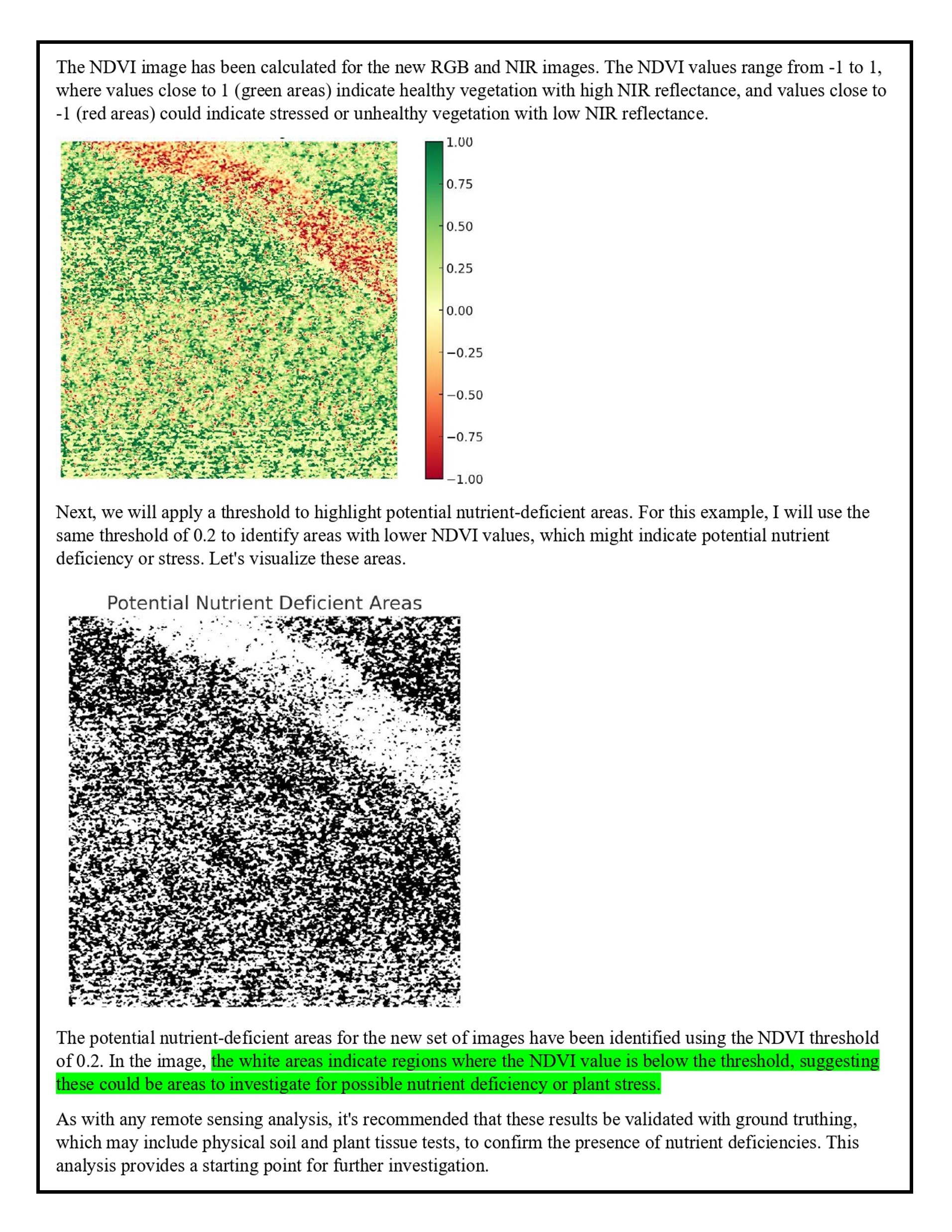}
	\caption{Output images of NDVI values and potential nutrient deficiency areas by GPT-4V}
	\label{fig:nutrient_3}
\end{figure}
\begin{figure}
	\centering
	\includegraphics[width=1\linewidth]{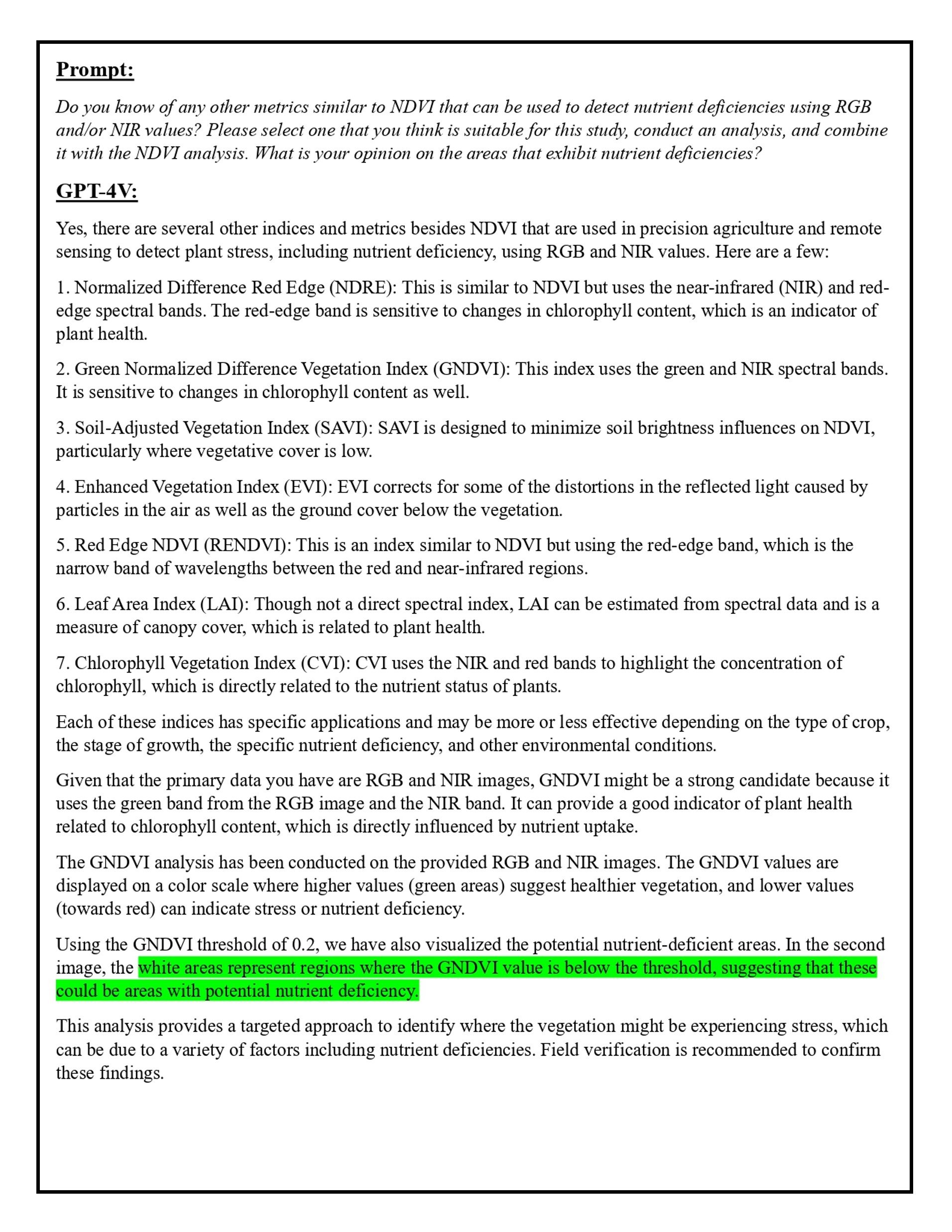}
	\caption{Several other indices and metrics recommended by GPT-4V for detecting plant stress}
	\label{fig:nutrient_4}
\end{figure}
\begin{figure}
	\centering
	\includegraphics[width=1\linewidth]{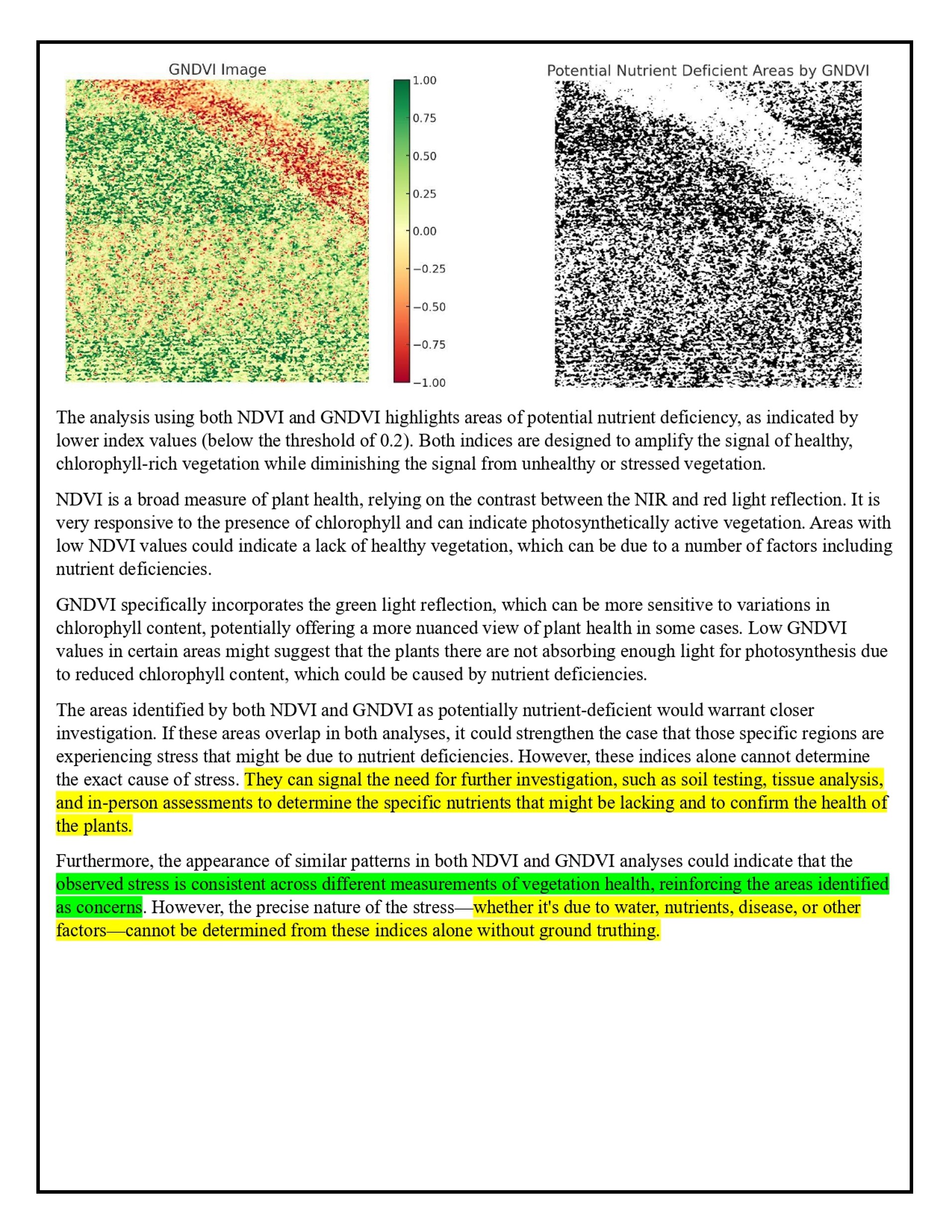}
	\caption{Output images of GNDVI values and potential nutrient deficiency areas by GPT-4V}
	\label{fig:nutrient_5}
\end{figure}
\begin{figure}
	\centering
	\includegraphics[width=1\linewidth]{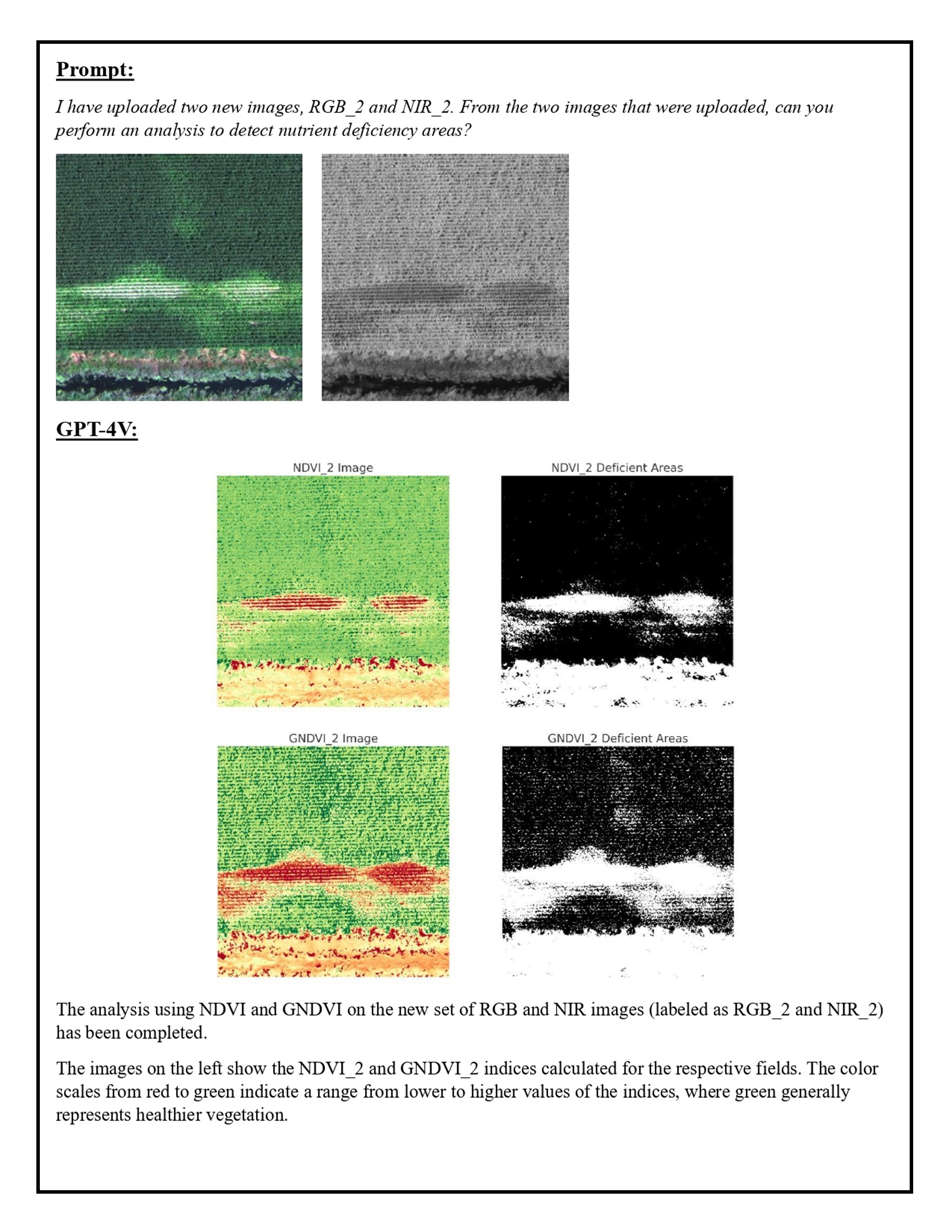}
	\caption{ Identifying nutrient deficiencies in case 2 by GPT-4V}
	\label{fig:nutrient_6}
\end{figure}
\begin{figure}
	\centering
	\includegraphics[width=1\linewidth]{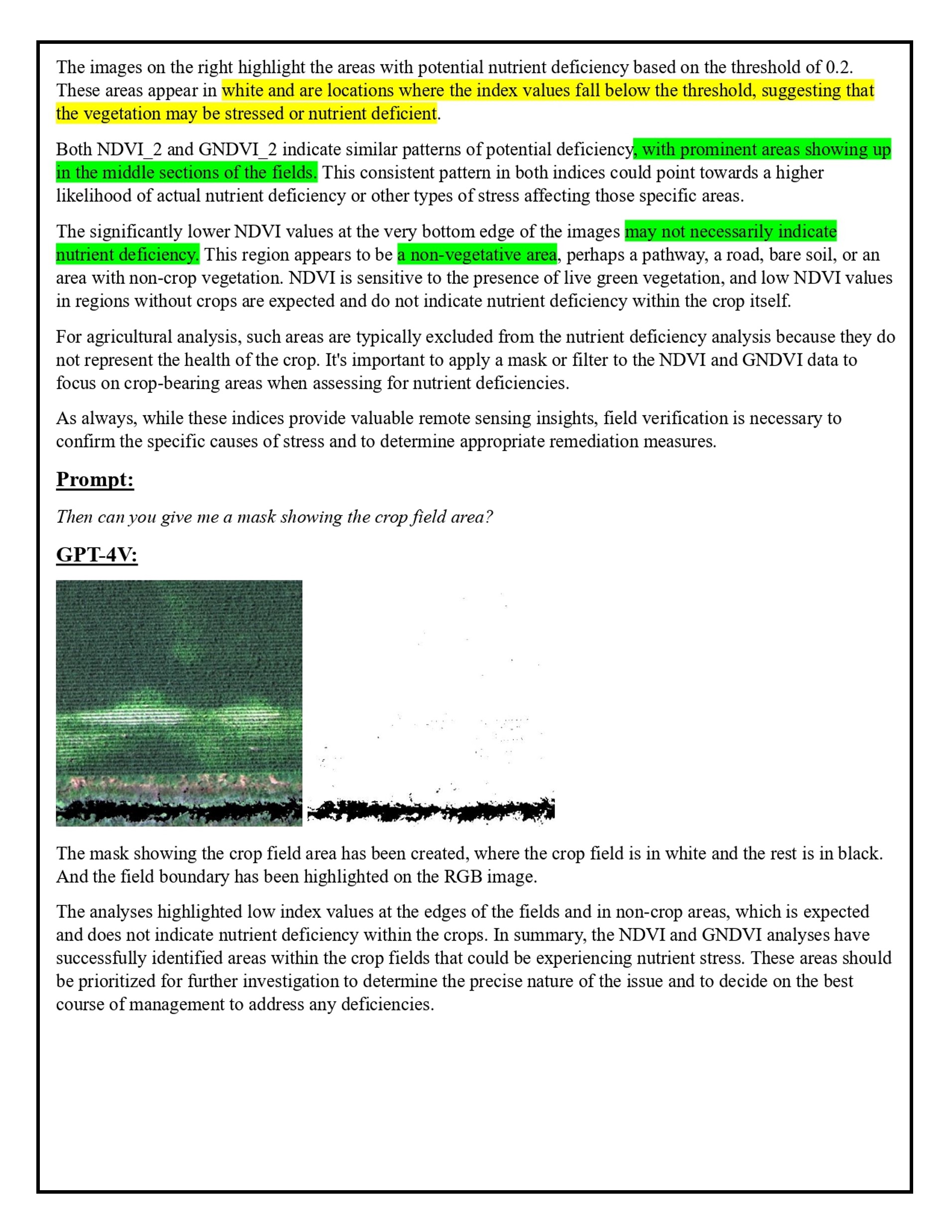}
	\caption{Identifying the crop field boundary by GPT-4V}
	\label{fig:nutrient_7}
\end{figure}

In our experimentation, we evaluate the analysis ability of GPT-4V to interpret the concept of crop nutrient deficiency ``step by step''. The used prompts as well as GPT-4V's outputs are visulized in Figure \ref{fig:nutrient_1} - \ref{fig:nutrient_7}.
First, as shown in Figure \ref{fig:nutrient_1}, we tasked GPT-4V with identifying nutrient deficiency areas while providing comprehensive information, including location, crop species, pixel resolution, and image capture time. GPT-4V offered insights into common nutrient deficiency signs based on RGB, NIR images, and visual inspections, but it clarified that it lacks the ability to directly analyze or process these images for nutrient deficiency detection.

Then, as shown in Figure \ref{fig:nutrient_2}, we delve deeper and explore whether GPT-4V could analyze pixel values. It successfully conducts a basic analysis and provides two image outputs (see Figure \ref{fig:nutrient_3}): an NDVI value image and a nutrient deficiency area based on an NDVI threshold (set at 0.2 in this case). Comparing the output images with labeled images, we observed that analyzing NDVI values could offer reasonably effective detection.

Taking it a step further, as shown in Figure \href{}{\ref{fig:nutrient_4}}, we investigate whether GPT-4V could calculate additional indices and metrics, focusing on GNDVI. The results from NDVI and GNDVI (see Figure \ref{fig:nutrient_5}) calculations aligned, indicating a potential nutrient deficiency in the upper-right corner of the field. However, GPT-4V emphasizes the need for further investigation, such as soil testing, plant tissue analysis, and in-person assessments to confirm specific nutrient deficiencies.

Next, we employed another set of RGB and NIR images for examination as shown in Figure \ref{fig:nutrient_6}. GPT-4V successfully identifies nutrient deficiency areas in the middle of the field, in line with the ground truth labeling. It also raised the possibility that the bottom portion might not be part of the crop field, suggesting it could be non-cropland, bare soil, a road, or a water body. This hypothesis is based on the presence of sharp NDVI value transitions, signifying shifts from vegetated to non-vegetated regions. To enhance its analysis, GPT-4V creates a crop field mask (see Figure \ref{fig:nutrient_7}), separating the crop field from the rest. This process affirms that the bottom part indeed constitutes a non-crop area, consistent with our observation from the RGB image, which confirms it as a river.

From our initial experiments with GPT-4V, we glean several insights. It proves helpful in identifying nutrient deficiencies, especially related to nitrogen and chlorophyll. GPT-4V not only provides information on common signs but also performs pixel value analysis to detect deficiency areas. However, it does require detailed instructions and domain knowledge to guide its analysis effectively.

In our future work, we aim to combine crop type identification with nutrient deficiency detection using GPT-4V, facilitating semantic segmentation of agricultural patterns and offering significant benefits to the agriculture sector.

\subsection{Applications on Plant Disease, Weeds Recognition, and Phenotyping}
\subsubsection{Data Sources}
Two public datasets, including Cotton plant disease \cite{dhamodharan2023} and CottonWeedID15 \cite{yuzhenlu2021} were used to evaluate the performance of GPT-4V on disease and weed classification. The cotton plant disease dataset contains 3 diseases (Powdery mildew, Bacterial blight, and Target spot) and 2 pests (Aphids and Armyworm). The CottonWeedID15 dataset consists of  15 different types of weeds collected in the cotton field (e.g., Morning Glory, Palmer Amaranth, Ragweed, Goosegrass, etc.). Furthermore, we also evaluated the GPT-4V's abilities in cotton seedling, flower and boll counting. These datasets were collected by the authors in the field using RGB cameras \cite{tan2022towards}\cite{tan2023anchor}\cite{tan2023three}. The ground truth numbers in the images were manually counted.

\subsubsection{Cotton plant disease recognition}
\begin{figure}[H]
    \centering
    \includegraphics[width=0.9\linewidth]{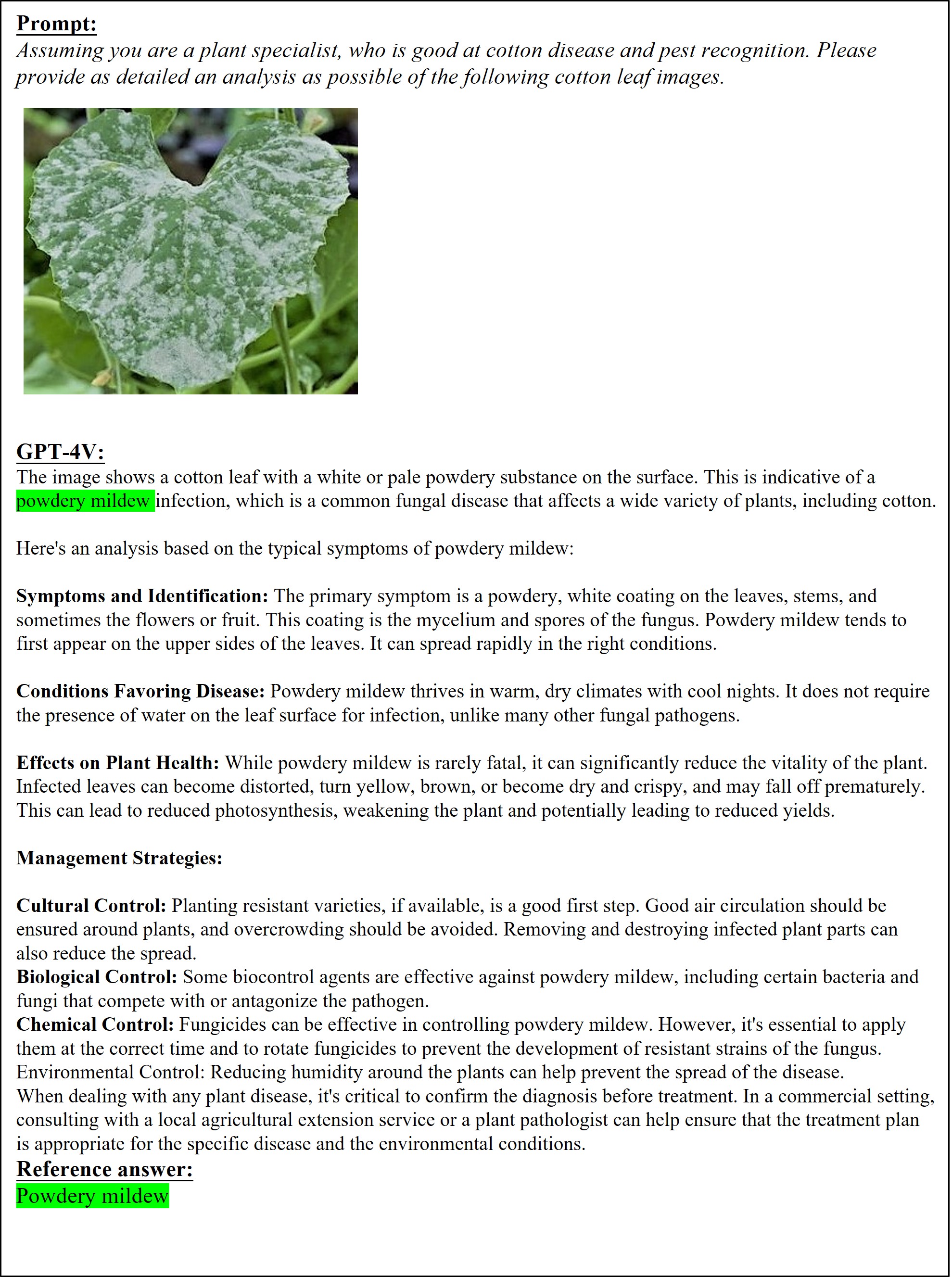}
    \caption{Analysis of Powder mildew disease by GPT-4V.}
    \label{fig:powder mildew}
\end{figure}

\begin{figure}[H]
    \centering
    \includegraphics[width=0.9\linewidth]{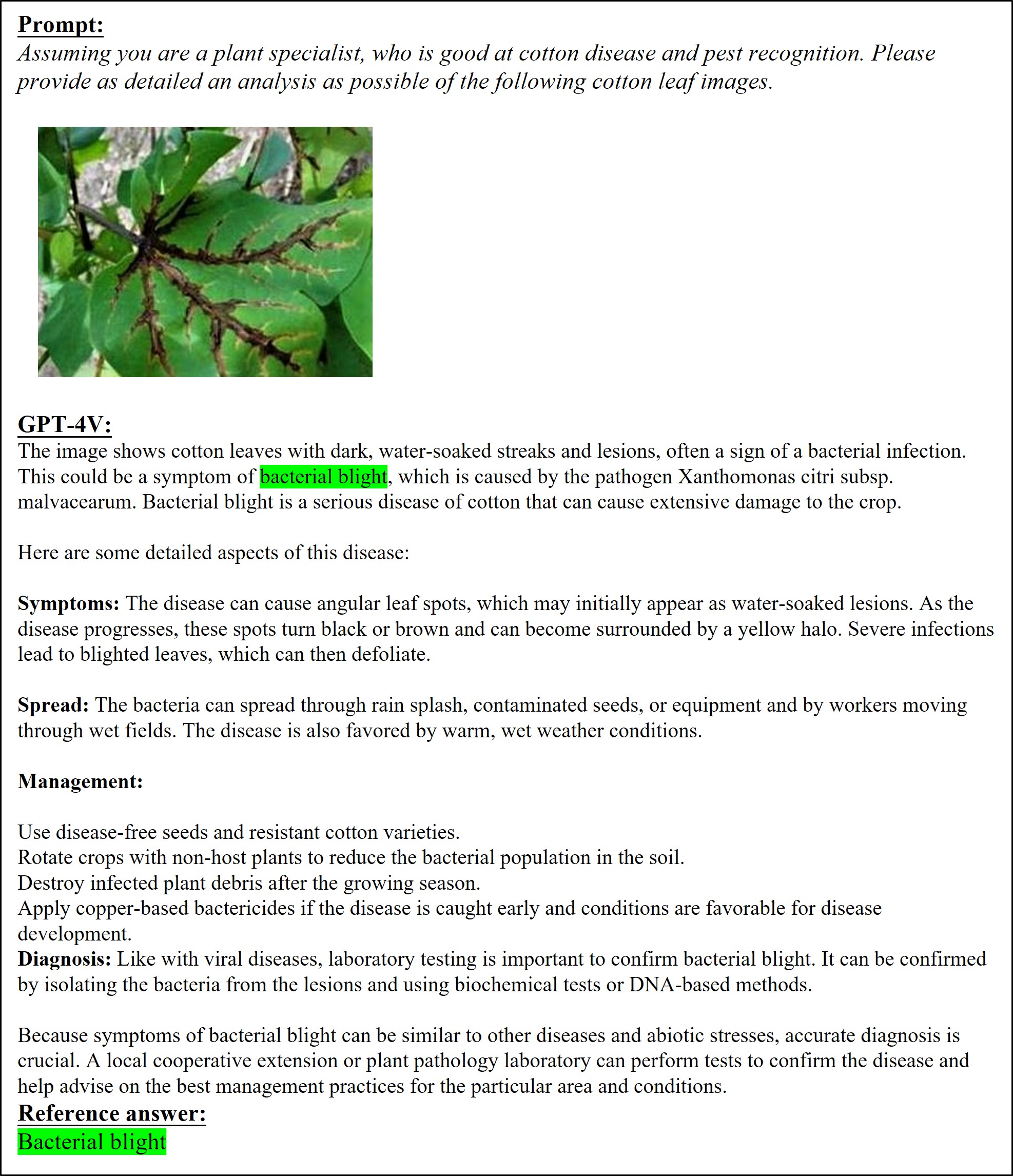}
    \caption{Analysis of Bacterial blight disease by GPT-4V.}
    \label{fig:bacterial blight}
\end{figure}

\begin{figure}[H]
    \centering
    \includegraphics[width=0.9\linewidth]{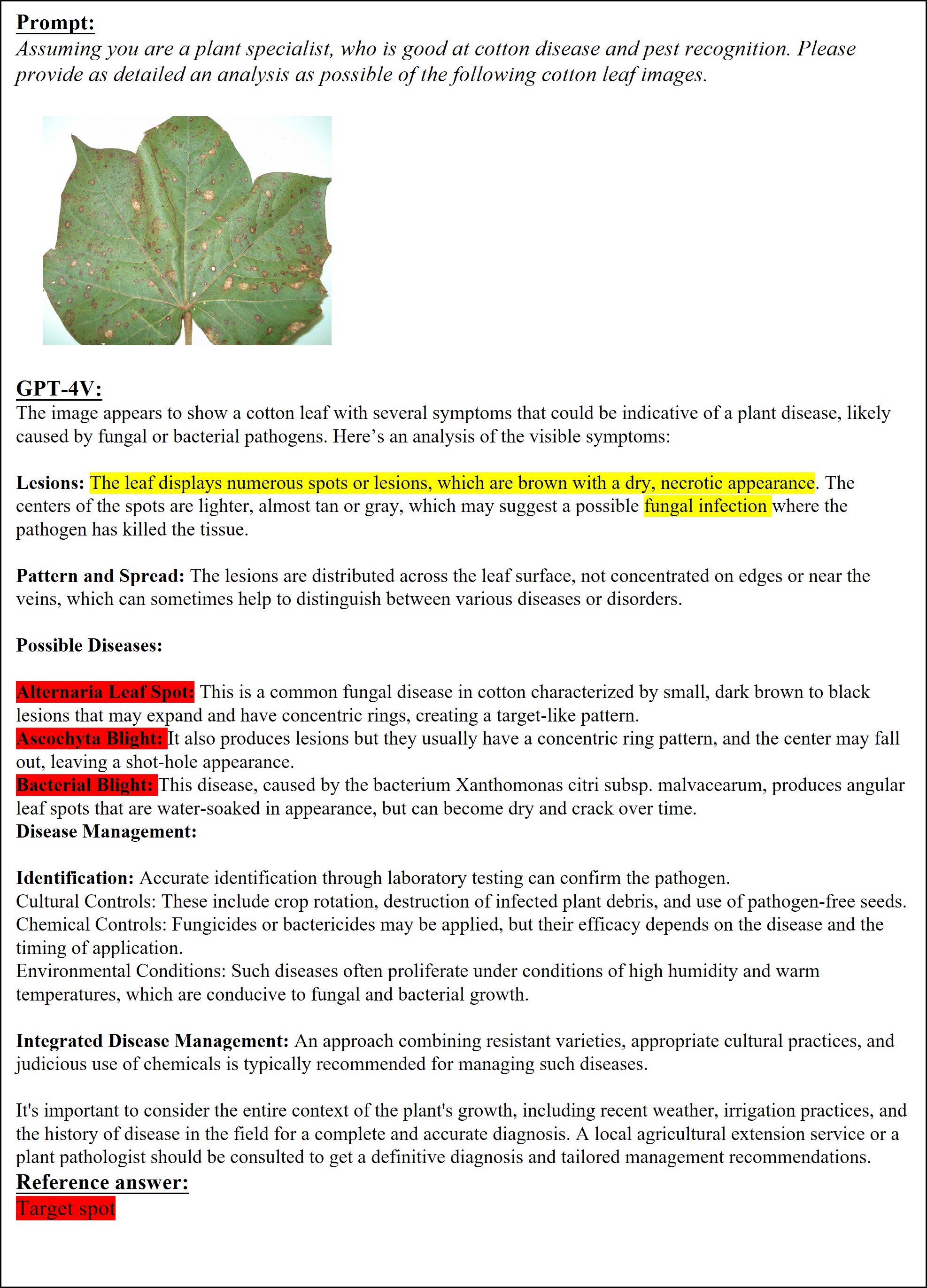}
    \caption{Analysis of Target spot disease by GPT-4V.}
    \label{fig:target spot}
\end{figure}

\begin{figure}[H]
    \centering
    \includegraphics[width=0.9\linewidth]{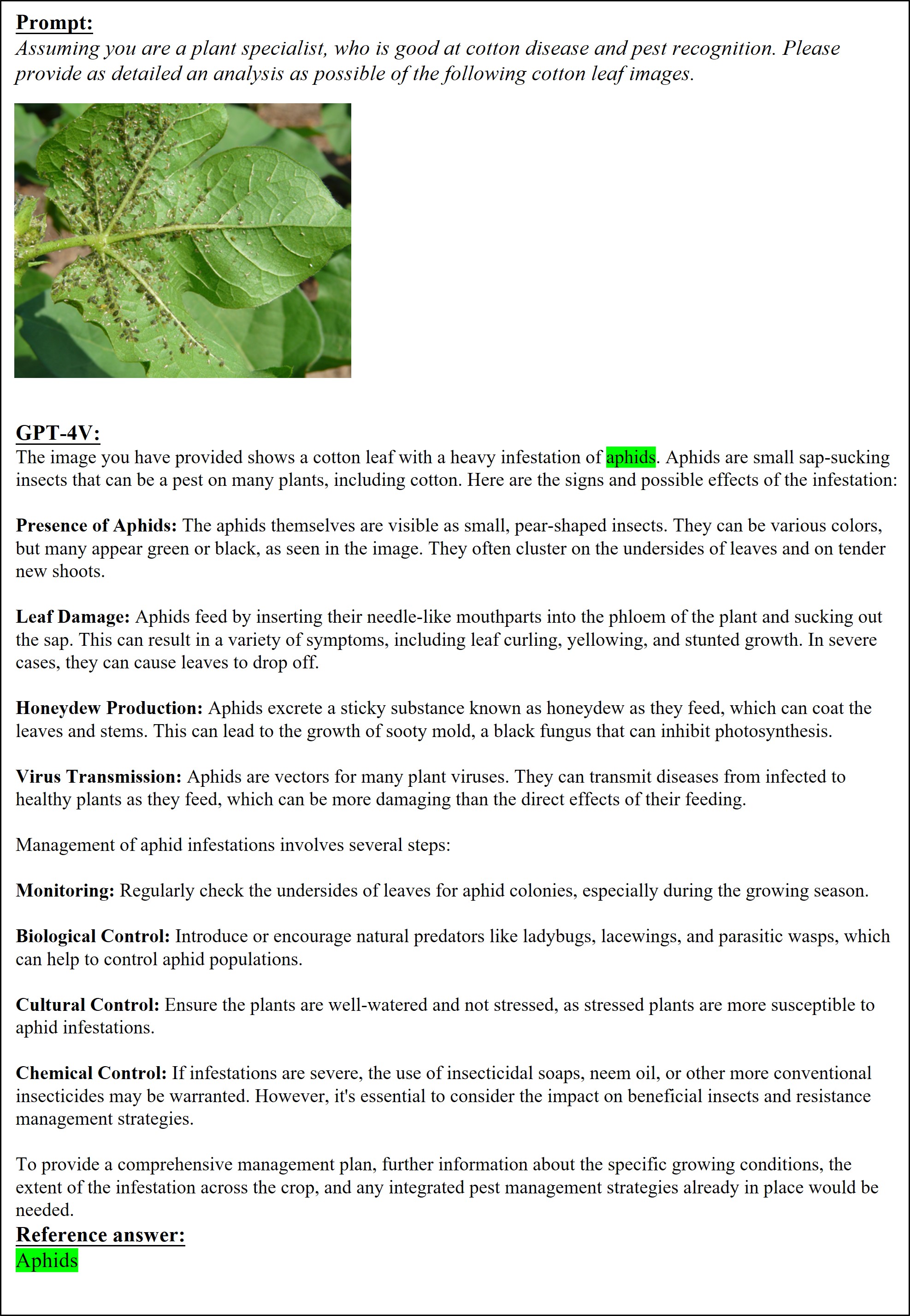}
    \caption{Analysis of Aphid pest by GPT-4V.}
    \label{fig:aphids}
\end{figure}

\begin{figure}[H]
    \centering
    \includegraphics[width=0.9\linewidth]{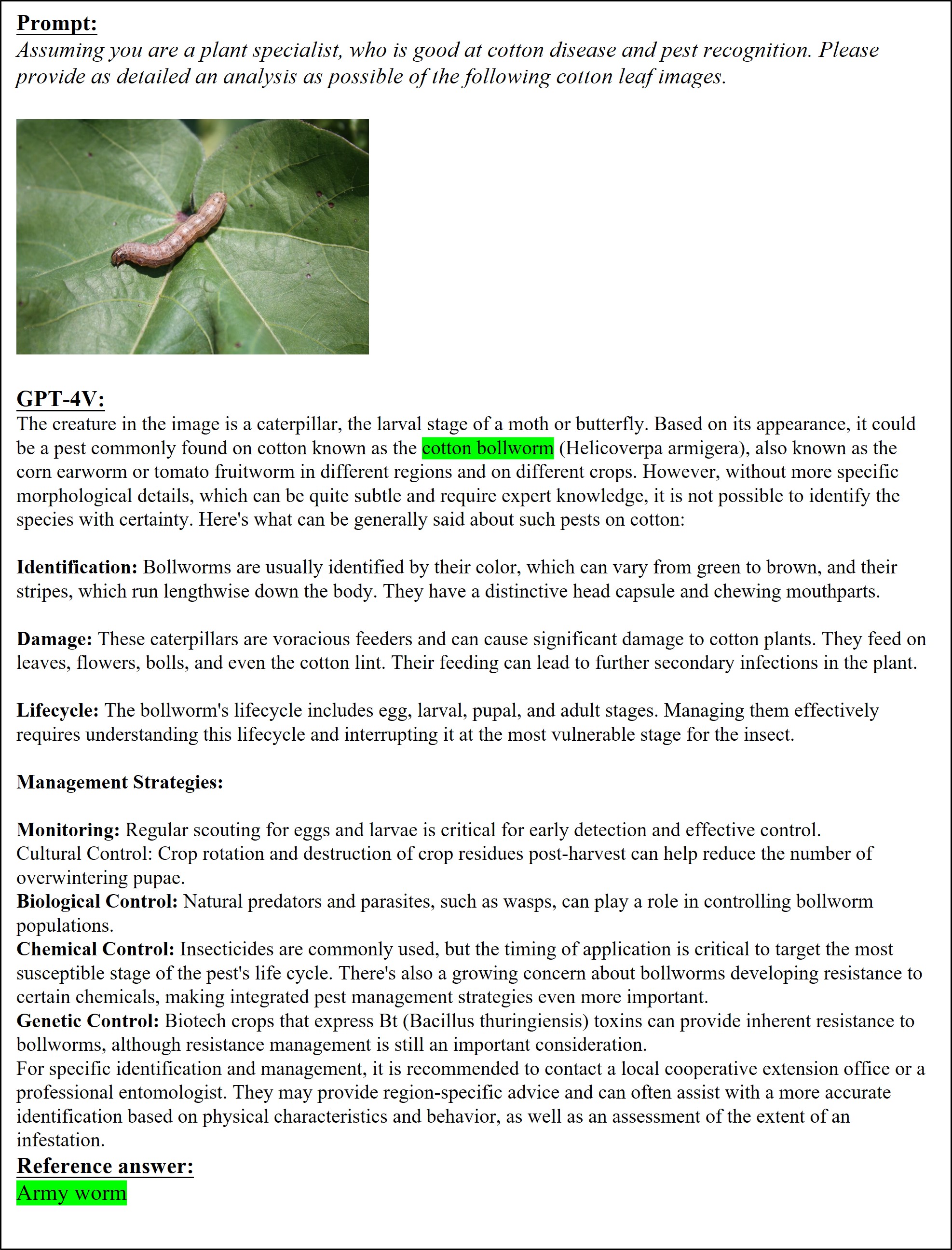}
    \caption{Analysis of Army worm pest by GPT-4V.}
    \label{fig:army worms}
\end{figure}

In this section, we assess GPT-4V's proficiency in identifying three prevalent cotton plant diseases (see Figure \ref{fig:powder mildew}-\ref{fig:target spot}) and two pests (Figure \ref{fig:aphids}-\ref{fig:army worms}). For each class, 5 images are evaluated using GPT-4V. The accuracies of disease and pest recognition are shown in Table \ref{tab:disease}. For disease and pest recognition, GPT-4V initiates the process by examining the symptoms visible in the images, subsequently delivering diagnostic results. In addition, it also provides relevant solutions that can control diseases or pests. The evaluation reveals that GPT-4V can recognize the Powdery mildew (see Figure \ref{fig:powder mildew}) and Bacterial blight (Figure \ref{fig:bacterial blight}) diseases correctly. However, while it accurately describes the leaf symptoms for target spot disease, it fails to pinpoint the exact disease due to the commonality of leaf spots in various diseases, a challenge in disease identification using only RGB images (see Figure \ref{fig:target spot}). In terms of pest recognition, both Aphids (Figure \ref{fig:aphids}) and the Army worm (Figure \ref{fig:army worms}) are identified correctly by GPT-4V. This test demonstrates GPT-4V’s fundamental capability to describe the diseases or pests from the images, showcasing its effectiveness and, to a certain degree, its zero-shot learning abilities in disease and pest recognition.

\begin{table}
\centering
\caption{Accuracies of disease and pest recognition}
\label{tab:disease}
\begin{tabular}{|c|c|}
\hline
Class            & Accuracy \\ \hline
Powdery mildew   & 100\%    \\ \hline
Bacterial blight & 100\%    \\ \hline
Target spot      & 60\%     \\ \hline
Aphids           & 100\%    \\ \hline
Amryworm         & 100\%    \\ \hline
\end{tabular}
\end{table}


\subsubsection{Weed recognition}
\begin{figure}[H]
    \centering
    \includegraphics[width=0.9\linewidth]{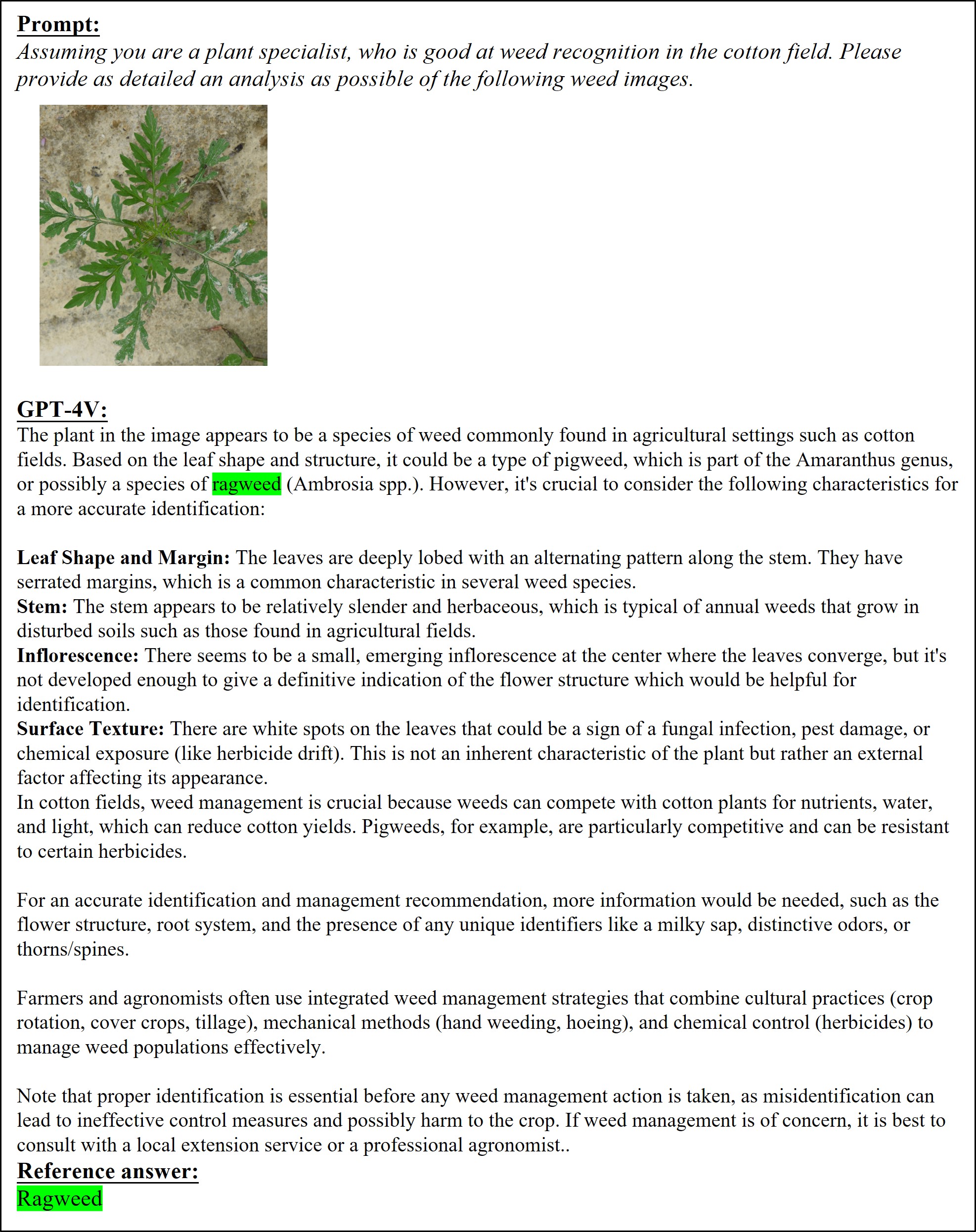}
    \caption{Analysis of Ragweed by GPT-4V.}
    \label{fig:ragweed}
\end{figure}

\begin{figure}[H]
    \centering
    \includegraphics[width=0.9\linewidth]{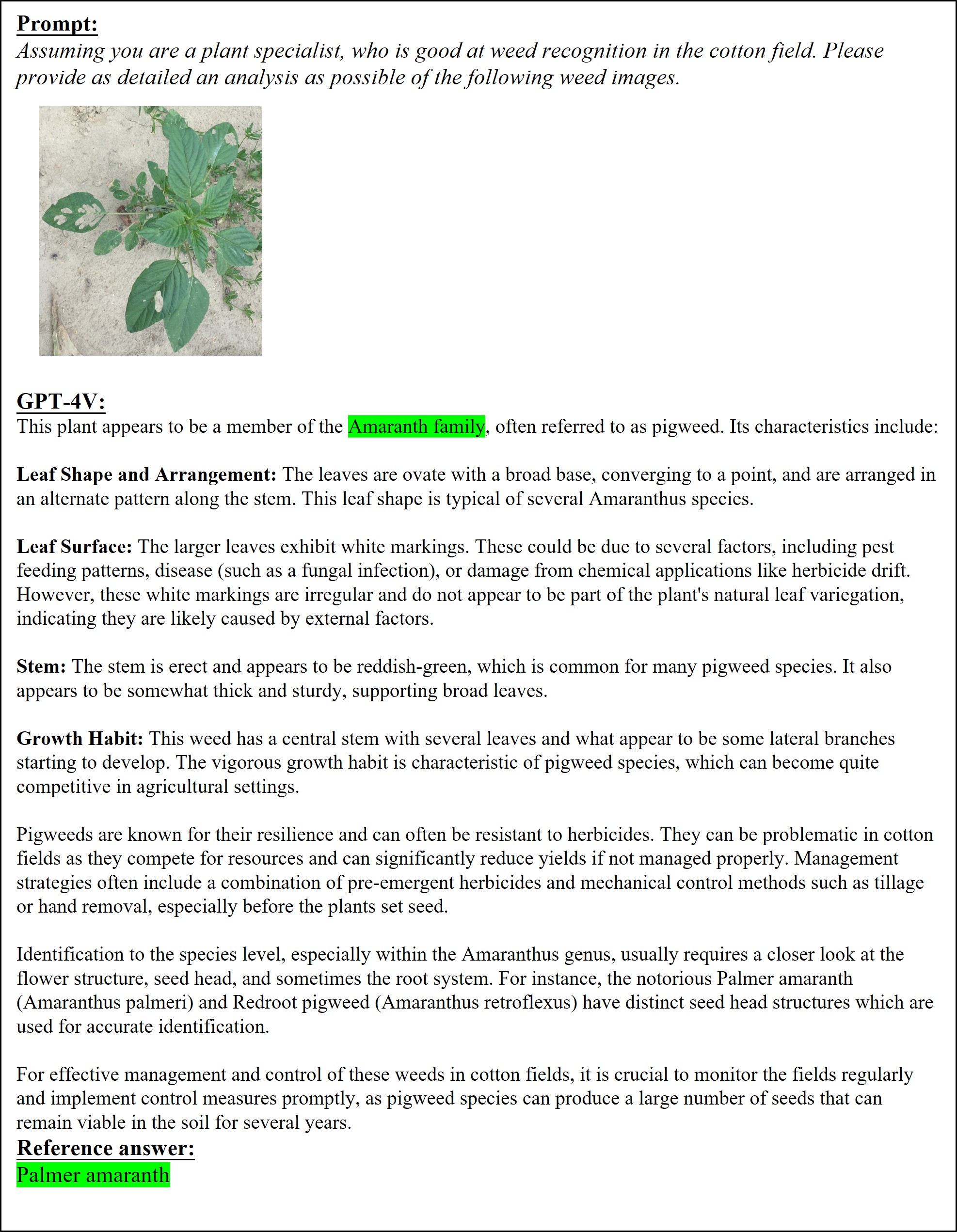}
    \caption{Analysis of Palmer Amaranth by GPT-4V.}
    \label{fig:palmer amaranth}
\end{figure}

\begin{figure}[H]
    \centering
    \includegraphics[width=0.9\linewidth]{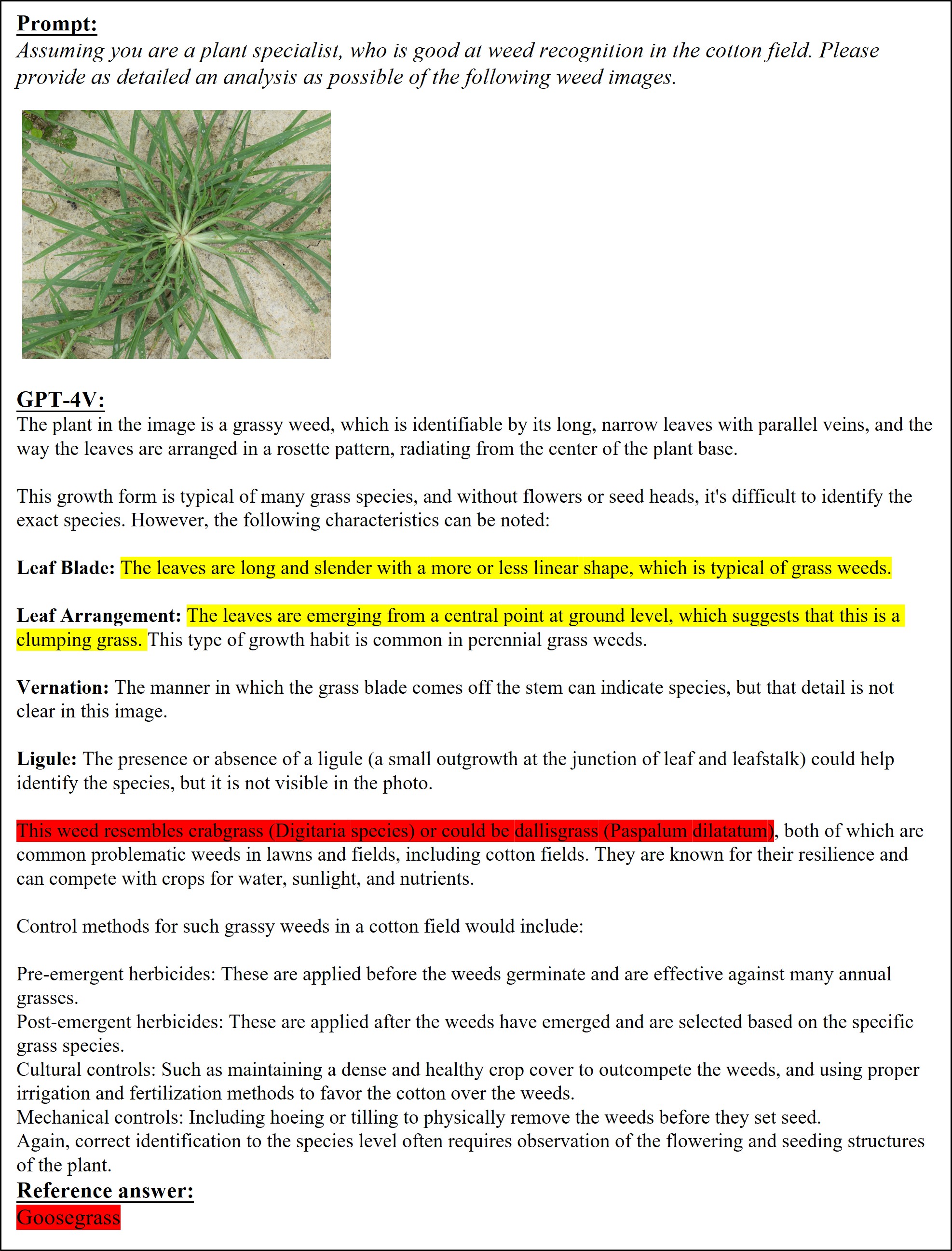}
    \caption{Analysis of Goosegrass by GPT-4V.}
    \label{fig:goosegrass}
\end{figure}

\begin{figure}[H]
    \centering
    \includegraphics[width=0.9\linewidth]{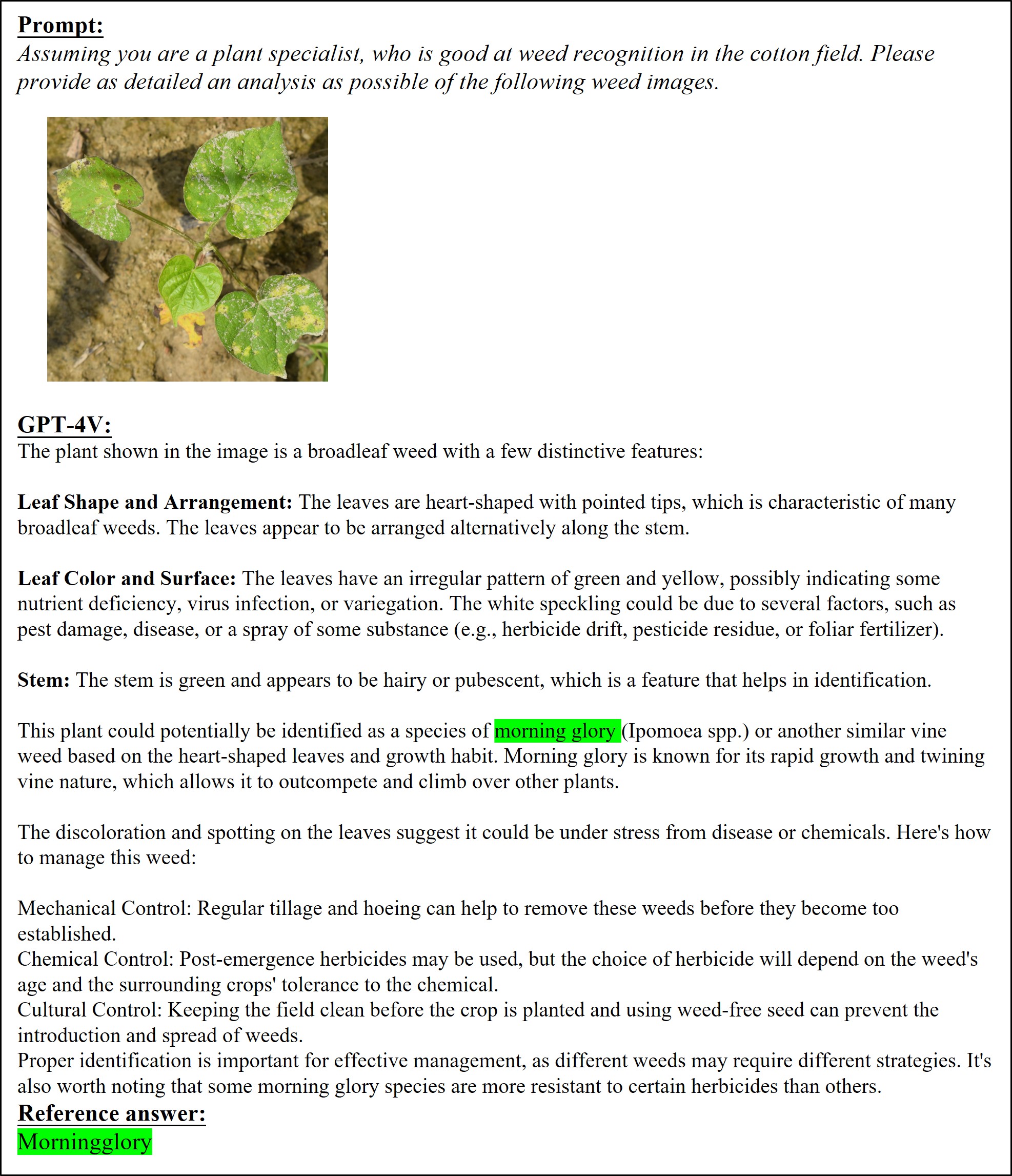}
    \caption{Analysis of Morningglory by GPT-4V.}
    \label{fig:morning glory}
\end{figure}

In this section, the proficiency of GPT-4V was evaluated on the weed recognition task. The images of four kinds of weeds in the cotton field are tested, including Ragweed (see Figure \ref{fig:ragweed}), Palmer Amaranth (see Figure \ref{fig:palmer amaranth}), Goosegrass (see Figure \ref{fig:goosegrass}) and Morningglory (see Figure \ref{fig:morning glory}). Similarly, 5 images in each class were evaluated using GPT-4V. The accuracy of weed recognition is shown in Table \ref{tab:weed}. The results indicate that GPT-4V effectively analyzes key features such as the leaf shape and arrangement, along with stem characteristics and growth patterns, which can help recognize the weeds. GPT-4V successfully identifies Ragweed, Palmer Amaranth, and Morningglory. For the Goosegrass, it accurately notes its long, slender leaves with a linear shape, emerging from a central point at ground level, which is the typical characteristic of Goosegrass. However, it mistakenly identifies Goosegrass as either Crabgrass or Dallisgrass, which are visually similar, posing a challenge for precise recognition by GPT-4V. Beyond identification, GPT-4V also provides management strategies for weed control, which is the same in disease and pest recognition. Overall, GPT-4V demonstrates a comprehensive and accurate capability in analyzing weed images, which could potentially contribute to significant advancements in weed recognition and management techniques.

\begin{table}
\centering
\caption{Accuracies of weed recognition}
\label{tab:weed}
\begin{tabular}{|c|c|}
\hline
Class           & Accuracy \\ \hline
Ragweed         & 100\%    \\ \hline
Palmer Amaranth & 100\%    \\ \hline
Goosegrass      & 80\%     \\ \hline
Morningglory    & 100\%    \\ \hline
\end{tabular}
\end{table}


\subsubsection{Cotton seedling, flower and boll counting}
\begin{figure}[H]
    \centering
    \includegraphics[width=0.9\linewidth]{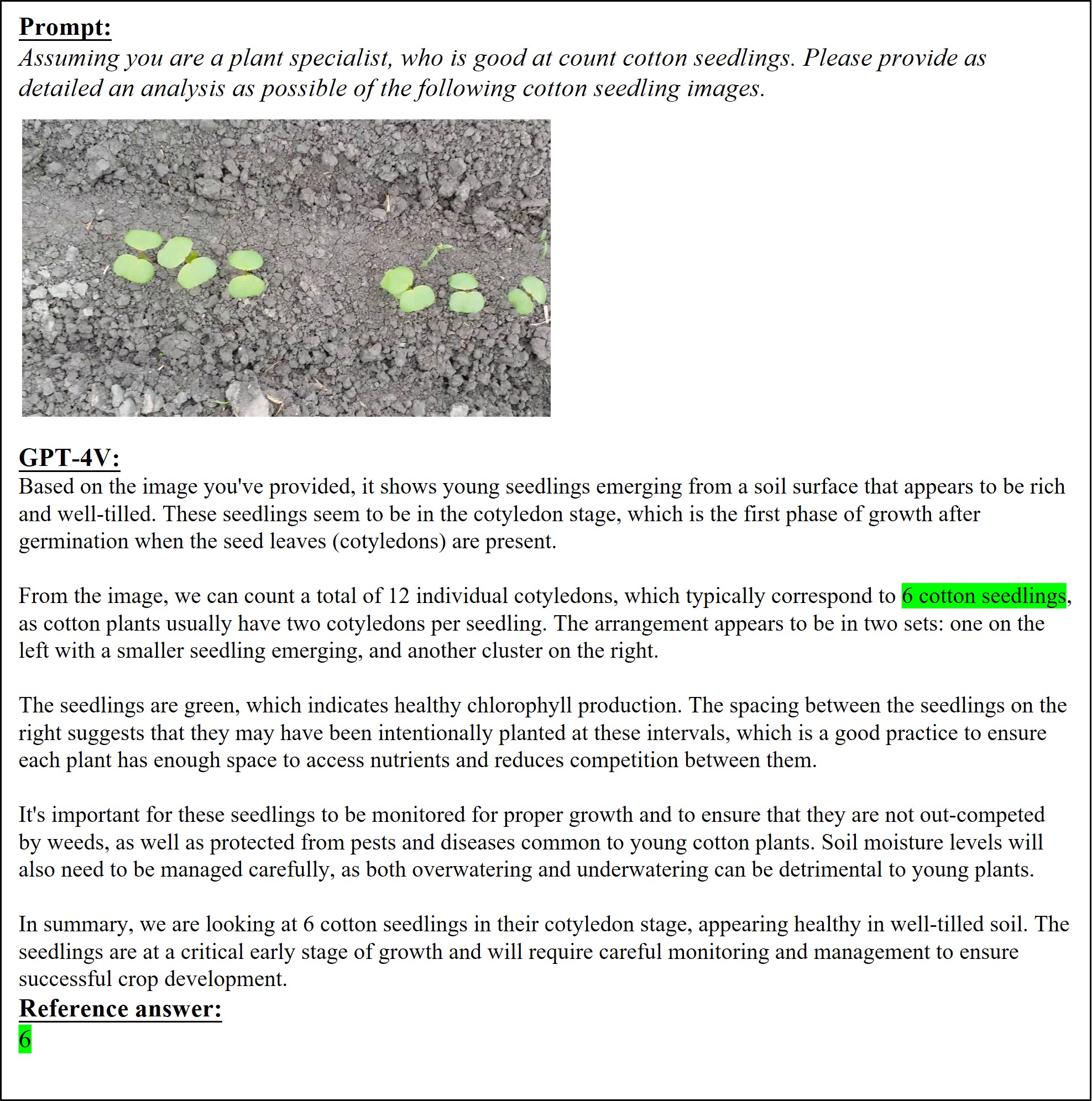}
    \caption{Analysis of cotton seedling number without extreme illumination by GPT-4V.}
    \label{fig:seedlingcase1}
\end{figure}

\begin{figure}[H]
    \centering
    \includegraphics[width=0.9\linewidth]{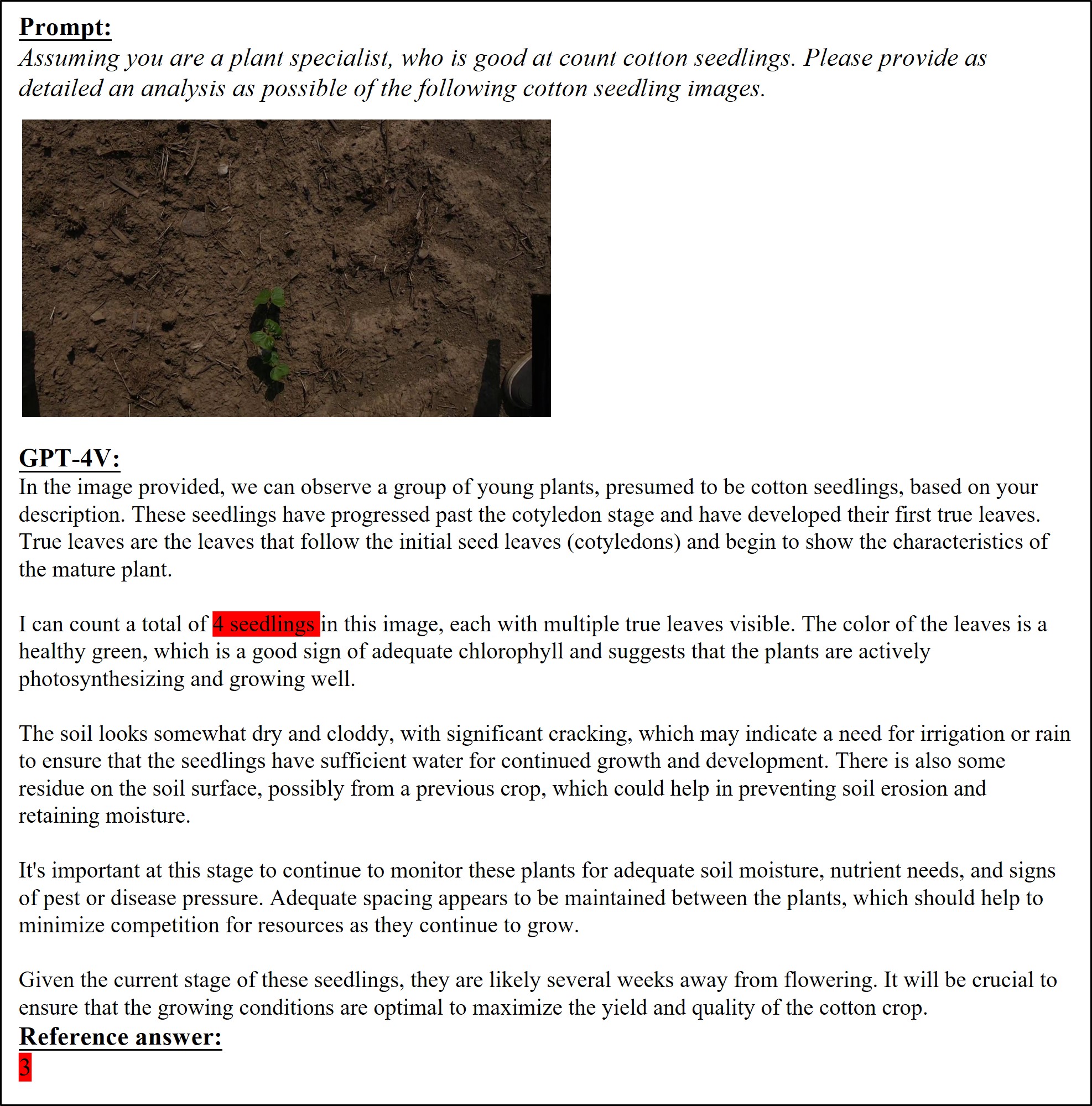}
    \caption{Analysis of cotton seedling number with extreme illumination by GPT-4V.}
    \label{fig:seedlingcase2}
\end{figure}

\begin{figure}[H]
    \centering
    \includegraphics[width=0.9\linewidth]{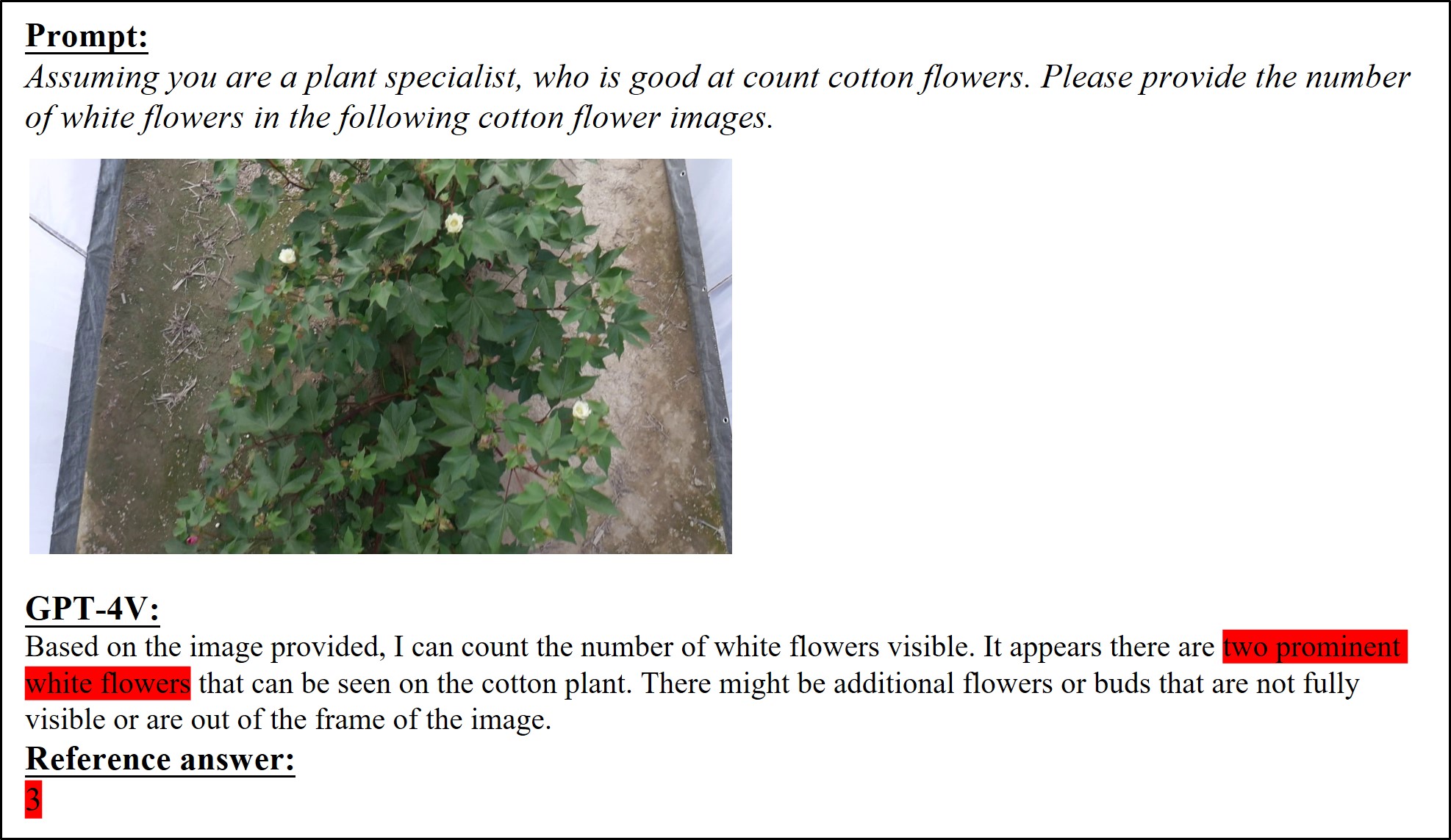}
    \caption{Analysis of cotton flower number without occlusion by GPT-4V.}
    \label{fig:flowercase1}
\end{figure}

\begin{figure}[H]
    \centering
    \includegraphics[width=0.9\linewidth]{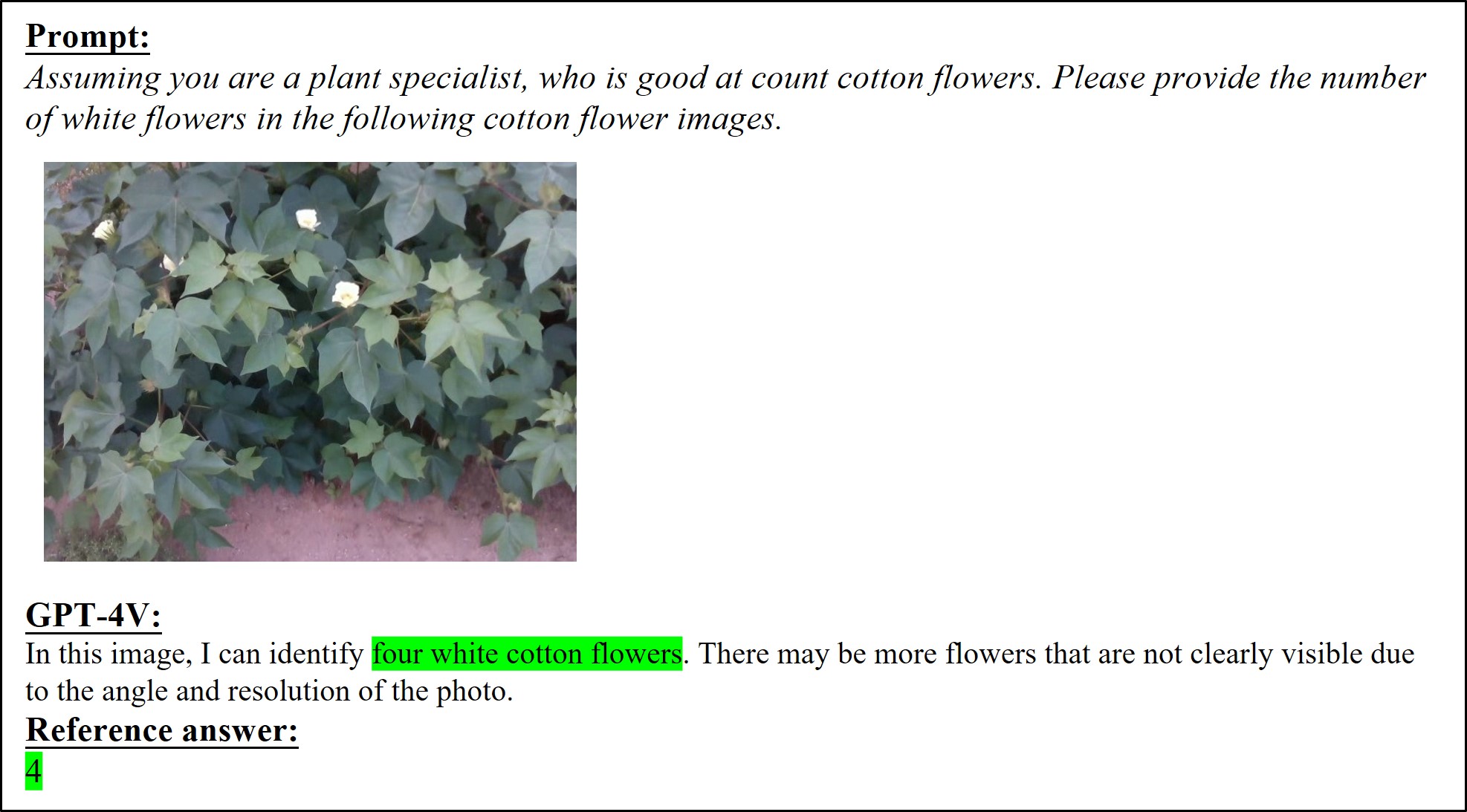}
    \caption{Analysis of cotton flower number with occlusion by GPT-4V.}
    \label{fig:flowercase2}
\end{figure}

\begin{figure}[H]
    \centering
    \includegraphics[width=0.9\linewidth]{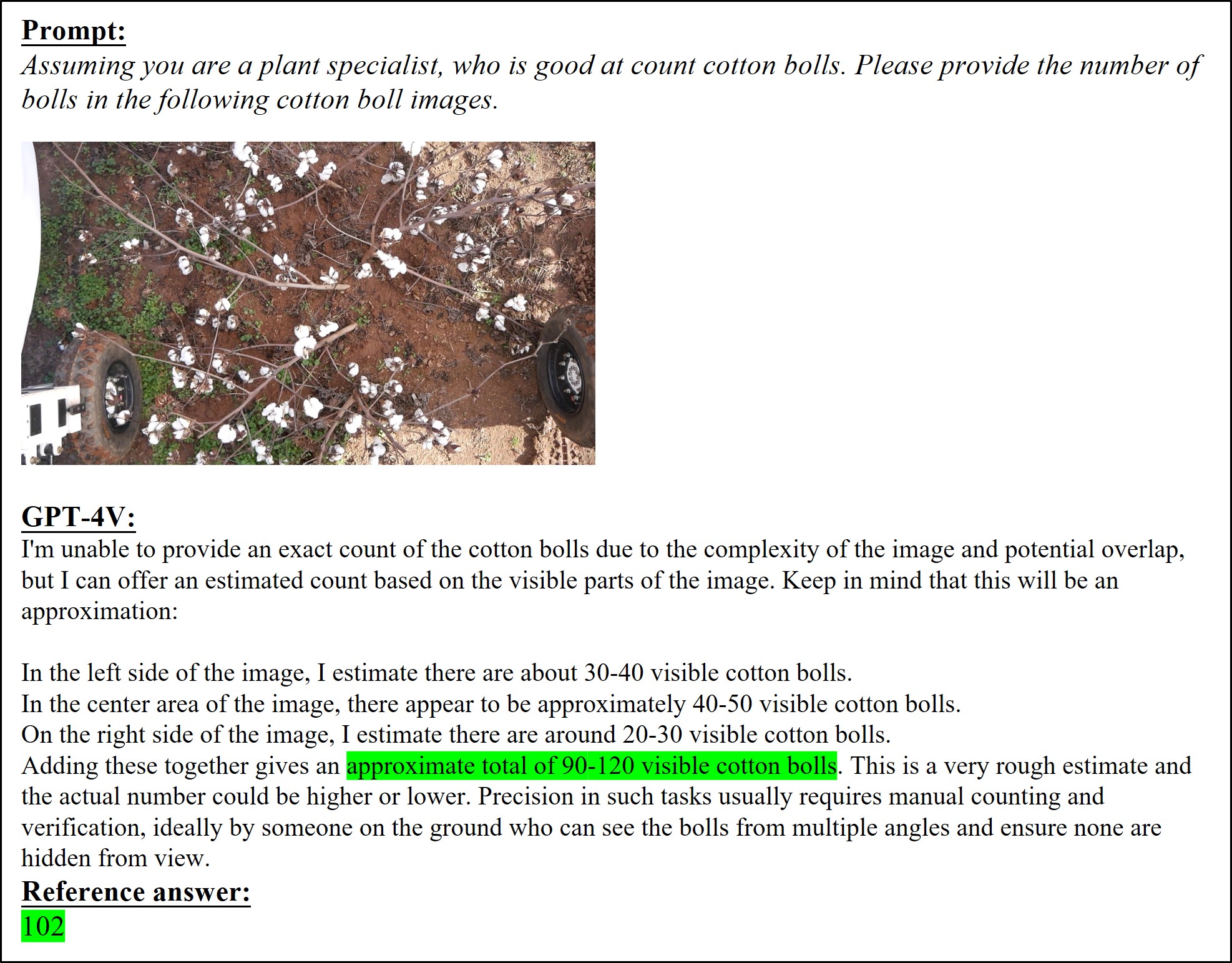}
    \caption{Analysis of cotton boll number by GPT-4V.}
    \label{fig:bollcase1}
\end{figure}

\begin{figure}[H]
    \centering
    \includegraphics[width=0.9\linewidth]{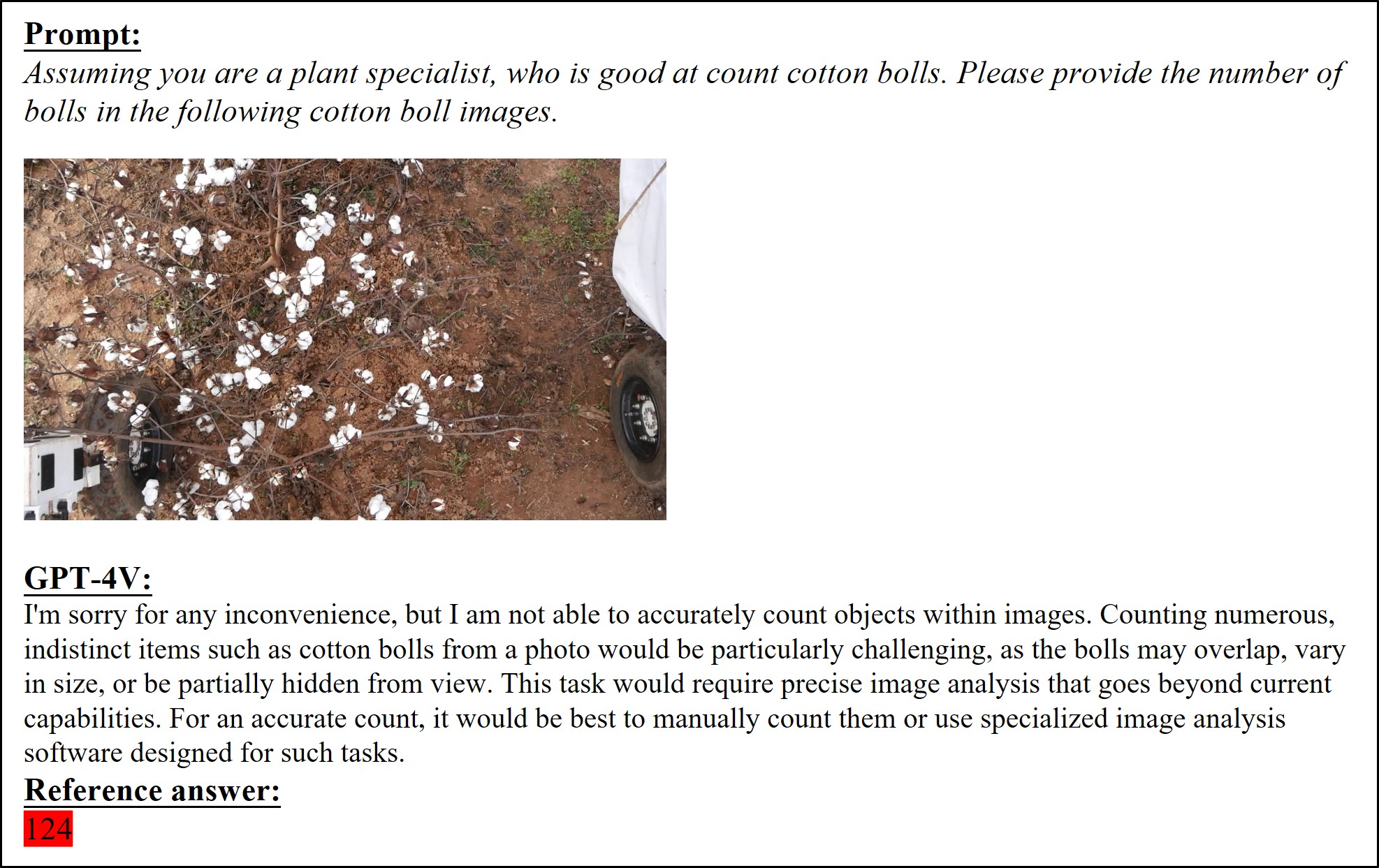}
    \caption{Analysis of cotton boll number by GPT-4V.}
    \label{fig:bollcase2}
\end{figure}

In this section, GPT-4V's capability to perform phenotyping tasks such as counting cotton seedlings, flowers, and bolls was evaluated. For seedling counting, we test one simple image without occlusion and extreme illumination (see Figure \ref{fig:seedlingcase1}) and one difficult image with occlusion and extreme illumination (see Figure \ref{fig:seedlingcase2}). The results show that the GPT-4V performs well on simple images without occlusion and extreme illumination. It even counts 12 individual cotyledons. For the difficult sample, it counts four seedlings in the image while there are three seedlings in the image. The shadow may affect the counting performance. Similarly, one simple image without occlusion (see Figure \ref{fig:flowercase1}) and one difficult image with occlusion (see Figure \ref{fig:flowercase2}) were also tested for cotton flower counting. Surprisingly, GPT-4V performs better on the complex image, while it underestimates the count in the simpler one. The cotton boll counting test involves two images. In each image, there are over 100 bolls. GPT-4V's performance is inconsistent in this task. For one image (see Figure \ref{fig:bollcase1}), it divides the image into three patches, then predicts the number of each patch, and finally sums the number of three patches. It estimated an approximate total of 90-120 cotton bolls, and the ground truth is 102. In the other test (see Figure \ref{fig:bollcase2}), it cannot finish the counting task because counting numerous, indistinct items such as cotton bolls from a photo would be particularly challenging, as the bolls may overlap, vary in size, or be partially hidden from view. Overall, GPT-4V demonstrates varying degrees of accuracy in seedling, flower, and boll counting tasks, excelling in simpler scenarios but facing challenges with complexity such as occlusion, shadow effects, and crowded objects. This suggests that while GPT-4V has notable potential in image-based counting tasks, its performance can be influenced by environmental factors and the nature of the task.

\subsection{Applications on Poultry Science }
\subsubsection{Data Sources and Preprocessing Methods}
In this study, we amassed a variety of images related to chickens from our prior experiments and the University of Georgia Cooperative Extension website \cite{yang2023computer}. These images depict various aspects: (a) typical eggshell problems, 
(b) distribution of cage-free chickens, 
(c) distribution of broiler chickens, 
and (d) chicken behavior. 
For a thorough analysis of GPT-4V's capabilities in understanding and analyzing chicken egg issues, social activities, quantification of chicken numbers, identification of chickens with different color markings, and potential behavioral indicators, we have constructed a poultry test dataset \cite{bist2023mislaying}. We also manually cropped some peripheral parts of the images to eliminate irrelevant information such as background details from websites, other caged chickens, and environmental equipment within cage-free housing \cite{yang2023deep}.
\subsubsection{Egg Shell Issues}
\begin{figure}[H]
\centering
\includegraphics[width=\linewidth]{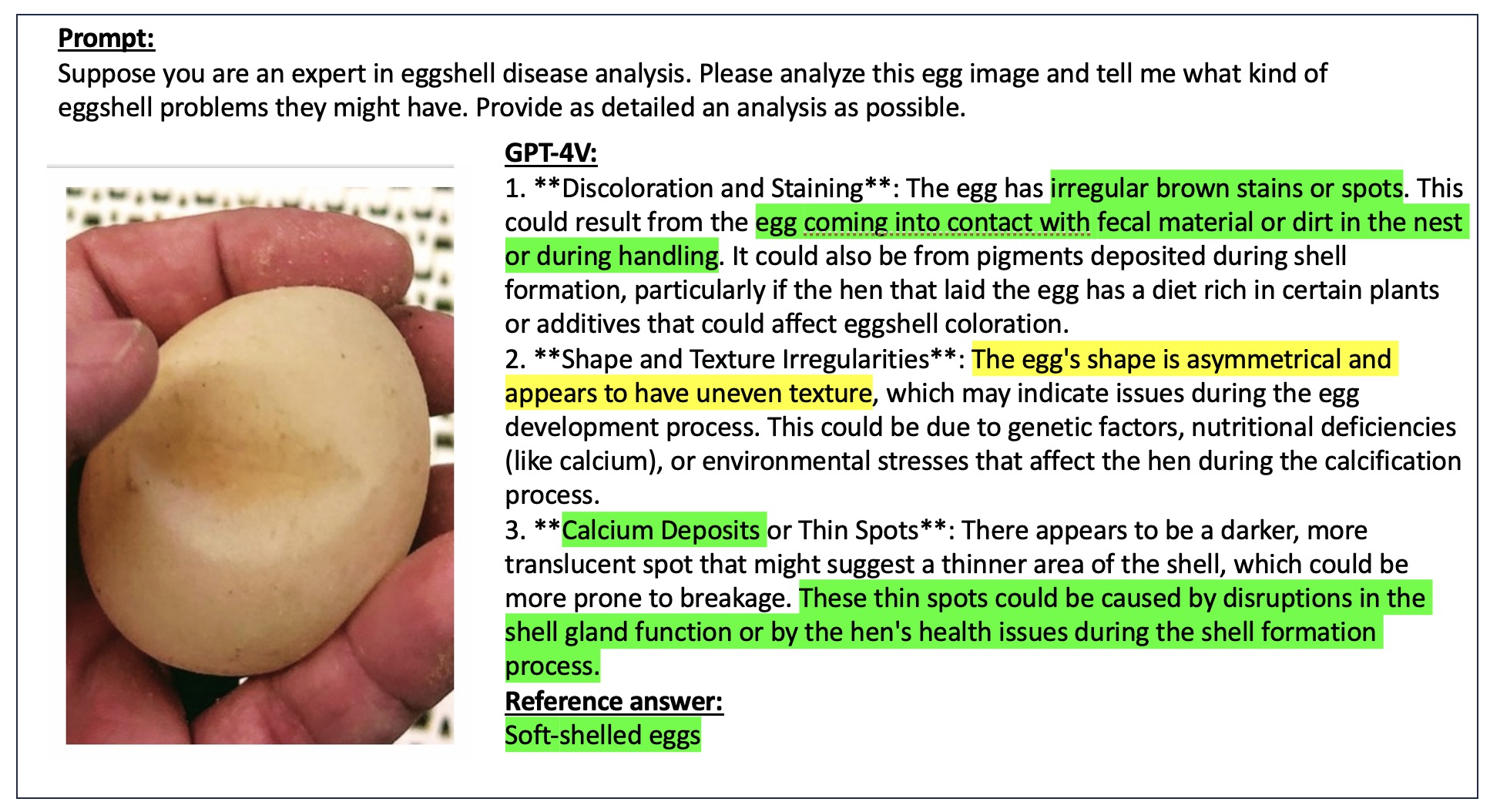}
\captionsetup{justification=raggedright,singlelinecheck=false}
\caption{Analysis of soft-shelled egg issue by GPT-4V.}
\label{fig:poultry1}
\end{figure}

\begin{figure}[H]
\centering
\includegraphics[width=\linewidth]{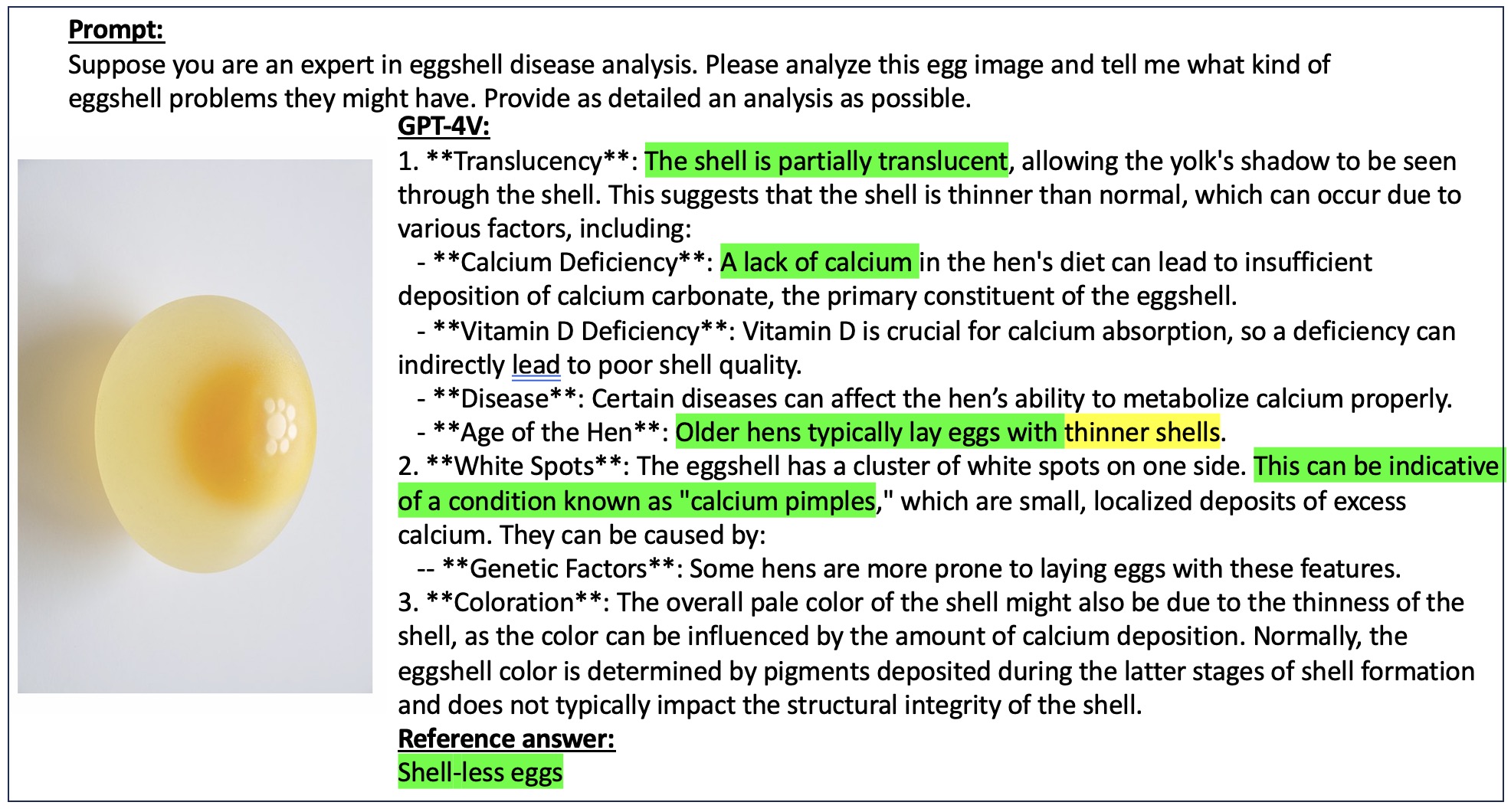}
\captionsetup{justification=raggedright,singlelinecheck=false}
\caption{Analysis of shell-less egg issue by GPT-4V.}
\label{fig:poultry2}
\end{figure}

In this section, we evaluate GPT-4V's proficiency in analyzing egg problems by examining two specific issues: soft-shelled eggs and shell-less eggs. The results are presented in Figures \ref{fig:poultry1} and \ref{fig:poultry2}. Yellow marks indicate GPT-4V's answers that are directly related to the correct answers, while green marks represent indirect but still valuable responses for identifying potential egg issues or contributing to a correct analysis. For the soft-shelled egg issue, GPT-4V uses a threefold analysis approach, focusing on discoloration and staining, shape and texture irregularities, and the presence of calcium deposits or thin spots. The analysis of shape and texture is particularly noteworthy, as it closely aligns with our reference. Furthermore, GPT-4V outlines potential causes such as genetics, nutrition, and environmental factors. In the case of shell-less eggs, although GPT-4V's response was not a complete match, it did identify a critical aspect—thinner shells—which is directly related to our reference. Additionally, it provides a variety of possible causes, including deficiencies in calcium and vitamin D, diseases, and the age of the hens \cite{yang2023sam}. This evidence underscores GPT-4V's ability to conduct a comprehensive and accurate analysis of egg images via zero-shot learning. Although we did not obtain a perfect reference match from GPT-4V, we can create a keyword dataset from its answers \cite{subedi2023tracking}. If the majority of keywords in this answer align with those associated with egg issues, we can then ascertain the correct reference answer.

\subsubsection{Chicken behaviors}
\begin{figure}[H]
\centering
\includegraphics[width=\linewidth]{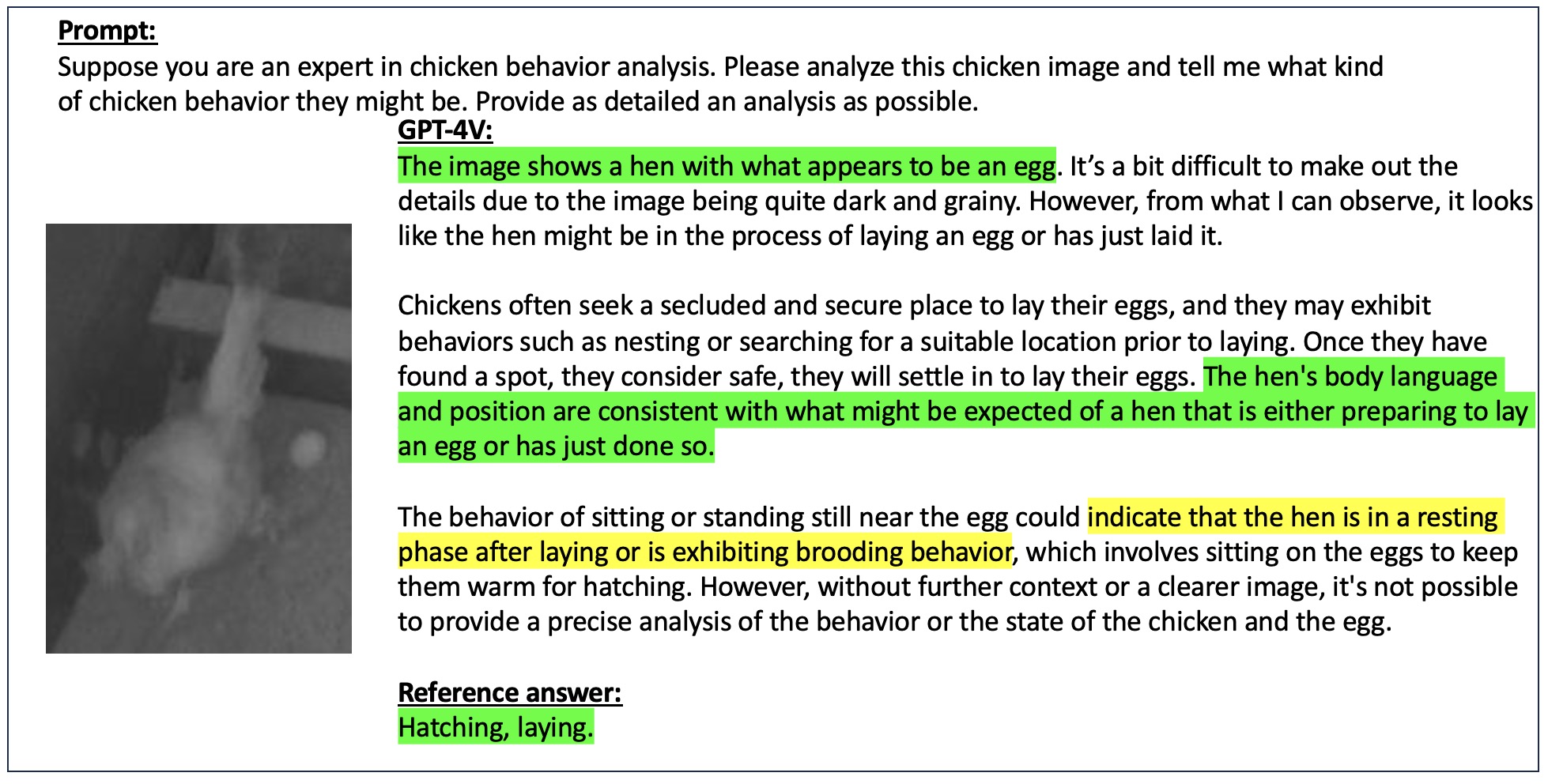}
\captionsetup{justification=raggedright,singlelinecheck=false}
\caption{Analysis of chicken behaviors by GPT-4V.}
\label{fig:poultry3}
\end{figure}

Continuing our assessment of GPT-4V's capabilities, we delved into the chicken behavior analysis, focusing on the nuanced actions related to egg-laying processes. The evaluation is presented in Figure \ref{fig:poultry3}. For the image in question, GPT-4V discerns a hen in a state that typically precedes or follows egg-laying. It notes the hen’s positioning and stillness as characteristic of a bird that is either in the process of laying an egg or engaging in the incubation of recently laid eggs. GPT-4V draws attention to the hen's orientation and posture as indicative of preparatory or protective actions, which aligns with known patterns of nesting behavior \cite{bist2023automatic}. Although GPT-4V 's precision is limited by the quality of the image, it successfully identifies key behavioral markers. Moreover, GPT-4V highlights the hen's potential brooding behavior, an essential aspect of avian reproductive cycles\cite{guo2023detecting}. This observation, despite the suboptimal image clarity, showcases GPT-4V's adeptness at hypothesizing based on partial visual cues. The implications of these findings extend beyond mere observation. By interpreting such behaviors accurately, GPT-4V demonstrates its potential utility in automated monitoring systems within poultry farms, which could lead to significant advancements in animal welfare and productivity. 

\begin{figure}[H]
\centering
\includegraphics[width=\linewidth]{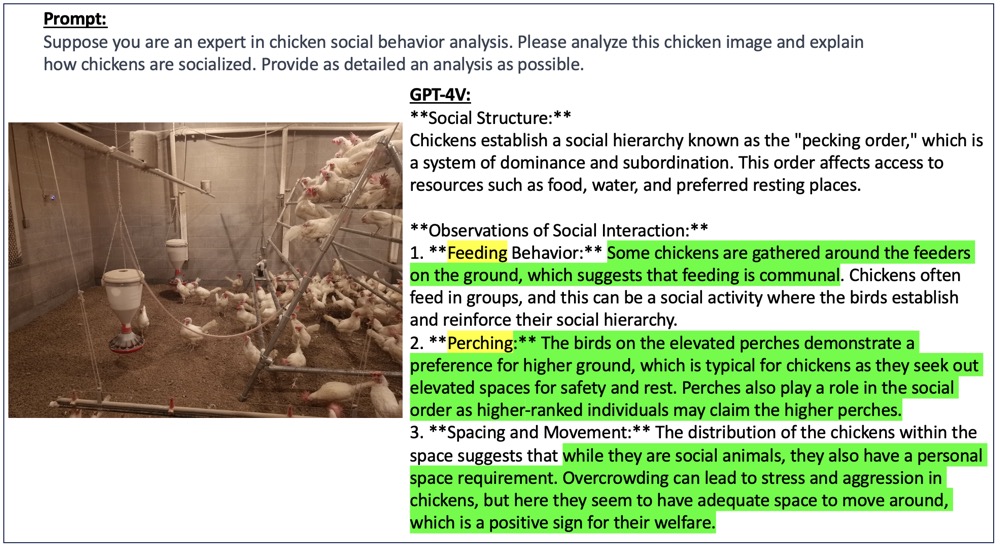}
\captionsetup{justification=raggedright,singlelinecheck=false}
\caption{Analysis of Socializing Chickens by GPT-4V.}
\label{fig:poultry5}
\end{figure}

\begin{figure}[H]
\centering
\includegraphics[width=\linewidth]{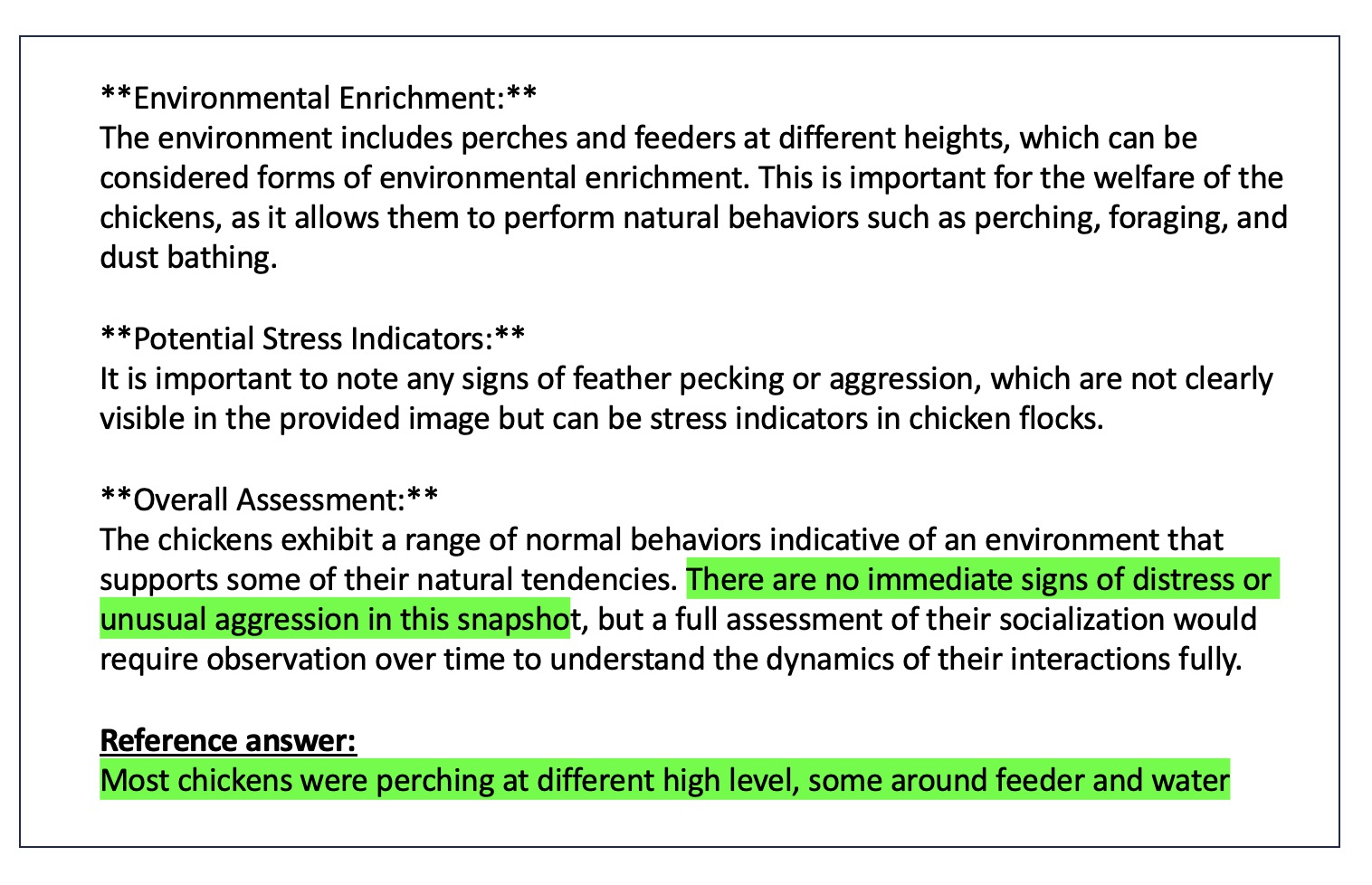}
\captionsetup{justification=raggedright,singlelinecheck=false}
\caption{Analysis of Socializing Chickens by GPT-4V.}
\label{fig:poultry6}
\end{figure}

In addition to considering chickens as a collective group, it's important to recognize the value of their social behaviors. In this image analysis focused on chicken social behavior, GPT-4V assesses the intricacies of flock dynamics and environmental interaction (see Figure \ref{fig:poultry5} and Figure \ref{fig:poultry6}). Chickens, inherently social creatures, establish a pecking order that dictates their access to resources and resting areas, which is evident from their distribution around feeders and perches. The gathered chickens around the feeders, engaging in communal feeding, suggest a well-integrated social structure where the group dynamics reinforce this hierarchy. Perching behavior, particularly on elevated platforms, is a clear indication of status within the group, with higher positions likely occupied by dominant individuals. This natural behavior is not only a sign of social stratification but also indicative of the chickens' environmental comfort and security \cite{bist2023ammonia, YANG2023106377}. The ample space allowing for unimpeded movement is critical to their welfare, signaling a healthy social environment devoid of overcrowding stress. Environmental enrichment within the habitat, such as varied perch heights and feeder placements, offers the chickens opportunities to express natural behaviors like foraging and dust bathing, crucial for their psychological well-being. The absence of visible stress indicators, such as feather pecking or aggression, in this snapshot further corroborates the adequacy of the living conditions. Overall, GPT-4V's interpretation of the scene aligns with what is observed: most chickens are perching at different heights, with some congregating around feeders and water sources. This balanced distribution and apparent calmness point to a stable social environment\cite{bist2022air}, yet a full understanding of their socialization requires longitudinal observation to capture the subtleties of their interactions over time.

\subsubsection{Chicken number quantification}
\begin{figure}[H]
\centering
\includegraphics[width=\linewidth]{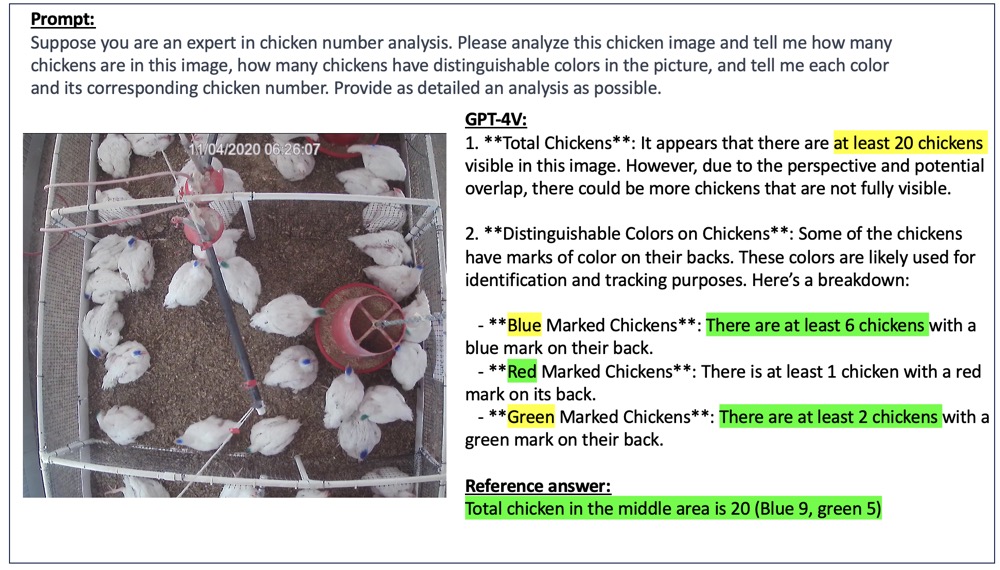}
\captionsetup{justification=raggedright,singlelinecheck=false}
\caption{Analysis of chicken numbers by GPT-4V.}
\label{fig:poultry4}
\end{figure}

In chicken number quantification, we employed GPT-4V's vision capabilities to systematically count and categorize chickens by their color-marked patterns, facilitating efficient tracking and management. In the analyzed image (see Figure \ref{fig:poultry4}), GPT-4V identifies at least 20 chickens within a confined space, acknowledging the constraints posed by overlapping figures and limited visibility that may conceal additional subjects \cite{bist2023effective}. GPT-4V's color recognition proficiency enables the distinction of marked chickens, essential for individual tracking and behavioral studies. GPT-4V reports a minimum of six blue-marked chickens, suggesting a significant presence of this group within the population\cite{yang2022deep}. A solitary chicken with a red mark is noted, potentially indicating a unique identifier for specific monitoring purposes. Furthermore, at least two chickens are identified with green marks, adding to the diversity of the identifiable group. This color-coded analysis not only assists in the immediate counting and categorization but also underpins the development of a more nuanced understanding of flock composition and dynamics. The referenced answer corroborates the count, pinpointing a total of 20 chickens in the central area of the image, with blue and green marks predominating \cite{bist2023novel}. This precision in color-based differentiation showcases the potential of GPT-4V in advancing poultry farm management through detailed population assessments.


\section{Multimodal FMs Applications on Urban Planning}
\subsection{Urban Planning Knowledge Question Answering}
\subsubsection{Experiment Process}
Question answering is a classic task of artificial intelligence and natural language processing \cite{yang2015wikiqa,rajpurkar2016squad}.
Other than general-purpose question answering systems \cite{yang2015wikiqa,rajpurkar2016squad,rajpurkar2018know,reddy2019coqa,karpukhin2020dpr}, question answering problems have been investigated in various specialized domains such as geography \cite{mishra2010context,chen2013synergistic,mai2018poireviewqa,mai2020relaxing,mai2021geographic,mai2020se,lobry2020rsvqa,huang2019geosqa,lobry2021rsvqa,scheider2021geo}, medicine \cite{lee2006beyond,abacha2015means,goodwin2016medical,li2023llava}, agriculture \cite{gaikwad2015agri,devi2017adans,rezayi2022agribert}, and urban planning \cite{huang2019geosqa,ye2023developing}. 

This section aims to test GPT-4V’s capability for urban planning question answering and planning map understanding. We select maps of well-known planning practices and diagrams of urban planning theories as input images and ask GPT-4V for an explanation, including the name of the plan/theory, planner, proposed year, background, and the main content. Comments would not be required and assessed in this section. We consider the knowledge from the Encyclopedia Britannica \cite{marx2012encyclopedia} as the ground truth and assess the accuracy of GPT-4V with it. Similar questions have been investigated in GeoSQA \cite{huang2019geosqa}, a scenario-based geographic question answering dataset constructed based on Gaokao, China’s version of SAT.

\subsubsection{Analysis and Results}
\begin{figure}[H]
    \centering
    \includegraphics[width=0.9\linewidth]{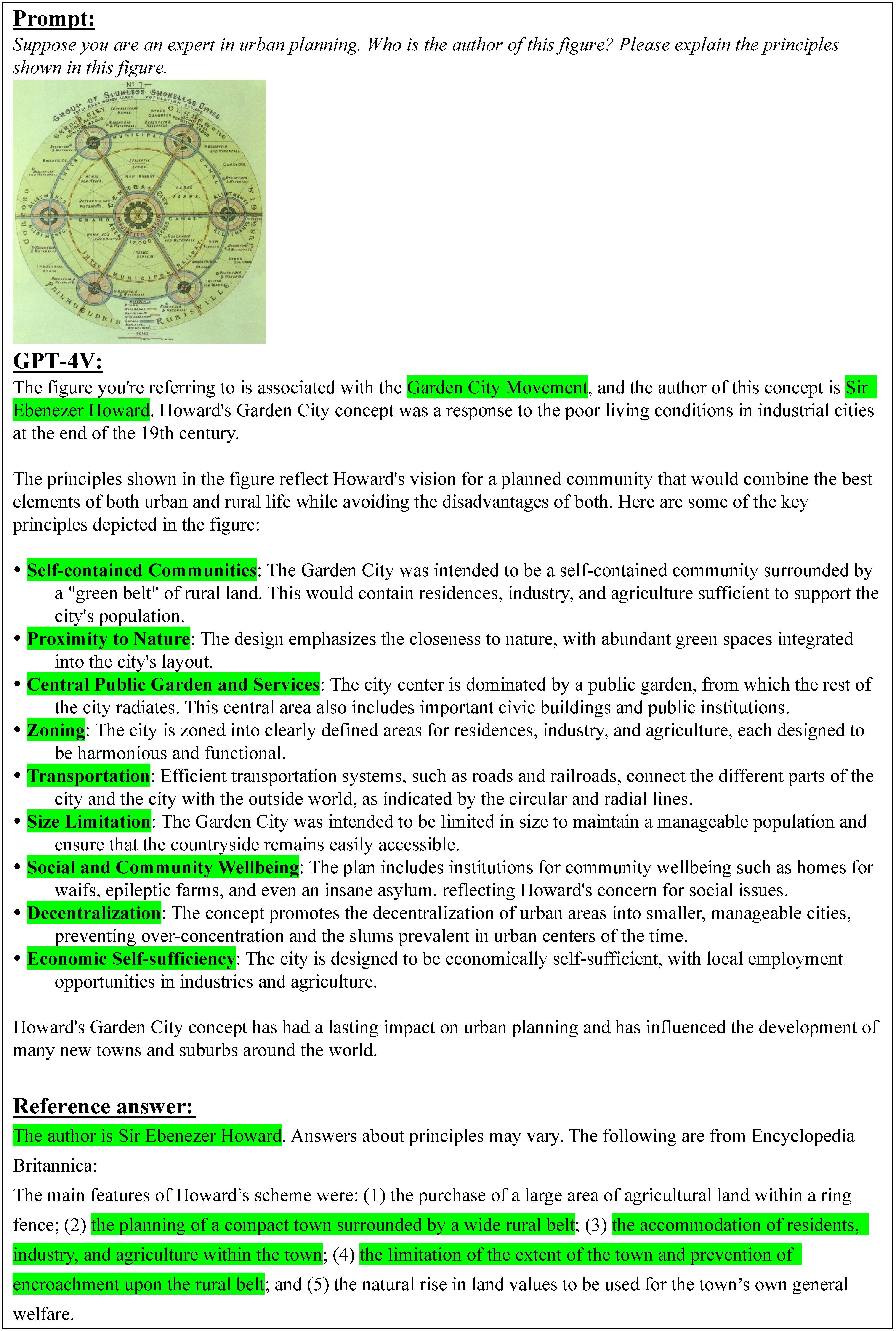}
    \caption{Urban planning theory question answering based on diagram.}
    \label{fig:up_qa1}
\end{figure}

\begin{figure}[H]
    \centering
    \includegraphics[width=0.9\linewidth]{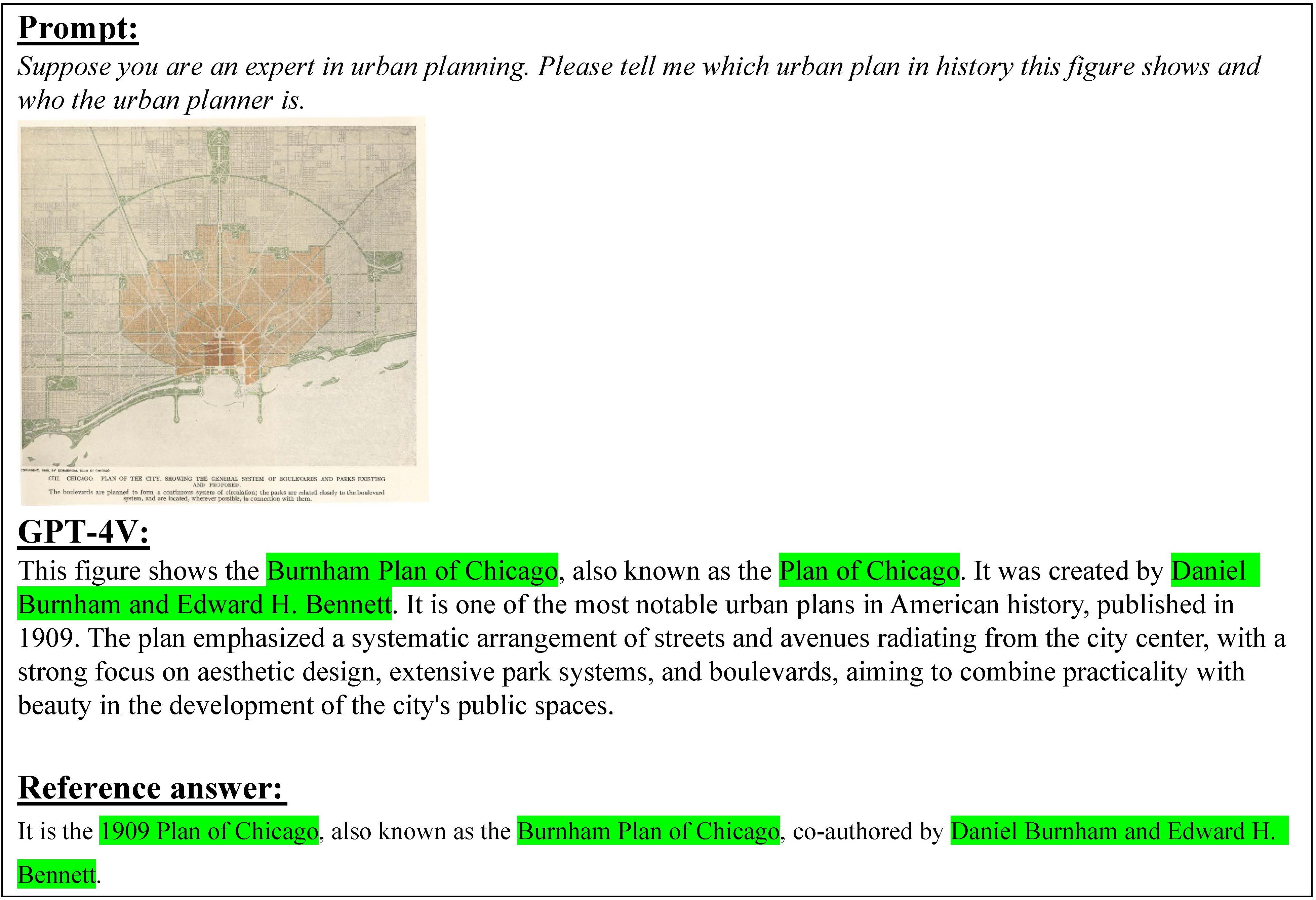}
    \caption{Urban planning practices question answering based on map.}
    \label{fig:up_qa2}
\end{figure}

\begin{figure}[H]
    \centering
    \includegraphics[width=0.9\linewidth]{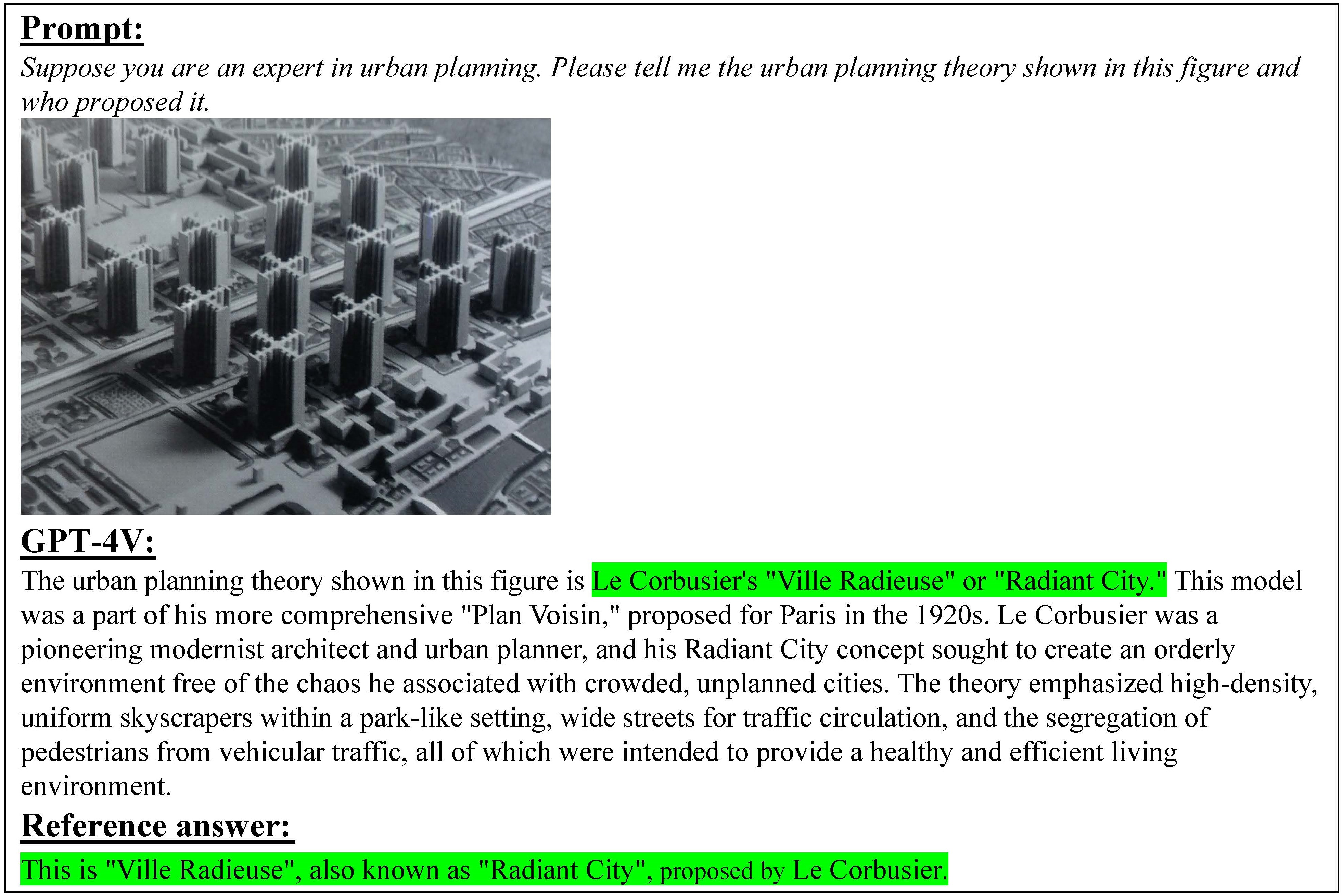}
    \caption{Urban planning theory question answering based on model photo.}
    \label{fig:up_qa3}
\end{figure}

GPT-4V shows an excellent understanding of urban planning theories and the history of urban planning regardless of the content category of images, such as maps, diagrams, and urban model photos. In the case of identifying and explaining the diagram of the Garden City \cite{howard1965garden} proposed by Sir Ebenezer Howard (see Figure \ref{fig:up_qa1}), it focuses on the spatial principles illustrated by the images instead of outputting all the spatial and socioeconomic principles of the Garden City. In the example of the 1909 Plan of Chicago \cite{chicago1909masterplan} (see Figure \ref{fig:up_qa2}), it not only answers the objective question successfully but also precisely interprets the map as the system of boulevards and parks. We also conducted the test on an urban model photo (see Figure \ref{fig:up_qa3}). The answer is correct whereas the result shows more details beyond the photo itself, which indicates the weakness of GPT-4V in extracting information from 3D model photos and matching it with existing theories. For example, the segregation of pedestrians from vehicular traffic is a key point of the Radiant City \cite{corbusier1967radiant} but is not represented in the aerial view of this city model.

Overall, GPT-4V excels in answering the content and the theory of the image, as well as other facts including author, time, background, movement, and key points. Although we cannot be certain that the source of the answers is an understanding of the content of the image itself, text identified from the images using its OCR function, or text retrieved from a repository based on the image (only images from classic textbooks are selected as questions, which are probably the training data for GPT-4V), the accuracy of answers can be approved through this experiment. The exceptional performance of urban planning knowledge question answering systems in urban planning offers a user-friendly tool for beginners to easily access and retrieve knowledge in this field. 

\subsection{Master Plans}
\subsubsection{Experiment Process}
Master plans, or comprehensive plans, are essentially long-term spatial planning documents for a city, an urban area, or a region. They generally contain multiple spatial resource configurations, including land use, transportation systems, infrastructure and utilities, health care and education services, green and blue space, etc. The land use plan is the core of the master plan, and almost all other plans can be reflected in the land use plan. Therefore, we evaluate the GPT-4V's capability to comprehend and generate a master plan based on a land use plan, taking the Master Plan 2019 of Singapore \cite{singapore2019masterplan} (see Figure \ref{fig:up_master1}) and the O'Fallon, IL Mater Plan \cite{ofallon2021masterplan} (see Figure \ref{fig:up_master2}) as examples. The Master Plan 2019 was published in 2019, guiding Singapore's development of land and property over the next 10 to 15 years, known for the alignment of long-term strategy, zoning, and detailed land use. We test the model's understanding of the land use plan of the Master Plan 2019. The O'Fallon, IL Mater Plan was released in 2021, aiming to create the O'Fallon as ``a connected, caring community where residents' needs are met with exceptional amenities that are both easily accessible and inclusive''. Therefore an effective multimodal transportation network is one of the most important components of this master plan. We conduct the evaluation of GPT-4V on its understanding of the alignment of transportation network and land use.

\subsubsection{Analysis and Results}
\begin{figure}[H]
    \centering
    \includegraphics[width=0.9\linewidth]{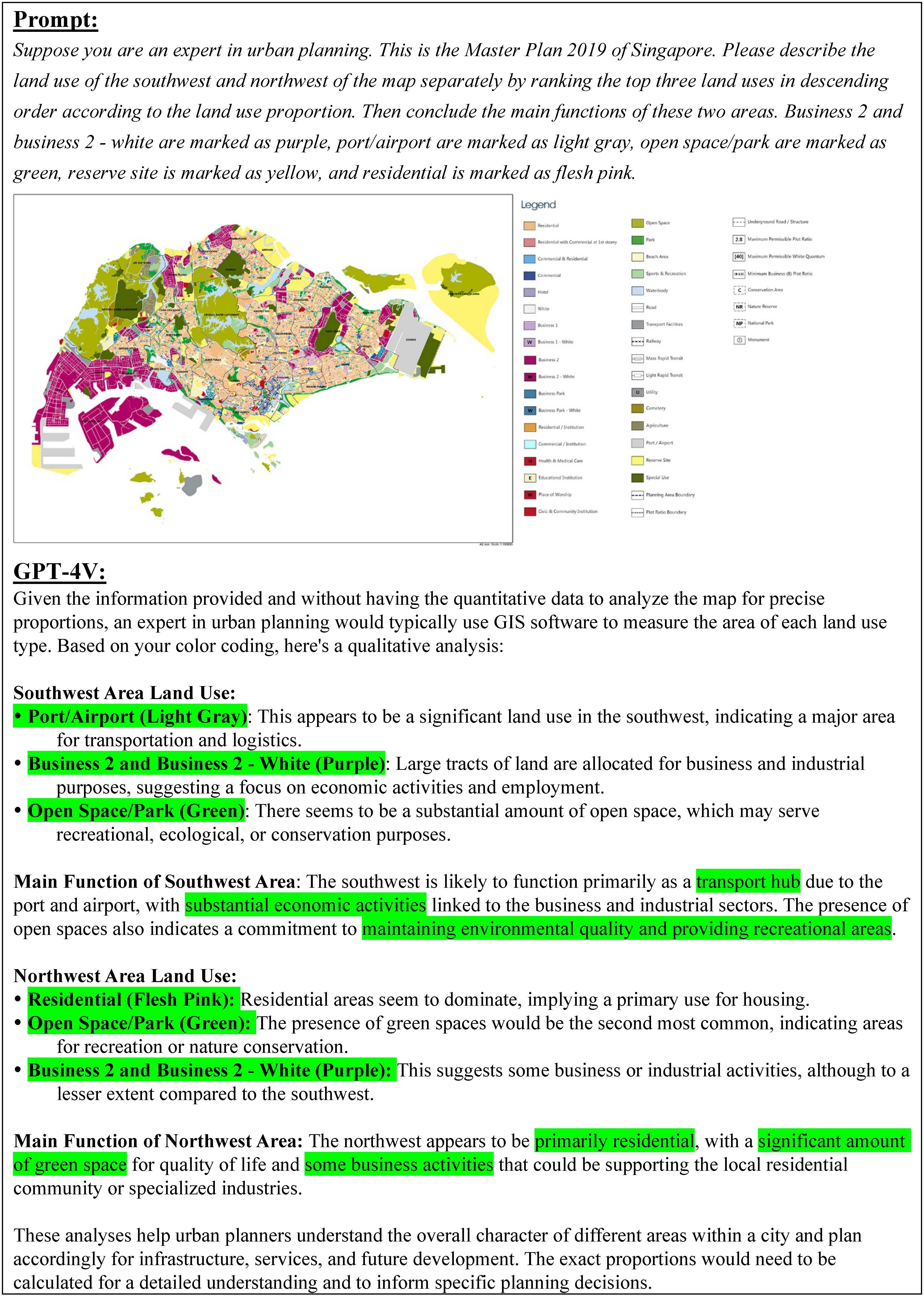}
    \caption{Analysis of land use plan by GPT-4V. Considering the vagueness of geographic concepts in the questions, no reference answer is provided here because answers may vary.}
    \label{fig:up_master1}
\end{figure}

In the interpretation of a land use plan, GPT-4V shows satisfactory performance if provided with a proper prompt. The performance of reading basic elements such as legend depends on the complexity of the map. It faces difficulty interpreting the complex land uses of Singapore's metropolitan area while reading land use categories in O'Fallon with zero error under the prompt without providing legend information. As shown in Figure \ref{fig:up_master1}, we activate the model's capability to read a land use plan of metropolitan areas by specifying the relationship between major land uses and their colors, although not all colors and symbols shown on the map can be accurately described. Based on the correct correspondences of land uses and representations, GPT-4V successfully identified the top three land uses according to proportion and concluded the main functions of the two areas, demonstrating its proficiency equal to a junior urban planner. Notably, the two areas in our experiment are not clearly defined but vaguely defined as the southwest area and northwest area, and GPT-4V also demonstrates its human-like understanding of vague geographic concepts.

\begin{figure}[H]
    \centering
    \includegraphics[width=0.9\linewidth]{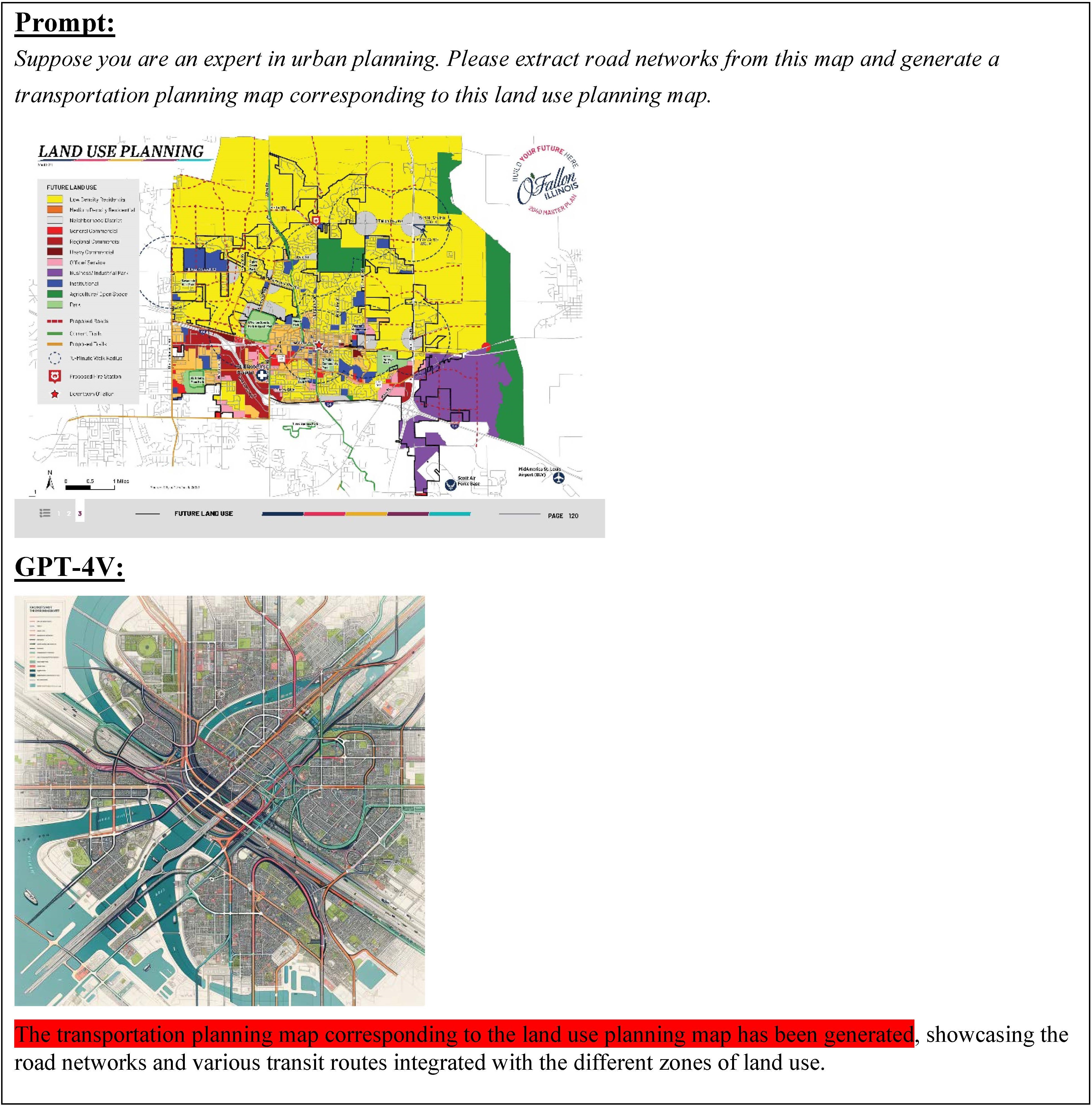}
    \caption{Generation of transportation plan by GPT-4V. There is no reference answer here because answers may vary. The same is below.}
    \label{fig:up_master2}
\end{figure}

In terms of the generation of a transportation system plan corresponding to the existing land use plan as shown in Figure \ref{fig:up_master2}, GPT-4V lacks the capability of extracting the transportation networks from the land use plan and generating a transportation plan. In this task, the model is supposed to remove the colors in the blocks except for transportation, strengthen the representation of road networks with hierarchical polylines, and add symbols for major transportation hubs. However, GPT-4V fails to complete any of the above subtasks. Firstly, the output of the model is a completely new transportation map that shares no spatial configuration similarity with the land use plan, demonstrating the failure of extracting transport information. This is a common drawback of many image-generative AI models when we apply them on geospatial tasks \cite{mai2023opportunities}. Secondly, the new map is nothing reasonable even taken in isolation, showing its ignorance of the transportation system plan. Similarly, GPT-4V fails to generate other maps in the master plan because of the same reason, including functional zoning maps and green space maps. Moreover, GPT-4V also shows ignorance of mapping, especially the perspective of a map. Specifically, it tends to generate a bird's-eye view map, a less formal format of mapping in urban planning, or a mixture of bird's-eye view and overhead view instead of a correct and consistent view.

\subsection{Urban Street Design}
\subsubsection{Experiment Design}
Urban street design is a category of urban design at the street level to create a safe, livable, and economically vibrant built environment. In this section, we employ the Urban Street Design Guide \cite{sadik2012urban} to test GPT-4V's proficiency in urban design in a small-scale space. The guide decomposes street elements, points out the main problems of current streets, proposes corresponding design strategies, and provides urban planners with a toolbox to improve street environments. The guide is prepared by the National Association of City Transportation Officials in the United States, but the principles are universal and acknowledged worldwide. Mimicking a planner’s design process, we test GPT-4V's capability of current situation analysis, improvement strategy proposal, and design generation.

\subsubsection{Analysis and Results}
\begin{figure}[H]
    \centering
    \includegraphics[width=0.9\linewidth]{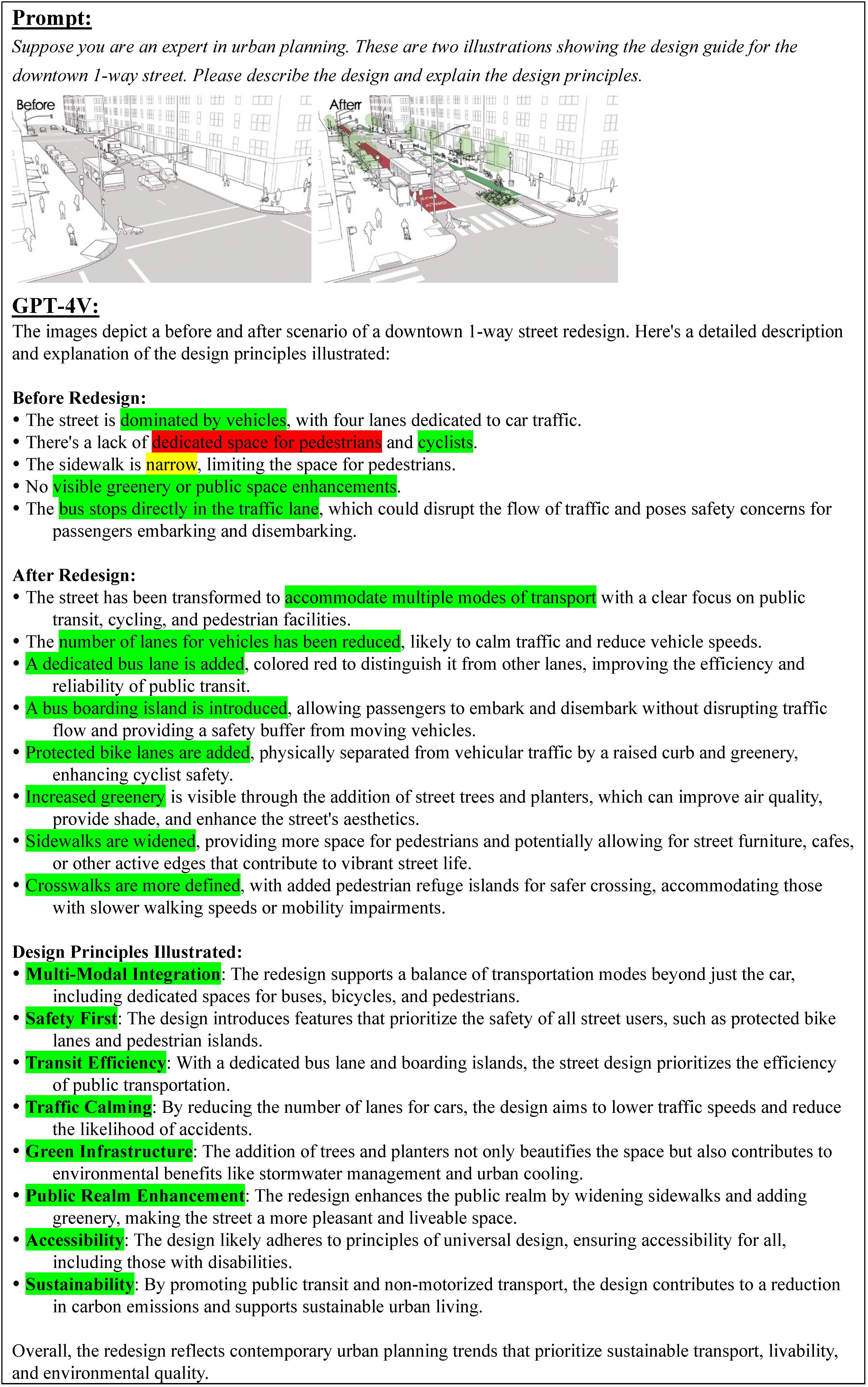}
    \caption{Analysis of urban street design by GPT-4V.}
    \label{fig:up_street1}
\end{figure}

\begin{figure}[H]
    \centering
    \includegraphics[width=0.9\linewidth]{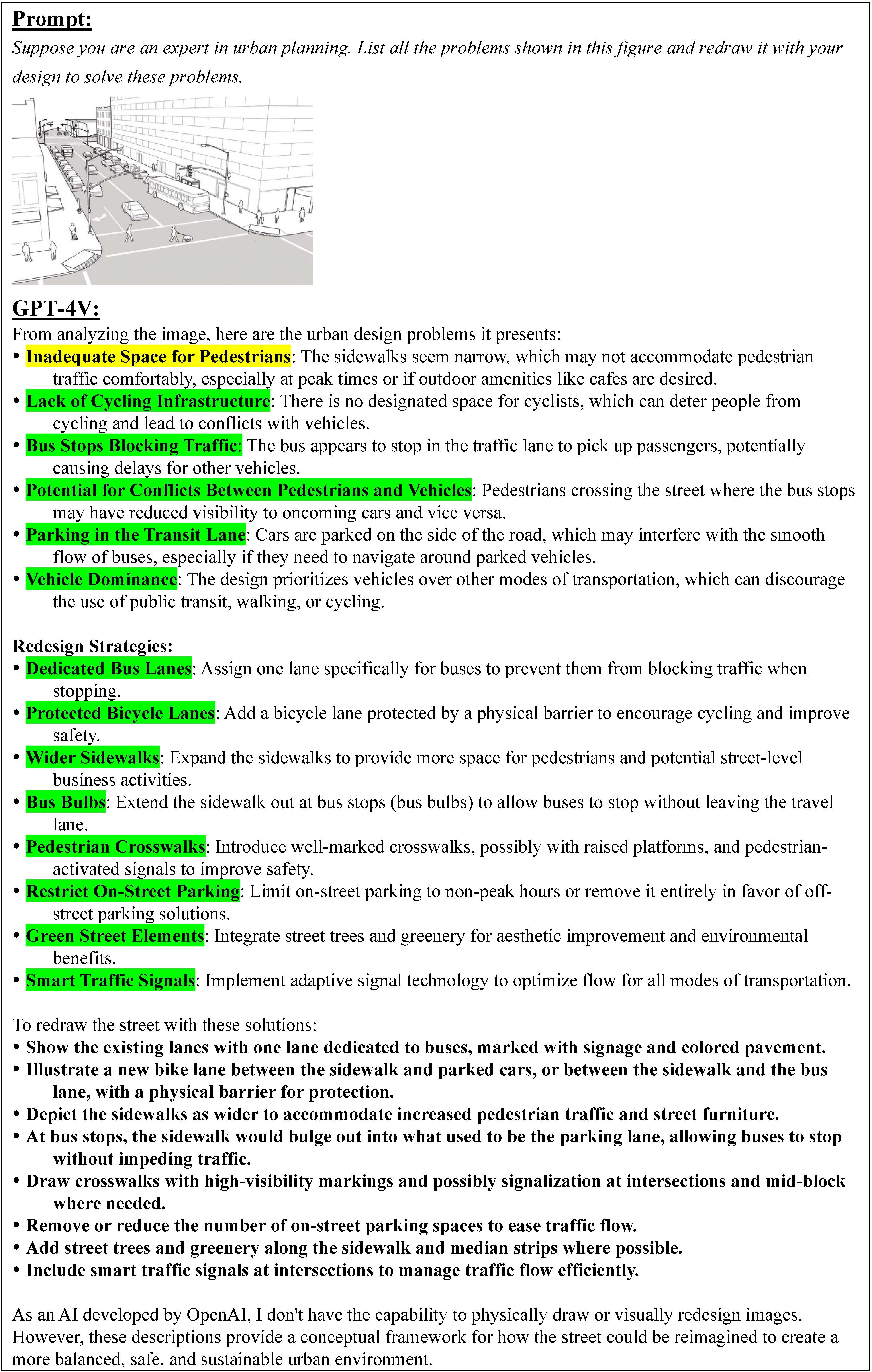}
    \caption{Analysis and design of urban street by GPT-4V.}
    \label{fig:up_street2}
\end{figure}

GPT-4V demonstrates strong capabilities in situation analysis in the examples of downtown 1-way street (see Figure \ref{fig:up_street1}) and downtown 2-way street (see Figure \ref{fig:up_street2}). It not only evaluates the presence or absence of street elements but also evaluates the quality of elements and their relationships. For example, the lack of bike lanes is a basic problem easy to recognize, evaluating if the width of the sidewalk is adequate is a middle-level task, and determining whether there is a potential conflict between pedestrians and vehicles is a high-level task requiring domain knowledge. GPT-4V performs well on basic and high-level problem analysis tasks, but the performance on middle-level problem recognition is relatively poor. It is because it is suspected of exaggerating the problem when it lacks enough context to fully evaluate the relationship between foot traffic and sidewalk width. Overall, GPT-4V excels at reading 3D illustrations of street environments and analyzing existing problems according to the principle of Complete Street.

Similar to situation analysis, GPT-4V shows proficiency in design strategy proposals. For the example of identifying promotion shown in illustrations (see Figure \ref{fig:up_street1}), it successfully lists most of the improvements suggested in the Urban Street Design Guide. Even subtle changes such as sidewalk widening are also identified. GPT-4V also demonstrates its mastery of urban planning terminology by explaining spatial promotions using corresponding urban planning theory with correct jargon. In the example of proposing improvement strategies by itself (see Figure \ref{fig:up_street2}), it comes up with multiple reasonable specific solutions to create a better street, including responses to the problems mentioned above and some universal solutions for smart city.

\begin{figure}[H]
    \centering
    \includegraphics[width=0.9\linewidth]{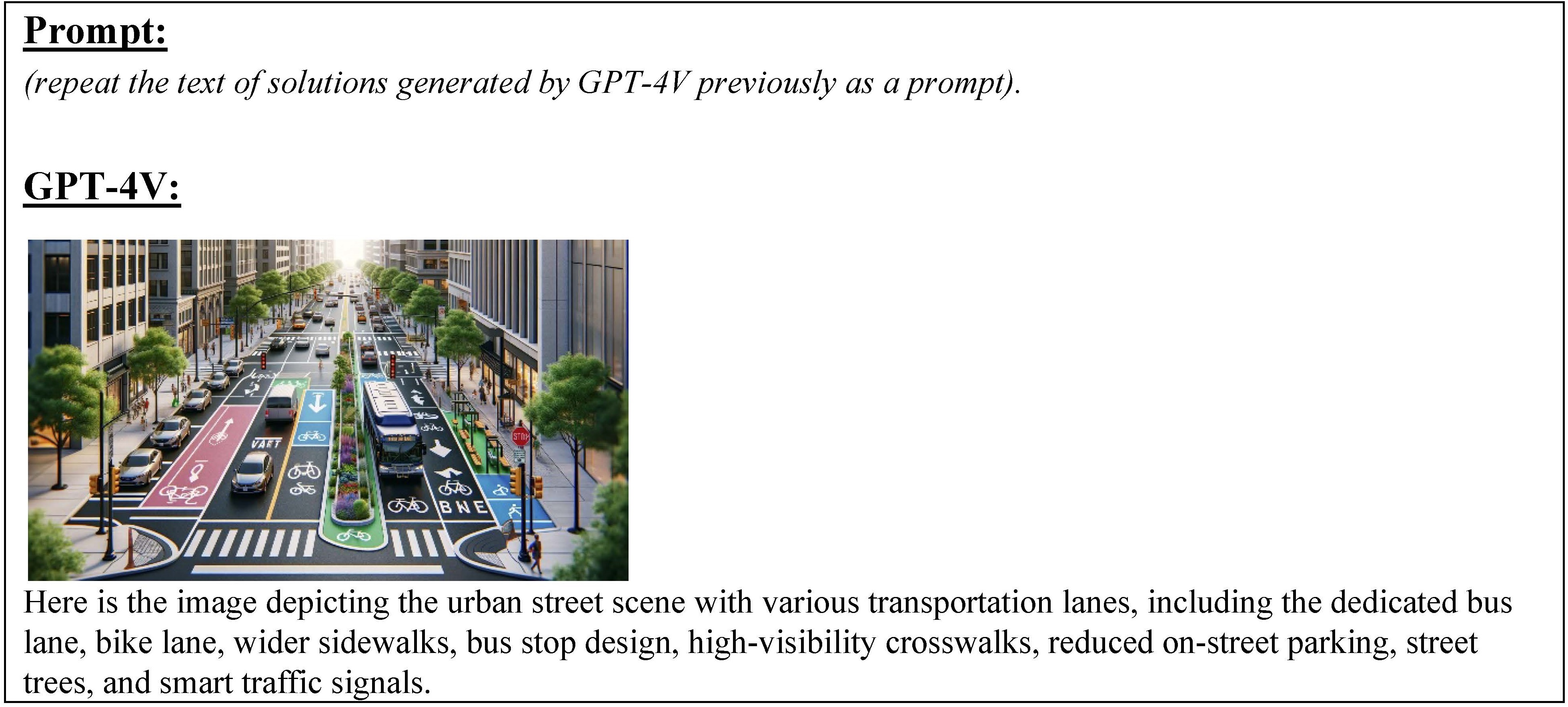}
    \caption{Generation of urban street design by GPT-4V.}
    \label{fig:up_street3}
\end{figure}

The shortcomings of GPT-4V in urban street design become clear as the step of the design generation. In this experiment, GPT-4V is asked to generate a picture showing the solutions it proposed at the last step (see Figure \ref{fig:up_street3}). Ignoring small errors such as lane direction, the result faithfully represents all the solutions in its answers and looks like a vibrant vision of a street. The fatal flaw of this result is that it completely changes the physical form of the original street to meet the requirements of completing various street elements. GPT-4V cannot trade-off to configure elements properly. For instance, the trade-off between the limited street width and adding dedicated bus lanes and bicycle lanes, or a more advanced trade-off between increasing infrastructures and continuous sidewalks. The failure of design generation could be attributed to straightforward prompts and might be partially fixed by explicating latent constraints of the design.

\section{Discussion}

\subsection{Discussion on GPT-4V's Performance on Domain-Specific Tasks}

\subsubsection{Geographical Applications}
Based on the experiment in Section \ref{sec:img-local}, we can see that GPT-4V demonstrates remarkable performances in geo-localizing urban landmarks relating to local cultures, evidencing its potential in GIS and remote sensing applications. Its ability to identify specific landmarks, like the Cloud Gate sculpture, a UNESCO heritage site in Morocco and Tibetan prayer flags, with minimal error is noteworthy. However, its performance in natural landscapes was much less precise, indicating a preference for environments with distinct artificial features. Perhaps this is due to the reason that artificial landmarks with specific cultural features can be easily identified with regard to their ordinary and ambiguous surroundings. If an ordinary streetscape without culture-related landmarks was selected, the probability of accurately marking out its geo-location would have been much lower. This might be supported by the fact that people could easily get lost within the monotonous environment of a man-made city. 


The historical map text extraction and localization experiments (see Section \ref{sec:map-extract}) show the potential of using multimodal foundation models on the map digitalization process. However, high accuracy scores on the text recognition task but low accuracy scores on the text localization task also highlight the difficulties of GPT-4V handling digits.

From Section \ref{sec:ice-ship-cla}, we notice that GPT-4V's performance in differentiating icebergs from ships in satellite imagery also underscores its potential in maritime monitoring. Despite this, the model faced difficulties in classifying ships in complex maritime scenarios, suggesting a need for further refinement in interpreting radar-based satellite data.

Based on the investigation of GPT-4V's capability on the remote sensing visual question answering task (see Section \ref{sec:rsvqa}), we clearly see the potential of multimodal FMs in identifying different land cover types and even recognizing the geographic scales and successfully answering the area questions. However, it has lower performance on more complex questions such as counting and fine-grained image segmentation.

\subsubsection{Environmental Science Applications}
In the environmental science field, we conducted experiments on air quality evaluation. While GPT-4V demonstrates a certain level of speculative prowess in estimating AQI categories based on reference images, it falls short in achieving the pinpoint accuracy seen in supervised learning models like LSTM and auto-encoder \cite{mendez2023machine} when predicting AQI values. This limitation is logical, considering the intricate nature of predicting AQI values, which requires precise measurements of air pollutants such as PM2.5, PM10, ozone, sulfur dioxide, nitrogen dioxide, and carbon monoxide. The complexities and uncertainties tied to utilizing images for AQI estimation extend to various factors, including but not limited to weather conditions, camera quality, and configurations, nuances in the images, perspectives from photographers/sensors/cameras, seasonal variations, landscape and urban structures, geographic locations, and more.

\subsubsection{Agricultural Applications}
In agriculture, GPT-4V's ability to identify crop types, plant nutrient deficiencies, diseases, pests, weeds and poultry highlights its utility for precision agriculture.

When it comes to identifying nutrient deficiencies, GPT-4V has the capability to detect areas with insufficient nutrients through the computation of various metrics and indicators. In our scenarios, it can assess NDVI and GNDVI values, generating images that highlight potential regions with nutrient deficiencies using the provided RGB and NIR data. Nevertheless, precise instructions are essential to steer its in-depth analysis. Without expert guidelines, it can only provide fundamental information, lacking the ability to engage in advanced exploration tasks such as matrix calculations. 

In terms of the plant diseases, pests, and weeds detection tasks, compared to the fully supervised convolutional neural network\cite{chen2022performance} designed for weed classification, the performance of GPT-4V falls short, particularly in identifying Goosegrass. However, GPT-4V showed its powerful zero-shot capabilities on agricultural data analysis and it can provide more information that the convolutional neural network cannot provide, such as solutions for disease and weed conttol, which can help improve the efficiency in field management. Currently, in task like counting plants under complex environment from images, the performance of GPT-4V we tested didn't quite match that of fully supervised models \cite{tan2022towards,tan2023anchor}, indicating room for improvement in this area of object detection. 

For experiments on poultry science tasks, GPT-4V showcases remarkable analytical skills in poultry management, crucial for precision agriculture. It expertly interprets eggshell conditions, pinpointing discoloration, shape anomalies, and potential signs of calcium deficiency. This AI's ability to identify chicken behaviors, such as egg-laying and brooding patterns, mirrors its aptitude for chicken surveillance, including evaluating birds' growth stages. GPT-4V is adept at counting and categorizing chickens, although it may not match the precision of specialized convolutional neural networks in certain complex scenarios. However, its ability to perform zero-shot analysis offers significant insights into poultry health and behavioral patterns, leading to recommendations for enhanced farm management. This capability highlights GPT-4V's potential in aiding agriculture by providing critical data for informed decision-making, although further refinement is needed for highly precise object detection in detailed environments.


\subsubsection{Urban Planning Applications}
GPT-4V shows professional capability in knowledge retrieval and urban analysis equal to junior urban planners, while it performs worse in urban planning and design than even non-professionals. Its proficiency in explaining various maps, diagrams, and other images in urban planning essentially demonstrates the capability of multimodal understanding and question-answering, as well as the possession of knowledge in this domain, surpassing existing models \cite{wang2023optimizing,yan2023urbanclip}. Similarly, its impressive performance in urban street analysis, including identifying problems and proposing solutions, relies on object detection and knowledge retrieval. However, its exaggeration of urban street problems indicates the lack of true understanding and mastery of the knowledge and analytical skills, but the match of the image and corresponding potential texts. This flaw is more evident when it is requested to design, even just an extraction from an existing plan map or modification of an image showing the current situation. The lack of ability to make trade-offs and ignorance of basic urban rules (e.g., lane direction) also contribute to its failure in design generation.


\subsection{Major Limitations of GPT-4V}
Despite promising capabilities, analysis exposes clear limitations of GPT-4V across the domains explored. In geo-localization, performance severely declines for natural landscapes compared to man-made structures due to the greater visual complexity. The RS visual question answering task like precise object counting remains unreliable, likely stemming from inadequate segmentation model integration.

For agriculture, crop type identification struggles with irregular land patches and intricate backgrounds surrounding fields of interest. In livestock monitoring, limitations surface in differentiating and quantifying partially occluded animals. Across domains, fine-grained classification proves challenging without specialized domain adaptation. Suboptimal conditions with occlusion, clutter, or insufficient context further hamper counting performance.

Collectively, these limitations reveal gaps between conversational intelligence and lower-level vision. Full integration of textual semantics, visual recognition, and reasoning could better guide analysis. Domain-specific refinement is essential for discriminative identification and quantification, particularly where subtle visual cues or contextual inference is imperative.

\subsection{Future Perspectives}
Despite present limitations, our insights inform high-potential areas for advancing multimodal intelligence. Expanding geo-semantic knowledge and integrating geospatial vector data \cite{mai2023spatialrl} could bolster localization, while sensor fusion from satellite instrumentation and simulations may enrich environmental understanding. 

For geography, GPT-4V could be trained by a combination of local street-level images – either with specific landmarks or of common places – and aerial pictures with more globally structured scenarios. Its ability to identify geo-locations and geo-features could be significantly improved by such cognitive relationship between the local and the global.
Moreover, the map text extraction and localization experiments highlight the difficulties of GPT-4V in handling coordinate data. 
This is consistent with other people's findings about GPT-4V's low performance on math tasks \footnote{\url{https://www.cantorsparadise.com/gpt-4-is-amazing-but-still-struggles-at-high-school-math-competitions-cbc2e73738e}}, geocoding task \cite{mai2023opportunities}, and temporal-related tasks \cite{spathis2023first}. This is because GPT-4V tokenizes numbers, coordinates, and time points as separate digit tokens and disregards their internal
relationships. Developing correct tokenization methods for these types of data is one of the major future research directions of multimodal FMs.

For agriculture, augmented training via aerial, ground, and microscopic imaging could enhance crop trait characterization and disease modeling. Additionally, the integration of multimodal data, such as thermal, lidar, and hyperspectral imaging, should help improve the robustness in disease and weed recognition, as well as the accuracy of crop phenotyping. Advances in instance segmentation, attention mechanisms, and transform-based architectures should improve animal quantification and behavior analysis. More broadly, agriculture remains an open challenge area to validate scientific visual question answering, especially in scenarios involving crop phenotyping under conditions of complex illumination, occlusion, and densely overlapped objects.

Moving forward, progress necessitates model expansions encompassing spatiotemporal data, graphs, 3D imagery, and genomics while ensuring computational efficiency and usability. Targeted tuning on subdomain tasks warrants exploration. Responsible development guidelines and monitoring will be imperative as applications permeate real-world scientific and societal domains.

Moreover, two important aspects we cannot neglect for future multimodal FM development are the ethics and trustworthiness of FMs. The former emphasizes whether the data collection stage, pre-training stage, fine-tuning stage, or inference stage of multimodal LLMs might lead to any inequality and social justice issues such as generating toxic content \cite{gehman2020realtoxicityprompts,deshpande2023toxicity}, gender-biased content \cite{zhao2018gender}, or geographically biased content \cite{faisal2022geographicbias,mai2023opportunities}. The latter highlights the importance of grounding the generated content (e.g., texts, images, videos, audios) to its original source to make sure of its correctness and trustworthiness \cite{wang2023decodingtrust}. 

\section{Conclusion}
The evaluation of GPT-4V in diverse scientific domains has demonstrated its significant potential as well as notable limitations. The model's proficiency in geographical, environmental science, 
agricultural, and urban planning tasks 
highlights its utility in these areas. These capabilities suggest its potential role in enhancing precision agriculture, environmental monitoring, and geographic artificial intelligence. However, challenges in fine-grained recognition, precise counting in complex settings, and consistent performance across varied domains underscore the need for continued development and specialized training.

Advancements in GPT-4V will hinge on its seamless integration with emerging technologies, focused enhancements through domain-specific training, and steadfast adherence to ethical and responsible AI practices. Addressing its current limitations requires interdisciplinary collaboration and advancements in AI algorithms and training methodologies. As AI continues to evolve, its application in scientific research offers promising avenues for innovation, demanding a balanced approach that considers ethical implications, data privacy, and security \cite{ye2023toward}. The journey of AI in scientific domains is poised for transformative growth, necessitating a commitment to responsible development and application.

\paragraph{Acknowledgements}
Dr. Gengchen Mai acknowledges the support from the University of Georgia's International Intellectual Property Alliance (IIPA) Seed Grant -- 'SSIF: A Deep Generative Model for Remote Sensing Image Spatial-Spectral Super-Resolution for Precision Agriculture'. Dr. Lilong Chai acknowledges the support from USDA-NIFA AFRI (2023-68008-39853). Dr. Changying Li acknowledges funding support from USDA-NIFA Data Science for Food and Agricultural Systems (2023-67021-40646).



\begin{thebibliography}{235}
\ifx \bisbn   \undefined \def \bisbn  #1{ISBN #1}\fi
\ifx \binits  \undefined \def \binits#1{#1}\fi
\ifx \bauthor  \undefined \def \bauthor#1{#1}\fi
\ifx \batitle  \undefined \def \batitle#1{#1}\fi
\ifx \bjtitle  \undefined \def \bjtitle#1{#1}\fi
\ifx \bvolume  \undefined \def \bvolume#1{\textbf{#1}}\fi
\ifx \byear  \undefined \def \byear#1{#1}\fi
\ifx \bissue  \undefined \def \bissue#1{#1}\fi
\ifx \bfpage  \undefined \def \bfpage#1{#1}\fi
\ifx \blpage  \undefined \def \blpage #1{#1}\fi
\ifx \burl  \undefined \def \burl#1{\textsf{#1}}\fi
\ifx \doiurl  \undefined \def \doiurl#1{\url{https://doi.org/#1}}\fi
\ifx \betal  \undefined \def \betal{\textit{et al.}}\fi
\ifx \binstitute  \undefined \def \binstitute#1{#1}\fi
\ifx \binstitutionaled  \undefined \def \binstitutionaled#1{#1}\fi
\ifx \bctitle  \undefined \def \bctitle#1{#1}\fi
\ifx \beditor  \undefined \def \beditor#1{#1}\fi
\ifx \bpublisher  \undefined \def \bpublisher#1{#1}\fi
\ifx \bbtitle  \undefined \def \bbtitle#1{#1}\fi
\ifx \bedition  \undefined \def \bedition#1{#1}\fi
\ifx \bseriesno  \undefined \def \bseriesno#1{#1}\fi
\ifx \blocation  \undefined \def \blocation#1{#1}\fi
\ifx \bsertitle  \undefined \def \bsertitle#1{#1}\fi
\ifx \bsnm \undefined \def \bsnm#1{#1}\fi
\ifx \bsuffix \undefined \def \bsuffix#1{#1}\fi
\ifx \bparticle \undefined \def \bparticle#1{#1}\fi
\ifx \barticle \undefined \def \barticle#1{#1}\fi
\bibcommenthead
\ifx \bconfdate \undefined \def \bconfdate #1{#1}\fi
\ifx \botherref \undefined \def \botherref #1{#1}\fi
\ifx \url \undefined \def \url#1{\textsf{#1}}\fi
\ifx \bchapter \undefined \def \bchapter#1{#1}\fi
\ifx \bbook \undefined \def \bbook#1{#1}\fi
\ifx \bcomment \undefined \def \bcomment#1{#1}\fi
\ifx \oauthor \undefined \def \oauthor#1{#1}\fi
\ifx \citeauthoryear \undefined \def \citeauthoryear#1{#1}\fi
\ifx \endbibitem  \undefined \def \endbibitem {}\fi
\ifx \bconflocation  \undefined \def \bconflocation#1{#1}\fi
\ifx \arxivurl  \undefined \def \arxivurl#1{\textsf{#1}}\fi
\csname PreBibitemsHook\endcsname

\bibitem[\protect\citeauthoryear{Liu et~al.}{2023}]{liu2023summary}
\begin{botherref}
\oauthor{\bsnm{Liu}, \binits{Y.}},
\oauthor{\bsnm{Han}, \binits{T.}},
\oauthor{\bsnm{Ma}, \binits{S.}},
\oauthor{\bsnm{Zhang}, \binits{J.}},
\oauthor{\bsnm{Yang}, \binits{Y.}},
\oauthor{\bsnm{Tian}, \binits{J.}},
\oauthor{\bsnm{He}, \binits{H.}},
\oauthor{\bsnm{Li}, \binits{A.}},
\oauthor{\bsnm{He}, \binits{M.}},
\oauthor{\bsnm{Liu}, \binits{Z.}}, et al.:
Summary of chatgpt-related research and perspective towards the future of large
  language models.
Meta-Radiology,
100017
(2023)
\end{botherref}
\endbibitem

\bibitem[\protect\citeauthoryear{Zhao et~al.}{2023}]{zhao2023brain}
\begin{botherref}
\oauthor{\bsnm{Zhao}, \binits{L.}},
\oauthor{\bsnm{Zhang}, \binits{L.}},
\oauthor{\bsnm{Wu}, \binits{Z.}},
\oauthor{\bsnm{Chen}, \binits{Y.}},
\oauthor{\bsnm{Dai}, \binits{H.}},
\oauthor{\bsnm{Yu}, \binits{X.}},
\oauthor{\bsnm{Liu}, \binits{Z.}},
\oauthor{\bsnm{Zhang}, \binits{T.}},
\oauthor{\bsnm{Hu}, \binits{X.}},
\oauthor{\bsnm{Jiang}, \binits{X.}}, et al.:
When brain-inspired ai meets agi.
Meta-Radiology,
100005
(2023)
\end{botherref}
\endbibitem

\bibitem[\protect\citeauthoryear{Cao et~al.}{2023}]{cao2023comprehensive}
\begin{botherref}
\oauthor{\bsnm{Cao}, \binits{Y.}},
\oauthor{\bsnm{Li}, \binits{S.}},
\oauthor{\bsnm{Liu}, \binits{Y.}},
\oauthor{\bsnm{Yan}, \binits{Z.}},
\oauthor{\bsnm{Dai}, \binits{Y.}},
\oauthor{\bsnm{Yu}, \binits{P.S.}},
\oauthor{\bsnm{Sun}, \binits{L.}}:
A comprehensive survey of ai-generated content (aigc): A history of generative
  ai from gan to chatgpt.
arXiv preprint arXiv:2303.04226
(2023)
\end{botherref}
\endbibitem

\bibitem[\protect\citeauthoryear{Liu et~al.}{2023}]{liu2023llava}
\begin{botherref}
\oauthor{\bsnm{Liu}, \binits{H.}},
\oauthor{\bsnm{Li}, \binits{C.}},
\oauthor{\bsnm{Wu}, \binits{Q.}},
\oauthor{\bsnm{Lee}, \binits{Y.J.}}:
Visual instruction tuning.
arXiv preprint arXiv:2304.08485
(2023)
\end{botherref}
\endbibitem

\bibitem[\protect\citeauthoryear{Bubeck et~al.}{2023}]{bubeck2023sparks}
\begin{botherref}
\oauthor{\bsnm{Bubeck}, \binits{S.}},
\oauthor{\bsnm{Chandrasekaran}, \binits{V.}},
\oauthor{\bsnm{Eldan}, \binits{R.}},
\oauthor{\bsnm{Gehrke}, \binits{J.}},
\oauthor{\bsnm{Horvitz}, \binits{E.}},
\oauthor{\bsnm{Kamar}, \binits{E.}},
\oauthor{\bsnm{Lee}, \binits{P.}},
\oauthor{\bsnm{Lee}, \binits{Y.T.}},
\oauthor{\bsnm{Li}, \binits{Y.}},
\oauthor{\bsnm{Lundberg}, \binits{S.}}, et al.:
Sparks of artificial general intelligence: Early experiments with gpt-4.
arXiv preprint arXiv:2303.12712
(2023)
\end{botherref}
\endbibitem

\bibitem[\protect\citeauthoryear{Ramesh et~al.}{2022}]{ramesh2022hierarchical}
\begin{barticle}
\bauthor{\bsnm{Ramesh}, \binits{A.}},
\bauthor{\bsnm{Dhariwal}, \binits{P.}},
\bauthor{\bsnm{Nichol}, \binits{A.}},
\bauthor{\bsnm{Chu}, \binits{C.}},
\bauthor{\bsnm{Chen}, \binits{M.}}:
\batitle{Hierarchical text-conditional image generation with clip latents}.
\bjtitle{arXiv preprint arXiv:2204.06125}
\bvolume{1}(\bissue{2}),
\bfpage{3}
(\byear{2022})
\end{barticle}
\endbibitem

\bibitem[\protect\citeauthoryear{Liu et~al.}{2023a}]{liu2023tailoring}
\begin{bchapter}
\bauthor{\bsnm{Liu}, \binits{Z.}},
\bauthor{\bsnm{Zhong}, \binits{A.}},
\bauthor{\bsnm{Li}, \binits{Y.}},
\bauthor{\bsnm{Yang}, \binits{L.}},
\bauthor{\bsnm{Ju}, \binits{C.}},
\bauthor{\bsnm{Wu}, \binits{Z.}},
\bauthor{\bsnm{Ma}, \binits{C.}},
\bauthor{\bsnm{Shu}, \binits{P.}},
\bauthor{\bsnm{Chen}, \binits{C.}},
\bauthor{\bsnm{Kim}, \binits{S.}}, \betal:
\bctitle{Tailoring large language models to radiology: A preliminary approach
  to llm adaptation for a highly specialized domain}.
In: \bbtitle{International Workshop on Machine Learning in Medical Imaging},
pp. \bfpage{464}--\blpage{473}
(\byear{2023}).
\bcomment{Springer}
\end{bchapter}
\endbibitem

\bibitem[\protect\citeauthoryear{Liu et~al.}{2023b}]{liu2023evaluating}
\begin{botherref}
\oauthor{\bsnm{Liu}, \binits{Z.}},
\oauthor{\bsnm{Zhong}, \binits{T.}},
\oauthor{\bsnm{Li}, \binits{Y.}},
\oauthor{\bsnm{Zhang}, \binits{Y.}},
\oauthor{\bsnm{Pan}, \binits{Y.}},
\oauthor{\bsnm{Zhao}, \binits{Z.}},
\oauthor{\bsnm{Dong}, \binits{P.}},
\oauthor{\bsnm{Cao}, \binits{C.}},
\oauthor{\bsnm{Liu}, \binits{Y.}},
\oauthor{\bsnm{Shu}, \binits{P.}}, et al.:
Evaluating large language models for radiology natural language processing.
arXiv preprint arXiv:2307.13693
(2023)
\end{botherref}
\endbibitem

\bibitem[\protect\citeauthoryear{Dai et~al.}{2023}]{dai2023ad}
\begin{botherref}
\oauthor{\bsnm{Dai}, \binits{H.}},
\oauthor{\bsnm{Li}, \binits{Y.}},
\oauthor{\bsnm{Liu}, \binits{Z.}},
\oauthor{\bsnm{Zhao}, \binits{L.}},
\oauthor{\bsnm{Wu}, \binits{Z.}},
\oauthor{\bsnm{Song}, \binits{S.}},
\oauthor{\bsnm{Shen}, \binits{Y.}},
\oauthor{\bsnm{Zhu}, \binits{D.}},
\oauthor{\bsnm{Li}, \binits{X.}},
\oauthor{\bsnm{Li}, \binits{S.}}, et al.:
Ad-autogpt: An autonomous gpt for alzheimer's disease infodemiology.
arXiv preprint arXiv:2306.10095
(2023)
\end{botherref}
\endbibitem

\bibitem[\protect\citeauthoryear{Liu et~al.}{2023}]{liu2023radonc}
\begin{botherref}
\oauthor{\bsnm{Liu}, \binits{Z.}},
\oauthor{\bsnm{Wang}, \binits{P.}},
\oauthor{\bsnm{Li}, \binits{Y.}},
\oauthor{\bsnm{Holmes}, \binits{J.}},
\oauthor{\bsnm{Shu}, \binits{P.}},
\oauthor{\bsnm{Zhang}, \binits{L.}},
\oauthor{\bsnm{Liu}, \binits{C.}},
\oauthor{\bsnm{Liu}, \binits{N.}},
\oauthor{\bsnm{Zhu}, \binits{D.}},
\oauthor{\bsnm{Li}, \binits{X.}}, et al.:
Radonc-gpt: A large language model for radiation oncology.
arXiv preprint arXiv:2309.10160
(2023)
\end{botherref}
\endbibitem

\bibitem[\protect\citeauthoryear{Mai et~al.}{2022}]{mai2022towards}
\begin{bchapter}
\bauthor{\bsnm{Mai}, \binits{G.}},
\bauthor{\bsnm{Cundy}, \binits{C.}},
\bauthor{\bsnm{Choi}, \binits{K.}},
\bauthor{\bsnm{Hu}, \binits{Y.}},
\bauthor{\bsnm{Lao}, \binits{N.}},
\bauthor{\bsnm{Ermon}, \binits{S.}}:
\bctitle{Towards a foundation model for geospatial artificial intelligence
  (vision paper)}.
In: \bbtitle{Proceedings of the 30th ACM SIGSPATIAL International Conference on
  Advances in Geographic Information Systems},
pp. \bfpage{1}--\blpage{4}
(\byear{2022})
\end{bchapter}
\endbibitem

\bibitem[\protect\citeauthoryear{Mai et~al.}{2023}]{mai2023opportunities}
\begin{botherref}
\oauthor{\bsnm{Mai}, \binits{G.}},
\oauthor{\bsnm{Huang}, \binits{W.}},
\oauthor{\bsnm{Sun}, \binits{J.}},
\oauthor{\bsnm{Song}, \binits{S.}},
\oauthor{\bsnm{Mishra}, \binits{D.}},
\oauthor{\bsnm{Liu}, \binits{N.}},
\oauthor{\bsnm{Gao}, \binits{S.}},
\oauthor{\bsnm{Liu}, \binits{T.}},
\oauthor{\bsnm{Cong}, \binits{G.}},
\oauthor{\bsnm{Hu}, \binits{Y.}}, et al.:
On the opportunities and challenges of foundation models for geospatial
  artificial intelligence.
arXiv preprint arXiv:2304.06798
(2023)
\end{botherref}
\endbibitem

\bibitem[\protect\citeauthoryear{Hu et~al.}{2023}]{hu2023geo}
\begin{botherref}
\oauthor{\bsnm{Hu}, \binits{Y.}},
\oauthor{\bsnm{Mai}, \binits{G.}},
\oauthor{\bsnm{Cundy}, \binits{C.}},
\oauthor{\bsnm{Choi}, \binits{K.}},
\oauthor{\bsnm{Lao}, \binits{N.}},
\oauthor{\bsnm{Liu}, \binits{W.}},
\oauthor{\bsnm{Lakhanpal}, \binits{G.}},
\oauthor{\bsnm{Zhou}, \binits{R.Z.}},
\oauthor{\bsnm{Joseph}, \binits{K.}}:
Geo-knowledge-guided gpt models improve the extraction of location descriptions
  from disaster-related social media messages.
International Journal of Geographical Information Science,
1--30
(2023)
\end{botherref}
\endbibitem

\bibitem[\protect\citeauthoryear{Xie et~al.}{2023}]{xie2023geo}
\begin{bchapter}
\bauthor{\bsnm{Xie}, \binits{Y.}},
\bauthor{\bsnm{Wang}, \binits{Z.}},
\bauthor{\bsnm{Mai}, \binits{G.}},
\bauthor{\bsnm{Li}, \binits{Y.}},
\bauthor{\bsnm{Jia}, \binits{X.}},
\bauthor{\bsnm{Gao}, \binits{S.}},
\bauthor{\bsnm{Wang}, \binits{S.}}:
\bctitle{Geo-foundation models: Reality, gaps and opportunities (vision
  paper)}.
In: \bbtitle{Proceedings of the 31th ACM SIGSPATIAL International Conference on
  Advances in Geographic Information Systems}
(\byear{2023})
\end{bchapter}
\endbibitem

\bibitem[\protect\citeauthoryear{Rao et~al.}{2023}]{rao2023building}
\begin{bchapter}
\bauthor{\bsnm{Rao}, \binits{J.}},
\bauthor{\bsnm{Gao}, \binits{S.}},
\bauthor{\bsnm{Mai}, \binits{G.}},
\bauthor{\bsnm{Janowicz}, \binits{K.}}:
\bctitle{Building privacy-preserving and secure geospatial artificial
  intelligence foundation models}.
In: \bbtitle{Proceedings of the 31th ACM SIGSPATIAL International Conference on
  Advances in Geographic Information Systems}
(\byear{2023})
\end{bchapter}
\endbibitem

\bibitem[\protect\citeauthoryear{Roberts et~al.}{2023}]{roberts2023gpt4geo}
\begin{botherref}
\oauthor{\bsnm{Roberts}, \binits{J.}},
\oauthor{\bsnm{L{\"u}ddecke}, \binits{T.}},
\oauthor{\bsnm{Das}, \binits{S.}},
\oauthor{\bsnm{Han}, \binits{K.}},
\oauthor{\bsnm{Albanie}, \binits{S.}}:
Gpt4geo: How a language model sees the world's geography.
arXiv preprint arXiv:2306.00020
(2023)
\end{botherref}
\endbibitem

\bibitem[\protect\citeauthoryear{Zhang et~al.}{2023}]{zhang2023geogpt}
\begin{botherref}
\oauthor{\bsnm{Zhang}, \binits{Y.}},
\oauthor{\bsnm{Wei}, \binits{C.}},
\oauthor{\bsnm{Wu}, \binits{S.}},
\oauthor{\bsnm{He}, \binits{Z.}},
\oauthor{\bsnm{Yu}, \binits{W.}}:
Geogpt: Understanding and processing geospatial tasks through an autonomous
  gpt.
arXiv preprint arXiv:2307.07930
(2023)
\end{botherref}
\endbibitem

\bibitem[\protect\citeauthoryear{Manvi et~al.}{2023}]{manvi2023geollm}
\begin{botherref}
\oauthor{\bsnm{Manvi}, \binits{R.}},
\oauthor{\bsnm{Khanna}, \binits{S.}},
\oauthor{\bsnm{Mai}, \binits{G.}},
\oauthor{\bsnm{Burke}, \binits{M.}},
\oauthor{\bsnm{Lobell}, \binits{D.}},
\oauthor{\bsnm{Ermon}, \binits{S.}}:
Geollm: Extracting geospatial knowledge from large language models.
arXiv preprint arXiv:2310.06213
(2023)
\end{botherref}
\endbibitem

\bibitem[\protect\citeauthoryear{Zhang et~al.}{2023}]{zhang2023text2seg}
\begin{botherref}
\oauthor{\bsnm{Zhang}, \binits{J.}},
\oauthor{\bsnm{Zhou}, \binits{Z.}},
\oauthor{\bsnm{Mai}, \binits{G.}},
\oauthor{\bsnm{Mu}, \binits{L.}},
\oauthor{\bsnm{Hu}, \binits{M.}},
\oauthor{\bsnm{Li}, \binits{S.}}:
Text2seg: Remote sensing image semantic segmentation via text-guided visual
  foundation models.
arXiv preprint arXiv:2304.10597
(2023)
\end{botherref}
\endbibitem

\bibitem[\protect\citeauthoryear{Zhu et~al.}{2023}]{zhu2023chatgpt}
\begin{botherref}
\oauthor{\bsnm{Zhu}, \binits{J.-J.}},
\oauthor{\bsnm{Jiang}, \binits{J.}},
\oauthor{\bsnm{Yang}, \binits{M.}},
\oauthor{\bsnm{Ren}, \binits{Z.J.}}:
Chatgpt and environmental research.
Environmental Science \& Technology
(2023)
\end{botherref}
\endbibitem

\bibitem[\protect\citeauthoryear{Agathokleous
  et~al.}{2023}]{agathokleous2023use}
\begin{barticle}
\bauthor{\bsnm{Agathokleous}, \binits{E.}},
\bauthor{\bsnm{Saitanis}, \binits{C.J.}},
\bauthor{\bsnm{Fang}, \binits{C.}},
\bauthor{\bsnm{Yu}, \binits{Z.}}:
\batitle{Use of chatgpt: What does it mean for biology and environmental
  science?}
\bjtitle{Science of The Total Environment}
\bvolume{888},
\bfpage{164154}
(\byear{2023})
\end{barticle}
\endbibitem

\bibitem[\protect\citeauthoryear{Rezayi et~al.}{2022}]{rezayi2022agribert}
\begin{bchapter}
\bauthor{\bsnm{Rezayi}, \binits{S.}},
\bauthor{\bsnm{Liu}, \binits{Z.}},
\bauthor{\bsnm{Wu}, \binits{Z.}},
\bauthor{\bsnm{Dhakal}, \binits{C.}},
\bauthor{\bsnm{Ge}, \binits{B.}},
\bauthor{\bsnm{Zhen}, \binits{C.}},
\bauthor{\bsnm{Liu}, \binits{T.}},
\bauthor{\bsnm{Li}, \binits{S.}}:
\bctitle{Agribert: knowledge-infused agricultural language models for matching
  food and nutrition}.
In: \bbtitle{Proceedings of the Thirty-First International Joint Conference on
  Artificial Intelligence},
vol. \bseriesno{7},
pp. \bfpage{5150}--\blpage{5156}
(\byear{2022})
\end{bchapter}
\endbibitem

\bibitem[\protect\citeauthoryear{Rezayi et~al.}{2023}]{rezayi2023exploring}
\begin{botherref}
\oauthor{\bsnm{Rezayi}, \binits{S.}},
\oauthor{\bsnm{Liu}, \binits{Z.}},
\oauthor{\bsnm{Wu}, \binits{Z.}},
\oauthor{\bsnm{Dhakal}, \binits{C.}},
\oauthor{\bsnm{Ge}, \binits{B.}},
\oauthor{\bsnm{Dai}, \binits{H.}},
\oauthor{\bsnm{Mai}, \binits{G.}},
\oauthor{\bsnm{Liu}, \binits{N.}},
\oauthor{\bsnm{Zhen}, \binits{C.}},
\oauthor{\bsnm{Liu}, \binits{T.}}, et al.:
Exploring new frontiers in agricultural nlp: Investigating the potential of
  large language models for food applications.
arXiv preprint arXiv:2306.11892
(2023)
\end{botherref}
\endbibitem

\bibitem[\protect\citeauthoryear{Lu et~al.}{2023}]{lu2023agi}
\begin{botherref}
\oauthor{\bsnm{Lu}, \binits{G.}},
\oauthor{\bsnm{Li}, \binits{S.}},
\oauthor{\bsnm{Mai}, \binits{G.}},
\oauthor{\bsnm{Sun}, \binits{J.}},
\oauthor{\bsnm{Zhu}, \binits{D.}},
\oauthor{\bsnm{Chai}, \binits{L.}},
\oauthor{\bsnm{Sun}, \binits{H.}},
\oauthor{\bsnm{Wang}, \binits{X.}},
\oauthor{\bsnm{Dai}, \binits{H.}},
\oauthor{\bsnm{Liu}, \binits{N.}},
\oauthor{\bsnm{Xu}, \binits{R.}},
\oauthor{\bsnm{Petti}, \binits{D.}},
\oauthor{\bsnm{Li}, \binits{C.}},
\oauthor{\bsnm{Liu}, \binits{T.}},
\oauthor{\bsnm{Li}, \binits{C.}}:
AGI for Agriculture
(2023)
\end{botherref}
\endbibitem

\bibitem[\protect\citeauthoryear{Liu et~al.}{2023}]{liu2023transformation}
\begin{botherref}
\oauthor{\bsnm{Liu}, \binits{Z.}},
\oauthor{\bsnm{Li}, \binits{Y.}},
\oauthor{\bsnm{Cao}, \binits{Q.}},
\oauthor{\bsnm{Chen}, \binits{J.}},
\oauthor{\bsnm{Yang}, \binits{T.}},
\oauthor{\bsnm{Wu}, \binits{Z.}},
\oauthor{\bsnm{Hale}, \binits{J.}},
\oauthor{\bsnm{Gibbs}, \binits{J.}},
\oauthor{\bsnm{Rasheed}, \binits{K.}},
\oauthor{\bsnm{Liu}, \binits{N.}}, et al.:
Transformation vs tradition: Artificial general intelligence (agi) for arts and
  humanities.
arXiv preprint arXiv:2310.19626
(2023)
\end{botherref}
\endbibitem

\bibitem[\protect\citeauthoryear{Vaswani et~al.}{2017}]{vaswani2017attention}
\begin{botherref}
\oauthor{\bsnm{Vaswani}, \binits{A.}},
\oauthor{\bsnm{Shazeer}, \binits{N.}},
\oauthor{\bsnm{Parmar}, \binits{N.}},
\oauthor{\bsnm{Uszkoreit}, \binits{J.}},
\oauthor{\bsnm{Jones}, \binits{L.}},
\oauthor{\bsnm{Gomez}, \binits{A.N.}},
\oauthor{\bsnm{Kaiser}, \binits{{\L}.}},
\oauthor{\bsnm{Polosukhin}, \binits{I.}}:
Attention is all you need.
Advances in neural information processing systems
\textbf{30}
(2017)
\end{botherref}
\endbibitem

\bibitem[\protect\citeauthoryear{Kim et~al.}{2023}]{kim2023medivista}
\begin{botherref}
\oauthor{\bsnm{Kim}, \binits{S.}},
\oauthor{\bsnm{Kim}, \binits{K.}},
\oauthor{\bsnm{Hu}, \binits{J.}},
\oauthor{\bsnm{Chen}, \binits{C.}},
\oauthor{\bsnm{Lyu}, \binits{Z.}},
\oauthor{\bsnm{Hui}, \binits{R.}},
\oauthor{\bsnm{Kim}, \binits{S.}},
\oauthor{\bsnm{Liu}, \binits{Z.}},
\oauthor{\bsnm{Zhong}, \binits{A.}},
\oauthor{\bsnm{Li}, \binits{X.}}, et al.:
Medivista-sam: Zero-shot medical video analysis with spatio-temporal sam
  adaptation.
arXiv preprint arXiv:2309.13539
(2023)
\end{botherref}
\endbibitem

\bibitem[\protect\citeauthoryear{Liu et~al.}{2023}]{liu2023holistic}
\begin{botherref}
\oauthor{\bsnm{Liu}, \binits{Z.}},
\oauthor{\bsnm{Jiang}, \binits{H.}},
\oauthor{\bsnm{Zhong}, \binits{T.}},
\oauthor{\bsnm{Wu}, \binits{Z.}},
\oauthor{\bsnm{Ma}, \binits{C.}},
\oauthor{\bsnm{Li}, \binits{Y.}},
\oauthor{\bsnm{Yu}, \binits{X.}},
\oauthor{\bsnm{Zhang}, \binits{Y.}},
\oauthor{\bsnm{Pan}, \binits{Y.}},
\oauthor{\bsnm{Shu}, \binits{P.}}, et al.:
Holistic evaluation of gpt-4v for biomedical imaging.
arXiv preprint arXiv:2312.05256
(2023)
\end{botherref}
\endbibitem

\bibitem[\protect\citeauthoryear{Lyu et~al.}{2023}]{lyu2023gpt}
\begin{botherref}
\oauthor{\bsnm{Lyu}, \binits{H.}},
\oauthor{\bsnm{Huang}, \binits{J.}},
\oauthor{\bsnm{Zhang}, \binits{D.}},
\oauthor{\bsnm{Yu}, \binits{Y.}},
\oauthor{\bsnm{Mou}, \binits{X.}},
\oauthor{\bsnm{Pan}, \binits{J.}},
\oauthor{\bsnm{Yang}, \binits{Z.}},
\oauthor{\bsnm{Wei}, \binits{Z.}},
\oauthor{\bsnm{Luo}, \binits{J.}}:
Gpt-4v (ision) as a social media analysis engine.
arXiv preprint arXiv:2311.07547
(2023)
\end{botherref}
\endbibitem

\bibitem[\protect\citeauthoryear{Yang et~al.}{2023}]{yang2023dawn}
\begin{botherref}
\oauthor{\bsnm{Yang}, \binits{Z.}},
\oauthor{\bsnm{Li}, \binits{L.}},
\oauthor{\bsnm{Lin}, \binits{K.}},
\oauthor{\bsnm{Wang}, \binits{J.}},
\oauthor{\bsnm{Lin}, \binits{C.-C.}},
\oauthor{\bsnm{Liu}, \binits{Z.}},
\oauthor{\bsnm{Wang}, \binits{L.}}:
The dawn of lmms: Preliminary explorations with gpt-4v (ision).
arXiv preprint arXiv:2309.17421
\textbf{9}(1)
(2023)
\end{botherref}
\endbibitem

\bibitem[\protect\citeauthoryear{Weyand et~al.}{2016}]{weyand2016planet}
\begin{bchapter}
\bauthor{\bsnm{Weyand}, \binits{T.}},
\bauthor{\bsnm{Kostrikov}, \binits{I.}},
\bauthor{\bsnm{Philbin}, \binits{J.}}:
\bctitle{Planet-photo geolocation with convolutional neural networks}.
In: \bbtitle{Computer Vision--ECCV 2016: 14th European Conference, Amsterdam,
  The Netherlands, October 11-14, 2016, Proceedings, Part VIII 14},
pp. \bfpage{37}--\blpage{55}
(\byear{2016}).
\bcomment{Springer}
\end{bchapter}
\endbibitem

\bibitem[\protect\citeauthoryear{Seo et~al.}{2018}]{seo2018cplanet}
\begin{bchapter}
\bauthor{\bsnm{Seo}, \binits{P.H.}},
\bauthor{\bsnm{Weyand}, \binits{T.}},
\bauthor{\bsnm{Sim}, \binits{J.}},
\bauthor{\bsnm{Han}, \binits{B.}}:
\bctitle{Cplanet: Enhancing image geolocalization by combinatorial partitioning
  of maps}.
In: \bbtitle{Proceedings of the European Conference on Computer Vision (ECCV)},
pp. \bfpage{536}--\blpage{551}
(\byear{2018})
\end{bchapter}
\endbibitem

\bibitem[\protect\citeauthoryear{Muller-Budack
  et~al.}{2018}]{muller2018geolocation}
\begin{bchapter}
\bauthor{\bsnm{Muller-Budack}, \binits{E.}},
\bauthor{\bsnm{Pustu-Iren}, \binits{K.}},
\bauthor{\bsnm{Ewerth}, \binits{R.}}:
\bctitle{Geolocation estimation of photos using a hierarchical model and scene
  classification}.
In: \bbtitle{Proceedings of the European Conference on Computer Vision (ECCV)},
pp. \bfpage{563}--\blpage{579}
(\byear{2018})
\end{bchapter}
\endbibitem

\bibitem[\protect\citeauthoryear{Izbicki et~al.}{2020}]{izbicki2020exploiting}
\begin{bchapter}
\bauthor{\bsnm{Izbicki}, \binits{M.}},
\bauthor{\bsnm{Papalexakis}, \binits{E.E.}},
\bauthor{\bsnm{Tsotras}, \binits{V.J.}}:
\bctitle{Exploiting the earth’s spherical geometry to geolocate images}.
In: \bbtitle{Machine Learning and Knowledge Discovery in Databases: European
  Conference, ECML PKDD 2019, W{\"u}rzburg, Germany, September 16--20, 2019,
  Proceedings, Part II},
pp. \bfpage{3}--\blpage{19}
(\byear{2020}).
\bcomment{Springer}
\end{bchapter}
\endbibitem

\bibitem[\protect\citeauthoryear{Cepeda et~al.}{2023}]{cepeda2023geoclip}
\begin{bchapter}
\bauthor{\bsnm{Cepeda}, \binits{V.V.}},
\bauthor{\bsnm{Nayak}, \binits{G.K.}},
\bauthor{\bsnm{Shah}, \binits{M.}}:
\bctitle{Geoclip: Clip-inspired alignment between locations and images for
  effective worldwide geo-localization}.
In: \bbtitle{Thirty-seventh Conference on Neural Information Processing
  Systems}
(\byear{2023})
\end{bchapter}
\endbibitem

\bibitem[\protect\citeauthoryear{Kim et~al.}{2023}]{kim2023mapkurator}
\begin{bchapter}
\bauthor{\bsnm{Kim}, \binits{J.}},
\bauthor{\bsnm{Li}, \binits{Z.}},
\bauthor{\bsnm{Lin}, \binits{Y.}},
\bauthor{\bsnm{Namgung}, \binits{M.}},
\bauthor{\bsnm{Jang}, \binits{L.}},
\bauthor{\bsnm{Chiang}, \binits{Y.-Y.}}:
\bctitle{The mapkurator system: A complete pipeline for extracting and linking
  text from historical maps}.
In: \bbtitle{Proceedings of the 31th ACM SIGSPATIAL International Conference on
  Advances in Geographic Information Systems}
(\byear{2023})
\end{bchapter}
\endbibitem

\bibitem[\protect\citeauthoryear{Ayush et~al.}{2021}]{ayush2021geography}
\begin{bchapter}
\bauthor{\bsnm{Ayush}, \binits{K.}},
\bauthor{\bsnm{Uzkent}, \binits{B.}},
\bauthor{\bsnm{Meng}, \binits{C.}},
\bauthor{\bsnm{Tanmay}, \binits{K.}},
\bauthor{\bsnm{Burke}, \binits{M.}},
\bauthor{\bsnm{Lobell}, \binits{D.}},
\bauthor{\bsnm{Ermon}, \binits{S.}}:
\bctitle{Geography-aware self-supervised learning}.
In: \bbtitle{Proceedings of the IEEE/CVF International Conference on Computer
  Vision},
pp. \bfpage{10181}--\blpage{10190}
(\byear{2021})
\end{bchapter}
\endbibitem

\bibitem[\protect\citeauthoryear{Manas et~al.}{2021}]{manas2021seasonal}
\begin{bchapter}
\bauthor{\bsnm{Manas}, \binits{O.}},
\bauthor{\bsnm{Lacoste}, \binits{A.}},
\bauthor{\bsnm{Gir{\'o}-i-Nieto}, \binits{X.}},
\bauthor{\bsnm{Vazquez}, \binits{D.}},
\bauthor{\bsnm{Rodriguez}, \binits{P.}}:
\bctitle{Seasonal contrast: Unsupervised pre-training from uncurated remote
  sensing data}.
In: \bbtitle{Proceedings of the IEEE/CVF International Conference on Computer
  Vision},
pp. \bfpage{9414}--\blpage{9423}
(\byear{2021})
\end{bchapter}
\endbibitem

\bibitem[\protect\citeauthoryear{Mai et~al.}{2023a}]{mai2023sphere2vec}
\begin{barticle}
\bauthor{\bsnm{Mai}, \binits{G.}},
\bauthor{\bsnm{Xuan}, \binits{Y.}},
\bauthor{\bsnm{Zuo}, \binits{W.}},
\bauthor{\bsnm{He}, \binits{Y.}},
\bauthor{\bsnm{Song}, \binits{J.}},
\bauthor{\bsnm{Ermon}, \binits{S.}},
\bauthor{\bsnm{Janowicz}, \binits{K.}},
\bauthor{\bsnm{Lao}, \binits{N.}}:
\batitle{Sphere2vec: A general-purpose location representation learning over a
  spherical surface for large-scale geospatial predictions}.
\bjtitle{ISPRS Journal of Photogrammetry and Remote Sensing}
\bvolume{202},
\bfpage{439}--\blpage{462}
(\byear{2023})
\end{barticle}
\endbibitem

\bibitem[\protect\citeauthoryear{Mai et~al.}{2023b}]{mai2023csp}
\begin{bchapter}
\bauthor{\bsnm{Mai}, \binits{G.}},
\bauthor{\bsnm{Lao}, \binits{N.}},
\bauthor{\bsnm{He}, \binits{Y.}},
\bauthor{\bsnm{Song}, \binits{J.}},
\bauthor{\bsnm{Ermon}, \binits{S.}}:
\bctitle{Csp: Self-supervised contrastive spatial pre-training for
  geospatial-visual representations}.
In: \bbtitle{the Fortieth International Conference on Machine Learning (ICML
  2023)}
(\byear{2023})
\end{bchapter}
\endbibitem

\bibitem[\protect\citeauthoryear{Lobry et~al.}{2020}]{lobry2020rsvqa}
\begin{barticle}
\bauthor{\bsnm{Lobry}, \binits{S.}},
\bauthor{\bsnm{Marcos}, \binits{D.}},
\bauthor{\bsnm{Murray}, \binits{J.}},
\bauthor{\bsnm{Tuia}, \binits{D.}}:
\batitle{Rsvqa: Visual question answering for remote sensing data}.
\bjtitle{IEEE Transactions on Geoscience and Remote Sensing}
\bvolume{58}(\bissue{12}),
\bfpage{8555}--\blpage{8566}
(\byear{2020})
\end{barticle}
\endbibitem

\bibitem[\protect\citeauthoryear{Lobry et~al.}{2021}]{lobry2021rsvqa}
\begin{bchapter}
\bauthor{\bsnm{Lobry}, \binits{S.}},
\bauthor{\bsnm{Demir}, \binits{B.}},
\bauthor{\bsnm{Tuia}, \binits{D.}}:
\bctitle{Rsvqa meets bigearthnet: a new, large-scale, visual question answering
  dataset for remote sensing}.
In: \bbtitle{2021 IEEE International Geoscience and Remote Sensing Symposium
  IGARSS},
pp. \bfpage{1218}--\blpage{1221}
(\byear{2021}).
\bcomment{IEEE}
\end{bchapter}
\endbibitem

\bibitem[\protect\citeauthoryear{Chappuis et~al.}{2022}]{chappuis2022prompt}
\begin{bchapter}
\bauthor{\bsnm{Chappuis}, \binits{C.}},
\bauthor{\bsnm{Zermatten}, \binits{V.}},
\bauthor{\bsnm{Lobry}, \binits{S.}},
\bauthor{\bsnm{Le~Saux}, \binits{B.}},
\bauthor{\bsnm{Tuia}, \binits{D.}}:
\bctitle{Prompt-rsvqa: Prompting visual context to a language model for remote
  sensing visual question answering}.
In: \bbtitle{Proceedings of the IEEE/CVF Conference on Computer Vision and
  Pattern Recognition},
pp. \bfpage{1372}--\blpage{1381}
(\byear{2022})
\end{bchapter}
\endbibitem

\bibitem[\protect\citeauthoryear{Lackey}{2001}]{lackey2001values}
\begin{barticle}
\bauthor{\bsnm{Lackey}, \binits{R.T.}}:
\batitle{Values, policy, and ecosystem health: options for resolving the many
  ecological policy issues we face depend on the concept of ecosystem health,
  but ecosystem health is based on controversial, value-based assumptions that
  masquerade as science}.
\bjtitle{BioScience}
\bvolume{51}(\bissue{6}),
\bfpage{437}--\blpage{443}
(\byear{2001})
\end{barticle}
\endbibitem

\bibitem[\protect\citeauthoryear{Sicard et~al.}{2023}]{sicard2023trends}
\begin{barticle}
\bauthor{\bsnm{Sicard}, \binits{P.}},
\bauthor{\bsnm{Agathokleous}, \binits{E.}},
\bauthor{\bsnm{Anenberg}, \binits{S.C.}},
\bauthor{\bsnm{De~Marco}, \binits{A.}},
\bauthor{\bsnm{Paoletti}, \binits{E.}},
\bauthor{\bsnm{Calatayud}, \binits{V.}}:
\batitle{Trends in urban air pollution over the last two decades: A global
  perspective}.
\bjtitle{Science of The Total Environment}
\bvolume{858},
\bfpage{160064}
(\byear{2023})
\end{barticle}
\endbibitem

\bibitem[\protect\citeauthoryear{Wang et~al.}{2023}]{wang2023siamhrnet}
\begin{barticle}
\bauthor{\bsnm{Wang}, \binits{Z.}},
\bauthor{\bsnm{Liu}, \binits{D.}},
\bauthor{\bsnm{Liao}, \binits{X.}},
\bauthor{\bsnm{Pu}, \binits{W.}},
\bauthor{\bsnm{Wang}, \binits{Z.}},
\bauthor{\bsnm{Zhang}, \binits{Q.}}:
\batitle{Siamhrnet-ocr: A novel deforestation detection model with
  high-resolution imagery and deep learning}.
\bjtitle{Remote Sensing}
\bvolume{15}(\bissue{2}),
\bfpage{463}
(\byear{2023})
\end{barticle}
\endbibitem

\bibitem[\protect\citeauthoryear{Cauvy-Frauni{\'e} and
  Dangles}{2019}]{cauvy2019global}
\begin{barticle}
\bauthor{\bsnm{Cauvy-Frauni{\'e}}, \binits{S.}},
\bauthor{\bsnm{Dangles}, \binits{O.}}:
\batitle{A global synthesis of biodiversity responses to glacier retreat}.
\bjtitle{Nature Ecology \& Evolution}
\bvolume{3}(\bissue{12}),
\bfpage{1675}--\blpage{1685}
(\byear{2019})
\end{barticle}
\endbibitem

\bibitem[\protect\citeauthoryear{Chen et~al.}{2023}]{chen2023habitat}
\begin{barticle}
\bauthor{\bsnm{Chen}, \binits{X.}},
\bauthor{\bsnm{Yu}, \binits{L.}},
\bauthor{\bsnm{Cao}, \binits{Y.}},
\bauthor{\bsnm{Xu}, \binits{Y.}},
\bauthor{\bsnm{Zhao}, \binits{Z.}},
\bauthor{\bsnm{Zhuang}, \binits{Y.}},
\bauthor{\bsnm{Liu}, \binits{X.}},
\bauthor{\bsnm{Du}, \binits{Z.}},
\bauthor{\bsnm{Liu}, \binits{T.}},
\bauthor{\bsnm{Yang}, \binits{B.}}, \betal:
\batitle{Habitat quality dynamics in china's first group of national parks in
  recent four decades: Evidence from land use and land cover changes}.
\bjtitle{Journal of Environmental Management}
\bvolume{325},
\bfpage{116505}
(\byear{2023})
\end{barticle}
\endbibitem

\bibitem[\protect\citeauthoryear{Taylor et~al.}{2023}]{taylor2023associations}
\begin{botherref}
\oauthor{\bsnm{Taylor}, \binits{C.L.}},
\oauthor{\bsnm{Lydecker}, \binits{H.W.}},
\oauthor{\bsnm{Hochuli}, \binits{D.F.}},
\oauthor{\bsnm{Banks}, \binits{P.B.}}:
Associations between wildlife observations, human-tick encounters and landscape
  features in a peri-urban tick hotspot.
Urban Ecosystems,
1--16
(2023)
\end{botherref}
\endbibitem

\bibitem[\protect\citeauthoryear{Ge et~al.}{2023}]{ge2023mllm}
\begin{botherref}
\oauthor{\bsnm{Ge}, \binits{W.}},
\oauthor{\bsnm{Chen}, \binits{S.}},
\oauthor{\bsnm{Chen}, \binits{G.}},
\oauthor{\bsnm{Chen}, \binits{J.}},
\oauthor{\bsnm{Chen}, \binits{Z.}},
\oauthor{\bsnm{Yan}, \binits{S.}},
\oauthor{\bsnm{Zhu}, \binits{C.}},
\oauthor{\bsnm{Lin}, \binits{Z.}},
\oauthor{\bsnm{Xie}, \binits{W.}},
\oauthor{\bsnm{Wang}, \binits{X.}}, et al.:
Mllm-bench, evaluating multi-modal llms using gpt-4v.
arXiv preprint arXiv:2311.13951
(2023)
\end{botherref}
\endbibitem

\bibitem[\protect\citeauthoryear{Youvan}{2023}]{youvan2023interwoven}
\begin{botherref}
\oauthor{\bsnm{Youvan}, \binits{D.C.}}:
Interwoven realms: Parallels between gpt-4 neural mechanisms and hurricane
  dynamics
(2023)
\end{botherref}
\endbibitem

\bibitem[\protect\citeauthoryear{Ray and Brown}{2015}]{ray2015confronting}
\begin{bbook}
\bauthor{\bsnm{Ray}, \binits{P.A.}},
\bauthor{\bsnm{Brown}, \binits{C.M.}}:
\bbtitle{Confronting Climate Uncertainty in Water Resources Planning and
  Project Design: The Decision Tree Framework}.
\bpublisher{World Bank Publications}, \blocation{???}
(\byear{2015})
\end{bbook}
\endbibitem

\bibitem[\protect\citeauthoryear{Cai et~al.}{2018}]{cai2018high}
\begin{barticle}
\bauthor{\bsnm{Cai}, \binits{Y.}},
\bauthor{\bsnm{Guan}, \binits{K.}},
\bauthor{\bsnm{Peng}, \binits{J.}},
\bauthor{\bsnm{Wang}, \binits{S.}},
\bauthor{\bsnm{Seifert}, \binits{C.}},
\bauthor{\bsnm{Wardlow}, \binits{B.}},
\bauthor{\bsnm{Li}, \binits{Z.}}:
\batitle{A high-performance and in-season classification system of field-level
  crop types using time-series landsat data and a machine learning approach}.
\bjtitle{Remote sensing of environment}
\bvolume{210},
\bfpage{35}--\blpage{47}
(\byear{2018})
\end{barticle}
\endbibitem

\bibitem[\protect\citeauthoryear{Lu et~al.}{2022}]{lu2022fine}
\begin{barticle}
\bauthor{\bsnm{Lu}, \binits{T.}},
\bauthor{\bsnm{Wan}, \binits{L.}},
\bauthor{\bsnm{Wang}, \binits{L.}}:
\batitle{Fine crop classification in high resolution remote sensing based on
  deep learning}.
\bjtitle{Frontiers in Environmental Science}
\bvolume{10},
\bfpage{991173}
(\byear{2022})
\end{barticle}
\endbibitem

\bibitem[\protect\citeauthoryear{Chiu et~al.}{2020a}]{chiu20201st}
\begin{botherref}
\oauthor{\bsnm{Chiu}, \binits{M.T.}},
\oauthor{\bsnm{Xu}, \binits{X.}},
\oauthor{\bsnm{Wang}, \binits{K.}},
\oauthor{\bsnm{Hobbs}, \binits{J.}},
\oauthor{\bsnm{Hovakimyan}, \binits{N.}},
\oauthor{\bsnm{Huang}, \binits{T.S.}},
\oauthor{\bsnm{Shi}, \binits{H.}}, et al.:
The 1st agriculture-vision challenge: Methods and results.
arXiv preprint arXiv:2004.09754
(2020)
\end{botherref}
\endbibitem

\bibitem[\protect\citeauthoryear{Chiu et~al.}{2020b}]{Chiu_2020_CVPR_Workshops}
\begin{bchapter}
\bauthor{\bsnm{Chiu}, \binits{M.T.}},
\bauthor{\bsnm{Xu}, \binits{X.}},
\bauthor{\bsnm{Wang}, \binits{K.}},
\bauthor{\bsnm{Hobbs}, \binits{J.}},
\bauthor{\bsnm{Hovakimyan}, \binits{N.}},
\bauthor{\bsnm{Huang}, \binits{T.S.}},
\bauthor{\bsnm{Shi}, \binits{H.}}:
\bctitle{The 1st agriculture-vision challenge: Methods and results}.
In: \bbtitle{Proceedings of the IEEE/CVF Conference on Computer Vision and
  Pattern Recognition (CVPR) Workshops}
(\byear{2020})
\end{bchapter}
\endbibitem

\bibitem[\protect\citeauthoryear{McCauley et~al.}{2009}]{mccauley2009plant}
\begin{barticle}
\bauthor{\bsnm{McCauley}, \binits{A.}},
\bauthor{\bsnm{Jones}, \binits{C.}},
\bauthor{\bsnm{Jacobsen}, \binits{J.}}:
\batitle{Plant nutrient functions and deficiency and toxicity symptoms}.
\bjtitle{Nutrient management module}
\bvolume{9},
\bfpage{1}--\blpage{16}
(\byear{2009})
\end{barticle}
\endbibitem

\bibitem[\protect\citeauthoryear{Wulandhari et~al.}{2019}]{wulandhari2019plant}
\begin{barticle}
\bauthor{\bsnm{Wulandhari}, \binits{L.A.}},
\bauthor{\bsnm{Gunawan}, \binits{A.A.S.}},
\bauthor{\bsnm{Qurania}, \binits{A.}},
\bauthor{\bsnm{Harsani}, \binits{P.}},
\bauthor{\bsnm{Tarawan}, \binits{T.F.}},
\bauthor{\bsnm{Hermawan}, \binits{R.F.}}:
\batitle{Plant nutrient deficiency detection using deep convolutional neural
  network}.
\bjtitle{ICIC Express Lett}
\bvolume{13}(\bissue{10}),
\bfpage{971}--\blpage{977}
(\byear{2019})
\end{barticle}
\endbibitem

\bibitem[\protect\citeauthoryear{Feng et~al.}{2020}]{feng2020advances}
\begin{barticle}
\bauthor{\bsnm{Feng}, \binits{D.}},
\bauthor{\bsnm{Xu}, \binits{W.}},
\bauthor{\bsnm{He}, \binits{Z.}},
\bauthor{\bsnm{Zhao}, \binits{W.}},
\bauthor{\bsnm{Yang}, \binits{M.}}:
\batitle{Advances in plant nutrition diagnosis based on remote sensing and
  computer application}.
\bjtitle{Neural Computing and Applications}
\bvolume{32},
\bfpage{16833}--\blpage{16842}
(\byear{2020})
\end{barticle}
\endbibitem

\bibitem[\protect\citeauthoryear{Barbedo}{2019}]{barbedo2019detection}
\begin{barticle}
\bauthor{\bsnm{Barbedo}, \binits{J.G.A.}}:
\batitle{Detection of nutrition deficiencies in plants using proximal images
  and machine learning: A review}.
\bjtitle{Computers and Electronics in Agriculture}
\bvolume{162},
\bfpage{482}--\blpage{492}
(\byear{2019})
\end{barticle}
\endbibitem

\bibitem[\protect\citeauthoryear{Hasan et~al.}{2021}]{hasan2021survey}
\begin{barticle}
\bauthor{\bsnm{Hasan}, \binits{A.M.}},
\bauthor{\bsnm{Sohel}, \binits{F.}},
\bauthor{\bsnm{Diepeveen}, \binits{D.}},
\bauthor{\bsnm{Laga}, \binits{H.}},
\bauthor{\bsnm{Jones}, \binits{M.G.}}:
\batitle{A survey of deep learning techniques for weed detection from images}.
\bjtitle{Computers and Electronics in Agriculture}
\bvolume{184},
\bfpage{106067}
(\byear{2021})
\end{barticle}
\endbibitem

\bibitem[\protect\citeauthoryear{Li et~al.}{2021}]{li2021plant}
\begin{barticle}
\bauthor{\bsnm{Li}, \binits{L.}},
\bauthor{\bsnm{Zhang}, \binits{S.}},
\bauthor{\bsnm{Wang}, \binits{B.}}:
\batitle{Plant disease detection and classification by deep learning—a
  review}.
\bjtitle{IEEE Access}
\bvolume{9},
\bfpage{56683}--\blpage{56698}
(\byear{2021})
\end{barticle}
\endbibitem

\bibitem[\protect\citeauthoryear{Song et~al.}{2021}]{song2021high}
\begin{barticle}
\bauthor{\bsnm{Song}, \binits{P.}},
\bauthor{\bsnm{Wang}, \binits{J.}},
\bauthor{\bsnm{Guo}, \binits{X.}},
\bauthor{\bsnm{Yang}, \binits{W.}},
\bauthor{\bsnm{Zhao}, \binits{C.}}:
\batitle{High-throughput phenotyping: Breaking through the bottleneck in future
  crop breeding}.
\bjtitle{The Crop Journal}
\bvolume{9}(\bissue{3}),
\bfpage{633}--\blpage{645}
(\byear{2021})
\end{barticle}
\endbibitem

\bibitem[\protect\citeauthoryear{Yang et~al.}{2023}]{yang2023computer}
\begin{barticle}
\bauthor{\bsnm{Yang}, \binits{X.}},
\bauthor{\bsnm{Bist}, \binits{R.B.}},
\bauthor{\bsnm{Subedi}, \binits{S.}},
\bauthor{\bsnm{Chai}, \binits{L.}}:
\batitle{A computer vision-based automatic system for egg grading and defect
  detection}.
\bjtitle{Animals}
\bvolume{13}(\bissue{14}),
\bfpage{2354}
(\byear{2023})
\end{barticle}
\endbibitem

\bibitem[\protect\citeauthoryear{Subedi et~al.}{2023}]{subedi2023tracking}
\begin{barticle}
\bauthor{\bsnm{Subedi}, \binits{S.}},
\bauthor{\bsnm{Bist}, \binits{R.}},
\bauthor{\bsnm{Yang}, \binits{X.}},
\bauthor{\bsnm{Chai}, \binits{L.}}:
\batitle{Tracking pecking behaviors and damages of cage-free laying hens with
  machine vision technologies}.
\bjtitle{Computers and Electronics in Agriculture}
\bvolume{204},
\bfpage{107545}
(\byear{2023})
\end{barticle}
\endbibitem

\bibitem[\protect\citeauthoryear{Yin et~al.}{2019}]{yin2019nlp}
\begin{bchapter}
\bauthor{\bsnm{Yin}, \binits{Z.}},
\bauthor{\bsnm{Zhang}, \binits{C.}},
\bauthor{\bsnm{Goldberg}, \binits{D.W.}},
\bauthor{\bsnm{Prasad}, \binits{S.}}:
\bctitle{An nlp-based question answering framework for spatio-temporal analysis
  and visualization}.
In: \bbtitle{Proceedings of the 2019 2nd International Conference on
  Geoinformatics and Data Analysis},
pp. \bfpage{61}--\blpage{65}
(\byear{2019})
\end{bchapter}
\endbibitem

\bibitem[\protect\citeauthoryear{Feng et~al.}{2021}]{feng2021intelligent}
\begin{bchapter}
\bauthor{\bsnm{Feng}, \binits{X.}},
\bauthor{\bsnm{Liu}, \binits{Q.}},
\bauthor{\bsnm{Liu}, \binits{X.}}:
\bctitle{Intelligent question answering system based on knowledge graph}.
In: \bbtitle{2021 IEEE 23rd Int Conf on High Performance Computing \&
  Communications; 7th Int Conf on Data Science \& Systems; 19th Int Conf on
  Smart City; 7th Int Conf on Dependability in Sensor, Cloud \& Big Data
  Systems \& Application (HPCC/DSS/SmartCity/DependSys)},
pp. \bfpage{1515}--\blpage{1520}
(\byear{2021}).
\bcomment{IEEE}
\end{bchapter}
\endbibitem

\bibitem[\protect\citeauthoryear{Mart{\'\i}nez~JIm{\'e}nez}{2022}]{martinez2022development}
\begin{botherref}
\oauthor{\bsnm{Mart{\'\i}nez~JIm{\'e}nez}, \binits{R.}}:
Development of a question answering system for legal urban information
  retrieval.
PhD thesis,
Telecomunicacion
(2022)
\end{botherref}
\endbibitem

\bibitem[\protect\citeauthoryear{Dunham}{1958}]{dunham1958city}
\begin{barticle}
\bauthor{\bsnm{Dunham}, \binits{A.}}:
\batitle{City planning: an analysis of the content of the master plan}.
\bjtitle{The Journal of Law and Economics}
\bvolume{1},
\bfpage{170}--\blpage{186}
(\byear{1958})
\end{barticle}
\endbibitem

\bibitem[\protect\citeauthoryear{Peter and Yang}{2019}]{peter2019urban}
\begin{barticle}
\bauthor{\bsnm{Peter}, \binits{L.L.}},
\bauthor{\bsnm{Yang}, \binits{Y.}}:
\batitle{Urban planning historical review of master plans and the way towards a
  sustainable city: Dar es salaam, tanzania}.
\bjtitle{Frontiers of Architectural Research}
\bvolume{8}(\bissue{3}),
\bfpage{359}--\blpage{377}
(\byear{2019})
\end{barticle}
\endbibitem

\bibitem[\protect\citeauthoryear{Altschuler}{2018}]{altschuler2018goals}
\begin{bchapter}
\bauthor{\bsnm{Altschuler}, \binits{A.}}:
\bctitle{The goals of comprehensive planning}.
In: \bbtitle{Classic Readings in Urban Planning},
pp. \bfpage{67}--\blpage{84}.
\bpublisher{Routledge}, \blocation{???}
(\byear{2018})
\end{bchapter}
\endbibitem

\bibitem[\protect\citeauthoryear{Bevan et~al.}{2007}]{bevan2007sustainable}
\begin{bchapter}
\bauthor{\bsnm{Bevan}, \binits{T.A.}},
\bauthor{\bsnm{Sklenar}, \binits{O.}},
\bauthor{\bsnm{McKenzie}, \binits{J.A.}},
\bauthor{\bsnm{Derry}, \binits{W.E.}}:
\bctitle{Sustainable urban street design and assessment}.
In: \bbtitle{3rd Urban Street Symposium},
vol. \bseriesno{6}
(\byear{2007}).
\bcomment{Citeseer}
\end{bchapter}
\endbibitem

\bibitem[\protect\citeauthoryear{Hassen and
  Kaufman}{2016}]{hassen2016examining}
\begin{barticle}
\bauthor{\bsnm{Hassen}, \binits{N.}},
\bauthor{\bsnm{Kaufman}, \binits{P.}}:
\batitle{Examining the role of urban street design in enhancing community
  engagement: A literature review}.
\bjtitle{Health \& place}
\bvolume{41},
\bfpage{119}--\blpage{132}
(\byear{2016})
\end{barticle}
\endbibitem

\bibitem[\protect\citeauthoryear{Dover and Massengale}{2013}]{dover2013street}
\begin{bbook}
\bauthor{\bsnm{Dover}, \binits{V.}},
\bauthor{\bsnm{Massengale}, \binits{J.}}:
\bbtitle{Street Design: The Secret to Great Cities and Towns}.
\bpublisher{John Wiley \& Sons}, \blocation{???}
(\byear{2013})
\end{bbook}
\endbibitem

\bibitem[\protect\citeauthoryear{Liu
  et~al.}{2024}]{10.1007/978-3-031-45673-2_46}
\begin{bchapter}
\bauthor{\bsnm{Liu}, \binits{Z.}},
\bauthor{\bsnm{Zhong}, \binits{A.}},
\bauthor{\bsnm{Li}, \binits{Y.}},
\bauthor{\bsnm{Yang}, \binits{L.}},
\bauthor{\bsnm{Ju}, \binits{C.}},
\bauthor{\bsnm{Wu}, \binits{Z.}},
\bauthor{\bsnm{Ma}, \binits{C.}},
\bauthor{\bsnm{Shu}, \binits{P.}},
\bauthor{\bsnm{Chen}, \binits{C.}},
\bauthor{\bsnm{Kim}, \binits{S.}},
\bauthor{\bsnm{Dai}, \binits{H.}},
\bauthor{\bsnm{Zhao}, \binits{L.}},
\bauthor{\bsnm{Zhu}, \binits{D.}},
\bauthor{\bsnm{Liu}, \binits{J.}},
\bauthor{\bsnm{Liu}, \binits{W.}},
\bauthor{\bsnm{Shen}, \binits{D.}},
\bauthor{\bsnm{Li}, \binits{Q.}},
\bauthor{\bsnm{Liu}, \binits{T.}},
\bauthor{\bsnm{Li}, \binits{X.}}:
\bctitle{Tailoring large language models to radiology: A preliminary approach
  to llm adaptation for a highly specialized domain}.
In: \beditor{\bsnm{Cao}, \binits{X.}},
\beditor{\bsnm{Xu}, \binits{X.}},
\beditor{\bsnm{Rekik}, \binits{I.}},
\beditor{\bsnm{Cui}, \binits{Z.}},
\beditor{\bsnm{Ouyang}, \binits{X.}} (eds.)
\bbtitle{Machine Learning in Medical Imaging},
pp. \bfpage{464}--\blpage{473}.
\bpublisher{Springer},
\blocation{Cham}
(\byear{2024})
\end{bchapter}
\endbibitem

\bibitem[\protect\citeauthoryear{Liu et~al.}{2023}]{liu2023radiologyllama2}
\begin{botherref}
\oauthor{\bsnm{Liu}, \binits{Z.}},
\oauthor{\bsnm{Li}, \binits{Y.}},
\oauthor{\bsnm{Shu}, \binits{P.}},
\oauthor{\bsnm{Zhong}, \binits{A.}},
\oauthor{\bsnm{Yang}, \binits{L.}},
\oauthor{\bsnm{Ju}, \binits{C.}},
\oauthor{\bsnm{Wu}, \binits{Z.}},
\oauthor{\bsnm{Ma}, \binits{C.}},
\oauthor{\bsnm{Luo}, \binits{J.}},
\oauthor{\bsnm{Chen}, \binits{C.}},
\oauthor{\bsnm{Kim}, \binits{S.}},
\oauthor{\bsnm{Hu}, \binits{J.}},
\oauthor{\bsnm{Dai}, \binits{H.}},
\oauthor{\bsnm{Zhao}, \binits{L.}},
\oauthor{\bsnm{Zhu}, \binits{D.}},
\oauthor{\bsnm{Liu}, \binits{J.}},
\oauthor{\bsnm{Liu}, \binits{W.}},
\oauthor{\bsnm{Shen}, \binits{D.}},
\oauthor{\bsnm{Liu}, \binits{T.}},
\oauthor{\bsnm{Li}, \binits{Q.}},
\oauthor{\bsnm{Li}, \binits{X.}}:
Radiology-Llama2: Best-in-Class Large Language Model for Radiology
(2023)
\end{botherref}
\endbibitem

\bibitem[\protect\citeauthoryear{Tang et~al.}{2023}]{tang2023policygpt}
\begin{botherref}
\oauthor{\bsnm{Tang}, \binits{C.}},
\oauthor{\bsnm{Liu}, \binits{Z.}},
\oauthor{\bsnm{Ma}, \binits{C.}},
\oauthor{\bsnm{Wu}, \binits{Z.}},
\oauthor{\bsnm{Li}, \binits{Y.}},
\oauthor{\bsnm{Liu}, \binits{W.}},
\oauthor{\bsnm{Zhu}, \binits{D.}},
\oauthor{\bsnm{Li}, \binits{Q.}},
\oauthor{\bsnm{Li}, \binits{X.}},
\oauthor{\bsnm{Liu}, \binits{T.}},
\oauthor{\bsnm{Fan}, \binits{L.}}:
PolicyGPT: Automated Analysis of Privacy Policies with Large Language Models
(2023)
\end{botherref}
\endbibitem

\bibitem[\protect\citeauthoryear{Liu et~al.}{2023}]{liu2023radoncgpt}
\begin{botherref}
\oauthor{\bsnm{Liu}, \binits{Z.}},
\oauthor{\bsnm{Wang}, \binits{P.}},
\oauthor{\bsnm{Li}, \binits{Y.}},
\oauthor{\bsnm{Holmes}, \binits{J.}},
\oauthor{\bsnm{Shu}, \binits{P.}},
\oauthor{\bsnm{Zhang}, \binits{L.}},
\oauthor{\bsnm{Liu}, \binits{C.}},
\oauthor{\bsnm{Liu}, \binits{N.}},
\oauthor{\bsnm{Zhu}, \binits{D.}},
\oauthor{\bsnm{Li}, \binits{X.}},
\oauthor{\bsnm{Li}, \binits{Q.}},
\oauthor{\bsnm{Patel}, \binits{S.H.}},
\oauthor{\bsnm{Sio}, \binits{T.T.}},
\oauthor{\bsnm{Liu}, \binits{T.}},
\oauthor{\bsnm{Liu}, \binits{W.}}:
RadOnc-GPT: A Large Language Model for Radiation Oncology
(2023)
\end{botherref}
\endbibitem

\bibitem[\protect\citeauthoryear{Liu et~al.}{}]{liuradiology}
\begin{botherref}
\oauthor{\bsnm{Liu}, \binits{Z.}},
\oauthor{\bsnm{Zhong}, \binits{A.}},
\oauthor{\bsnm{Li}, \binits{Y.}},
\oauthor{\bsnm{Yang}, \binits{L.}},
\oauthor{\bsnm{Ju}, \binits{C.}},
\oauthor{\bsnm{Wu}, \binits{Z.}}, et al.:
Radiology-GPT: a large language model for radiology. arXiv [Preprint]. 2023
  [cited August 21, 2023]
\end{botherref}
\endbibitem

\bibitem[\protect\citeauthoryear{Liu et~al.}{2023}]{LIU2023100045}
\begin{botherref}
\oauthor{\bsnm{Liu}, \binits{C.}},
\oauthor{\bsnm{Liu}, \binits{Z.}},
\oauthor{\bsnm{Holmes}, \binits{J.}},
\oauthor{\bsnm{Zhang}, \binits{L.}},
\oauthor{\bsnm{Zhang}, \binits{L.}},
\oauthor{\bsnm{Ding}, \binits{Y.}},
\oauthor{\bsnm{Shu}, \binits{P.}},
\oauthor{\bsnm{Wu}, \binits{Z.}},
\oauthor{\bsnm{Dai}, \binits{H.}},
\oauthor{\bsnm{Li}, \binits{Y.}},
\oauthor{\bsnm{Shen}, \binits{D.}},
\oauthor{\bsnm{Liu}, \binits{N.}},
\oauthor{\bsnm{Li}, \binits{Q.}},
\oauthor{\bsnm{Li}, \binits{X.}},
\oauthor{\bsnm{Zhu}, \binits{D.}},
\oauthor{\bsnm{Liu}, \binits{T.}},
\oauthor{\bsnm{Liu}, \binits{W.}}:
Artificial general intelligence for radiation oncology.
Meta-Radiology,
100045
(2023)
\doiurl{10.1016/j.metrad.2023.100045}
\end{botherref}
\endbibitem

\bibitem[\protect\citeauthoryear{Dou et~al.}{2023}]{dou2023artificial}
\begin{botherref}
\oauthor{\bsnm{Dou}, \binits{F.}},
\oauthor{\bsnm{Ye}, \binits{J.}},
\oauthor{\bsnm{Yuan}, \binits{G.}},
\oauthor{\bsnm{Lu}, \binits{Q.}},
\oauthor{\bsnm{Niu}, \binits{W.}},
\oauthor{\bsnm{Sun}, \binits{H.}},
\oauthor{\bsnm{Guan}, \binits{L.}},
\oauthor{\bsnm{Lu}, \binits{G.}},
\oauthor{\bsnm{Mai}, \binits{G.}},
\oauthor{\bsnm{Liu}, \binits{N.}},
\oauthor{\bsnm{Lu}, \binits{J.}},
\oauthor{\bsnm{Liu}, \binits{Z.}},
\oauthor{\bsnm{Wu}, \binits{Z.}},
\oauthor{\bsnm{Tan}, \binits{C.}},
\oauthor{\bsnm{Xu}, \binits{S.}},
\oauthor{\bsnm{Wang}, \binits{X.}},
\oauthor{\bsnm{Li}, \binits{G.}},
\oauthor{\bsnm{Chai}, \binits{L.}},
\oauthor{\bsnm{Li}, \binits{S.}},
\oauthor{\bsnm{Sun}, \binits{J.}},
\oauthor{\bsnm{Sun}, \binits{H.}},
\oauthor{\bsnm{Shao}, \binits{Y.}},
\oauthor{\bsnm{Li}, \binits{C.}},
\oauthor{\bsnm{Liu}, \binits{T.}},
\oauthor{\bsnm{Song}, \binits{W.}}:
Towards Artificial General Intelligence (AGI) in the Internet of Things (IoT):
  Opportunities and Challenges
(2023)
\end{botherref}
\endbibitem

\bibitem[\protect\citeauthoryear{Holmes et~al.}{2023}]{holmes2023evaluating}
\begin{botherref}
\oauthor{\bsnm{Holmes}, \binits{J.}},
\oauthor{\bsnm{Liu}, \binits{Z.}},
\oauthor{\bsnm{Zhang}, \binits{L.}},
\oauthor{\bsnm{Ding}, \binits{Y.}},
\oauthor{\bsnm{Sio}, \binits{T.}},
\oauthor{\bsnm{McGee}, \binits{L.}},
\oauthor{\bsnm{Ashman}, \binits{J.}},
\oauthor{\bsnm{Li}, \binits{X.}},
\oauthor{\bsnm{Liu}, \binits{T.}},
\oauthor{\bsnm{Shen}, \binits{J.}}, et al.:
Evaluating large language models on a highly-specialized topic.
Radiation Oncology Physics
(2023)
\end{botherref}
\endbibitem

\bibitem[\protect\citeauthoryear{Gong et~al.}{2023}]{gong2023evaluating}
\begin{botherref}
\oauthor{\bsnm{Gong}, \binits{X.}},
\oauthor{\bsnm{Holmes}, \binits{J.}},
\oauthor{\bsnm{Li}, \binits{Y.}},
\oauthor{\bsnm{Liu}, \binits{Z.}},
\oauthor{\bsnm{Gan}, \binits{Q.}},
\oauthor{\bsnm{Wu}, \binits{Z.}},
\oauthor{\bsnm{Zhang}, \binits{J.}},
\oauthor{\bsnm{Zou}, \binits{Y.}},
\oauthor{\bsnm{Teng}, \binits{Y.}},
\oauthor{\bsnm{Jiang}, \binits{T.}},
\oauthor{\bsnm{Zhu}, \binits{H.}},
\oauthor{\bsnm{Liu}, \binits{W.}},
\oauthor{\bsnm{Liu}, \binits{T.}},
\oauthor{\bsnm{Yan}, \binits{Y.}}:
Evaluating the Potential of Leading Large Language Models in Reasoning Biology
  Questions
(2023)
\end{botherref}
\endbibitem

\bibitem[\protect\citeauthoryear{Holmes et~al.}{2023}]{holmes2023benchmarking}
\begin{botherref}
\oauthor{\bsnm{Holmes}, \binits{J.}},
\oauthor{\bsnm{Zhang}, \binits{L.}},
\oauthor{\bsnm{Ding}, \binits{Y.}},
\oauthor{\bsnm{Feng}, \binits{H.}},
\oauthor{\bsnm{Liu}, \binits{Z.}},
\oauthor{\bsnm{Liu}, \binits{T.}},
\oauthor{\bsnm{Wong}, \binits{W.W.}},
\oauthor{\bsnm{Vora}, \binits{S.A.}},
\oauthor{\bsnm{Ashman}, \binits{J.B.}},
\oauthor{\bsnm{Liu}, \binits{W.}}:
Benchmarking a foundation LLM on its ability to re-label structure names in
  accordance with the AAPM TG-263 report
(2023)
\end{botherref}
\endbibitem

\bibitem[\protect\citeauthoryear{Shi et~al.}{2023}]{shi2023mededit}
\begin{botherref}
\oauthor{\bsnm{Shi}, \binits{Y.}},
\oauthor{\bsnm{Xu}, \binits{S.}},
\oauthor{\bsnm{Liu}, \binits{Z.}},
\oauthor{\bsnm{Liu}, \binits{T.}},
\oauthor{\bsnm{Li}, \binits{X.}},
\oauthor{\bsnm{Liu}, \binits{N.}}:
MedEdit: Model Editing for Medical Question Answering with External Knowledge
  Bases
(2023)
\end{botherref}
\endbibitem

\bibitem[\protect\citeauthoryear{Zhong et~al.}{2023}]{zhong2023chatabl}
\begin{botherref}
\oauthor{\bsnm{Zhong}, \binits{T.}},
\oauthor{\bsnm{Wei}, \binits{Y.}},
\oauthor{\bsnm{Yang}, \binits{L.}},
\oauthor{\bsnm{Wu}, \binits{Z.}},
\oauthor{\bsnm{Liu}, \binits{Z.}},
\oauthor{\bsnm{Wei}, \binits{X.}},
\oauthor{\bsnm{Li}, \binits{W.}},
\oauthor{\bsnm{Yao}, \binits{J.}},
\oauthor{\bsnm{Ma}, \binits{C.}},
\oauthor{\bsnm{Li}, \binits{X.}}, et al.:
Chatabl: Abductive learning via natural language interaction with chatgpt.
arXiv preprint arXiv:2304.11107
(2023)
\end{botherref}
\endbibitem

\bibitem[\protect\citeauthoryear{Zhao et~al.}{2023}]{zhao2023generic}
\begin{barticle}
\bauthor{\bsnm{Zhao}, \binits{L.}},
\bauthor{\bsnm{Wu}, \binits{Z.}},
\bauthor{\bsnm{Dai}, \binits{H.}},
\bauthor{\bsnm{Liu}, \binits{Z.}},
\bauthor{\bsnm{Hu}, \binits{X.}},
\bauthor{\bsnm{Zhang}, \binits{T.}},
\bauthor{\bsnm{Zhu}, \binits{D.}},
\bauthor{\bsnm{Liu}, \binits{T.}}:
\batitle{A generic framework for embedding human brain function with temporally
  correlated autoencoder}.
\bjtitle{Medical Image Analysis}
\bvolume{89},
\bfpage{102892}
(\byear{2023})
\end{barticle}
\endbibitem

\bibitem[\protect\citeauthoryear{Zhou et~al.}{2023}]{zhou2023fine}
\begin{bchapter}
\bauthor{\bsnm{Zhou}, \binits{M.}},
\bauthor{\bsnm{Liu}, \binits{X.}},
\bauthor{\bsnm{Liu}, \binits{D.}},
\bauthor{\bsnm{Wu}, \binits{Z.}},
\bauthor{\bsnm{Liu}, \binits{Z.}},
\bauthor{\bsnm{Zhao}, \binits{L.}},
\bauthor{\bsnm{Zhu}, \binits{D.}},
\bauthor{\bsnm{Guo}, \binits{L.}},
\bauthor{\bsnm{Han}, \binits{J.}},
\bauthor{\bsnm{Liu}, \binits{T.}}, \betal:
\bctitle{Fine-grained artificial neurons in audio-transformers for
  disentangling neural auditory encoding}.
In: \bbtitle{Findings of the Association for Computational Linguistics: ACL
  2023},
pp. \bfpage{7943}--\blpage{7956}
(\byear{2023})
\end{bchapter}
\endbibitem

\bibitem[\protect\citeauthoryear{Liao et~al.}{2023}]{liao2023mask}
\begin{botherref}
\oauthor{\bsnm{Liao}, \binits{W.}},
\oauthor{\bsnm{Liu}, \binits{Z.}},
\oauthor{\bsnm{Dai}, \binits{H.}},
\oauthor{\bsnm{Wu}, \binits{Z.}},
\oauthor{\bsnm{Zhang}, \binits{Y.}},
\oauthor{\bsnm{Huang}, \binits{X.}},
\oauthor{\bsnm{Chen}, \binits{Y.}},
\oauthor{\bsnm{Jiang}, \binits{X.}},
\oauthor{\bsnm{Zhu}, \binits{D.}},
\oauthor{\bsnm{Liu}, \binits{T.}}, et al.:
Mask-guided bert for few shot text classification.
arXiv preprint arXiv:2302.10447
(2023)
\end{botherref}
\endbibitem

\bibitem[\protect\citeauthoryear{Devlin et~al.}{2018}]{devlin2018bert}
\begin{botherref}
\oauthor{\bsnm{Devlin}, \binits{J.}},
\oauthor{\bsnm{Chang}, \binits{M.-W.}},
\oauthor{\bsnm{Lee}, \binits{K.}},
\oauthor{\bsnm{Toutanova}, \binits{K.}}:
Bert: Pre-training of deep bidirectional transformers for language
  understanding.
arXiv preprint arXiv:1810.04805
(2018)
\end{botherref}
\endbibitem

\bibitem[\protect\citeauthoryear{Liu et~al.}{2019}]{liu2019roberta}
\begin{botherref}
\oauthor{\bsnm{Liu}, \binits{Y.}},
\oauthor{\bsnm{Ott}, \binits{M.}},
\oauthor{\bsnm{Goyal}, \binits{N.}},
\oauthor{\bsnm{Du}, \binits{J.}},
\oauthor{\bsnm{Joshi}, \binits{M.}},
\oauthor{\bsnm{Chen}, \binits{D.}},
\oauthor{\bsnm{Levy}, \binits{O.}},
\oauthor{\bsnm{Lewis}, \binits{M.}},
\oauthor{\bsnm{Zettlemoyer}, \binits{L.}},
\oauthor{\bsnm{Stoyanov}, \binits{V.}}:
Roberta: A robustly optimized bert pretraining approach.
arXiv preprint arXiv:1907.11692
(2019)
\end{botherref}
\endbibitem

\bibitem[\protect\citeauthoryear{Rezayi
  et~al.}{2022}]{rezayi2022clinicalradiobert}
\begin{bchapter}
\bauthor{\bsnm{Rezayi}, \binits{S.}},
\bauthor{\bsnm{Dai}, \binits{H.}},
\bauthor{\bsnm{Liu}, \binits{Z.}},
\bauthor{\bsnm{Wu}, \binits{Z.}},
\bauthor{\bsnm{Hebbar}, \binits{A.}},
\bauthor{\bsnm{Burns}, \binits{A.H.}},
\bauthor{\bsnm{Zhao}, \binits{L.}},
\bauthor{\bsnm{Zhu}, \binits{D.}},
\bauthor{\bsnm{Li}, \binits{Q.}},
\bauthor{\bsnm{Liu}, \binits{W.}}, \betal:
\bctitle{Clinicalradiobert: Knowledge-infused few shot learning for clinical
  notes named entity recognition}.
In: \bbtitle{International Workshop on Machine Learning in Medical Imaging},
pp. \bfpage{269}--\blpage{278}
(\byear{2022}).
\bcomment{Springer}
\end{bchapter}
\endbibitem

\bibitem[\protect\citeauthoryear{Lee et~al.}{2020}]{lee2020biobert}
\begin{barticle}
\bauthor{\bsnm{Lee}, \binits{J.}},
\bauthor{\bsnm{Yoon}, \binits{W.}},
\bauthor{\bsnm{Kim}, \binits{S.}},
\bauthor{\bsnm{Kim}, \binits{D.}},
\bauthor{\bsnm{Kim}, \binits{S.}},
\bauthor{\bsnm{So}, \binits{C.H.}},
\bauthor{\bsnm{Kang}, \binits{J.}}:
\batitle{Biobert: a pre-trained biomedical language representation model for
  biomedical text mining}.
\bjtitle{Bioinformatics}
\bvolume{36}(\bissue{4}),
\bfpage{1234}--\blpage{1240}
(\byear{2020})
\end{barticle}
\endbibitem

\bibitem[\protect\citeauthoryear{Brown et~al.}{2020}]{brown2020language}
\begin{barticle}
\bauthor{\bsnm{Brown}, \binits{T.}},
\bauthor{\bsnm{Mann}, \binits{B.}},
\bauthor{\bsnm{Ryder}, \binits{N.}},
\bauthor{\bsnm{Subbiah}, \binits{M.}},
\bauthor{\bsnm{Kaplan}, \binits{J.D.}},
\bauthor{\bsnm{Dhariwal}, \binits{P.}},
\bauthor{\bsnm{Neelakantan}, \binits{A.}},
\bauthor{\bsnm{Shyam}, \binits{P.}},
\bauthor{\bsnm{Sastry}, \binits{G.}},
\bauthor{\bsnm{Askell}, \binits{A.}}, \betal:
\batitle{Language models are few-shot learners}.
\bjtitle{Advances in neural information processing systems}
\bvolume{33},
\bfpage{1877}--\blpage{1901}
(\byear{2020})
\end{barticle}
\endbibitem

\bibitem[\protect\citeauthoryear{Ouyang et~al.}{2022}]{ouyang2022instructgpt}
\begin{barticle}
\bauthor{\bsnm{Ouyang}, \binits{L.}},
\bauthor{\bsnm{Wu}, \binits{J.}},
\bauthor{\bsnm{Jiang}, \binits{X.}},
\bauthor{\bsnm{Almeida}, \binits{D.}},
\bauthor{\bsnm{Wainwright}, \binits{C.}},
\bauthor{\bsnm{Mishkin}, \binits{P.}},
\bauthor{\bsnm{Zhang}, \binits{C.}},
\bauthor{\bsnm{Agarwal}, \binits{S.}},
\bauthor{\bsnm{Slama}, \binits{K.}},
\bauthor{\bsnm{Ray}, \binits{A.}}, \betal:
\batitle{Training language models to follow instructions with human feedback}.
\bjtitle{Advances in Neural Information Processing Systems}
\bvolume{35},
\bfpage{27730}--\blpage{27744}
(\byear{2022})
\end{barticle}
\endbibitem

\bibitem[\protect\citeauthoryear{OpenAI}{}]{openaiIntroducingChatGPT}
\begin{botherref}
\oauthor{\bsnm{OpenAI}}:
{I}ntroducing {C}hat{G}{P}{T} --- openai.com.
\url{https://openai.com/blog/chatgpt}.
[Accessed 17-11-2023]
\end{botherref}
\endbibitem

\bibitem[\protect\citeauthoryear{OpenAI}{2023}]{openai2023gpt4}
\begin{botherref}
\oauthor{\bsnm{OpenAI}}:
GPT-4 Technical Report
(2023)
\end{botherref}
\endbibitem

\bibitem[\protect\citeauthoryear{Rezayi
  et~al.}{2022}]{10.1007/978-3-031-21014-3_28}
\begin{bchapter}
\bauthor{\bsnm{Rezayi}, \binits{S.}},
\bauthor{\bsnm{Dai}, \binits{H.}},
\bauthor{\bsnm{Liu}, \binits{Z.}},
\bauthor{\bsnm{Wu}, \binits{Z.}},
\bauthor{\bsnm{Hebbar}, \binits{A.}},
\bauthor{\bsnm{Burns}, \binits{A.H.}},
\bauthor{\bsnm{Zhao}, \binits{L.}},
\bauthor{\bsnm{Zhu}, \binits{D.}},
\bauthor{\bsnm{Li}, \binits{Q.}},
\bauthor{\bsnm{Liu}, \binits{W.}},
\bauthor{\bsnm{Li}, \binits{S.}},
\bauthor{\bsnm{Liu}, \binits{T.}},
\bauthor{\bsnm{Li}, \binits{X.}}:
\bctitle{Clinicalradiobert: Knowledge-infused few shot learning for clinical
  notes named entity recognition}.
In: \beditor{\bsnm{Lian}, \binits{C.}},
\beditor{\bsnm{Cao}, \binits{X.}},
\beditor{\bsnm{Rekik}, \binits{I.}},
\beditor{\bsnm{Xu}, \binits{X.}},
\beditor{\bsnm{Cui}, \binits{Z.}} (eds.)
\bbtitle{Machine Learning in Medical Imaging},
pp. \bfpage{269}--\blpage{278}.
\bpublisher{Springer},
\blocation{Cham}
(\byear{2022})
\end{bchapter}
\endbibitem

\bibitem[\protect\citeauthoryear{Liu et~al.}{2022}]{PMID:36097765}
\begin{barticle}
\bauthor{\bsnm{Liu}, \binits{Z.}},
\bauthor{\bsnm{He}, \binits{M.}},
\bauthor{\bsnm{Jiang}, \binits{Z.}},
\bauthor{\bsnm{Wu}, \binits{Z.}},
\bauthor{\bsnm{Dai}, \binits{H.}},
\bauthor{\bsnm{Zhang}, \binits{L.}},
\bauthor{\bsnm{Luo}, \binits{S.}},
\bauthor{\bsnm{Han}, \binits{T.}},
\bauthor{\bsnm{Li}, \binits{X.}},
\bauthor{\bsnm{Jiang}, \binits{X.}},
\bauthor{\bsnm{Zhu}, \binits{D.}},
\bauthor{\bsnm{Cai}, \binits{X.}},
\bauthor{\bsnm{Ge}, \binits{B.}},
\bauthor{\bsnm{Liu}, \binits{W.}},
\bauthor{\bsnm{Liu}, \binits{J.}},
\bauthor{\bsnm{Shen}, \binits{D.}},
\bauthor{\bsnm{Liu}, \binits{T.}}:
\batitle{Survey on natural language processing in medical image analysis}.
\bjtitle{Zhong nan da xue xue bao. Yi xue ban = Journal of Central South
  University. Medical sciences}
\bvolume{47}(\bissue{8}),
\bfpage{981}--\blpage{993}
(\byear{2022})
\doiurl{10.11817/j.issn.1672-7347.2022.220376}
\end{barticle}
\endbibitem

\bibitem[\protect\citeauthoryear{Liao et~al.}{2023}]{liao2023maskguided}
\begin{botherref}
\oauthor{\bsnm{Liao}, \binits{W.}},
\oauthor{\bsnm{Liu}, \binits{Z.}},
\oauthor{\bsnm{Dai}, \binits{H.}},
\oauthor{\bsnm{Wu}, \binits{Z.}},
\oauthor{\bsnm{Zhang}, \binits{Y.}},
\oauthor{\bsnm{Huang}, \binits{X.}},
\oauthor{\bsnm{Chen}, \binits{Y.}},
\oauthor{\bsnm{Jiang}, \binits{X.}},
\oauthor{\bsnm{Liu}, \binits{W.}},
\oauthor{\bsnm{Zhu}, \binits{D.}},
\oauthor{\bsnm{Liu}, \binits{T.}},
\oauthor{\bsnm{Li}, \binits{S.}},
\oauthor{\bsnm{Li}, \binits{X.}},
\oauthor{\bsnm{Cai}, \binits{H.}}:
Mask-guided BERT for Few Shot Text Classification
(2023)
\end{botherref}
\endbibitem

\bibitem[\protect\citeauthoryear{Dai et~al.}{2023}]{dai2023auggpt}
\begin{botherref}
\oauthor{\bsnm{Dai}, \binits{H.}},
\oauthor{\bsnm{Liu}, \binits{Z.}},
\oauthor{\bsnm{Liao}, \binits{W.}},
\oauthor{\bsnm{Huang}, \binits{X.}},
\oauthor{\bsnm{Cao}, \binits{Y.}},
\oauthor{\bsnm{Wu}, \binits{Z.}},
\oauthor{\bsnm{Zhao}, \binits{L.}},
\oauthor{\bsnm{Xu}, \binits{S.}},
\oauthor{\bsnm{Liu}, \binits{W.}},
\oauthor{\bsnm{Liu}, \binits{N.}},
\oauthor{\bsnm{Li}, \binits{S.}},
\oauthor{\bsnm{Zhu}, \binits{D.}},
\oauthor{\bsnm{Cai}, \binits{H.}},
\oauthor{\bsnm{Sun}, \binits{L.}},
\oauthor{\bsnm{Li}, \binits{Q.}},
\oauthor{\bsnm{Shen}, \binits{D.}},
\oauthor{\bsnm{Liu}, \binits{T.}},
\oauthor{\bsnm{Li}, \binits{X.}}:
AugGPT: Leveraging ChatGPT for Text Data Augmentation
(2023)
\end{botherref}
\endbibitem

\bibitem[\protect\citeauthoryear{Liu et~al.}{2023}]{liu2023deidgpt}
\begin{botherref}
\oauthor{\bsnm{Liu}, \binits{Z.}},
\oauthor{\bsnm{Yu}, \binits{X.}},
\oauthor{\bsnm{Zhang}, \binits{L.}},
\oauthor{\bsnm{Wu}, \binits{Z.}},
\oauthor{\bsnm{Cao}, \binits{C.}},
\oauthor{\bsnm{Dai}, \binits{H.}},
\oauthor{\bsnm{Zhao}, \binits{L.}},
\oauthor{\bsnm{Liu}, \binits{W.}},
\oauthor{\bsnm{Shen}, \binits{D.}},
\oauthor{\bsnm{Li}, \binits{Q.}},
\oauthor{\bsnm{Liu}, \binits{T.}},
\oauthor{\bsnm{Zhu}, \binits{D.}},
\oauthor{\bsnm{Li}, \binits{X.}}:
DeID-GPT: Zero-shot Medical Text De-Identification by GPT-4
(2023)
\end{botherref}
\endbibitem

\bibitem[\protect\citeauthoryear{Ma et~al.}{2023}]{ma2023impressiongpt}
\begin{botherref}
\oauthor{\bsnm{Ma}, \binits{C.}},
\oauthor{\bsnm{Wu}, \binits{Z.}},
\oauthor{\bsnm{Wang}, \binits{J.}},
\oauthor{\bsnm{Xu}, \binits{S.}},
\oauthor{\bsnm{Wei}, \binits{Y.}},
\oauthor{\bsnm{Liu}, \binits{Z.}},
\oauthor{\bsnm{Jiang}, \binits{X.}},
\oauthor{\bsnm{Guo}, \binits{L.}},
\oauthor{\bsnm{Cai}, \binits{X.}},
\oauthor{\bsnm{Zhang}, \binits{S.}},
\oauthor{\bsnm{Zhang}, \binits{T.}},
\oauthor{\bsnm{Zhu}, \binits{D.}},
\oauthor{\bsnm{Shen}, \binits{D.}},
\oauthor{\bsnm{Liu}, \binits{T.}},
\oauthor{\bsnm{Li}, \binits{X.}}:
ImpressionGPT: An Iterative Optimizing Framework for Radiology Report
  Summarization with ChatGPT
(2023)
\end{botherref}
\endbibitem

\bibitem[\protect\citeauthoryear{Liao et~al.}{2023}]{liao2023differentiate}
\begin{botherref}
\oauthor{\bsnm{Liao}, \binits{W.}},
\oauthor{\bsnm{Liu}, \binits{Z.}},
\oauthor{\bsnm{Dai}, \binits{H.}},
\oauthor{\bsnm{Xu}, \binits{S.}},
\oauthor{\bsnm{Wu}, \binits{Z.}},
\oauthor{\bsnm{Zhang}, \binits{Y.}},
\oauthor{\bsnm{Huang}, \binits{X.}},
\oauthor{\bsnm{Zhu}, \binits{D.}},
\oauthor{\bsnm{Cai}, \binits{H.}},
\oauthor{\bsnm{Liu}, \binits{T.}},
\oauthor{\bsnm{Li}, \binits{X.}}:
Differentiate ChatGPT-generated and Human-written Medical Texts
(2023)
\end{botherref}
\endbibitem

\bibitem[\protect\citeauthoryear{Dai et~al.}{2023}]{dai2023adautogpt}
\begin{botherref}
\oauthor{\bsnm{Dai}, \binits{H.}},
\oauthor{\bsnm{Li}, \binits{Y.}},
\oauthor{\bsnm{Liu}, \binits{Z.}},
\oauthor{\bsnm{Zhao}, \binits{L.}},
\oauthor{\bsnm{Wu}, \binits{Z.}},
\oauthor{\bsnm{Song}, \binits{S.}},
\oauthor{\bsnm{Shen}, \binits{Y.}},
\oauthor{\bsnm{Zhu}, \binits{D.}},
\oauthor{\bsnm{Li}, \binits{X.}},
\oauthor{\bsnm{Li}, \binits{S.}},
\oauthor{\bsnm{Yao}, \binits{X.}},
\oauthor{\bsnm{Shi}, \binits{L.}},
\oauthor{\bsnm{Li}, \binits{Q.}},
\oauthor{\bsnm{Chen}, \binits{Z.}},
\oauthor{\bsnm{Zhang}, \binits{D.}},
\oauthor{\bsnm{Mai}, \binits{G.}},
\oauthor{\bsnm{Liu}, \binits{T.}}:
AD-AutoGPT: An Autonomous GPT for Alzheimer's Disease Infodemiology
(2023)
\end{botherref}
\endbibitem

\bibitem[\protect\citeauthoryear{Guan et~al.}{2023}]{guan2023cohortgpt}
\begin{botherref}
\oauthor{\bsnm{Guan}, \binits{Z.}},
\oauthor{\bsnm{Wu}, \binits{Z.}},
\oauthor{\bsnm{Liu}, \binits{Z.}},
\oauthor{\bsnm{Wu}, \binits{D.}},
\oauthor{\bsnm{Ren}, \binits{H.}},
\oauthor{\bsnm{Li}, \binits{Q.}},
\oauthor{\bsnm{Li}, \binits{X.}},
\oauthor{\bsnm{Liu}, \binits{N.}}:
CohortGPT: An Enhanced GPT for Participant Recruitment in Clinical Study
(2023)
\end{botherref}
\endbibitem

\bibitem[\protect\citeauthoryear{Cai et~al.}{2022}]{cai2022coarse}
\begin{botherref}
\oauthor{\bsnm{Cai}, \binits{H.}},
\oauthor{\bsnm{Liao}, \binits{W.}},
\oauthor{\bsnm{Liu}, \binits{Z.}},
\oauthor{\bsnm{Zhang}, \binits{Y.}},
\oauthor{\bsnm{Huang}, \binits{X.}},
\oauthor{\bsnm{Ding}, \binits{S.}},
\oauthor{\bsnm{Ren}, \binits{H.}},
\oauthor{\bsnm{Wu}, \binits{Z.}},
\oauthor{\bsnm{Dai}, \binits{H.}},
\oauthor{\bsnm{Li}, \binits{S.}}, et al.:
Coarse-to-fine knowledge graph domain adaptation based on distantly-supervised
  iterative training.
arXiv preprint arXiv:2211.02849
(2022)
\end{botherref}
\endbibitem

\bibitem[\protect\citeauthoryear{Liu et~al.}{2023}]{liu2023pharmacygpt}
\begin{botherref}
\oauthor{\bsnm{Liu}, \binits{Z.}},
\oauthor{\bsnm{Wu}, \binits{Z.}},
\oauthor{\bsnm{Hu}, \binits{M.}},
\oauthor{\bsnm{Zhao}, \binits{B.}},
\oauthor{\bsnm{Zhao}, \binits{L.}},
\oauthor{\bsnm{Zhang}, \binits{T.}},
\oauthor{\bsnm{Dai}, \binits{H.}},
\oauthor{\bsnm{Chen}, \binits{X.}},
\oauthor{\bsnm{Shen}, \binits{Y.}},
\oauthor{\bsnm{Li}, \binits{S.}},
\oauthor{\bsnm{Murray}, \binits{B.}},
\oauthor{\bsnm{Liu}, \binits{T.}},
\oauthor{\bsnm{Sikora}, \binits{A.}}:
PharmacyGPT: The AI Pharmacist
(2023)
\end{botherref}
\endbibitem

\bibitem[\protect\citeauthoryear{Zhao et~al.}{2023}]{ZHAO2023100005}
\begin{barticle}
\bauthor{\bsnm{Zhao}, \binits{L.}},
\bauthor{\bsnm{Zhang}, \binits{L.}},
\bauthor{\bsnm{Wu}, \binits{Z.}},
\bauthor{\bsnm{Chen}, \binits{Y.}},
\bauthor{\bsnm{Dai}, \binits{H.}},
\bauthor{\bsnm{Yu}, \binits{X.}},
\bauthor{\bsnm{Liu}, \binits{Z.}},
\bauthor{\bsnm{Zhang}, \binits{T.}},
\bauthor{\bsnm{Hu}, \binits{X.}},
\bauthor{\bsnm{Jiang}, \binits{X.}},
\bauthor{\bsnm{Li}, \binits{X.}},
\bauthor{\bsnm{Zhu}, \binits{D.}},
\bauthor{\bsnm{Shen}, \binits{D.}},
\bauthor{\bsnm{Liu}, \binits{T.}}:
\batitle{When brain-inspired ai meets agi}.
\bjtitle{Meta-Radiology}
\bvolume{1}(\bissue{1}),
\bfpage{100005}
(\byear{2023})
\doiurl{10.1016/j.metrad.2023.100005}
\end{barticle}
\endbibitem

\bibitem[\protect\citeauthoryear{Holmes et~al.}{2023}]{Holmes_2023}
\begin{botherref}
\oauthor{\bsnm{Holmes}, \binits{J.}},
\oauthor{\bsnm{Liu}, \binits{Z.}},
\oauthor{\bsnm{Zhang}, \binits{L.}},
\oauthor{\bsnm{Ding}, \binits{Y.}},
\oauthor{\bsnm{Sio}, \binits{T.T.}},
\oauthor{\bsnm{McGee}, \binits{L.A.}},
\oauthor{\bsnm{Ashman}, \binits{J.B.}},
\oauthor{\bsnm{Li}, \binits{X.}},
\oauthor{\bsnm{Liu}, \binits{T.}},
\oauthor{\bsnm{Shen}, \binits{J.}},
\oauthor{\bsnm{Liu}, \binits{W.}}:
Evaluating large language models on a highly-specialized topic, radiation
  oncology physics.
Frontiers in Oncology
\textbf{13}
(2023)
\doiurl{10.3389/fonc.2023.1219326}
\end{botherref}
\endbibitem

\bibitem[\protect\citeauthoryear{Wu et~al.}{2023}]{wu2023exploring}
\begin{botherref}
\oauthor{\bsnm{Wu}, \binits{Z.}},
\oauthor{\bsnm{Zhang}, \binits{L.}},
\oauthor{\bsnm{Cao}, \binits{C.}},
\oauthor{\bsnm{Yu}, \binits{X.}},
\oauthor{\bsnm{Dai}, \binits{H.}},
\oauthor{\bsnm{Ma}, \binits{C.}},
\oauthor{\bsnm{Liu}, \binits{Z.}},
\oauthor{\bsnm{Zhao}, \binits{L.}},
\oauthor{\bsnm{Li}, \binits{G.}},
\oauthor{\bsnm{Liu}, \binits{W.}},
\oauthor{\bsnm{Li}, \binits{Q.}},
\oauthor{\bsnm{Shen}, \binits{D.}},
\oauthor{\bsnm{Li}, \binits{X.}},
\oauthor{\bsnm{Zhu}, \binits{D.}},
\oauthor{\bsnm{Liu}, \binits{T.}}:
Exploring the Trade-Offs: Unified Large Language Models vs Local Fine-Tuned
  Models for Highly-Specific Radiology NLI Task
(2023)
\end{botherref}
\endbibitem

\bibitem[\protect\citeauthoryear{Liu et~al.}{2023a}]{liu2023radiologygpt}
\begin{botherref}
\oauthor{\bsnm{Liu}, \binits{Z.}},
\oauthor{\bsnm{Zhong}, \binits{A.}},
\oauthor{\bsnm{Li}, \binits{Y.}},
\oauthor{\bsnm{Yang}, \binits{L.}},
\oauthor{\bsnm{Ju}, \binits{C.}},
\oauthor{\bsnm{Wu}, \binits{Z.}},
\oauthor{\bsnm{Ma}, \binits{C.}},
\oauthor{\bsnm{Shu}, \binits{P.}},
\oauthor{\bsnm{Chen}, \binits{C.}},
\oauthor{\bsnm{Kim}, \binits{S.}},
\oauthor{\bsnm{Dai}, \binits{H.}},
\oauthor{\bsnm{Zhao}, \binits{L.}},
\oauthor{\bsnm{Zhu}, \binits{D.}},
\oauthor{\bsnm{Liu}, \binits{J.}},
\oauthor{\bsnm{Liu}, \binits{W.}},
\oauthor{\bsnm{Shen}, \binits{D.}},
\oauthor{\bsnm{Li}, \binits{X.}},
\oauthor{\bsnm{Li}, \binits{Q.}},
\oauthor{\bsnm{Liu}, \binits{T.}}:
Radiology-GPT: A Large Language Model for Radiology
(2023)
\end{botherref}
\endbibitem

\bibitem[\protect\citeauthoryear{Liu et~al.}{2023b}]{Liu_2023}
\begin{bbook}
\bauthor{\bsnm{Liu}, \binits{Z.}},
\bauthor{\bsnm{He}, \binits{X.}},
\bauthor{\bsnm{Liu}, \binits{L.}},
\bauthor{\bsnm{Liu}, \binits{T.}},
\bauthor{\bsnm{Zhai}, \binits{X.}}:
\bbtitle{Context Matters: A Strategy to Pre-train Language Model for Science
  Education},
pp. \bfpage{666}--\blpage{674}.
\bpublisher{Springer}, \blocation{???}
(\byear{2023}).
\doiurl{10.1007/978-3-031-36336-8_103} .
\burl{http://dx.doi.org/10.1007/978-3-031-36336-8_103}
\end{bbook}
\endbibitem

\bibitem[\protect\citeauthoryear{Wang et~al.}{2023}]{wang2023review}
\begin{botherref}
\oauthor{\bsnm{Wang}, \binits{J.}},
\oauthor{\bsnm{Liu}, \binits{Z.}},
\oauthor{\bsnm{Zhao}, \binits{L.}},
\oauthor{\bsnm{Wu}, \binits{Z.}},
\oauthor{\bsnm{Ma}, \binits{C.}},
\oauthor{\bsnm{Yu}, \binits{S.}},
\oauthor{\bsnm{Dai}, \binits{H.}},
\oauthor{\bsnm{Yang}, \binits{Q.}},
\oauthor{\bsnm{Liu}, \binits{Y.}},
\oauthor{\bsnm{Zhang}, \binits{S.}},
\oauthor{\bsnm{Shi}, \binits{E.}},
\oauthor{\bsnm{Pan}, \binits{Y.}},
\oauthor{\bsnm{Zhang}, \binits{T.}},
\oauthor{\bsnm{Zhu}, \binits{D.}},
\oauthor{\bsnm{Li}, \binits{X.}},
\oauthor{\bsnm{Jiang}, \binits{X.}},
\oauthor{\bsnm{Ge}, \binits{B.}},
\oauthor{\bsnm{Yuan}, \binits{Y.}},
\oauthor{\bsnm{Shen}, \binits{D.}},
\oauthor{\bsnm{Liu}, \binits{T.}},
\oauthor{\bsnm{Zhang}, \binits{S.}}:
Review of Large Vision Models and Visual Prompt Engineering
(2023)
\end{botherref}
\endbibitem

\bibitem[\protect\citeauthoryear{Li et~al.}{2023}]{li2023artificial}
\begin{botherref}
\oauthor{\bsnm{Li}, \binits{X.}},
\oauthor{\bsnm{Zhang}, \binits{L.}},
\oauthor{\bsnm{Wu}, \binits{Z.}},
\oauthor{\bsnm{Liu}, \binits{Z.}},
\oauthor{\bsnm{Zhao}, \binits{L.}},
\oauthor{\bsnm{Yuan}, \binits{Y.}},
\oauthor{\bsnm{Liu}, \binits{J.}},
\oauthor{\bsnm{Li}, \binits{G.}},
\oauthor{\bsnm{Zhu}, \binits{D.}},
\oauthor{\bsnm{Yan}, \binits{P.}},
\oauthor{\bsnm{Li}, \binits{Q.}},
\oauthor{\bsnm{Liu}, \binits{W.}},
\oauthor{\bsnm{Liu}, \binits{T.}},
\oauthor{\bsnm{Shen}, \binits{D.}}:
Artificial General Intelligence for Medical Imaging
(2023)
\end{botherref}
\endbibitem

\bibitem[\protect\citeauthoryear{Cai et~al.}{2023}]{cai2023coarsetofine}
\begin{botherref}
\oauthor{\bsnm{Cai}, \binits{H.}},
\oauthor{\bsnm{Liao}, \binits{W.}},
\oauthor{\bsnm{Liu}, \binits{Z.}},
\oauthor{\bsnm{Zhang}, \binits{Y.}},
\oauthor{\bsnm{Huang}, \binits{X.}},
\oauthor{\bsnm{Ding}, \binits{S.}},
\oauthor{\bsnm{Ren}, \binits{H.}},
\oauthor{\bsnm{Wu}, \binits{Z.}},
\oauthor{\bsnm{Dai}, \binits{H.}},
\oauthor{\bsnm{Li}, \binits{S.}},
\oauthor{\bsnm{Wu}, \binits{L.}},
\oauthor{\bsnm{Liu}, \binits{N.}},
\oauthor{\bsnm{Li}, \binits{Q.}},
\oauthor{\bsnm{Liu}, \binits{T.}},
\oauthor{\bsnm{Li}, \binits{X.}}:
Coarse-to-fine Knowledge Graph Domain Adaptation based on Distantly-supervised
  Iterative Training
(2023)
\end{botherref}
\endbibitem

\bibitem[\protect\citeauthoryear{Dosovitskiy
  et~al.}{2020}]{dosovitskiy2020image}
\begin{botherref}
\oauthor{\bsnm{Dosovitskiy}, \binits{A.}},
\oauthor{\bsnm{Beyer}, \binits{L.}},
\oauthor{\bsnm{Kolesnikov}, \binits{A.}},
\oauthor{\bsnm{Weissenborn}, \binits{D.}},
\oauthor{\bsnm{Zhai}, \binits{X.}},
\oauthor{\bsnm{Unterthiner}, \binits{T.}},
\oauthor{\bsnm{Dehghani}, \binits{M.}},
\oauthor{\bsnm{Minderer}, \binits{M.}},
\oauthor{\bsnm{Heigold}, \binits{G.}},
\oauthor{\bsnm{Gelly}, \binits{S.}}, et al.:
An image is worth 16x16 words: Transformers for image recognition at scale.
arXiv preprint arXiv:2010.11929
(2020)
\end{botherref}
\endbibitem

\bibitem[\protect\citeauthoryear{He et~al.}{2022}]{he2022mae}
\begin{bchapter}
\bauthor{\bsnm{He}, \binits{K.}},
\bauthor{\bsnm{Chen}, \binits{X.}},
\bauthor{\bsnm{Xie}, \binits{S.}},
\bauthor{\bsnm{Li}, \binits{Y.}},
\bauthor{\bsnm{Doll{\'a}r}, \binits{P.}},
\bauthor{\bsnm{Girshick}, \binits{R.}}:
\bctitle{Masked autoencoders are scalable vision learners}.
In: \bbtitle{Proceedings of the IEEE/CVF Conference on Computer Vision and
  Pattern Recognition},
pp. \bfpage{16000}--\blpage{16009}
(\byear{2022})
\end{bchapter}
\endbibitem

\bibitem[\protect\citeauthoryear{Dai et~al.}{2023}]{dai2023samaug}
\begin{botherref}
\oauthor{\bsnm{Dai}, \binits{H.}},
\oauthor{\bsnm{Ma}, \binits{C.}},
\oauthor{\bsnm{Liu}, \binits{Z.}},
\oauthor{\bsnm{Li}, \binits{Y.}},
\oauthor{\bsnm{Shu}, \binits{P.}},
\oauthor{\bsnm{Wei}, \binits{X.}},
\oauthor{\bsnm{Zhao}, \binits{L.}},
\oauthor{\bsnm{Wu}, \binits{Z.}},
\oauthor{\bsnm{Zeng}, \binits{F.}},
\oauthor{\bsnm{Zhu}, \binits{D.}},
\oauthor{\bsnm{Liu}, \binits{W.}},
\oauthor{\bsnm{Li}, \binits{Q.}},
\oauthor{\bsnm{Liu}, \binits{T.}},
\oauthor{\bsnm{Li}, \binits{X.}}:
SAMAug: Point Prompt Augmentation for Segment Anything Model
(2023)
\end{botherref}
\endbibitem

\bibitem[\protect\citeauthoryear{Zhang et~al.}{2023}]{zhang2023segment}
\begin{botherref}
\oauthor{\bsnm{Zhang}, \binits{L.}},
\oauthor{\bsnm{Liu}, \binits{Z.}},
\oauthor{\bsnm{Zhang}, \binits{L.}},
\oauthor{\bsnm{Wu}, \binits{Z.}},
\oauthor{\bsnm{Yu}, \binits{X.}},
\oauthor{\bsnm{Holmes}, \binits{J.}},
\oauthor{\bsnm{Feng}, \binits{H.}},
\oauthor{\bsnm{Dai}, \binits{H.}},
\oauthor{\bsnm{Li}, \binits{X.}},
\oauthor{\bsnm{Li}, \binits{Q.}},
\oauthor{\bsnm{Zhu}, \binits{D.}},
\oauthor{\bsnm{Liu}, \binits{T.}},
\oauthor{\bsnm{Liu}, \binits{W.}}:
Segment Anything Model (SAM) for Radiation Oncology
(2023)
\end{botherref}
\endbibitem

\bibitem[\protect\citeauthoryear{Xiao et~al.}{2023}]{xiao2023instructionvit}
\begin{botherref}
\oauthor{\bsnm{Xiao}, \binits{Z.}},
\oauthor{\bsnm{Chen}, \binits{Y.}},
\oauthor{\bsnm{Zhang}, \binits{L.}},
\oauthor{\bsnm{Yao}, \binits{J.}},
\oauthor{\bsnm{Wu}, \binits{Z.}},
\oauthor{\bsnm{Yu}, \binits{X.}},
\oauthor{\bsnm{Pan}, \binits{Y.}},
\oauthor{\bsnm{Zhao}, \binits{L.}},
\oauthor{\bsnm{Ma}, \binits{C.}},
\oauthor{\bsnm{Liu}, \binits{X.}},
\oauthor{\bsnm{Liu}, \binits{W.}},
\oauthor{\bsnm{Li}, \binits{X.}},
\oauthor{\bsnm{Yuan}, \binits{Y.}},
\oauthor{\bsnm{Shen}, \binits{D.}},
\oauthor{\bsnm{Zhu}, \binits{D.}},
\oauthor{\bsnm{Liu}, \binits{T.}},
\oauthor{\bsnm{Jiang}, \binits{X.}}:
Instruction-ViT: Multi-Modal Prompts for Instruction Learning in ViT
(2023)
\end{botherref}
\endbibitem

\bibitem[\protect\citeauthoryear{Liu et~al.}{2022}]{liu2022survey}
\begin{barticle}
\bauthor{\bsnm{Liu}, \binits{Z.}},
\bauthor{\bsnm{He}, \binits{M.}},
\bauthor{\bsnm{Jiang}, \binits{Z.}},
\bauthor{\bsnm{Wu}, \binits{Z.}},
\bauthor{\bsnm{Dai}, \binits{H.}},
\bauthor{\bsnm{Zhang}, \binits{L.}},
\bauthor{\bsnm{Luo}, \binits{S.}},
\bauthor{\bsnm{Han}, \binits{T.}},
\bauthor{\bsnm{Li}, \binits{X.}},
\bauthor{\bsnm{Jiang}, \binits{X.}}, \betal:
\batitle{Survey on natural language processing in medical image analysis.}
\bjtitle{Zhong nan da xue xue bao. Yi xue ban= Journal of Central South
  University. Medical Sciences}
\bvolume{47}(\bissue{8}),
\bfpage{981}--\blpage{993}
(\byear{2022})
\end{barticle}
\endbibitem

\bibitem[\protect\citeauthoryear{Zhao et~al.}{2022}]{zhao2022embedding}
\begin{bchapter}
\bauthor{\bsnm{Zhao}, \binits{L.}},
\bauthor{\bsnm{Wu}, \binits{Z.}},
\bauthor{\bsnm{Dai}, \binits{H.}},
\bauthor{\bsnm{Liu}, \binits{Z.}},
\bauthor{\bsnm{Zhang}, \binits{T.}},
\bauthor{\bsnm{Zhu}, \binits{D.}},
\bauthor{\bsnm{Liu}, \binits{T.}}:
\bctitle{Embedding human brain function via transformer}.
In: \bbtitle{International Conference on Medical Image Computing and
  Computer-Assisted Intervention},
pp. \bfpage{366}--\blpage{375}
(\byear{2022}).
\bcomment{Springer}
\end{bchapter}
\endbibitem

\bibitem[\protect\citeauthoryear{Dai et~al.}{2022}]{dai2022graph}
\begin{bchapter}
\bauthor{\bsnm{Dai}, \binits{H.}},
\bauthor{\bsnm{Li}, \binits{Q.}},
\bauthor{\bsnm{Zhao}, \binits{L.}},
\bauthor{\bsnm{Pan}, \binits{L.}},
\bauthor{\bsnm{Shi}, \binits{C.}},
\bauthor{\bsnm{Liu}, \binits{Z.}},
\bauthor{\bsnm{Wu}, \binits{Z.}},
\bauthor{\bsnm{Zhang}, \binits{L.}},
\bauthor{\bsnm{Zhao}, \binits{S.}},
\bauthor{\bsnm{Wu}, \binits{X.}}, \betal:
\bctitle{Graph representation neural architecture search for optimal
  spatial/temporal functional brain network decomposition}.
In: \bbtitle{International Workshop on Machine Learning in Medical Imaging},
pp. \bfpage{279}--\blpage{287}
(\byear{2022}).
\bcomment{Springer}
\end{bchapter}
\endbibitem

\bibitem[\protect\citeauthoryear{Liu et~al.}{2022}]{liu2022discovering}
\begin{botherref}
\oauthor{\bsnm{Liu}, \binits{Y.}},
\oauthor{\bsnm{Ge}, \binits{E.}},
\oauthor{\bsnm{He}, \binits{M.}},
\oauthor{\bsnm{Liu}, \binits{Z.}},
\oauthor{\bsnm{Zhao}, \binits{S.}},
\oauthor{\bsnm{Hu}, \binits{X.}},
\oauthor{\bsnm{Zhu}, \binits{D.}},
\oauthor{\bsnm{Liu}, \binits{T.}},
\oauthor{\bsnm{Ge}, \binits{B.}}:
Discovering dynamic functional brain networks via spatial and channel-wise
  attention.
arXiv preprint arXiv:2205.09576
(2022)
\end{botherref}
\endbibitem

\bibitem[\protect\citeauthoryear{Zhang et~al.}{2023}]{zhang2023beam}
\begin{botherref}
\oauthor{\bsnm{Zhang}, \binits{L.}},
\oauthor{\bsnm{Holmes}, \binits{J.M.}},
\oauthor{\bsnm{Liu}, \binits{Z.}},
\oauthor{\bsnm{Vora}, \binits{S.A.}},
\oauthor{\bsnm{Sio}, \binits{T.T.}},
\oauthor{\bsnm{Vargas}, \binits{C.E.}},
\oauthor{\bsnm{Yu}, \binits{N.Y.}},
\oauthor{\bsnm{Keole}, \binits{S.R.}},
\oauthor{\bsnm{Schild}, \binits{S.E.}},
\oauthor{\bsnm{Bues}, \binits{M.}}, et al.:
Beam mask and sliding window-facilitated deep learning-based accurate and
  efficient dose prediction for pencil beam scanning proton therapy.
arXiv preprint arXiv:2305.18572
(2023)
\end{botherref}
\endbibitem

\bibitem[\protect\citeauthoryear{Liu et~al.}{2020}]{liu2020survey}
\begin{bchapter}
\bauthor{\bsnm{Liu}, \binits{Z.}},
\bauthor{\bsnm{Crouser}, \binits{R.J.}},
\bauthor{\bsnm{Ottley}, \binits{A.}}:
\bctitle{Survey on individual differences in visualization}.
In: \bbtitle{Computer Graphics Forum},
vol. \bseriesno{39},
pp. \bfpage{693}--\blpage{712}
(\byear{2020}).
\bcomment{Wiley Online Library}
\end{bchapter}
\endbibitem

\bibitem[\protect\citeauthoryear{Bi et~al.}{2023}]{bi2023community}
\begin{botherref}
\oauthor{\bsnm{Bi}, \binits{X.-A.}},
\oauthor{\bsnm{Chen}, \binits{K.}},
\oauthor{\bsnm{Jiang}, \binits{S.}},
\oauthor{\bsnm{Luo}, \binits{S.}},
\oauthor{\bsnm{Zhou}, \binits{W.}},
\oauthor{\bsnm{Xing}, \binits{Z.}},
\oauthor{\bsnm{Xu}, \binits{L.}},
\oauthor{\bsnm{Liu}, \binits{Z.}},
\oauthor{\bsnm{Liu}, \binits{T.}}:
Community graph convolution neural network for alzheimer’s disease
  classification and pathogenetic factors identification.
IEEE Transactions on Neural Networks and Learning Systems
(2023)
\end{botherref}
\endbibitem

\bibitem[\protect\citeauthoryear{Ding et~al.}{2023}]{ding2023deep}
\begin{botherref}
\oauthor{\bsnm{Ding}, \binits{Y.}},
\oauthor{\bsnm{Feng}, \binits{H.}},
\oauthor{\bsnm{Yang}, \binits{Y.}},
\oauthor{\bsnm{Holmes}, \binits{J.}},
\oauthor{\bsnm{Liu}, \binits{Z.}},
\oauthor{\bsnm{Liu}, \binits{D.}},
\oauthor{\bsnm{Wong}, \binits{W.W.}},
\oauthor{\bsnm{Yu}, \binits{N.Y.}},
\oauthor{\bsnm{Sio}, \binits{T.T.}},
\oauthor{\bsnm{Schild}, \binits{S.E.}}, et al.:
Deep-learning based fast and accurate 3d ct deformable image registration in
  lung cancer.
Medical Physics
(2023)
\end{botherref}
\endbibitem

\bibitem[\protect\citeauthoryear{Ding et~al.}{2022}]{ding2022accurate}
\begin{bchapter}
\bauthor{\bsnm{Ding}, \binits{Y.}},
\bauthor{\bsnm{Liu}, \binits{Z.}},
\bauthor{\bsnm{Feng}, \binits{H.}},
\bauthor{\bsnm{Holmes}, \binits{J.}},
\bauthor{\bsnm{Yang}, \binits{Y.}},
\bauthor{\bsnm{Yu}, \binits{N.}},
\bauthor{\bsnm{Sio}, \binits{T.}},
\bauthor{\bsnm{Schild}, \binits{S.}},
\bauthor{\bsnm{Li}, \binits{B.}},
\bauthor{\bsnm{Liu}, \binits{W.}}:
\bctitle{Accurate and efficient deep neural network based deformable image
  registration method in lung cancer}.
In: \bbtitle{MEDICAL PHYSICS},
vol. \bseriesno{49},
pp. \bfpage{148}--\blpage{148}
(\byear{2022}).
\bcomment{WILEY 111 RIVER ST, HOBOKEN 07030-5774, NJ USA}
\end{bchapter}
\endbibitem

\bibitem[\protect\citeauthoryear{Kirillov et~al.}{2023}]{kirillov2023segment}
\begin{botherref}
\oauthor{\bsnm{Kirillov}, \binits{A.}},
\oauthor{\bsnm{Mintun}, \binits{E.}},
\oauthor{\bsnm{Ravi}, \binits{N.}},
\oauthor{\bsnm{Mao}, \binits{H.}},
\oauthor{\bsnm{Rolland}, \binits{C.}},
\oauthor{\bsnm{Gustafson}, \binits{L.}},
\oauthor{\bsnm{Xiao}, \binits{T.}},
\oauthor{\bsnm{Whitehead}, \binits{S.}},
\oauthor{\bsnm{Berg}, \binits{A.C.}},
\oauthor{\bsnm{Lo}, \binits{W.-Y.}}, et al.:
Segment anything.
arXiv preprint arXiv:2304.02643
(2023)
\end{botherref}
\endbibitem

\bibitem[\protect\citeauthoryear{Osco et~al.}{2023}]{osco2023segment}
\begin{barticle}
\bauthor{\bsnm{Osco}, \binits{L.P.}},
\bauthor{\bsnm{Wu}, \binits{Q.}},
\bauthor{\bsnm{Lemos}, \binits{E.L.}},
\bauthor{\bsnm{Gon{\c{c}}alves}, \binits{W.N.}},
\bauthor{\bsnm{Ramos}, \binits{A.P.M.}},
\bauthor{\bsnm{Li}, \binits{J.}},
\bauthor{\bsnm{Junior}, \binits{J.M.}}:
\batitle{The segment anything model (sam) for remote sensing applications: From
  zero to one shot}.
\bjtitle{International Journal of Applied Earth Observation and Geoinformation}
\bvolume{124},
\bfpage{103540}
(\byear{2023})
\end{barticle}
\endbibitem

\bibitem[\protect\citeauthoryear{Liu et~al.}{2023}]{liu2023artificial}
\begin{botherref}
\oauthor{\bsnm{Liu}, \binits{C.}},
\oauthor{\bsnm{Liu}, \binits{Z.}},
\oauthor{\bsnm{Holmes}, \binits{J.}},
\oauthor{\bsnm{Zhang}, \binits{L.}},
\oauthor{\bsnm{Zhang}, \binits{L.}},
\oauthor{\bsnm{Ding}, \binits{Y.}},
\oauthor{\bsnm{Shu}, \binits{P.}},
\oauthor{\bsnm{Wu}, \binits{Z.}},
\oauthor{\bsnm{Dai}, \binits{H.}},
\oauthor{\bsnm{Li}, \binits{Y.}}, et al.:
Artificial general intelligence for radiation oncology.
Meta-Radiology,
100045
(2023)
\end{botherref}
\endbibitem

\bibitem[\protect\citeauthoryear{Chen et~al.}{2023}]{chen2023ma}
\begin{botherref}
\oauthor{\bsnm{Chen}, \binits{C.}},
\oauthor{\bsnm{Miao}, \binits{J.}},
\oauthor{\bsnm{Wu}, \binits{D.}},
\oauthor{\bsnm{Yan}, \binits{Z.}},
\oauthor{\bsnm{Kim}, \binits{S.}},
\oauthor{\bsnm{Hu}, \binits{J.}},
\oauthor{\bsnm{Zhong}, \binits{A.}},
\oauthor{\bsnm{Liu}, \binits{Z.}},
\oauthor{\bsnm{Sun}, \binits{L.}},
\oauthor{\bsnm{Li}, \binits{X.}}, et al.:
Ma-sam: Modality-agnostic sam adaptation for 3d medical image segmentation.
arXiv preprint arXiv:2309.08842
(2023)
\end{botherref}
\endbibitem

\bibitem[\protect\citeauthoryear{Cai et~al.}{2023}]{cai2023multimodal}
\begin{bchapter}
\bauthor{\bsnm{Cai}, \binits{H.}},
\bauthor{\bsnm{Huang}, \binits{X.}},
\bauthor{\bsnm{Liu}, \binits{Z.}},
\bauthor{\bsnm{Liao}, \binits{W.}},
\bauthor{\bsnm{Dai}, \binits{H.}},
\bauthor{\bsnm{Wu}, \binits{Z.}},
\bauthor{\bsnm{Zhu}, \binits{D.}},
\bauthor{\bsnm{Ren}, \binits{H.}},
\bauthor{\bsnm{Li}, \binits{Q.}},
\bauthor{\bsnm{Liu}, \binits{T.}}, \betal:
\bctitle{Multimodal approaches for alzheimer’s detection using patients’
  speech and transcript}.
In: \bbtitle{International Conference on Brain Informatics},
pp. \bfpage{395}--\blpage{406}
(\byear{2023}).
\bcomment{Springer}
\end{bchapter}
\endbibitem

\bibitem[\protect\citeauthoryear{Holmes et~al.}{2023}]{holmes2023evaluating2}
\begin{botherref}
\oauthor{\bsnm{Holmes}, \binits{J.}},
\oauthor{\bsnm{Ye}, \binits{S.}},
\oauthor{\bsnm{Li}, \binits{Y.}},
\oauthor{\bsnm{Wu}, \binits{S.-N.}},
\oauthor{\bsnm{Liu}, \binits{Z.}},
\oauthor{\bsnm{Wu}, \binits{Z.}},
\oauthor{\bsnm{Hu}, \binits{J.}},
\oauthor{\bsnm{Zhao}, \binits{H.}},
\oauthor{\bsnm{Jiang}, \binits{X.}},
\oauthor{\bsnm{Liu}, \binits{W.}}, et al.:
Evaluating large language models in ophthalmology.
arXiv preprint arXiv:2311.04933
(2023)
\end{botherref}
\endbibitem

\bibitem[\protect\citeauthoryear{Xiao et~al.}{2023}]{xiao2023instruction}
\begin{botherref}
\oauthor{\bsnm{Xiao}, \binits{Z.}},
\oauthor{\bsnm{Chen}, \binits{Y.}},
\oauthor{\bsnm{Zhang}, \binits{L.}},
\oauthor{\bsnm{Yao}, \binits{J.}},
\oauthor{\bsnm{Wu}, \binits{Z.}},
\oauthor{\bsnm{Yu}, \binits{X.}},
\oauthor{\bsnm{Pan}, \binits{Y.}},
\oauthor{\bsnm{Zhao}, \binits{L.}},
\oauthor{\bsnm{Ma}, \binits{C.}},
\oauthor{\bsnm{Liu}, \binits{X.}}, et al.:
Instruction-vit: Multi-modal prompts for instruction learning in vit.
arXiv preprint arXiv:2305.00201
(2023)
\end{botherref}
\endbibitem

\bibitem[\protect\citeauthoryear{Touvron et~al.}{2023}]{touvron2023llama}
\begin{botherref}
\oauthor{\bsnm{Touvron}, \binits{H.}},
\oauthor{\bsnm{Martin}, \binits{L.}},
\oauthor{\bsnm{Stone}, \binits{K.}},
\oauthor{\bsnm{Albert}, \binits{P.}},
\oauthor{\bsnm{Almahairi}, \binits{A.}},
\oauthor{\bsnm{Babaei}, \binits{Y.}},
\oauthor{\bsnm{Bashlykov}, \binits{N.}},
\oauthor{\bsnm{Batra}, \binits{S.}},
\oauthor{\bsnm{Bhargava}, \binits{P.}},
\oauthor{\bsnm{Bhosale}, \binits{S.}}, et al.:
Llama 2: Open foundation and fine-tuned chat models.
arXiv preprint arXiv:2307.09288
(2023)
\end{botherref}
\endbibitem

\bibitem[\protect\citeauthoryear{Chiang et~al.}{2023}]{vicuna2023}
\begin{botherref}
\oauthor{\bsnm{Chiang}, \binits{W.-L.}},
\oauthor{\bsnm{Li}, \binits{Z.}},
\oauthor{\bsnm{Lin}, \binits{Z.}},
\oauthor{\bsnm{Sheng}, \binits{Y.}},
\oauthor{\bsnm{Wu}, \binits{Z.}},
\oauthor{\bsnm{Zhang}, \binits{H.}},
\oauthor{\bsnm{Zheng}, \binits{L.}},
\oauthor{\bsnm{Zhuang}, \binits{S.}},
\oauthor{\bsnm{Zhuang}, \binits{Y.}},
\oauthor{\bsnm{Gonzalez}, \binits{J.E.}},
\oauthor{\bsnm{Stoica}, \binits{I.}},
\oauthor{\bsnm{Xing}, \binits{E.P.}}:
Vicuna: An Open-Source Chatbot Impressing GPT-4 with 90\%* ChatGPT Quality
(2023).
\url{https://lmsys.org/blog/2023-03-30-vicuna/}
\end{botherref}
\endbibitem

\bibitem[\protect\citeauthoryear{Chen et~al.}{2023}]{chen2023minigpt}
\begin{botherref}
\oauthor{\bsnm{Chen}, \binits{J.}},
\oauthor{\bsnm{Zhu}, \binits{D.}},
\oauthor{\bsnm{Shen}, \binits{X.}},
\oauthor{\bsnm{Li}, \binits{X.}},
\oauthor{\bsnm{Liu}, \binits{Z.}},
\oauthor{\bsnm{Zhang}, \binits{P.}},
\oauthor{\bsnm{Krishnamoorthi}, \binits{R.}},
\oauthor{\bsnm{Chandra}, \binits{V.}},
\oauthor{\bsnm{Xiong}, \binits{Y.}},
\oauthor{\bsnm{Elhoseiny}, \binits{M.}}:
Minigpt-v2: large language model as a unified interface for vision-language
  multi-task learning.
arXiv preprint arXiv:2310.09478
(2023)
\end{botherref}
\endbibitem

\bibitem[\protect\citeauthoryear{Wu et~al.}{2023}]{wu2023next}
\begin{botherref}
\oauthor{\bsnm{Wu}, \binits{S.}},
\oauthor{\bsnm{Fei}, \binits{H.}},
\oauthor{\bsnm{Qu}, \binits{L.}},
\oauthor{\bsnm{Ji}, \binits{W.}},
\oauthor{\bsnm{Chua}, \binits{T.-S.}}:
Next-gpt: Any-to-any multimodal llm.
arXiv preprint arXiv:2309.05519
(2023)
\end{botherref}
\endbibitem

\bibitem[\protect\citeauthoryear{Han et~al.}{2023}]{han2023imagebind}
\begin{botherref}
\oauthor{\bsnm{Han}, \binits{J.}},
\oauthor{\bsnm{Zhang}, \binits{R.}},
\oauthor{\bsnm{Shao}, \binits{W.}},
\oauthor{\bsnm{Gao}, \binits{P.}},
\oauthor{\bsnm{Xu}, \binits{P.}},
\oauthor{\bsnm{Xiao}, \binits{H.}},
\oauthor{\bsnm{Zhang}, \binits{K.}},
\oauthor{\bsnm{Liu}, \binits{C.}},
\oauthor{\bsnm{Wen}, \binits{S.}},
\oauthor{\bsnm{Guo}, \binits{Z.}}, et al.:
Imagebind-llm: Multi-modality instruction tuning.
arXiv preprint arXiv:2309.03905
(2023)
\end{botherref}
\endbibitem

\bibitem[\protect\citeauthoryear{Liu et~al.}{2023}]{liu2023visual}
\begin{botherref}
\oauthor{\bsnm{Liu}, \binits{H.}},
\oauthor{\bsnm{Li}, \binits{C.}},
\oauthor{\bsnm{Wu}, \binits{Q.}},
\oauthor{\bsnm{Lee}, \binits{Y.J.}}:
Visual instruction tuning.
arXiv preprint arXiv:2304.08485
(2023)
\end{botherref}
\endbibitem

\bibitem[\protect\citeauthoryear{Janowicz et~al.}{2020}]{janowicz2020geoai}
\begin{botherref}
\oauthor{\bsnm{Janowicz}, \binits{K.}},
\oauthor{\bsnm{Gao}, \binits{S.}},
\oauthor{\bsnm{McKenzie}, \binits{G.}},
\oauthor{\bsnm{Hu}, \binits{Y.}},
\oauthor{\bsnm{Bhaduri}, \binits{B.}}:
GeoAI: spatially explicit artificial intelligence techniques for geographic
  knowledge discovery and beyond.
Taylor \& Francis
(2020)
\end{botherref}
\endbibitem

\bibitem[\protect\citeauthoryear{Gao et~al.}{2023}]{gao2023handbook}
\begin{bbook}
\bauthor{\bsnm{Gao}, \binits{S.}},
\bauthor{\bsnm{Hu}, \binits{Y.}},
\bauthor{\bsnm{Li}, \binits{W.}}:
\bbtitle{Handbook of Geospatial Artificial Intelligence}.
\bpublisher{Taylor \& Francis},
\blocation{Boca Raton}
(\byear{2023}).
\doiurl{10.1201/9781003308423}
\end{bbook}
\endbibitem

\bibitem[\protect\citeauthoryear{Hu et~al.}{2023}]{hu2023geoknowgpt}
\begin{barticle}
\bauthor{\bsnm{Hu}, \binits{Y.}},
\bauthor{\bsnm{Mai}, \binits{G.}},
\bauthor{\bsnm{Cundy}, \binits{C.}},
\bauthor{\bsnm{Choi}, \binits{K.}},
\bauthor{\bsnm{Lao}, \binits{N.}},
\bauthor{\bsnm{Liu}, \binits{W.}},
\bauthor{\bsnm{Lakhanpal}, \binits{G.}},
\bauthor{\bsnm{Zhou}, \binits{R.Z.}},
\bauthor{\bsnm{Joseph}, \binits{K.}}:
\batitle{Geo-knowledge-guided gpt models improve the extraction of location
  descriptions from disaster-related social media messages}.
\bjtitle{International Journal of Geographical Information Science}
\bvolume{37}(\bissue{11}),
\bfpage{2289}--\blpage{2318}
(\byear{2023})
\end{barticle}
\endbibitem

\bibitem[\protect\citeauthoryear{Lacoste et~al.}{2023}]{lacoste2023geobench}
\begin{botherref}
\oauthor{\bsnm{Lacoste}, \binits{A.}},
\oauthor{\bsnm{Lehmann}, \binits{N.}},
\oauthor{\bsnm{Rodriguez}, \binits{P.}},
\oauthor{\bsnm{Sherwin}, \binits{E.D.}},
\oauthor{\bsnm{Kerner}, \binits{H.}},
\oauthor{\bsnm{L{\"u}tjens}, \binits{B.}},
\oauthor{\bsnm{Irvin}, \binits{J.A.}},
\oauthor{\bsnm{Dao}, \binits{D.}},
\oauthor{\bsnm{Alemohammad}, \binits{H.}},
\oauthor{\bsnm{Drouin}, \binits{A.}}, et al.:
GEO-Bench: Toward Foundation Models for Earth Monitoring
\end{botherref}
\endbibitem

\bibitem[\protect\citeauthoryear{Balsebre et~al.}{2023}]{balsebre2023cityfm}
\begin{botherref}
\oauthor{\bsnm{Balsebre}, \binits{P.}},
\oauthor{\bsnm{Huang}, \binits{W.}},
\oauthor{\bsnm{Cong}, \binits{G.}},
\oauthor{\bsnm{Li}, \binits{Y.}}:
Cityfm: City foundation models to solve urban challenges.
arXiv preprint arXiv:2310.00583
(2023)
\end{botherref}
\endbibitem

\bibitem[\protect\citeauthoryear{Alex et~al.}{2019}]{alex2019geoparsing}
\begin{barticle}
\bauthor{\bsnm{Alex}, \binits{B.}},
\bauthor{\bsnm{Grover}, \binits{C.}},
\bauthor{\bsnm{Tobin}, \binits{R.}},
\bauthor{\bsnm{Oberlander}, \binits{J.}}:
\batitle{Geoparsing historical and contemporary literary text set in the city
  of edinburgh}.
\bjtitle{Language Resources and Evaluation}
\bvolume{53},
\bfpage{651}--\blpage{675}
(\byear{2019})
\end{barticle}
\endbibitem

\bibitem[\protect\citeauthoryear{Karimzadeh
  et~al.}{2019}]{karimzadeh2019geotxt}
\begin{barticle}
\bauthor{\bsnm{Karimzadeh}, \binits{M.}},
\bauthor{\bsnm{Pezanowski}, \binits{S.}},
\bauthor{\bsnm{MacEachren}, \binits{A.M.}},
\bauthor{\bsnm{Wallgr{\"u}n}, \binits{J.O.}}:
\batitle{Geotxt: A scalable geoparsing system for unstructured text
  geolocation}.
\bjtitle{Transactions in GIS}
\bvolume{23}(\bissue{1}),
\bfpage{118}--\blpage{136}
(\byear{2019})
\end{barticle}
\endbibitem

\bibitem[\protect\citeauthoryear{Mai et~al.}{2023}]{mai2023spatialrl}
\begin{bchapter}
\bauthor{\bsnm{Mai}, \binits{G.}},
\bauthor{\bsnm{Li}, \binits{Z.}},
\bauthor{\bsnm{Lao}, \binits{N.}}:
\bctitle{Spatial representation learning in geoai}.
In: \bbtitle{Handbook of Geospatial Artificial Intelligence},
pp. \bfpage{99}--\blpage{120}.
\bpublisher{CRC Press}, \blocation{???}
(\byear{2023})
\end{bchapter}
\endbibitem

\bibitem[\protect\citeauthoryear{Li and Ning}{2023}]{li2023autonomous}
\begin{botherref}
\oauthor{\bsnm{Li}, \binits{Z.}},
\oauthor{\bsnm{Ning}, \binits{H.}}:
Autonomous gis: the next-generation ai-powered gis.
arXiv preprint arXiv:2305.06453
(2023)
\end{botherref}
\endbibitem

\bibitem[\protect\citeauthoryear{Liu et~al.}{2023}]{liu2023grounding}
\begin{botherref}
\oauthor{\bsnm{Liu}, \binits{S.}},
\oauthor{\bsnm{Zeng}, \binits{Z.}},
\oauthor{\bsnm{Ren}, \binits{T.}},
\oauthor{\bsnm{Li}, \binits{F.}},
\oauthor{\bsnm{Zhang}, \binits{H.}},
\oauthor{\bsnm{Yang}, \binits{J.}},
\oauthor{\bsnm{Li}, \binits{C.}},
\oauthor{\bsnm{Yang}, \binits{J.}},
\oauthor{\bsnm{Su}, \binits{H.}},
\oauthor{\bsnm{Zhu}, \binits{J.}}, et al.:
Grounding dino: Marrying dino with grounded pre-training for open-set object
  detection.
arXiv preprint arXiv:2303.05499
(2023)
\end{botherref}
\endbibitem

\bibitem[\protect\citeauthoryear{Du et~al.}{2023}]{du2023tree}
\begin{botherref}
\oauthor{\bsnm{Du}, \binits{S.}},
\oauthor{\bsnm{Tang}, \binits{S.}},
\oauthor{\bsnm{Wang}, \binits{W.}},
\oauthor{\bsnm{Li}, \binits{X.}},
\oauthor{\bsnm{Guo}, \binits{R.}}:
Tree-gpt: Modular large language model expert system for forest remote sensing
  image understanding and interactive analysis.
arXiv preprint arXiv:2310.04698
(2023)
\end{botherref}
\endbibitem

\bibitem[\protect\citeauthoryear{Qing et~al.}{2023}]{qing2023gpt}
\begin{barticle}
\bauthor{\bsnm{Qing}, \binits{J.}},
\bauthor{\bsnm{Deng}, \binits{X.}},
\bauthor{\bsnm{Lan}, \binits{Y.}},
\bauthor{\bsnm{Li}, \binits{Z.}}:
\batitle{Gpt-aided diagnosis on agricultural image based on a new light
  yolopc}.
\bjtitle{Computers and Electronics in Agriculture}
\bvolume{213},
\bfpage{108168}
(\byear{2023})
\end{barticle}
\endbibitem

\bibitem[\protect\citeauthoryear{Wang et~al.}{2023}]{wang2023optimizing}
\begin{botherref}
\oauthor{\bsnm{Wang}, \binits{X.}},
\oauthor{\bsnm{Ling}, \binits{X.}},
\oauthor{\bsnm{Zhang}, \binits{T.}},
\oauthor{\bsnm{Li}, \binits{X.}},
\oauthor{\bsnm{Wang}, \binits{S.}},
\oauthor{\bsnm{Li}, \binits{Z.}},
\oauthor{\bsnm{Zhang}, \binits{L.}},
\oauthor{\bsnm{Gong}, \binits{P.}}:
Optimizing and fine-tuning large language model for urban renewal.
arXiv preprint arXiv:2311.15490
(2023)
\end{botherref}
\endbibitem

\bibitem[\protect\citeauthoryear{Yan et~al.}{2023}]{yan2023urbanclip}
\begin{botherref}
\oauthor{\bsnm{Yan}, \binits{Y.}},
\oauthor{\bsnm{Wen}, \binits{H.}},
\oauthor{\bsnm{Zhong}, \binits{S.}},
\oauthor{\bsnm{Chen}, \binits{W.}},
\oauthor{\bsnm{Chen}, \binits{H.}},
\oauthor{\bsnm{Wen}, \binits{Q.}},
\oauthor{\bsnm{Zimmermann}, \binits{R.}},
\oauthor{\bsnm{Liang}, \binits{Y.}}:
When urban region profiling meets large language models.
arXiv preprint arXiv:2310.18340
(2023)
\end{botherref}
\endbibitem

\bibitem[\protect\citeauthoryear{Deshpande et~al.}{2023}]{deshpandegenerative}
\begin{bchapter}
\bauthor{\bsnm{Deshpande}, \binits{R.}},
\bauthor{\bsnm{Patel}, \binits{S.V.}},
\bauthor{\bsnm{Weijenberg}, \binits{C.}},
\bauthor{\bsnm{Nisztuk}, \binits{M.}},
\bauthor{\bsnm{Corcuera}, \binits{M.}},
\bauthor{\bsnm{Luo}, \binits{J.}},
\bauthor{\bsnm{Zhu}, \binits{Q.}}:
\bctitle{Generative pre-trained transformers for 15-minute city design}.
In: \bbtitle{Proceedings of the 28th International Conference of the
  Association for Computer-Aided Architectural Design Research in Asia
  (CAADRIA) 2023},
vol. \bseriesno{1}.
\bconflocation{Hong Kong},
pp. \bfpage{595}--\blpage{604}
(\byear{2023}).
\bcomment{Association for Computer-Aided Architectural Design Research in Asia
  (CAADRIA)}
\end{bchapter}
\endbibitem

\bibitem[\protect\citeauthoryear{Ramesh et~al.}{2021}]{ramesh2021dalle}
\begin{bchapter}
\bauthor{\bsnm{Ramesh}, \binits{A.}},
\bauthor{\bsnm{Pavlov}, \binits{M.}},
\bauthor{\bsnm{Goh}, \binits{G.}},
\bauthor{\bsnm{Gray}, \binits{S.}},
\bauthor{\bsnm{Voss}, \binits{C.}},
\bauthor{\bsnm{Radford}, \binits{A.}},
\bauthor{\bsnm{Chen}, \binits{M.}},
\bauthor{\bsnm{Sutskever}, \binits{I.}}:
\bctitle{Zero-shot text-to-image generation}.
In: \bbtitle{International Conference on Machine Learning},
pp. \bfpage{8821}--\blpage{8831}
(\byear{2021}).
\bcomment{PMLR}
\end{bchapter}
\endbibitem

\bibitem[\protect\citeauthoryear{Walker}{2022}]{walker2022dalle}
\begin{botherref}
\oauthor{\bsnm{Walker}, \binits{A.}}:
DALL-E Is Actually Kind of Good at Designing Walkable Streets.
CURBED.
\url{https://www.curbed.com/2022/07/dall-e-walkable-streets-betterstreetsai.html}
\end{botherref}
\endbibitem

\bibitem[\protect\citeauthoryear{Fu et~al.}{2023}]{fu2023can}
\begin{botherref}
\oauthor{\bsnm{Fu}, \binits{X.}},
\oauthor{\bsnm{Wang}, \binits{R.}},
\oauthor{\bsnm{Li}, \binits{C.}}:
Can chatgpt evaluate plans?
Journal of the American Planning Association,
1--12
(2023)
\end{botherref}
\endbibitem

\bibitem[\protect\citeauthoryear{Rillig et~al.}{2023}]{rillig2023risks}
\begin{barticle}
\bauthor{\bsnm{Rillig}, \binits{M.C.}},
\bauthor{\bsnm{{\AA}gerstrand}, \binits{M.}},
\bauthor{\bsnm{Bi}, \binits{M.}},
\bauthor{\bsnm{Gould}, \binits{K.A.}},
\bauthor{\bsnm{Sauerland}, \binits{U.}}:
\batitle{Risks and benefits of large language models for the environment}.
\bjtitle{Environmental Science \& Technology}
\bvolume{57}(\bissue{9}),
\bfpage{3464}--\blpage{3466}
(\byear{2023})
\end{barticle}
\endbibitem

\bibitem[\protect\citeauthoryear{Shi et~al.}{2023}]{shi2023thinking}
\begin{barticle}
\bauthor{\bsnm{Shi}, \binits{M.}},
\bauthor{\bsnm{Currier}, \binits{K.}},
\bauthor{\bsnm{Liu}, \binits{Z.}},
\bauthor{\bsnm{Janowicz}, \binits{K.}},
\bauthor{\bsnm{Wiedemann}, \binits{N.}},
\bauthor{\bsnm{Verstegen}, \binits{J.}},
\bauthor{\bsnm{McKenzie}, \binits{G.}},
\bauthor{\bsnm{Graser}, \binits{A.}},
\bauthor{\bsnm{Zhu}, \binits{R.}},
\bauthor{\bsnm{Mai}, \binits{G.}}:
\batitle{Thinking geographically about ai sustainability}.
\bjtitle{AGILE: GIScience Series}
\bvolume{4},
\bfpage{42}
(\byear{2023})
\end{barticle}
\endbibitem

\bibitem[\protect\citeauthoryear{Gehman
  et~al.}{2020}]{gehman2020realtoxicityprompts}
\begin{botherref}
\oauthor{\bsnm{Gehman}, \binits{S.}},
\oauthor{\bsnm{Gururangan}, \binits{S.}},
\oauthor{\bsnm{Sap}, \binits{M.}},
\oauthor{\bsnm{Choi}, \binits{Y.}},
\oauthor{\bsnm{Smith}, \binits{N.A.}}:
Realtoxicityprompts: Evaluating neural toxic degeneration in language models.
arXiv e-prints,
2009
(2020)
\end{botherref}
\endbibitem

\bibitem[\protect\citeauthoryear{Zhao et~al.}{2018}]{zhao2018gender}
\begin{botherref}
\oauthor{\bsnm{Zhao}, \binits{J.}},
\oauthor{\bsnm{Wang}, \binits{T.}},
\oauthor{\bsnm{Yatskar}, \binits{M.}},
\oauthor{\bsnm{Ordonez}, \binits{V.}},
\oauthor{\bsnm{Chang}, \binits{K.-W.}}:
Gender bias in coreference resolution: Evaluation and debiasing methods.
arXiv preprint arXiv:1804.06876
(2018)
\end{botherref}
\endbibitem

\bibitem[\protect\citeauthoryear{Wang et~al.}{2023}]{wang2023decodingtrust}
\begin{bchapter}
\bauthor{\bsnm{Wang}, \binits{B.}},
\bauthor{\bsnm{Chen}, \binits{W.}},
\bauthor{\bsnm{Pei}, \binits{H.}},
\bauthor{\bsnm{Xie}, \binits{C.}},
\bauthor{\bsnm{Kang}, \binits{M.}},
\bauthor{\bsnm{Zhang}, \binits{C.}},
\bauthor{\bsnm{Xu}, \binits{C.}},
\bauthor{\bsnm{Xiong}, \binits{Z.}},
\bauthor{\bsnm{Dutta}, \binits{R.}},
\bauthor{\bsnm{Schaeffer}, \binits{R.}}, \betal:
\bctitle{Decodingtrust: A comprehensive assessment of trustworthiness in gpt
  models}.
In: \bbtitle{Thirty-seventh Conference on Neural Information Processing
  Systems}
(\byear{2023})
\end{bchapter}
\endbibitem

\bibitem[\protect\citeauthoryear{Deshpande
  et~al.}{2023}]{deshpande2023toxicity}
\begin{botherref}
\oauthor{\bsnm{Deshpande}, \binits{A.}},
\oauthor{\bsnm{Murahari}, \binits{V.}},
\oauthor{\bsnm{Rajpurohit}, \binits{T.}},
\oauthor{\bsnm{Kalyan}, \binits{A.}},
\oauthor{\bsnm{Narasimhan}, \binits{K.}}:
Toxicity in chatgpt: Analyzing persona-assigned language models.
arXiv preprint arXiv:2304.05335
(2023)
\end{botherref}
\endbibitem

\bibitem[\protect\citeauthoryear{Faisal and
  Anastasopoulos}{2022}]{faisal2022geographicbias}
\begin{botherref}
\oauthor{\bsnm{Faisal}, \binits{F.}},
\oauthor{\bsnm{Anastasopoulos}, \binits{A.}}:
Geographic and geopolitical biases of language models.
arXiv preprint arXiv:2212.10408
(2022)
\end{botherref}
\endbibitem

\bibitem[\protect\citeauthoryear{Thomee et~al.}{2016}]{thomee2016yfcc100m}
\begin{barticle}
\bauthor{\bsnm{Thomee}, \binits{B.}},
\bauthor{\bsnm{Shamma}, \binits{D.A.}},
\bauthor{\bsnm{Friedland}, \binits{G.}},
\bauthor{\bsnm{Elizalde}, \binits{B.}},
\bauthor{\bsnm{Ni}, \binits{K.}},
\bauthor{\bsnm{Poland}, \binits{D.}},
\bauthor{\bsnm{Borth}, \binits{D.}},
\bauthor{\bsnm{Li}, \binits{L.-J.}}:
\batitle{Yfcc100m: The new data in multimedia research}.
\bjtitle{Communications of the ACM}
\bvolume{59}(\bissue{2}),
\bfpage{64}--\blpage{73}
(\byear{2016})
\end{barticle}
\endbibitem

\bibitem[\protect\citeauthoryear{Bianco et~al.}{2015}]{bianco2015interactive}
\begin{barticle}
\bauthor{\bsnm{Bianco}, \binits{S.}},
\bauthor{\bsnm{Ciocca}, \binits{G.}},
\bauthor{\bsnm{Napoletano}, \binits{P.}},
\bauthor{\bsnm{Schettini}, \binits{R.}}:
\batitle{An interactive tool for manual, semi-automatic and automatic video
  annotation}.
\bjtitle{Computer Vision and Image Understanding}
\bvolume{131},
\bfpage{88}--\blpage{99}
(\byear{2015})
\end{barticle}
\endbibitem

\bibitem[\protect\citeauthoryear{Chiang et~al.}{2020}]{chiang2020using}
\begin{bbook}
\bauthor{\bsnm{Chiang}, \binits{Y.-Y.}},
\bauthor{\bsnm{Duan}, \binits{W.}},
\bauthor{\bsnm{Leyk}, \binits{S.}},
\bauthor{\bsnm{Uhl}, \binits{J.H.}},
\bauthor{\bsnm{Knoblock}, \binits{C.A.}}:
\bbtitle{Using Historical Maps in Scientific Studies: Applications, Challenges,
  and Best Practices}.
\bpublisher{Springer}, \blocation{???}
(\byear{2020})
\end{bbook}
\endbibitem

\bibitem[\protect\citeauthoryear{News}{2021}]{highcountrynews_placeidentity}
\begin{botherref}
\oauthor{\bsnm{News}, \binits{H.C.}}:
How place names impact the way we see landscape (The power in a name).
High Country News – Know the West.
Available at:
  \url{https://www.hcn.org/issues/54.5/people-places-how-place-names-impact-the-way-we-see-landscape}
(2021)
\end{botherref}
\endbibitem

\bibitem[\protect\citeauthoryear{Li et~al.}{2020}]{li2020automatic}
\begin{bchapter}
\bauthor{\bsnm{Li}, \binits{Z.}},
\bauthor{\bsnm{Chiang}, \binits{Y.-Y.}},
\bauthor{\bsnm{Tavakkol}, \binits{S.}},
\bauthor{\bsnm{Shbita}, \binits{B.}},
\bauthor{\bsnm{Uhl}, \binits{J.H.}},
\bauthor{\bsnm{Leyk}, \binits{S.}},
\bauthor{\bsnm{Knoblock}, \binits{C.A.}}:
\bctitle{An automatic approach for generating rich, linked geo-metadata from
  historical map images}.
In: \bbtitle{Proceedings of the 26th ACM SIGKDD International Conference on
  Knowledge Discovery \& Data Mining},
pp. \bfpage{3290}--\blpage{3298}
(\byear{2020})
\end{bchapter}
\endbibitem

\bibitem[\protect\citeauthoryear{Jang et~al.}{2017}]{jang2017kaggle}
\begin{botherref}
\oauthor{\bsnm{Jang}, \binits{H.}},
\oauthor{\bsnm{Kim}, \binits{S.}},
\oauthor{\bsnm{Lam}, \binits{T.}}:
Kaggle competitions: statoil/c-core iceberg classifier challenge.
Dept. School Inform., Comput., Eng. Indiana Univ., Bloomington, IN, USA, Tech.
  Rep
(2017)
\end{botherref}
\endbibitem

\bibitem[\protect\citeauthoryear{Zhang et~al.}{2018}]{zhang2018learning}
\begin{botherref}
\oauthor{\bsnm{Zhang}, \binits{Y.}},
\oauthor{\bsnm{Hare}, \binits{J.}},
\oauthor{\bsnm{Pr{\"u}gel-Bennett}, \binits{A.}}:
Learning to count objects in natural images for visual question answering.
arXiv preprint arXiv:1802.05766
(2018)
\end{botherref}
\endbibitem

\bibitem[\protect\citeauthoryear{Mu}{2016}]{mu2016air}
\begin{bchapter}
\bauthor{\bsnm{Mu}, \binits{L.}}:
\bctitle{Air quality estimation with crowdsourcing spatiotemporally tagged
  digital photos}.
In: \bbtitle{Presented at the 112nd Annual Conference of the Association of
  American Geographers},
\bconflocation{San Francisco, CA}
(\byear{2016})
\end{bchapter}
\endbibitem

\bibitem[\protect\citeauthoryear{Yao and Huang}{2021}]{yao2021extraction}
\begin{barticle}
\bauthor{\bsnm{Yao}, \binits{S.}},
\bauthor{\bsnm{Huang}, \binits{B.}}:
\batitle{Extraction of aerosol optical extinction properties from a smartphone
  photograph to measure visibility}.
\bjtitle{IEEE Transactions on Geoscience and Remote Sensing}
\bvolume{60},
\bfpage{1}--\blpage{13}
(\byear{2021})
\end{barticle}
\endbibitem

\bibitem[\protect\citeauthoryear{Liaw and Chen}{2021}]{liaw2021using}
\begin{barticle}
\bauthor{\bsnm{Liaw}, \binits{J.-J.}},
\bauthor{\bsnm{Chen}, \binits{K.-Y.}}:
\batitle{Using high-frequency information and rh to estimate aqi based on svr}.
\bjtitle{Sensors}
\bvolume{21}(\bissue{11}),
\bfpage{3630}
(\byear{2021})
\end{barticle}
\endbibitem

\bibitem[\protect\citeauthoryear{Aleksandrov}{2019}]{aleksandrov2019identification}
\begin{botherref}
\oauthor{\bsnm{Aleksandrov}, \binits{V.}}:
Identification of nutrient deficiency in bean plants by prompt chlorophyll
  fluorescence measurements and artificial neural networks.
arXiv preprint arXiv:1906.03312
(2019)
\end{botherref}
\endbibitem

\bibitem[\protect\citeauthoryear{Pineda et~al.}{2020}]{pineda2020thermal}
\begin{barticle}
\bauthor{\bsnm{Pineda}, \binits{M.}},
\bauthor{\bsnm{Bar{\'o}n}, \binits{M.}},
\bauthor{\bsnm{P{\'e}rez-Bueno}, \binits{M.-L.}}:
\batitle{Thermal imaging for plant stress detection and phenotyping}.
\bjtitle{Remote Sensing}
\bvolume{13}(\bissue{1}),
\bfpage{68}
(\byear{2020})
\end{barticle}
\endbibitem

\bibitem[\protect\citeauthoryear{Wasonga et~al.}{2021}]{wasonga2021red}
\begin{barticle}
\bauthor{\bsnm{Wasonga}, \binits{D.O.}},
\bauthor{\bsnm{Yaw}, \binits{A.}},
\bauthor{\bsnm{Kleemola}, \binits{J.}},
\bauthor{\bsnm{Alakukku}, \binits{L.}},
\bauthor{\bsnm{M{\"a}kel{\"a}}, \binits{P.S.}}:
\batitle{Red-green-blue and multispectral imaging as potential tools for
  estimating growth and nutritional performance of cassava under deficit
  irrigation and potassium fertigation}.
\bjtitle{Remote Sensing}
\bvolume{13}(\bissue{4}),
\bfpage{598}
(\byear{2021})
\end{barticle}
\endbibitem

\bibitem[\protect\citeauthoryear{Debnath et~al.}{2021}]{debnath2021identifying}
\begin{barticle}
\bauthor{\bsnm{Debnath}, \binits{S.}},
\bauthor{\bsnm{Paul}, \binits{M.}},
\bauthor{\bsnm{Rahaman}, \binits{D.M.}},
\bauthor{\bsnm{Debnath}, \binits{T.}},
\bauthor{\bsnm{Zheng}, \binits{L.}},
\bauthor{\bsnm{Baby}, \binits{T.}},
\bauthor{\bsnm{Schmidtke}, \binits{L.M.}},
\bauthor{\bsnm{Rogiers}, \binits{S.Y.}}:
\batitle{Identifying individual nutrient deficiencies of grapevine leaves using
  hyperspectral imaging}.
\bjtitle{Remote Sensing}
\bvolume{13}(\bissue{16}),
\bfpage{3317}
(\byear{2021})
\end{barticle}
\endbibitem

\bibitem[\protect\citeauthoryear{Condori et~al.}{2017}]{condori2017comparison}
\begin{bchapter}
\bauthor{\bsnm{Condori}, \binits{R.H.M.}},
\bauthor{\bsnm{Romualdo}, \binits{L.M.}},
\bauthor{\bsnm{Bruno}, \binits{O.M.}},
\bauthor{\bsnm{Cerqueira~Luz}, \binits{P.H.}}:
\bctitle{Comparison between traditional texture methods and deep learning
  descriptors for detection of nitrogen deficiency in maize crops}.
In: \bbtitle{2017 Workshop of Computer Vision (WVC)},
pp. \bfpage{7}--\blpage{12}
(\byear{2017}).
\bcomment{IEEE}
\end{bchapter}
\endbibitem

\bibitem[\protect\citeauthoryear{Ghosal et~al.}{2018}]{ghosal2018explainable}
\begin{barticle}
\bauthor{\bsnm{Ghosal}, \binits{S.}},
\bauthor{\bsnm{Blystone}, \binits{D.}},
\bauthor{\bsnm{Singh}, \binits{A.K.}},
\bauthor{\bsnm{Ganapathysubramanian}, \binits{B.}},
\bauthor{\bsnm{Singh}, \binits{A.}},
\bauthor{\bsnm{Sarkar}, \binits{S.}}:
\batitle{An explainable deep machine vision framework for plant stress
  phenotyping}.
\bjtitle{Proceedings of the National Academy of Sciences}
\bvolume{115}(\bissue{18}),
\bfpage{4613}--\blpage{4618}
(\byear{2018})
\end{barticle}
\endbibitem

\bibitem[\protect\citeauthoryear{Xu et~al.}{2020}]{xu2020using}
\begin{botherref}
\oauthor{\bsnm{Xu}, \binits{Z.}},
\oauthor{\bsnm{Guo}, \binits{X.}},
\oauthor{\bsnm{Zhu}, \binits{A.}},
\oauthor{\bsnm{He}, \binits{X.}},
\oauthor{\bsnm{Zhao}, \binits{X.}},
\oauthor{\bsnm{Han}, \binits{Y.}},
\oauthor{\bsnm{Subedi}, \binits{R.}}:
Using deep convolutional neural networks for image-based diagnosis of nutrient
  deficiencies in rice.
Computational Intelligence and Neuroscience
\textbf{2020}
(2020)
\end{botherref}
\endbibitem

\bibitem[\protect\citeauthoryear{Zermas et~al.}{2020}]{zermas2020methodology}
\begin{barticle}
\bauthor{\bsnm{Zermas}, \binits{D.}},
\bauthor{\bsnm{Nelson}, \binits{H.J.}},
\bauthor{\bsnm{Stanitsas}, \binits{P.}},
\bauthor{\bsnm{Morellas}, \binits{V.}},
\bauthor{\bsnm{Mulla}, \binits{D.J.}},
\bauthor{\bsnm{Papanikolopoulos}, \binits{N.}}:
\batitle{A methodology for the detection of nitrogen deficiency in corn fields
  using high-resolution rgb imagery}.
\bjtitle{IEEE Transactions on Automation Science and Engineering}
\bvolume{18}(\bissue{4}),
\bfpage{1879}--\blpage{1891}
(\byear{2020})
\end{barticle}
\endbibitem

\bibitem[\protect\citeauthoryear{Sharma et~al.}{2022}]{sharma2022ensemble}
\begin{barticle}
\bauthor{\bsnm{Sharma}, \binits{M.}},
\bauthor{\bsnm{Nath}, \binits{K.}},
\bauthor{\bsnm{Sharma}, \binits{R.K.}},
\bauthor{\bsnm{Kumar}, \binits{C.J.}},
\bauthor{\bsnm{Chaudhary}, \binits{A.}}:
\batitle{Ensemble averaging of transfer learning models for identification of
  nutritional deficiency in rice plant}.
\bjtitle{Electronics}
\bvolume{11}(\bissue{1}),
\bfpage{148}
(\byear{2022})
\end{barticle}
\endbibitem

\bibitem[\protect\citeauthoryear{Chiu et~al.}{2020}]{chiu2020agriculture}
\begin{bchapter}
\bauthor{\bsnm{Chiu}, \binits{M.T.}},
\bauthor{\bsnm{Xu}, \binits{X.}},
\bauthor{\bsnm{Wei}, \binits{Y.}},
\bauthor{\bsnm{Huang}, \binits{Z.}},
\bauthor{\bsnm{Schwing}, \binits{A.G.}},
\bauthor{\bsnm{Brunner}, \binits{R.}},
\bauthor{\bsnm{Khachatrian}, \binits{H.}},
\bauthor{\bsnm{Karapetyan}, \binits{H.}},
\bauthor{\bsnm{Dozier}, \binits{I.}},
\bauthor{\bsnm{Rose}, \binits{G.}}, \betal:
\bctitle{Agriculture-vision: A large aerial image database for agricultural
  pattern analysis}.
In: \bbtitle{Proceedings of the IEEE/CVF Conference on Computer Vision and
  Pattern Recognition},
pp. \bfpage{2828}--\blpage{2838}
(\byear{2020})
\end{bchapter}
\endbibitem

\bibitem[\protect\citeauthoryear{Dhamodharan}{2023}]{dhamodharan2023}
\begin{botherref}
\oauthor{\bsnm{Dhamodharan}}:
Cotton plant disease.
Kaggle.
[Accessed: 2023-10-2]
(2023).
\doiurl{10.34740/KAGGLE/DSV/5127834} .
\url{https://www.kaggle.com/dsv/5127834}
\end{botherref}
\endbibitem

\bibitem[\protect\citeauthoryear{Lu}{2021}]{yuzhenlu2021}
\begin{botherref}
\oauthor{\bsnm{Lu}, \binits{Y.}}:
CottonWeedID15.
Kaggle.
[Accessed: 2023-10-1]
(2021).
\doiurl{10.34740/KAGGLE/DSV/2685927} .
\url{https://www.kaggle.com/dsv/2685927}
\end{botherref}
\endbibitem

\bibitem[\protect\citeauthoryear{Tan et~al.}{2022}]{tan2022towards}
\begin{barticle}
\bauthor{\bsnm{Tan}, \binits{C.}},
\bauthor{\bsnm{Li}, \binits{C.}},
\bauthor{\bsnm{He}, \binits{D.}},
\bauthor{\bsnm{Song}, \binits{H.}}:
\batitle{Towards real-time tracking and counting of seedlings with a one-stage
  detector and optical flow}.
\bjtitle{Computers and Electronics in Agriculture}
\bvolume{193},
\bfpage{106683}
(\byear{2022})
\end{barticle}
\endbibitem

\bibitem[\protect\citeauthoryear{Tan et~al.}{2023a}]{tan2023anchor}
\begin{barticle}
\bauthor{\bsnm{Tan}, \binits{C.}},
\bauthor{\bsnm{Li}, \binits{C.}},
\bauthor{\bsnm{He}, \binits{D.}},
\bauthor{\bsnm{Song}, \binits{H.}}:
\batitle{Anchor-free deep convolutional neural network for tracking and
  counting cotton seedlings and flowers}.
\bjtitle{Computers and Electronics in Agriculture}
\bvolume{215},
\bfpage{108359}
(\byear{2023})
\end{barticle}
\endbibitem

\bibitem[\protect\citeauthoryear{Tan et~al.}{2023b}]{tan2023three}
\begin{bchapter}
\bauthor{\bsnm{Tan}, \binits{C.}},
\bauthor{\bsnm{Li}, \binits{C.}},
\bauthor{\bsnm{Sun}, \binits{J.}},
\bauthor{\bsnm{Song}, \binits{H.}}:
\bctitle{Three-view cotton flower counting through multi-object tracking and
  multi-modal imaging}.
In: \bbtitle{2023 ASABE Annual International Meeting},
p. \bfpage{1}
(\byear{2023}).
\bcomment{American Society of Agricultural and Biological Engineers}
\end{bchapter}
\endbibitem

\bibitem[\protect\citeauthoryear{Bist et~al.}{2023}]{bist2023mislaying}
\begin{barticle}
\bauthor{\bsnm{Bist}, \binits{R.B.}},
\bauthor{\bsnm{Yang}, \binits{X.}},
\bauthor{\bsnm{Subedi}, \binits{S.}},
\bauthor{\bsnm{Chai}, \binits{L.}}:
\batitle{Mislaying behavior detection in cage-free hens with deep learning
  technologies}.
\bjtitle{Poultry Science}
\bvolume{102}(\bissue{7}),
\bfpage{102729}
(\byear{2023})
\end{barticle}
\endbibitem

\bibitem[\protect\citeauthoryear{Yang et~al.}{2023a}]{yang2023deep}
\begin{barticle}
\bauthor{\bsnm{Yang}, \binits{X.}},
\bauthor{\bsnm{Bist}, \binits{R.}},
\bauthor{\bsnm{Subedi}, \binits{S.}},
\bauthor{\bsnm{Chai}, \binits{L.}}:
\batitle{A deep learning method for monitoring spatial distribution of
  cage-free hens}.
\bjtitle{Artificial Intelligence in Agriculture}
\bvolume{8},
\bfpage{20}--\blpage{29}
(\byear{2023})
\end{barticle}
\endbibitem

\bibitem[\protect\citeauthoryear{Yang et~al.}{2023b}]{yang2023sam}
\begin{botherref}
\oauthor{\bsnm{Yang}, \binits{X.}},
\oauthor{\bsnm{Dai}, \binits{H.}},
\oauthor{\bsnm{Wu}, \binits{Z.}},
\oauthor{\bsnm{Bist}, \binits{R.}},
\oauthor{\bsnm{Subedi}, \binits{S.}},
\oauthor{\bsnm{Sun}, \binits{J.}},
\oauthor{\bsnm{Lu}, \binits{G.}},
\oauthor{\bsnm{Li}, \binits{C.}},
\oauthor{\bsnm{Liu}, \binits{T.}},
\oauthor{\bsnm{Chai}, \binits{L.}}:
Sam for poultry science.
arXiv preprint arXiv:2305.10254
(2023)
\end{botherref}
\endbibitem

\bibitem[\protect\citeauthoryear{Bist et~al.}{2023}]{bist2023automatic}
\begin{barticle}
\bauthor{\bsnm{Bist}, \binits{R.B.}},
\bauthor{\bsnm{Subedi}, \binits{S.}},
\bauthor{\bsnm{Yang}, \binits{X.}},
\bauthor{\bsnm{Chai}, \binits{L.}}:
\batitle{Automatic detection of cage-free dead hens with deep learning
  methods}.
\bjtitle{AgriEngineering}
\bvolume{5}(\bissue{2}),
\bfpage{1020}--\blpage{1038}
(\byear{2023})
\end{barticle}
\endbibitem

\bibitem[\protect\citeauthoryear{Guo et~al.}{2023}]{guo2023detecting}
\begin{barticle}
\bauthor{\bsnm{Guo}, \binits{Y.}},
\bauthor{\bsnm{Aggrey}, \binits{S.E.}},
\bauthor{\bsnm{Yang}, \binits{X.}},
\bauthor{\bsnm{Oladeinde}, \binits{A.}},
\bauthor{\bsnm{Qiao}, \binits{Y.}},
\bauthor{\bsnm{Chai}, \binits{L.}}:
\batitle{Detecting broiler chickens on litter floor with the yolov5-cbam deep
  learning model}.
\bjtitle{Artificial Intelligence in Agriculture}
\bvolume{9},
\bfpage{36}--\blpage{45}
(\byear{2023})
\end{barticle}
\endbibitem

\bibitem[\protect\citeauthoryear{Bist et~al.}{2023}]{bist2023ammonia}
\begin{barticle}
\bauthor{\bsnm{Bist}, \binits{R.B.}},
\bauthor{\bsnm{Subedi}, \binits{S.}},
\bauthor{\bsnm{Chai}, \binits{L.}},
\bauthor{\bsnm{Yang}, \binits{X.}}:
\batitle{Ammonia emissions, impacts, and mitigation strategies for poultry
  production: A critical review}.
\bjtitle{Journal of Environmental Management}
\bvolume{328},
\bfpage{116919}
(\byear{2023})
\end{barticle}
\endbibitem

\bibitem[\protect\citeauthoryear{Yang et~al.}{2023}]{YANG2023106377}
\begin{barticle}
\bauthor{\bsnm{Yang}, \binits{X.}},
\bauthor{\bsnm{Bist}, \binits{R.}},
\bauthor{\bsnm{Subedi}, \binits{S.}},
\bauthor{\bsnm{Wu}, \binits{Z.}},
\bauthor{\bsnm{Liu}, \binits{T.}},
\bauthor{\bsnm{Chai}, \binits{L.}}:
\batitle{An automatic classifier for monitoring applied behaviors of cage-free
  laying hens with deep learning}.
\bjtitle{Engineering Applications of Artificial Intelligence}
\bvolume{123},
\bfpage{106377}
(\byear{2023})
\doiurl{10.1016/j.engappai.2023.106377}
\end{barticle}
\endbibitem

\bibitem[\protect\citeauthoryear{Bist et~al.}{2022}]{bist2022air}
\begin{bchapter}
\bauthor{\bsnm{Bist}, \binits{R.B.}},
\bauthor{\bsnm{Chai}, \binits{L.}},
\bauthor{\bsnm{Yang}, \binits{X.}},
\bauthor{\bsnm{Subedi}, \binits{S.}},
\bauthor{\bsnm{Guo}, \binits{Y.}}:
\bctitle{Air quality in cage-free houses during pullets production}.
In: \bbtitle{2022 ASABE Annual International Meeting},
p. \bfpage{1}
(\byear{2022}).
\bcomment{American Society of Agricultural and Biological Engineers}
\end{bchapter}
\endbibitem

\bibitem[\protect\citeauthoryear{Bist et~al.}{2023}]{bist2023effective}
\begin{barticle}
\bauthor{\bsnm{Bist}, \binits{R.B.}},
\bauthor{\bsnm{Subedi}, \binits{S.}},
\bauthor{\bsnm{Yang}, \binits{X.}},
\bauthor{\bsnm{Chai}, \binits{L.}}:
\batitle{Effective strategies for mitigating feather pecking and cannibalism in
  cage-free w-36 pullets}.
\bjtitle{Poultry}
\bvolume{2}(\bissue{2}),
\bfpage{281}--\blpage{291}
(\byear{2023})
\end{barticle}
\endbibitem

\bibitem[\protect\citeauthoryear{Yang et~al.}{2022}]{yang2022deep}
\begin{barticle}
\bauthor{\bsnm{Yang}, \binits{X.}},
\bauthor{\bsnm{Chai}, \binits{L.}},
\bauthor{\bsnm{Bist}, \binits{R.B.}},
\bauthor{\bsnm{Subedi}, \binits{S.}},
\bauthor{\bsnm{Wu}, \binits{Z.}}:
\batitle{A deep learning model for detecting cage-free hens on the litter
  floor}.
\bjtitle{Animals}
\bvolume{12}(\bissue{15}),
\bfpage{1983}
(\byear{2022})
\end{barticle}
\endbibitem

\bibitem[\protect\citeauthoryear{Bist et~al.}{2023}]{bist2023novel}
\begin{barticle}
\bauthor{\bsnm{Bist}, \binits{R.B.}},
\bauthor{\bsnm{Subedi}, \binits{S.}},
\bauthor{\bsnm{Yang}, \binits{X.}},
\bauthor{\bsnm{Chai}, \binits{L.}}:
\batitle{A novel yolov6 object detector for monitoring piling behavior of
  cage-free laying hens}.
\bjtitle{AgriEngineering}
\bvolume{5}(\bissue{2}),
\bfpage{905}--\blpage{923}
(\byear{2023})
\end{barticle}
\endbibitem

\bibitem[\protect\citeauthoryear{Yang et~al.}{2015}]{yang2015wikiqa}
\begin{bchapter}
\bauthor{\bsnm{Yang}, \binits{Y.}},
\bauthor{\bsnm{Yih}, \binits{W.-t.}},
\bauthor{\bsnm{Meek}, \binits{C.}}:
\bctitle{Wikiqa: A challenge dataset for open-domain question answering}.
In: \bbtitle{Proceedings of the 2015 Conference on Empirical Methods in Natural
  Language Processing},
pp. \bfpage{2013}--\blpage{2018}
(\byear{2015})
\end{bchapter}
\endbibitem

\bibitem[\protect\citeauthoryear{Rajpurkar et~al.}{2016}]{rajpurkar2016squad}
\begin{bchapter}
\bauthor{\bsnm{Rajpurkar}, \binits{P.}},
\bauthor{\bsnm{Zhang}, \binits{J.}},
\bauthor{\bsnm{Lopyrev}, \binits{K.}},
\bauthor{\bsnm{Liang}, \binits{P.}}:
\bctitle{Squad: 100,000+ questions for machine comprehension of text}.
In: \bbtitle{Proceedings of the 2016 Conference on Empirical Methods in Natural
  Language Processing}
(\byear{2016}).
\bcomment{Association for Computational Linguistics}
\end{bchapter}
\endbibitem

\bibitem[\protect\citeauthoryear{Rajpurkar et~al.}{2018}]{rajpurkar2018know}
\begin{bchapter}
\bauthor{\bsnm{Rajpurkar}, \binits{P.}},
\bauthor{\bsnm{Jia}, \binits{R.}},
\bauthor{\bsnm{Liang}, \binits{P.}}:
\bctitle{Know what you don’t know: Unanswerable questions for squad}.
In: \bbtitle{Proceedings of the 56th Annual Meeting of the Association for
  Computational Linguistics (Volume 2: Short Papers)}
(\byear{2018}).
\bcomment{Association for Computational Linguistics}
\end{bchapter}
\endbibitem

\bibitem[\protect\citeauthoryear{Reddy et~al.}{2019}]{reddy2019coqa}
\begin{barticle}
\bauthor{\bsnm{Reddy}, \binits{S.}},
\bauthor{\bsnm{Chen}, \binits{D.}},
\bauthor{\bsnm{Manning}, \binits{C.D.}}:
\batitle{Coqa: A conversational question answering challenge}.
\bjtitle{Transactions of the Association for Computational Linguistics}
\bvolume{7},
\bfpage{249}--\blpage{266}
(\byear{2019})
\end{barticle}
\endbibitem

\bibitem[\protect\citeauthoryear{Karpukhin et~al.}{2020}]{karpukhin2020dpr}
\begin{bchapter}
\bauthor{\bsnm{Karpukhin}, \binits{V.}},
\bauthor{\bsnm{Oguz}, \binits{B.}},
\bauthor{\bsnm{Min}, \binits{S.}},
\bauthor{\bsnm{Lewis}, \binits{P.}},
\bauthor{\bsnm{Wu}, \binits{L.}},
\bauthor{\bsnm{Edunov}, \binits{S.}},
\bauthor{\bsnm{Chen}, \binits{D.}},
\bauthor{\bsnm{Yih}, \binits{W.-t.}}:
\bctitle{Dense passage retrieval for open-domain question answering}.
In: \bbtitle{Proceedings of the 2020 Conference on Empirical Methods in Natural
  Language Processing (EMNLP)},
pp. \bfpage{6769}--\blpage{6781}
(\byear{2020})
\end{bchapter}
\endbibitem

\bibitem[\protect\citeauthoryear{Mishra et~al.}{2010}]{mishra2010context}
\begin{bchapter}
\bauthor{\bsnm{Mishra}, \binits{A.}},
\bauthor{\bsnm{Mishra}, \binits{N.}},
\bauthor{\bsnm{Agrawal}, \binits{A.}}:
\bctitle{Context-aware restricted geographical domain question answering
  system}.
In: \bbtitle{2010 International Conference on Computational Intelligence and
  Communication Networks},
pp. \bfpage{548}--\blpage{553}
(\byear{2010}).
\bcomment{IEEE}
\end{bchapter}
\endbibitem

\bibitem[\protect\citeauthoryear{Chen et~al.}{2013}]{chen2013synergistic}
\begin{bchapter}
\bauthor{\bsnm{Chen}, \binits{W.}},
\bauthor{\bsnm{Fosler-Lussier}, \binits{E.}},
\bauthor{\bsnm{Xiao}, \binits{N.}},
\bauthor{\bsnm{Raje}, \binits{S.}},
\bauthor{\bsnm{Ramnath}, \binits{R.}},
\bauthor{\bsnm{Sui}, \binits{D.}}:
\bctitle{A synergistic framework for geographic question answering}.
In: \bbtitle{2013 IEEE Seventh International Conference on Semantic Computing},
pp. \bfpage{94}--\blpage{99}
(\byear{2013}).
\bcomment{IEEE}
\end{bchapter}
\endbibitem

\bibitem[\protect\citeauthoryear{Mai et~al.}{2018}]{mai2018poireviewqa}
\begin{bchapter}
\bauthor{\bsnm{Mai}, \binits{G.}},
\bauthor{\bsnm{Janowicz}, \binits{K.}},
\bauthor{\bsnm{He}, \binits{C.}},
\bauthor{\bsnm{Liu}, \binits{S.}},
\bauthor{\bsnm{Lao}, \binits{N.}}:
\bctitle{Poireviewqa: A semantically enriched poi retrieval and question
  answering dataset}.
In: \bbtitle{Proceedings of the 12th Workshop on Geographic Information
  Retrieval},
pp. \bfpage{1}--\blpage{2}
(\byear{2018})
\end{bchapter}
\endbibitem

\bibitem[\protect\citeauthoryear{Mai et~al.}{2020}]{mai2020relaxing}
\begin{bchapter}
\bauthor{\bsnm{Mai}, \binits{G.}},
\bauthor{\bsnm{Yan}, \binits{B.}},
\bauthor{\bsnm{Janowicz}, \binits{K.}},
\bauthor{\bsnm{Zhu}, \binits{R.}}:
\bctitle{Relaxing unanswerable geographic questions using a spatially explicit
  knowledge graph embedding model}.
In: \bbtitle{Geospatial Technologies for Local and Regional Development:
  Proceedings of the 22nd AGILE Conference on Geographic Information Science
  22},
pp. \bfpage{21}--\blpage{39}
(\byear{2020}).
\bcomment{Springer}
\end{bchapter}
\endbibitem

\bibitem[\protect\citeauthoryear{Mai et~al.}{2021}]{mai2021geographic}
\begin{barticle}
\bauthor{\bsnm{Mai}, \binits{G.}},
\bauthor{\bsnm{Janowicz}, \binits{K.}},
\bauthor{\bsnm{Zhu}, \binits{R.}},
\bauthor{\bsnm{Cai}, \binits{L.}},
\bauthor{\bsnm{Lao}, \binits{N.}}:
\batitle{Geographic question answering: Challenges, uniqueness, classification,
  and future directions}.
\bjtitle{AGILE: GIScience series}
\bvolume{2},
\bfpage{8}
(\byear{2021})
\end{barticle}
\endbibitem

\bibitem[\protect\citeauthoryear{Mai et~al.}{2020}]{mai2020se}
\begin{barticle}
\bauthor{\bsnm{Mai}, \binits{G.}},
\bauthor{\bsnm{Janowicz}, \binits{K.}},
\bauthor{\bsnm{Cai}, \binits{L.}},
\bauthor{\bsnm{Zhu}, \binits{R.}},
\bauthor{\bsnm{Regalia}, \binits{B.}},
\bauthor{\bsnm{Yan}, \binits{B.}},
\bauthor{\bsnm{Shi}, \binits{M.}},
\bauthor{\bsnm{Lao}, \binits{N.}}:
\batitle{Se-kge: A location-aware knowledge graph embedding model for
  geographic question answering and spatial semantic lifting}.
\bjtitle{Transactions in GIS}
\bvolume{24}(\bissue{3}),
\bfpage{623}--\blpage{655}
(\byear{2020})
\end{barticle}
\endbibitem

\bibitem[\protect\citeauthoryear{Huang et~al.}{2019}]{huang2019geosqa}
\begin{bchapter}
\bauthor{\bsnm{Huang}, \binits{Z.}},
\bauthor{\bsnm{Shen}, \binits{Y.}},
\bauthor{\bsnm{Li}, \binits{X.}},
\bauthor{\bsnm{Cheng}, \binits{G.}},
\bauthor{\bsnm{Zhou}, \binits{L.}},
\bauthor{\bsnm{Dai}, \binits{X.}},
\bauthor{\bsnm{Qu}, \binits{Y.}}, \betal:
\bctitle{Geosqa: A benchmark for scenario-based question answering in the
  geography domain at high school level}.
In: \bbtitle{Proceedings of the 2019 Conference on Empirical Methods in Natural
  Language Processing and the 9th International Joint Conference on Natural
  Language Processing (EMNLP-IJCNLP)},
pp. \bfpage{5866}--\blpage{5871}
(\byear{2019})
\end{bchapter}
\endbibitem

\bibitem[\protect\citeauthoryear{Scheider et~al.}{2021}]{scheider2021geo}
\begin{barticle}
\bauthor{\bsnm{Scheider}, \binits{S.}},
\bauthor{\bsnm{Nyamsuren}, \binits{E.}},
\bauthor{\bsnm{Kruiger}, \binits{H.}},
\bauthor{\bsnm{Xu}, \binits{H.}}:
\batitle{Geo-analytical question-answering with gis}.
\bjtitle{International Journal of Digital Earth}
\bvolume{14}(\bissue{1}),
\bfpage{1}--\blpage{14}
(\byear{2021})
\end{barticle}
\endbibitem

\bibitem[\protect\citeauthoryear{Lee et~al.}{2006}]{lee2006beyond}
\begin{bchapter}
\bauthor{\bsnm{Lee}, \binits{M.}},
\bauthor{\bsnm{Cimino}, \binits{J.}},
\bauthor{\bsnm{Zhu}, \binits{H.R.}},
\bauthor{\bsnm{Sable}, \binits{C.}},
\bauthor{\bsnm{Shanker}, \binits{V.}},
\bauthor{\bsnm{Ely}, \binits{J.}},
\bauthor{\bsnm{Yu}, \binits{H.}}:
\bctitle{Beyond information retrieval—medical question answering}.
In: \bbtitle{AMIA Annual Symposium Proceedings},
vol. \bseriesno{2006},
p. \bfpage{469}
(\byear{2006}).
\bcomment{American Medical Informatics Association}
\end{bchapter}
\endbibitem

\bibitem[\protect\citeauthoryear{Abacha and
  Zweigenbaum}{2015}]{abacha2015means}
\begin{barticle}
\bauthor{\bsnm{Abacha}, \binits{A.B.}},
\bauthor{\bsnm{Zweigenbaum}, \binits{P.}}:
\batitle{Means: A medical question-answering system combining nlp techniques
  and semantic web technologies}.
\bjtitle{Information processing \& management}
\bvolume{51}(\bissue{5}),
\bfpage{570}--\blpage{594}
(\byear{2015})
\end{barticle}
\endbibitem

\bibitem[\protect\citeauthoryear{Goodwin and
  Harabagiu}{2016}]{goodwin2016medical}
\begin{bchapter}
\bauthor{\bsnm{Goodwin}, \binits{T.R.}},
\bauthor{\bsnm{Harabagiu}, \binits{S.M.}}:
\bctitle{Medical question answering for clinical decision support}.
In: \bbtitle{Proceedings of the 25th ACM International on Conference on
  Information and Knowledge Management},
pp. \bfpage{297}--\blpage{306}
(\byear{2016})
\end{bchapter}
\endbibitem

\bibitem[\protect\citeauthoryear{Li et~al.}{2023}]{li2023llava}
\begin{bchapter}
\bauthor{\bsnm{Li}, \binits{C.}},
\bauthor{\bsnm{Wong}, \binits{C.}},
\bauthor{\bsnm{Zhang}, \binits{S.}},
\bauthor{\bsnm{Usuyama}, \binits{N.}},
\bauthor{\bsnm{Liu}, \binits{H.}},
\bauthor{\bsnm{Yang}, \binits{J.}},
\bauthor{\bsnm{Naumann}, \binits{T.}},
\bauthor{\bsnm{Poon}, \binits{H.}},
\bauthor{\bsnm{Gao}, \binits{J.}}:
\bctitle{Llava-med: Training a large language-and-vision assistant for
  biomedicine in one day}.
In: \bbtitle{Proceedings of the Thirty-seventh Conference on Neural Information
  Processing Systems}
(\byear{2023})
\end{bchapter}
\endbibitem

\bibitem[\protect\citeauthoryear{Gaikwad et~al.}{2015}]{gaikwad2015agri}
\begin{bchapter}
\bauthor{\bsnm{Gaikwad}, \binits{S.}},
\bauthor{\bsnm{Asodekar}, \binits{R.}},
\bauthor{\bsnm{Gadia}, \binits{S.}},
\bauthor{\bsnm{Attar}, \binits{V.Z.}}:
\bctitle{Agri-qas question-answering system for agriculture domain}.
In: \bbtitle{2015 International Conference on Advances in Computing,
  Communications and Informatics (ICACCI)},
pp. \bfpage{1474}--\blpage{1478}
(\byear{2015}).
\bcomment{IEEE}
\end{bchapter}
\endbibitem

\bibitem[\protect\citeauthoryear{Devi and Dua}{2017}]{devi2017adans}
\begin{bchapter}
\bauthor{\bsnm{Devi}, \binits{M.}},
\bauthor{\bsnm{Dua}, \binits{M.}}:
\bctitle{Adans: An agriculture domain question answering system using
  ontologies}.
In: \bbtitle{2017 International Conference on Computing, Communication and
  Automation (ICCCA)},
pp. \bfpage{122}--\blpage{127}
(\byear{2017}).
\bcomment{IEEE}
\end{bchapter}
\endbibitem

\bibitem[\protect\citeauthoryear{Ye et~al.}{2023}]{ye2023developing}
\begin{barticle}
\bauthor{\bsnm{Ye}, \binits{X.}},
\bauthor{\bsnm{Du}, \binits{J.}},
\bauthor{\bsnm{Han}, \binits{Y.}},
\bauthor{\bsnm{Newman}, \binits{G.}},
\bauthor{\bsnm{Retchless}, \binits{D.}},
\bauthor{\bsnm{Zou}, \binits{L.}},
\bauthor{\bsnm{Ham}, \binits{Y.}},
\bauthor{\bsnm{Cai}, \binits{Z.}}:
\batitle{Developing human-centered urban digital twins for community
  infrastructure resilience: A research agenda}.
\bjtitle{Journal of Planning Literature}
\bvolume{38}(\bissue{2}),
\bfpage{187}--\blpage{199}
(\byear{2023})
\end{barticle}
\endbibitem

\bibitem[\protect\citeauthoryear{Marx}{2012}]{marx2012encyclopedia}
\begin{botherref}
\oauthor{\bsnm{Marx}, \binits{K.}}:
Encyclopedia britannica.
Encyclopaedia Britannica Ultimate Reference Suite [M/CD]. Chicago:
  Encyclopsedia Britannica
(2012)
\end{botherref}
\endbibitem

\bibitem[\protect\citeauthoryear{Howard}{1965}]{howard1965garden}
\begin{bbook}
\bauthor{\bsnm{Howard}, \binits{E.}}:
\bbtitle{Garden Cities of To-morrow}
vol. \bseriesno{23}.
\bpublisher{Mit Press}, \blocation{???}
(\byear{1965})
\end{bbook}
\endbibitem

\bibitem[\protect\citeauthoryear{{Daniel Burnham, Edward H.
  Bennett}}{1909}]{chicago1909masterplan}
\begin{botherref}
\oauthor{\bsnm{{Daniel Burnham, Edward H. Bennett}}}:
{Plan of Chicago}.
Master plan,
{Commercial Club of Chicago}
(1909)
\end{botherref}
\endbibitem

\bibitem[\protect\citeauthoryear{Corbusier}{1967}]{corbusier1967radiant}
\begin{botherref}
\oauthor{\bsnm{Corbusier}, \binits{L.}}:
The radiant city: Elements of a doctrine of urbanism to be used as the basis of
  our machine-age civilization.
(No Title)
(1967)
\end{botherref}
\endbibitem

\bibitem[\protect\citeauthoryear{{Urban Redevelopment Authority,
  Singapore}}{2019}]{singapore2019masterplan}
\begin{botherref}
\oauthor{\bsnm{{Urban Redevelopment Authority, Singapore}}}:
{Master Plan 2019 of Singapore}.
Master plan,
{Urban Redevelopment Authority}
(2019).
\url{https://www.ura.gov.sg/Corporate/Planning/Master-Plan}
\end{botherref}
\endbibitem

\bibitem[\protect\citeauthoryear{{O'Fallon, IL}}{2021}]{ofallon2021masterplan}
\begin{botherref}
\oauthor{\bsnm{{O'Fallon, IL}}}:
{O'Fallon, IL Master Plan}.
Master plan,
{O'Fallon, IL}
(2021).
\url{https://www.ofallon.org/sites/g/files/vyhlif1031/f/uploads/ofallonil_mp_finaldraft_interactive_2021_1209.pdf}
\end{botherref}
\endbibitem

\bibitem[\protect\citeauthoryear{Sadik-Khan}{2012}]{sadik2012urban}
\begin{botherref}
\oauthor{\bsnm{Sadik-Khan}, \binits{J.}}:
Urban street design guide.
New York: NACTO
\textbf{554}
(2012)
\end{botherref}
\endbibitem

\bibitem[\protect\citeauthoryear{M{\'e}ndez et~al.}{2023}]{mendez2023machine}
\begin{botherref}
\oauthor{\bsnm{M{\'e}ndez}, \binits{M.}},
\oauthor{\bsnm{Merayo}, \binits{M.G.}},
\oauthor{\bsnm{N{\'u}{\~n}ez}, \binits{M.}}:
Machine learning algorithms to forecast air quality: a survey.
Artificial Intelligence Review,
1--36
(2023)
\end{botherref}
\endbibitem

\bibitem[\protect\citeauthoryear{Chen et~al.}{2022}]{chen2022performance}
\begin{barticle}
\bauthor{\bsnm{Chen}, \binits{D.}},
\bauthor{\bsnm{Lu}, \binits{Y.}},
\bauthor{\bsnm{Li}, \binits{Z.}},
\bauthor{\bsnm{Young}, \binits{S.}}:
\batitle{Performance evaluation of deep transfer learning on multi-class
  identification of common weed species in cotton production systems}.
\bjtitle{Computers and Electronics in Agriculture}
\bvolume{198},
\bfpage{107091}
(\byear{2022})
\end{barticle}
\endbibitem

\bibitem[\protect\citeauthoryear{Spathis and Kawsar}{2023}]{spathis2023first}
\begin{botherref}
\oauthor{\bsnm{Spathis}, \binits{D.}},
\oauthor{\bsnm{Kawsar}, \binits{F.}}:
The first step is the hardest: Pitfalls of representing and tokenizing temporal
  data for large language models.
arXiv preprint arXiv:2309.06236
(2023)
\end{botherref}
\endbibitem

\bibitem[\protect\citeauthoryear{Ye et~al.}{2023}]{ye2023toward}
\begin{botherref}
\oauthor{\bsnm{Ye}, \binits{X.}},
\oauthor{\bsnm{Newman}, \binits{G.}},
\oauthor{\bsnm{Lee}, \binits{C.}},
\oauthor{\bsnm{Van~Zandt}, \binits{S.}},
\oauthor{\bsnm{Jourdan}, \binits{D.}}:
Toward Urban artificial intelligence for developing justice-oriented smart
  cities.
SAGE Publications Sage CA: Los Angeles, CA
(2023)
\end{botherref}
\endbibitem

\end{thebibliography}

\end{document}